\documentclass[letterpaper,twocolumn,10pt]{article}

\usepackage[letterpaper,left=0.75in,right=0.75in,top=0.9in,bottom=1in,columnsep=0.3in]{geometry}

\usepackage[hyphens]{url}
\usepackage{natbib}
\usepackage{graphicx}
\usepackage{caption}
\usepackage{subcaption}
\usepackage{booktabs}
\usepackage{adjustbox}

\usepackage{algorithm}
\usepackage{algorithmic}

\usepackage{newfloat}
\usepackage{listings}
\DeclareCaptionStyle{ruled}{labelfont=normalfont,labelsep=colon,strut=off}
\floatstyle{ruled}
\newfloat{listing}{tb}{lst}{}
\floatname{listing}{Listing}

\usepackage{tikz,pgfplots}
\pgfplotsset{compat=1.18}
\usetikzlibrary{external}
\usetikzlibrary{calc}
\usetikzlibrary{arrows.meta}
\usepackage{tcolorbox}

\usepackage{amsmath}
\usepackage{amsthm}
\usepackage{amssymb}
\usepackage{mathtools}
\DeclarePairedDelimiter\abs{\lvert}{\rvert}

\usepackage{newtxtext}
\usepackage{newtxmath}   
\usepackage{bm}          
\usepackage{upgreek}

\usepackage{microtype}
\usepackage{lipsum}
\usepackage{placeins}

\usepackage{titlesec}
\titleformat{\section}{\normalfont\large\bfseries}{\thesection}{0.6em}{}
\titleformat{\subsection}{\normalfont\normalsize\bfseries}{\thesubsection}{0.6em}{}
\titleformat{\subsubsection}{\normalfont\normalsize\bfseries\itshape}{\thesubsubsection}{0.6em}{}
\titlespacing*{\section}{0pt}{1.2ex plus .2ex}{0.6ex}
\titlespacing*{\subsection}{0pt}{1.0ex plus .2ex}{0.4ex}
\theoremstyle{plain}
\newtheorem{theorem}{Theorem}
\newtheorem{proposition}[theorem]{Proposition}

\theoremstyle{definition}

\theoremstyle{remark}

\usepackage{hyperref}
\usepackage[capitalize,noabbrev]{cleveref}

\usepackage[textsize=tiny]{todonotes}

\title{\bfseries Eigenvalues as a Metric for Memory Dynamics in Sequence Models}

\author{
  \textbf{Rahel Rickenbach}\textsuperscript{*,1},
  \textbf{Jelena Trisovic}\textsuperscript{*,1,2},\\
  \textbf{Alexandre Didier}\textsuperscript{1},
  \textbf{Jerome Sieber}\textsuperscript{\dag,1}, and
  \textbf{Melanie N. Zeilinger}\textsuperscript{\dag,1} \\[0.6em]
  \texttt{\{rrahel, tjelena, adidier, jsieber, mzeilinger\}@ethz.ch} \\[0.2em]
  \textsuperscript{1}ETH Zurich \textsuperscript{2}ETH AI Center
}
\date{}

\begin{document}

\maketitle

\renewcommand{\thefootnote}{\fnsymbol{footnote}}
\footnotetext[1]{Authors contributed equally (shared first authorship).}
\footnotetext[2]{Authors contributed equally (shared last authorship).}
\renewcommand{\thefootnote}{\arabic{footnote}}
\begin{abstract}
While softmax attention drives state-of-the-art performance in sequence modeling, its quadratic complexity motivates linear alternatives such as state space models (SSMs).  Structural differences between the two model classes, however, hinder direct comparisons of their memory dynamics, creating the need for a common metric to analyze, interpret, and improve their information processing capabilities. Inspired by recent advances in SSM performance driven by eigenvalue-guided insights, we leverage the dynamical systems framework to bring attention models into a unified analytical framework with SSMs. This allows us to perform a structured analysis, which investigates the applicability of an eigenvalue-spectrum memory dynamics metric to attention models. To this end, we first conduct an extensive empirical study across diverse attention-based models and SSMs on a range of benchmarks. We show that, for both model classes, eigenvalues influence key aspects of memory and long-range dependency modeling, revealing spectral signatures that align with task requirements. Building on these findings, we show how spectral signatures can motivate architectural modifications, how they can be guided through the training process, and how they can provide information about feature importance. The results thereby enable and emphasize the role of eigenvalue analysis as a principled metric for interpreting, explaining, and ultimately improving the capabilities of sequence models.\footnote[1]{Code is available at \href{https://github.com/IntelligentControlSystems/Task-Level-Insights-from-Eigenvalues-across-Sequence-Models}{github.com/IntelligentControlSystems/Task-Level-Insights-across-Sequence-Models}.}
\end{abstract}

\section{Introduction}
Deep sequence models are central to modern artificial intelligence, with transformers and softmax attention often forming their backbone and driving much of their success~\cite{bommasani2021opportunities,Transformer}. However, the quadratic complexity of softmax attention limits scalability~\cite{Tay2021} and motivated linear alternatives such as state space models (SSMs) that aim to preserve expressive power while enabling efficient long-context modeling~\cite{Katharopoulos2020,Gu2022,s5,Orvieto2023,mamba2,xlstm,schlag2021linear}.
Despite these advances, a fundamental question remains: \textit{how do different model classes process and retain information?} \\
For Recurrent Neural Networks (RNNs) and SSMs, this question has traditionally been approached through
linear systems theory (see the Appendix for an extended
discussion). Eigenvalue spectra characterize stability and the trade-off
between memory retention and information decay: in RNNs, they have been used
to mitigate vanishing gradients~\cite{helfrich2019eigenvaluenormalizedrecurrentneural}
and linked to task-specific structures~\cite{Jarne_2022,naiman2023operator},
while in SSMs, eigenvalue placement at initialization enables long-range
dependency modeling~\cite{Gu2020,Fu2023} and the transition-matrix eigenvalues
govern stability and memory length~\cite{wang2024stablessmalleviatingcursememory,grazzi2025unlockingstatetrackinglinearrnns}.
Closely related memory-function analyses~\cite{oppenheim1997signals} quantify
the influence of past inputs on outputs of SSMs~\cite{wang2023state} and
RNNs~\cite{Su_2019}, but require architecture-specific design choices and
generalize poorly to linear parameter-varying (LPV) systems such as attention
and Mamba~\cite{mamba2}. For attention-based models, analyses have instead
centered on the score matrix~\cite{bao2024self,bhojanapalli2021eigen} or on
memory-extension mechanisms~\cite{raffel2016feedforwardnetworksattentionsolve,sukhbaatar2019augmentingselfattentionpersistentmemory},
which do not translate into a framework permitting systematic comparison with
SSMs (see Appendix for details). A unified, architecture-agnostic metric for memory dynamics is thus
still missing.

\begin{figure*}[t]
    \centering
    \includegraphics[width=0.95\textwidth]{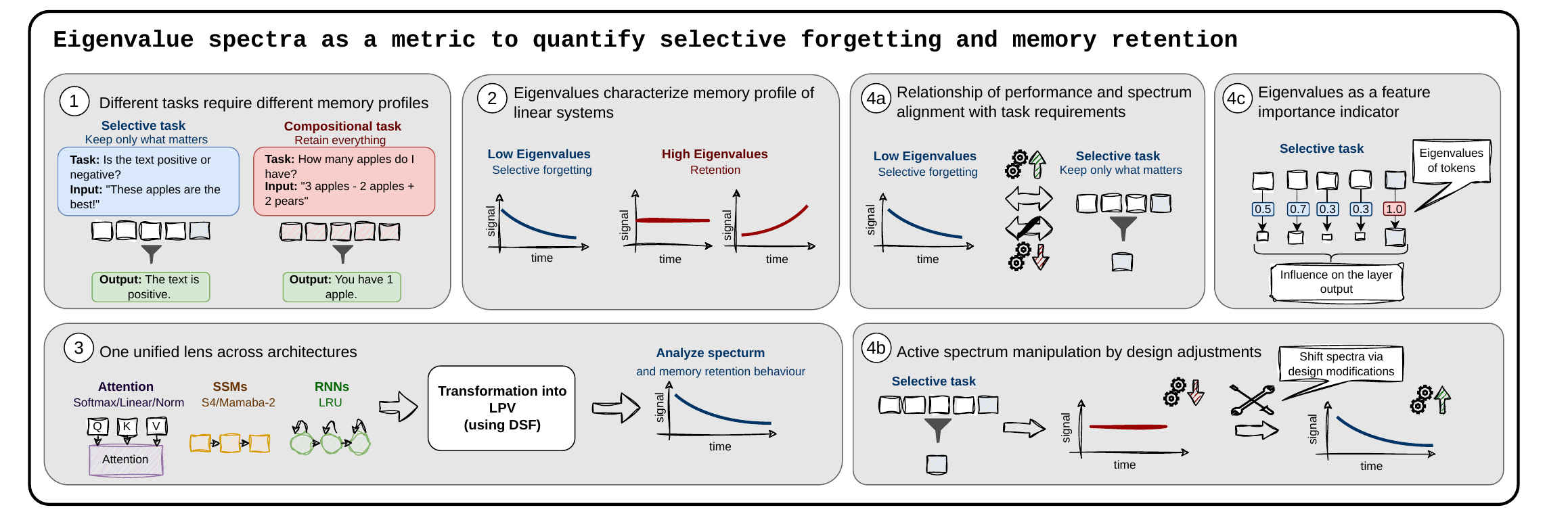}
    \caption{Tasks differ in the memory they demand \emph{(1)}, and in a
linear system this profile is set by the eigenvalue spectrum \emph{(2)}. Via the DSF, attention, SSMs, and RNNs are
all recast as LPV systems, so one spectral analysis applies
across architectures \emph{(3)}. The spectrum then serves as an analysis tool
\emph{(4a)}, a design target \emph{(4b)}, and a token-level feature-importance
measure \emph{(4c)}.}
    \label{fig:overview}
\end{figure*}

To close this gap, we leverage the dynamical systems framework
(DSF)~\cite{sieber2024understanding}, which provides a unified perspective for
analyzing masked attention and its linear alternatives as dynamical systems.
Understanding whether characteristic eigenvalue patterns still emerge in
attention-based models, despite eigenvalues not being explicitly placed or
constrained, can reveal fundamental insights into how different sequence
models perform long-range dependency modeling. To achieve this, we first formulate the connection between eigenvalues and memory in Proposition~\ref{prop:gains} and then
conduct a comprehensive empirical analysis of eigenvalue spectra across
different attention mechanisms, SSMs, and RNNs over a variety of tasks and model
sizes, empirically confirming a link between the eigenvalue distribution and good
performance on tasks with specific memory requirements. Specifically, we
observe a high concentration of eigenvalues close to one when long memory is
important, and similar peaks close to zero when memory selectiveness is
required by the task. Motivated by our findings, we then investigate the use
of eigenvalues as a metric that guides architectural changes, reveals
favorable eigenvalue placements during training and initialization, and
provides information about feature importance. Figure~\ref{fig:overview} illustrates our perspective: tasks differ in the
memory they demand, and a model's eigenvalue spectrum captures how well it meets
that demand. Through the DSF, this spectral view applies uniformly across
attention, SSMs, and RNNs, serving simultaneously as an analysis tool, a design target,
and a feature-importance measure.

\paragraph{Notation:} Hidden state and model size are denoted with $N$ and $d$. Subscripts, e.g., $\cdot_i$, denote the time index (or input dependency). Specifically, $v_i$ is the value of vector $v$ at time $i$, $\abs{v_i}_{\infty}$ its infinity norm, and $\operatorname{diag}(v_i)$ defines the diagonal matrix whose diagonal is the vector $v_i$. The identity matrix is denoted as $\mathbb{I}$ and $\bm{\uplambda}(A)$ indicates the eigenvalues of $A$.

\section{Sequence Models as LPV Systems}
\label{sec:lpv}
In this section, we review how the DSF~\cite{sieber2024understanding}
equivalently reformulates the input-output behavior of causal (i.e., masked)
attention models, SSMs and RNNs as linear parameter-varying (LPV) dynamical systems
of the form
\begin{equation}\label{eqn:LTV}
    h_{i} = \Lambda_{i}h_{i-1} + B_{i}u_{i}, \quad
    y_i = C_i h_i + D_iu_i,
\end{equation}
where $u_i\in\mathbb{R}^d$ is the $i$-th input token, $y_i\in\mathbb{R}^m$
the output, and $h_i \in \mathbb{R}^N$ an internal hidden state initialized
as $h_{-1}=0$. The parameter variation is governed by the input: the state transition matrix
$\Lambda_i \in \mathbb{R}^{N \times N}$, the input matrix
$B_i \in \mathbb{R}^{N \times d}$, the output matrix
$C_i \in \mathbb{R}^{m \times N}$, and $D_i \in \mathbb{R}^{m \times d}$, which acts as a skip connection, may each depend on the input, e.g., $\Lambda_i = f(u_i)$.
All model classes studied in this paper admit a representation of the
form in~\eqref{eqn:LTV}, differing in their structure specifics:
linear time-invariant (LTI) SSMs fit it directly with
$\Lambda_i \equiv \Lambda$, i.e., a non-diagonal, constant state
matrix, while RNNs can be translated into an LTI representation via basic
algebraic manipulations. SSM LPV systems such as Mamba~\cite{mamba} have state matrices that are input-dependent, but can be chosen diagonal, i.e., $\Lambda_i= \operatorname{diag}(\bm{\lambda}_i)$ and $\bm{\lambda}_i \in \mathbb{R}^{N}$. Finally, causal attention admits an LPV representation with
$\Lambda_i= \lambda_i\mathbb{I}$ and $\lambda_i \in \mathbb{R}$, i.e., also input-dependent and diagonal. This shared structure is the foundation of our analysis: since $\Lambda_i$
is either time-invariant or diagonal for every model considered, the
eigenvalues $\bm{\uplambda}(\Lambda_i)$ directly
govern the propagation of past information through the recursion
in~\eqref{eqn:LTV}. They thus provide a single, architecture-agnostic metric
for memory dynamics, which we formalize in Proposition~\ref{prop:gains} and
analyze in our empirical study.

\section{Eigenvalue Products Are the Gains of the Input-Output Map}
\label{sec:kernel}
For LTI systems, the eigenvalues of $\Lambda$ are known descriptors of
system behavior~\cite{kailath1980linear}: eigenvalues near zero
induce rapid forgetting, while eigenvalues close to the unit
circle\footnote{A discrete-time LTI system is stable if its eigenvalues lie
within the unit circle, i.e.,
$\abs{\bm{\uplambda}(\Lambda)}_{\infty} \leq 1$.} retain signal strength.
For general LPV systems, no such interpretation holds and the pointwise
eigenvalues of a time-varying $\Lambda_i$ do not, in general, determine the
input-output behavior~\cite{rugh2000research}. However, the DSF
reformulations all fall into two special cases for
which the connection can be recovered: $\Lambda_i$ is either time-invariant
(LTI SSMs, RNNs) or time-varying but diagonal (attention, Mamba). In both cases, the state matrices share a common eigenbasis $V$ over all $i$ and can be rewritten as $\Lambda_i = V^{-1}\operatorname{diag}(\bm{\lambda_i}) V$, and $\bm{\lambda}_i = \bm{\uplambda}(\Lambda_i)$. This allows the recursion in~\eqref{eqn:LTV} to be unrolled, yielding the following elementary but, to our knowledge,
previously unexploited identity (proof in the Appendix):

\begin{proposition}[Eigenvalue products are the input--output gains]
\label{prop:gains}
Given any model studied in the DSF as the LPV system in~\eqref{eqn:LTV}, with either a diagonal or time-invariant transition matrix $\Lambda_i$, for every
$t$, the following holds: 
\[
  y_t = D_t u_t \;+\; \sum_{j=0}^{t} C_tV^{-1} 
        \Big(\underbrace{\textstyle\prod_{i=j+1}^{t}\operatorname{diag}(\bm{\lambda}_i)}_{\textstyle g_{t,j}}\Big)
        VB_j\, u_j ,
\]
where $V$ defines the time-invariant and orthonormal eigenvector basis of $\Lambda_i$, and $\bm{\lambda}_i$ denotes the corresponding vector of eigenvalues.
\end{proposition}

Proposition~\ref{prop:gains} shows that the influence of input $u_j$ on output $y_t$ is
gated by the eigenvalue product $g_{t,j}$ between the
two steps. It is therefore the product, not any single eigenvalue,
that sets the gain from input $j$ to output $t$. The identity holds along the
realized trajectory of trained sequence models, whose eigenvalues emerge from
learned weights, establishing their spectra as a quantitative handle on memory
dynamics. Two consequences follow, separating attention from recurrent models: \emph{(i) Attention can re-amplify inputs through a single scalar gate.}
In the DSF, attention's eigenvalue is a scalar rather than a vector,
$\Lambda_i=\lambda_i\mathbb{I}$, so a single gain discounts the entire hidden
state at once, in contrast to a diagonal SSM, which forgets each dimension
selectively. Here $\lambda_i>1$ is admissible and signals re-amplification
rather than instability, as it re-weights previously encountered information. Because the gain is the product
$g_{t,j}$ rather than any single factor, steps with $\lambda_i>1$ can be offset
by steps with $\lambda_i<1$ along the same trajectory. \emph{(ii) Recurrent models are constrained and cannot re-amplify.}
Since their eigenvalues never exceed one, re-amplification cannot occur. For
Mamba, $\bm{\lambda_i}\in[0,1]$ prevents re-amplification, resulting in monotonic, though not strict, forgetting; for LTI SSMs and RNNs, $\Lambda$ is constant with its eigenvalues typically
bounded by one, so $|g_{t,j}|$ stays constant or decreases exponentially.

\section{Experiment Design} 
\label{subsec:methodology}

\paragraph{Task selection:}
We select five tasks designed to probe distinct model capabilities. Specifically, we use a subset of the Long Range Arena (LRA) benchmarks \cite{Tay2021}, consisting of (i) \textit{Long ListOps}, which tests reasoning over deeply nested structures where every token in the input sequence is essential; (ii) \textit{Byte-level text classification} (IMDb), which evaluates compositional generalization and the ability to process long natural-language sequences with sparse informative signals; and (iii) \textit{Image classification from pixel sequences} (CIFAR-10), which emphasizes learning sparse local and global spatial relationships.
In addition, we include (iv) the \textit{MQAR} task \cite{zoology2023}, which requires associative recall, stressing a model’s ability to retrieve specific elements of the input sequence with high fidelity, as well as (v) the next token prediction task on \textit{WikiText-103} \cite{merity2016pointer}, which underscores the model's natural language generation capabilities (the {WikiText} visualizations have been deferred to the Appendix). Together, these tasks capture a spectrum of challenges: long-context reasoning, compositionality, spatial structure learning, and selective memory. 

\paragraph{Model selection:} We evaluate 6 representative architectures: two LTI models, an SSM (S4 \cite{S4}) and an RNN (LRU \cite{LRU}), an LPV SSM (Mamba-2 \cite{mamba2}), and three causal attention models (softmax~\cite{Transformer}, norm ~\cite{sieber2024understanding} and linear attention~\cite{Katharopoulos2020}). To ensure a fair comparison, for each task we fix the overall architecture and tune the remaining hyperparameters to achieve a trade-off between competitive performance and comparable model sizes. Complete training details are provided in the Appendix. To extend our investigation to a larger model (7+ billion parameters), we also analyze the pretrained softmax attention-based Olmo-3 model \cite{olmo2025olmo3} on WikiText.
\paragraph{Presentation of Results:} In the upcoming experiments, the dimension of the network input is consistently set to 1, creating a clear assignment of eigenvalues to input tokens. Note that, given preprocessing and modifications through previous layers, the model input dimensions $d$ are generally different from 1 and are provided in Table \ref{tab:num_params}. We visualize the eigenvalue spectra over a sequence containing multiple test data inputs, e.g., all pixels of an image in CIFAR-10. Since the eigenvalues of Mamba-2 and attention-based models depend on the input, i.e., they emerge from an LPV system, we compute them over a batch of these sequences and report the batch-averaged distributions (for task-specific batch sizes, see Appendix). For multi-headed models, we present results from a single head and provide remaining heads and additional random seeds, as well as results averaged over heads or seeds, in the Appendix, Figures~\ref{fig:avg_heads} to~\ref{fig:avg_seeds3}, supporting our findings. Achieved performances are indicated next to the model label.

\section{Spectral Patterns Across Models and~Tasks}
\label{sec:empirical_study}
We conduct an ablation study across multiple benchmarks, analyzing DSF eigenvalue
distributions over models, tasks, layers, and heads to test whether spectral
signatures align with the required memory and the performance
obtained.

\begin {figure*}
\centering
\begin{adjustbox}{max height=0.95\textheight, max width=0.96\textwidth}
\begin{tikzpicture}[scale=1.0]

     \node (pic0) at (-10.0, 25.5) {\includegraphics[width=23cm]
    {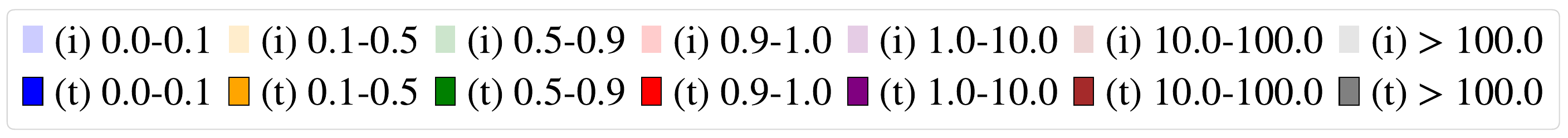}};
    
    \node (pic1) at (-10.0, 18.4) {\includegraphics[width=24.5cm]
    {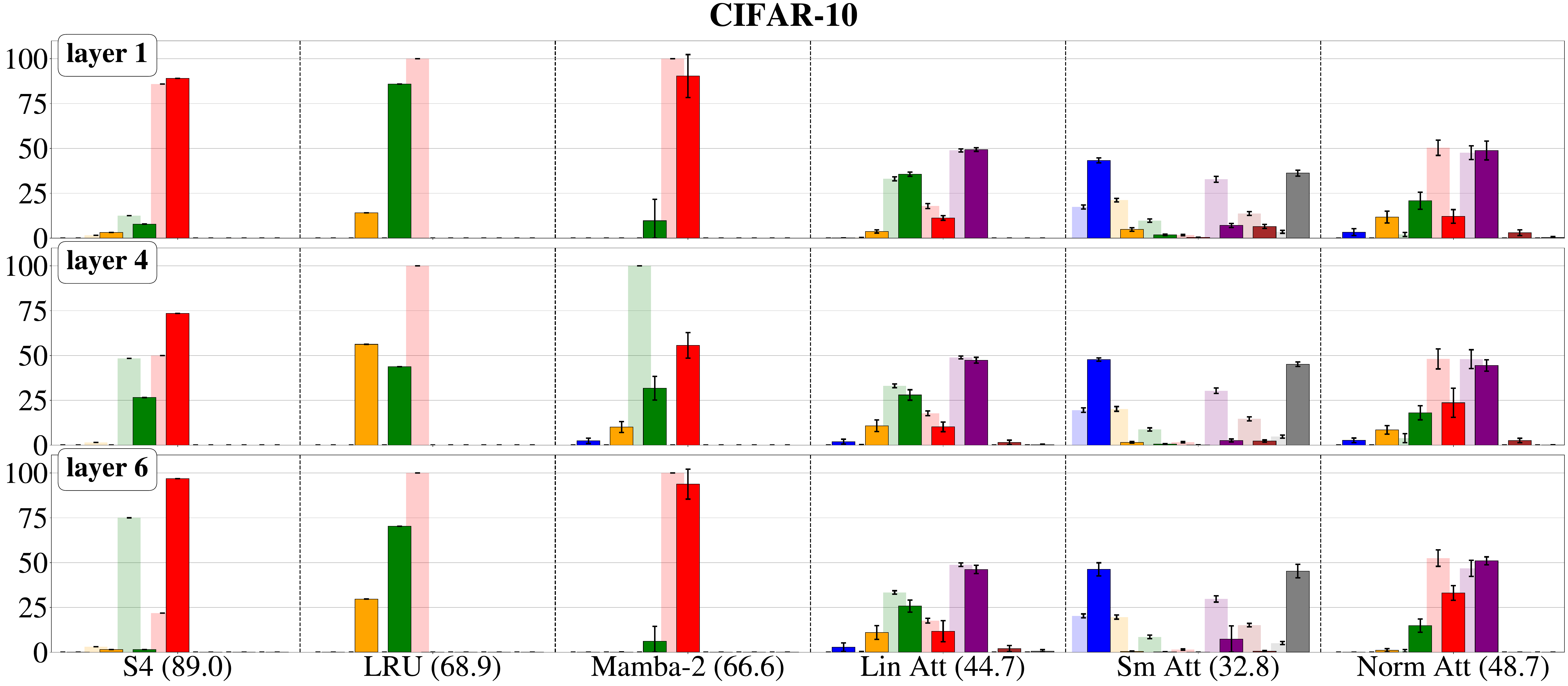}};
    
    \node (pic2) at (-10.0, 7.6) {\includegraphics[width=24.5cm]
    {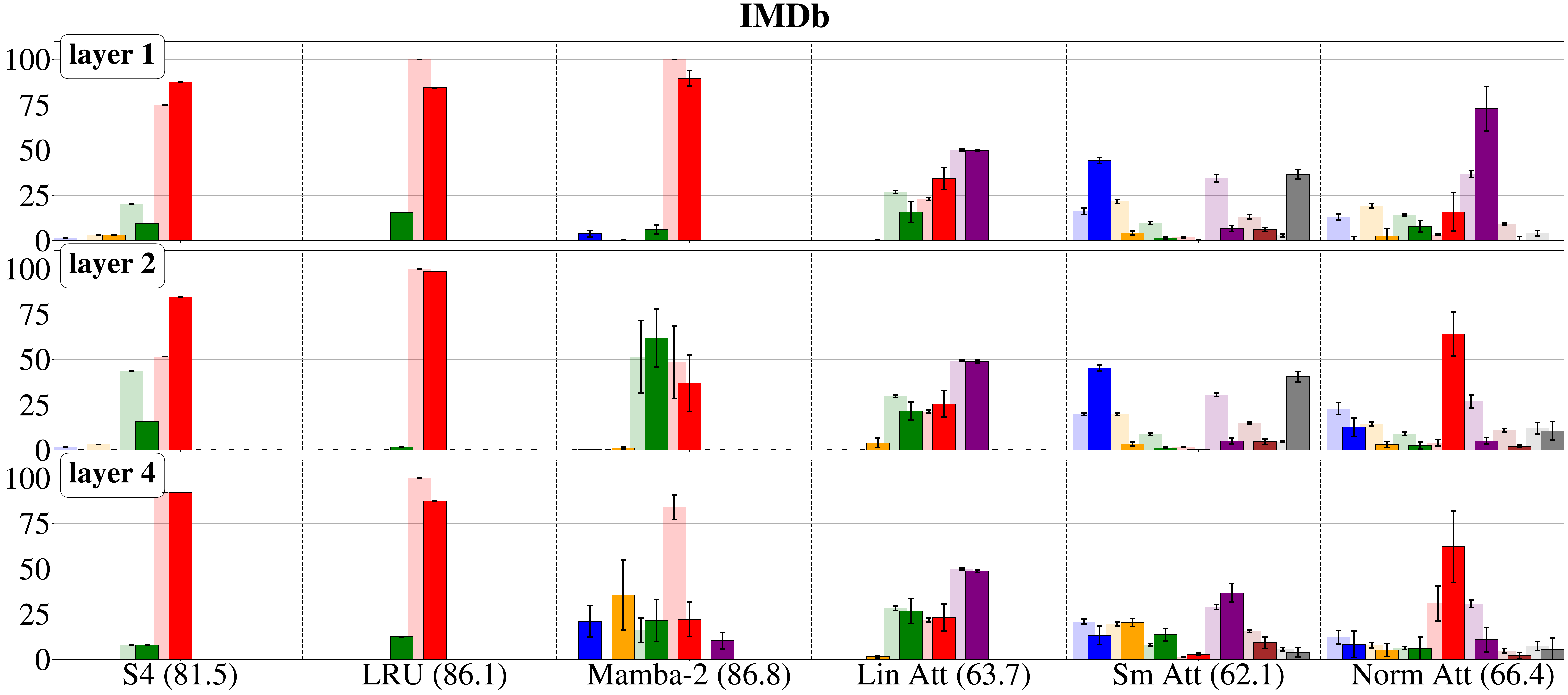}};
    
    \node (pic3) at (-10.0, -3.2) {\includegraphics[width=24.5cm]
    {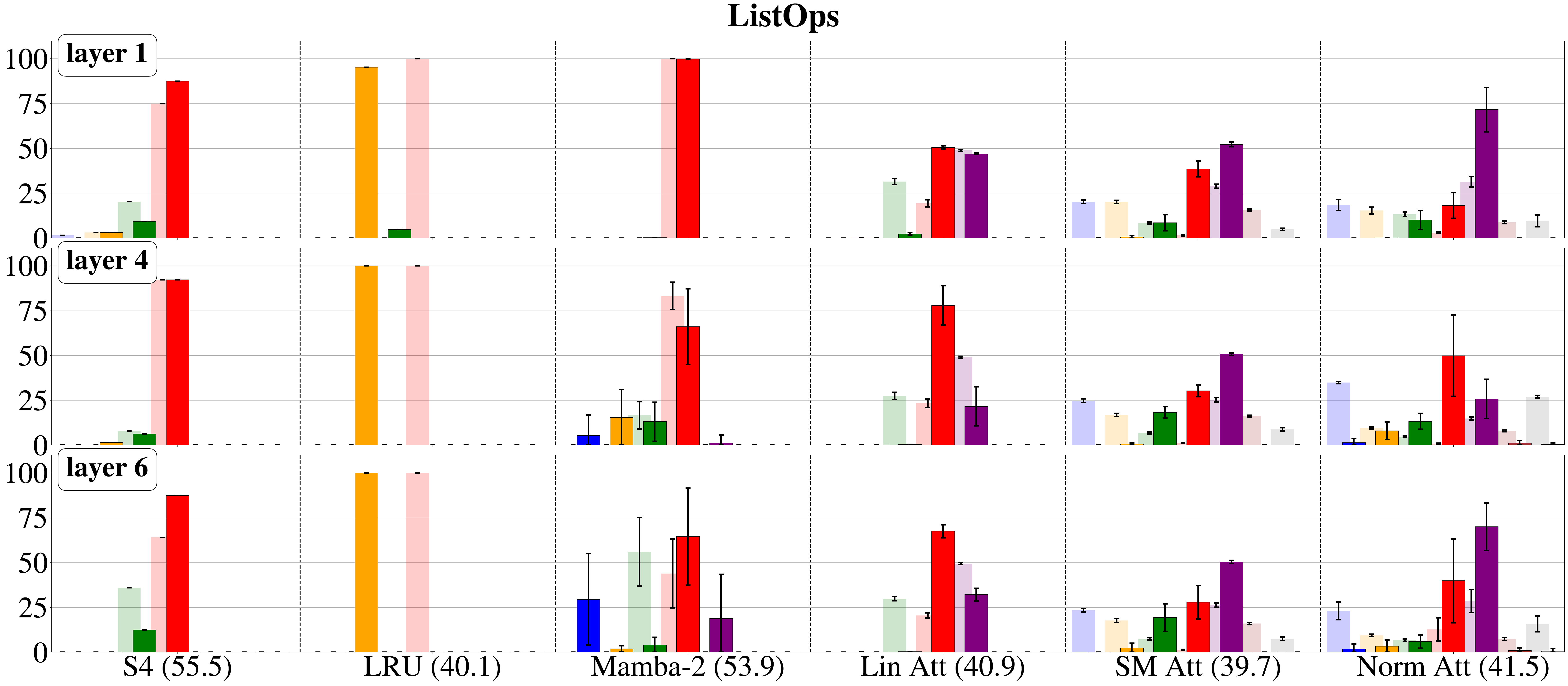}};

    \node (pic4) at (-10.0, -12.5) {\includegraphics[width=24.5cm]
    {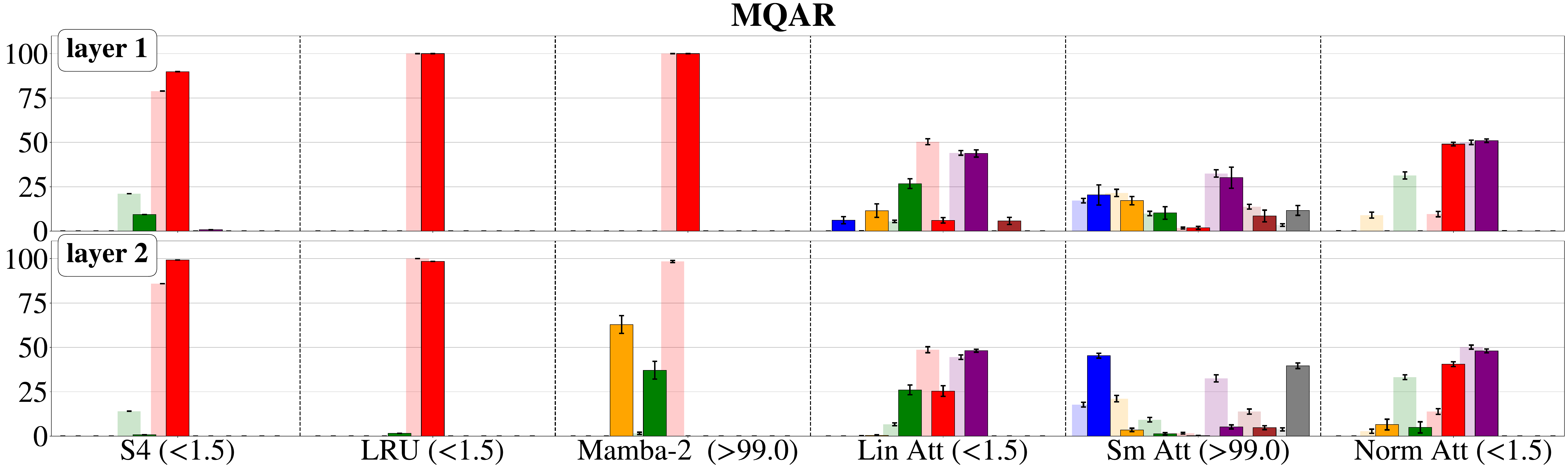}};

\end{tikzpicture}
\end{adjustbox}
\caption{Eigenvalue distributions for one head, across models, selected layers, and tasks. Bars show the percentage of eigenvalues within discretized ranges (chosen to emphasize eigenvalues near zero and near one). Light and dark bars indicate the distribution at initialization and after training, respectively. Error bars denote standard deviation across input sequences. Model performance, measured as percentage of correct output sequences for the other tasks (higher is better), is indicated in parentheses.}

 \label{fig:main_plot}
\end{figure*}

\subsection{Results and Discussion}

Task performance and the corresponding eigenvalue distributions for all models are shown in Figure~\ref{fig:main_plot} (for complex eigenvalues, magnitudes are reported, while their placement in the complex plane is provided in the Appendix, Figure~\ref{fig:complex_eigenvalues_appendix}). Based on our observations, we make the following statement: \textit{Eigenvalues capture essential aspects of memory and long-range dependency modeling, not only for LTI SSMs, but also LPV and attention models. Hence, downstream task requirements are reflected in spectral eigenvalue signatures of well-performing models.}

\paragraph{Patterns emerging across tasks:} The conducted study shows that task-dependent eigenvalue distributions reveal distinct memory retention and gating behaviors. When long-term memory is critical, we observe a strong concentration of eigenvalues near one across all investigated models, whereas tasks that require selective forgetting exhibit peaks near zero. This aligns with results derived for LTI SSMs using memory functions \cite{wang2023state} and our proposition rooted in linear system theory, that eigenvalues close to zero induce selective forgetting, and eigenvalues close to one are responsible for long-term memory. Specifically, on LRA tasks, which require long-term memory, all well-performing models not only avoid placing eigenvalues close to zero, but show prominent peaks around one, at least in the first layer and often across all layers. By contrast, attention models distribute eigenvalues both near zero and well above one, introducing a gating mechanism via a pattern of selective forgetting and re-amplification. Note that on ListOps, where every token is crucial, the performance of all models is relatively low. Hence, an ultimately favorable eigenvalue spectrum for solving this task may not yet have emerged during training, and spectral conclusions on this task have to be taken with a grain of salt. However, we still observe some trends. For example, we find that all models, including attention, avoid placing eigenvalues close to zero. It is interesting that this behavior extends to attention models, despite the fact that their eigenvalues are not explicitly designed but instead emerge from learned weight matrices and can hence take any value. The fact that attention still avoids near-zero eigenvalues in this setting reinforces our hypothesis linking small eigenvalues to selective forgetting: when forgetting is harmful, the model tries not to, albeit imperfectly. By contrast, on MQAR and WikiText tasks (see Appendix, Figure~\ref{fig:main_plot_appendix}), selective forgetting proves beneficial, since only highly specific information needs to be retained.

\paragraph{Correlation visualizations and CDF plots:} While the layer-wise spectrum visualizations provide detailed insights, they are less suitable for a quick assessment. For this purpose, Appendix Figure~\ref{fig:correlation_plot_appendix} presents correlation plots relating task performance with the average percentages of small and large eigenvalues across all heads and layers. Furthermore, to illustrate that the bin selection in our histograms is indeed representative and not selective in any way, in Figure~\ref{fig:cdf_appendix} in the Appendix, we also provide a cumulative density version of the results obtained in Figure~\ref{fig:main_plot}. 

\paragraph{Patterns emerging across models:} We observe that softmax attention is especially affected by gating: across all spectra, it shows a nearly balanced mix of very low and very high eigenvalues, which can be detrimental when not placed correctly, and important information gets lost. This is particularly visible on CIFAR-10, and to a lesser degree on IMDb, where the gating effect is less pronounced, and where due to the textual analysis nature of the IMDb task, not all tokens are crucial, implying selective forgetting can be favorable. In contrast, Mamba-2, as an LPV SSM, occupies a middle ground: its eigenvalue distribution avoids excessive gating, while still allowing selective placement near zero, enabling competitive performance not only on LRA benchmarks but also on MQAR and WikiText. LTI SSMs, by comparison, rarely place any eigenvalues close to zero, and consistently underperform attention on both of these tasks. This observation is further strengthened by the analysis of the investigated high-performing large-scale softmax attention model Olmo-3 on Wikitext (perplexity 13.41), where we observe that nearly a quarter of the heads in the initial layers perform the task of selective forgetting, a phenomenon that becomes more pronounced with increasing layer depth among all heads. The corresponding eigenvalue spectrum is provided in Figure~\ref{fig:olmo3}. 

\paragraph{Effects of initialization:} In LTI SSMs, eigenvalue placement is known to influence training stability \cite{LRU, Gu2020}, yet our analysis suggests that the initial eigenvalue configuration is not preserved during training and therefore cannot fully explain the final eigenvalue distributions. For instance, S4 is initialized with eigenvalues having the magnitude primarily in the range $(0.5,1)$, but this distribution consistently shifts towards a sharp concentration around~$1$ across tasks and layers. LRU, though also initialized close to~$1$, often drifts away from the initialization, most notably on ListOps and CIFAR-10. A similar departure from initialization is observed in both Mamba-2 and softmax attention, where the resulting eigenvalue spectra differ substantially from the starting distributions. In contrast, linear attention largely retains the overall shape of its initialization, raising the question of the potential importance of a task-dependent initialization. An illustration of the eigenvalue evolution during training is provided in Figures~\ref{fig:cifar_evolution} and \ref{fig:imdb_evolution} in the Appendix.

\section{Model Analysis and Design Guided by Eigenvalue Spectra}
\label{sec:model_analysis_and_design}

\begin {figure*} [!hbtp]
\centering
\begin{adjustbox}{max height=\textheight, max width=0.94\textwidth}
\begin{tikzpicture}[scale=1.0]

    \node (pic1) at (-13.0, 6.5) {\includegraphics[width=7.95cm]
    {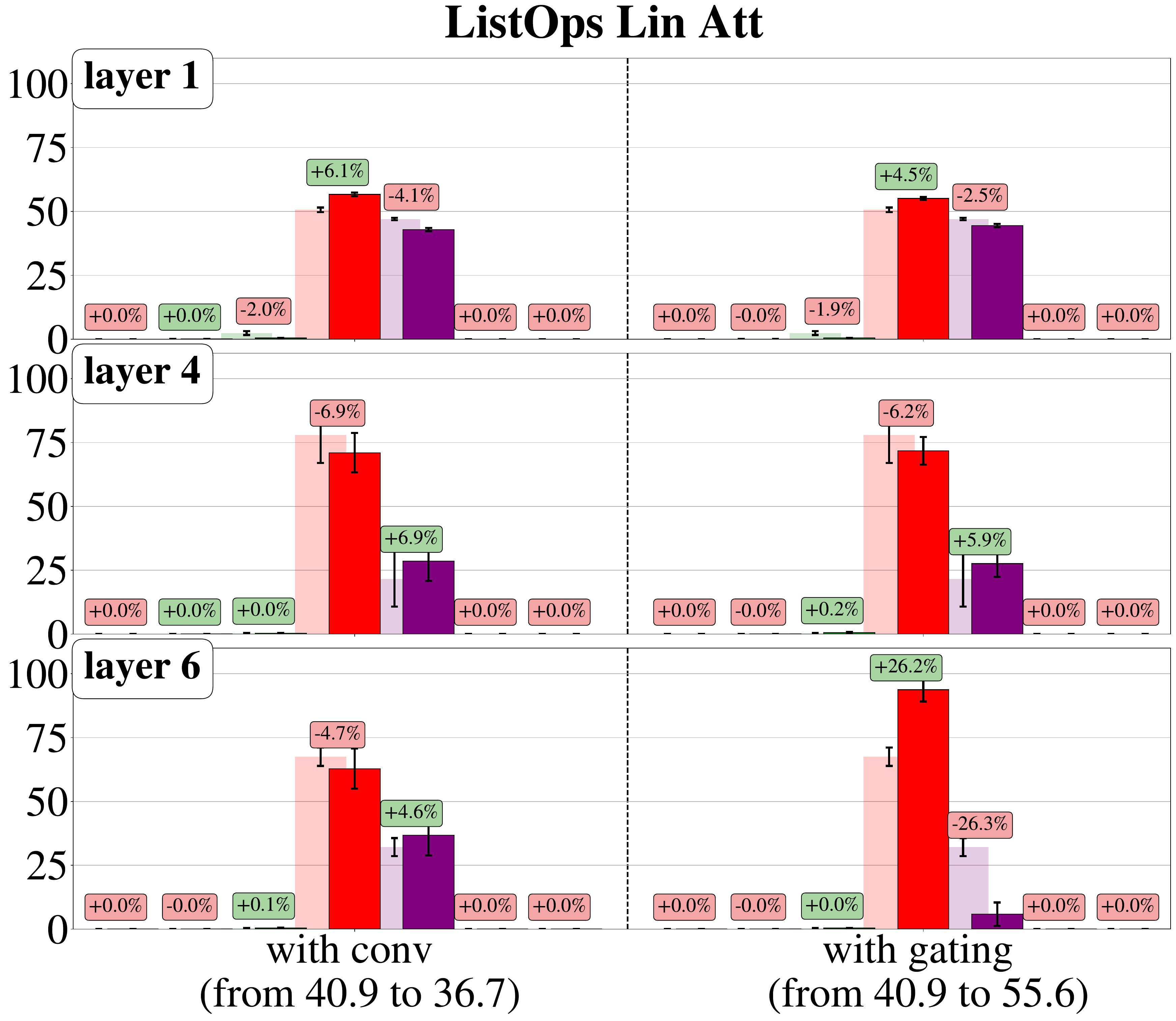}};

    \node (pic2) at (-4.9, 6.5) {\includegraphics[width=7.5cm]    {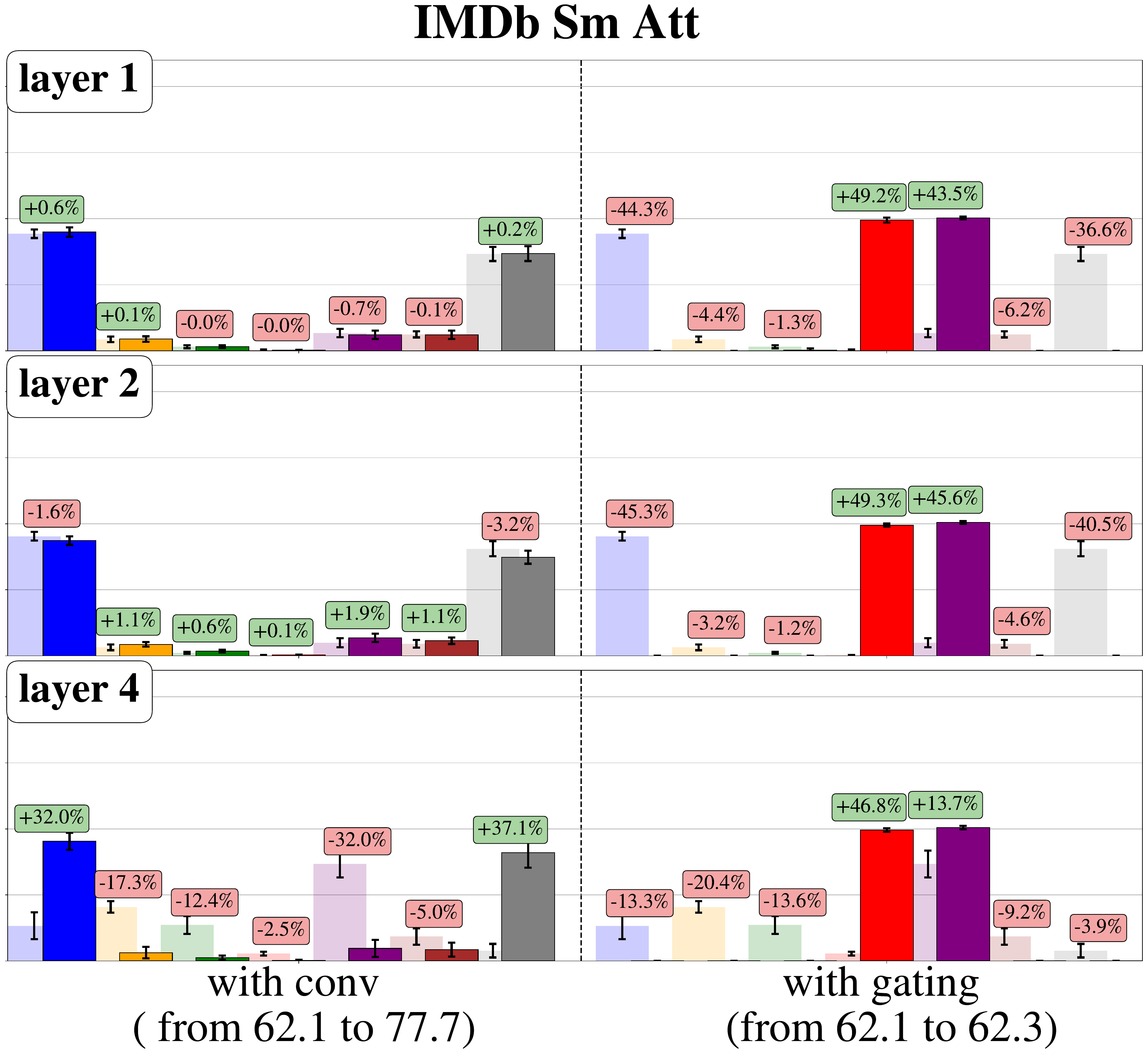}};

    \node (pic3) at (2.8, 6.5) {\includegraphics[width=7.5cm]{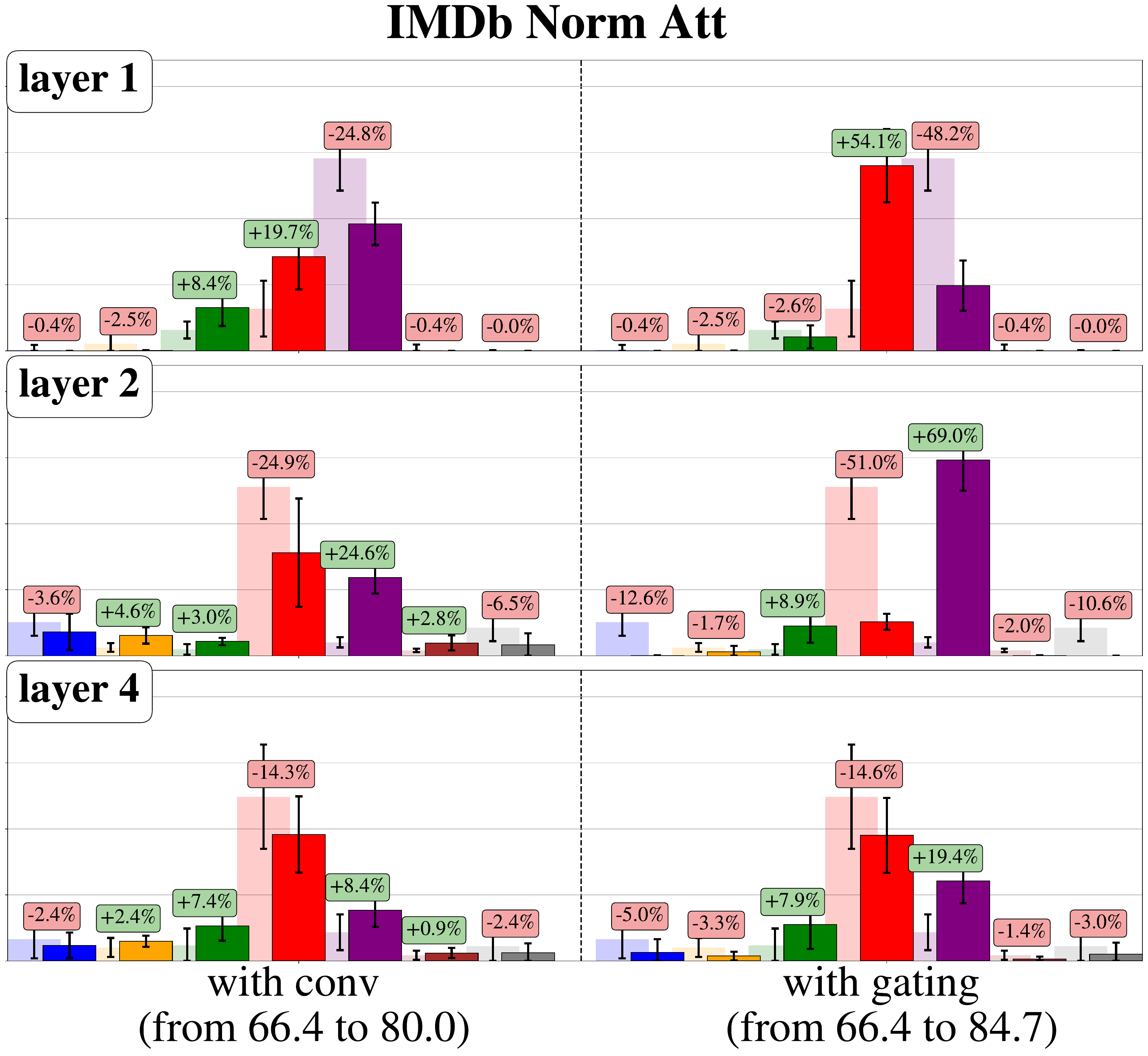}};
    
    \node (pic8) at (-13.0, 1.68) {\includegraphics[width=7.93cm]
    {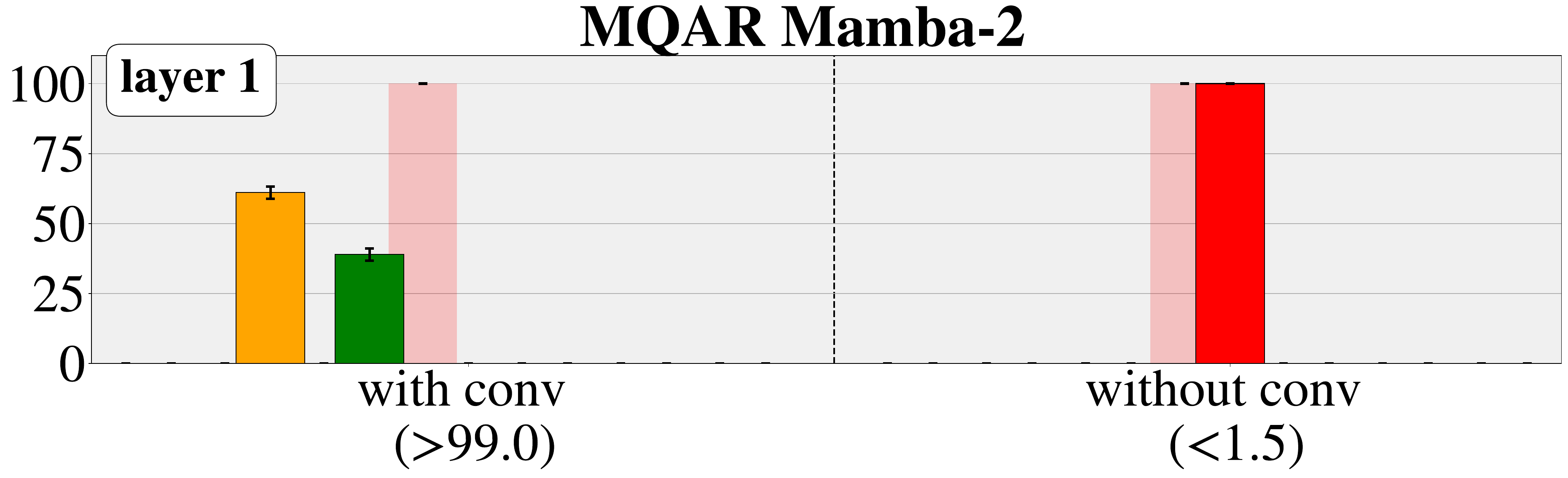}};
    \node (pic8) at (-13.0, -0.66) {\includegraphics[width=7.93cm]
    {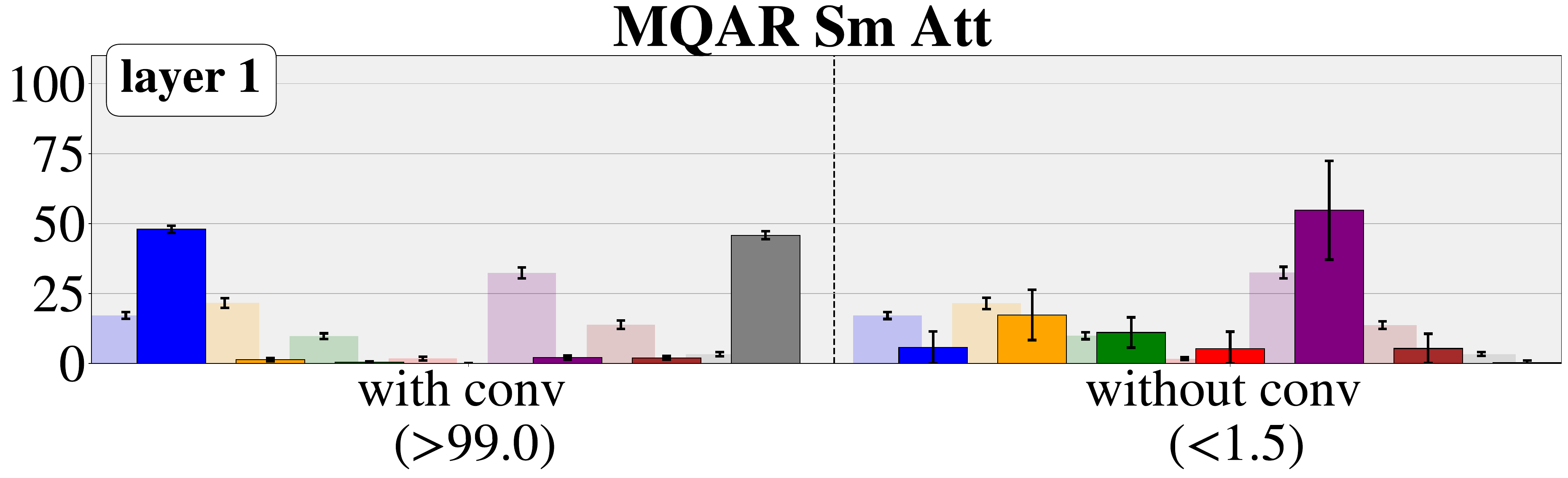}};

    \node (pic9) at (-4.9, 0.5) {\includegraphics[width=7.5cm]    
    {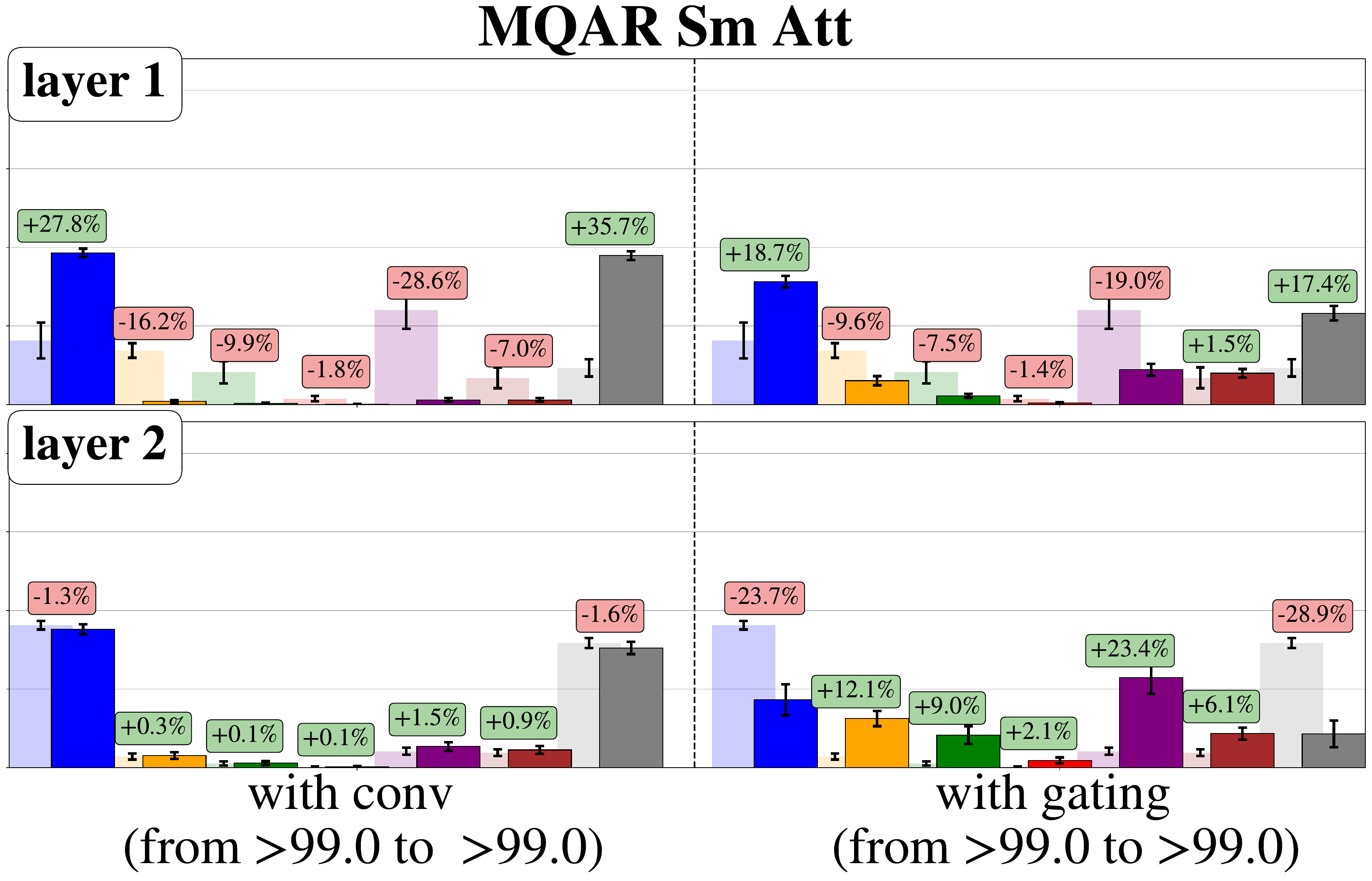}};

    \node (pic10) at (2.8, 0.5) {\includegraphics[width=7.5cm]
    {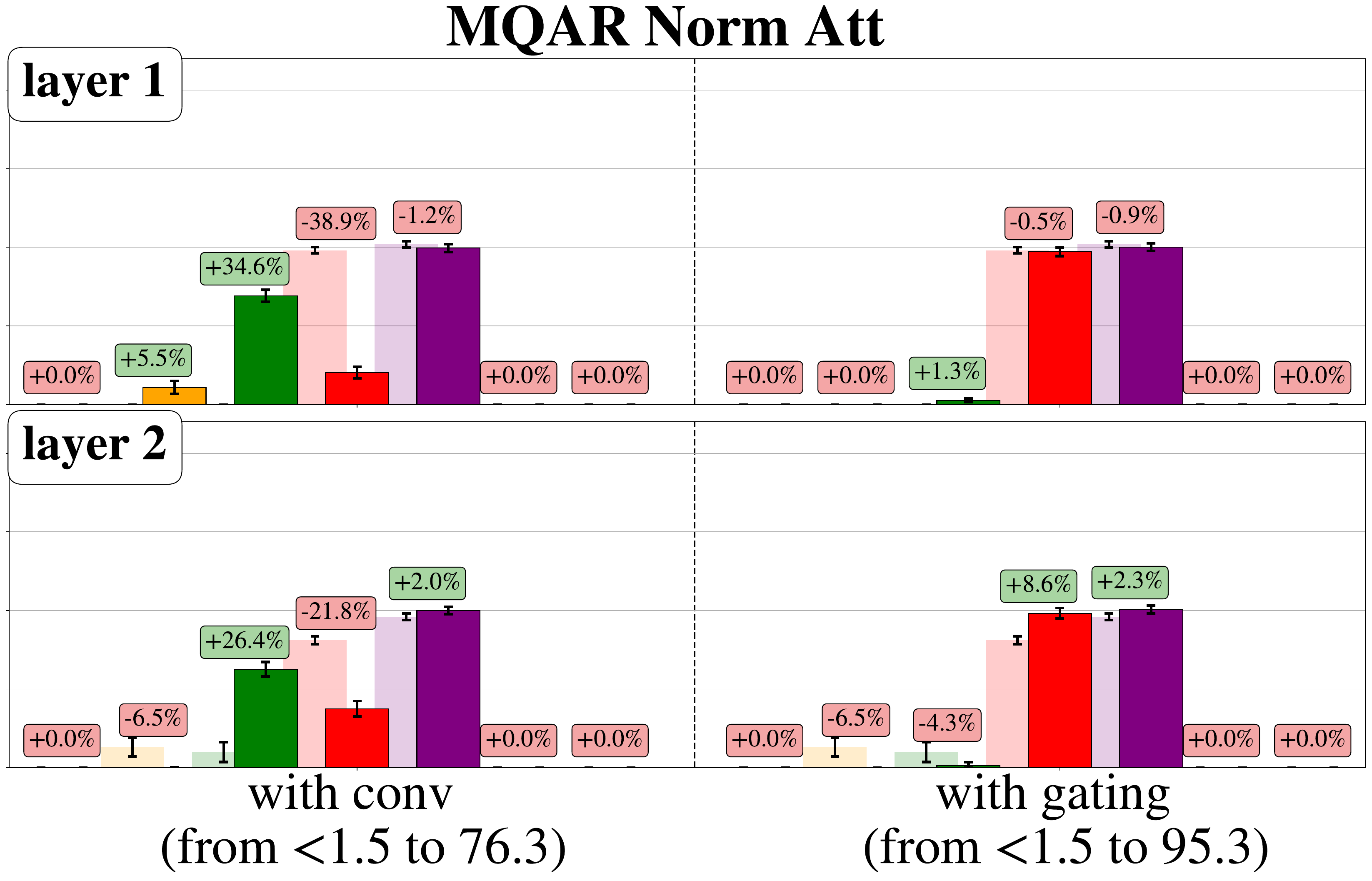}};
    
\end{tikzpicture}
\end{adjustbox}
\caption{Comparison of the effects of gating and convolution on the eigenvalue spectra for one head, across selected tasks, models, and layers. Complete plots for all layers and tasks are provided in the Appendix, Figures~\ref{fig:gating_conv_lra_appendix} and~\ref{fig:gating_conv_wikitext_appendix}.
}
  \label{fig:gating_conv}
\end{figure*}

In the following, we investigate how insights from the previous section can motivate architectural changes that relieve the model of a task-specific memory requirement, i.e., selective forgetting or memory retention. We also explore how these insights can shape the spectrum toward favorable distributions, and how eigenvalues relate to the importance of individual tokens, enabling importance-based feature reduction.

\subsection{Architectural Changes}
\label{subsec:architectural_changes}

In this subsection, we show how the eigenvalue distributions shift as a new component takes on its expected role, observing performance improvements across tasks (e.g., ListOps, IMDb) and models (e.g., norm attention), as shown in Figure \ref{fig:heatmaps} in the Appendix. 

\paragraph{Gating:} As discussed in the previous section, we observe a correlation between the gating behavior of a model and the selective-memory demands of the task. For example, on ListOps, where all input information is essential,  well-performing models tend not to perform gating, whereas on other tasks, they exhibit diverse gating strategies with varying success. Motivated by this, we investigate how attention models behave when equipped with an explicit gating mechanism. Following \citet{yang2024gated}, we augment each layer with a gating mechanism: the input to each layer is linearly projected, passed through a SiLU activation, and used to multiplicatively gate the layer’s output. This introduces a learnable, input-dependent modulation that can adaptively control the contribution of each layer to the forward pass. \\

Figure \ref{fig:gating_conv} illustrates that introducing an explicit gating mechanism alleviates the need for the dynamical system to implement gating implicitly, while still allowing it when beneficial. This effect is particularly pronounced on IMDb, but also visible on MQAR, where the eigenvalue distribution shifts away from zero, indicating that the dynamical system is primarily leveraged for memory preservation and state space expansion once gating is explicit. Using our novel metric, this shift reflects a redistribution of eigenvalues away from the selectivity regime (near zero) toward the memory regime (near one). An exception arises on the not well-performing combination of softmax attention and {ListOps}, as shown in the Appendix, Figure~\ref{fig:gating_conv_lra_appendix}, where the model does not exhibit low eigenvalues prior to gating but develops them once gating is introduced.

\paragraph{Convolution:} Motivated by the findings in \cite{okpekpe2025recalling} showing the connection of layer reduction and convolution introduction on MQAR, we investigate this further from the eigenvalue perspective. Specifically, we prepend each layer with a 1D convolution along the sequence dimension, applied to all hidden channels with kernel size $d_{\text{conv}}$ and appropriate padding to preserve length. This lightweight operation introduces local interactions among neighboring tokens prior to the main transformation, thereby enriching the input with short-range contextual information. From a spectral analysis point of view, we make the assumption that convolution alleviates the task of a  {memory preserving layer} in the recurrent dynamics by providing local context directly, which is supported by the following two observations from Figure \ref{fig:gating_conv}: (i) the introduction of convolution causes a shift in the eigenvalue spectrum, with eigenvalues occurring more frequently near zero and less frequently near one (see e.g., softmax attention on IMDb and MQAR in Figure \ref{fig:gating_conv}); (ii)  when using two layers for the MQAR task in Figure~\ref{fig:main_plot}, the second layer exhibits clear gating-like behavior, following a first layer characterized by stronger memory retention. With convolution offloading the memory-retaining layer, both softmax attention and Mamba-2 can be reduced to a single layer of selective forgetting and successfully learn the task, despite not being able to do so with one layer otherwise (as shown in the plots with darker background in Figure \ref{fig:gating_conv}). 

Both cases highlight how architectural modifications can reallocate functional responsibilities across layers and how this is reflected in changes to the eigenvalue distributions. These spectral shifts align with consistent performance gains on tasks requiring selective memory (as shown in Figure~\ref{fig:gating_conv} and Figure \ref{fig:heatmaps} in the Appendix), underscoring a link between introducing convolution, spectral structure, and task-level outcomes. Exceptions arise only in linear attention and {ListOps} task (provided in the Appendix), where memory preservation remains critical, and the benefits of selectivity diminish. 

\begin{figure}[t]
    \begin{adjustbox}{max width=\linewidth}
        \begin{tikzpicture}

            \node (pic1) at (-5.0, 10.5)
                {\includegraphics[width=7cm]
                {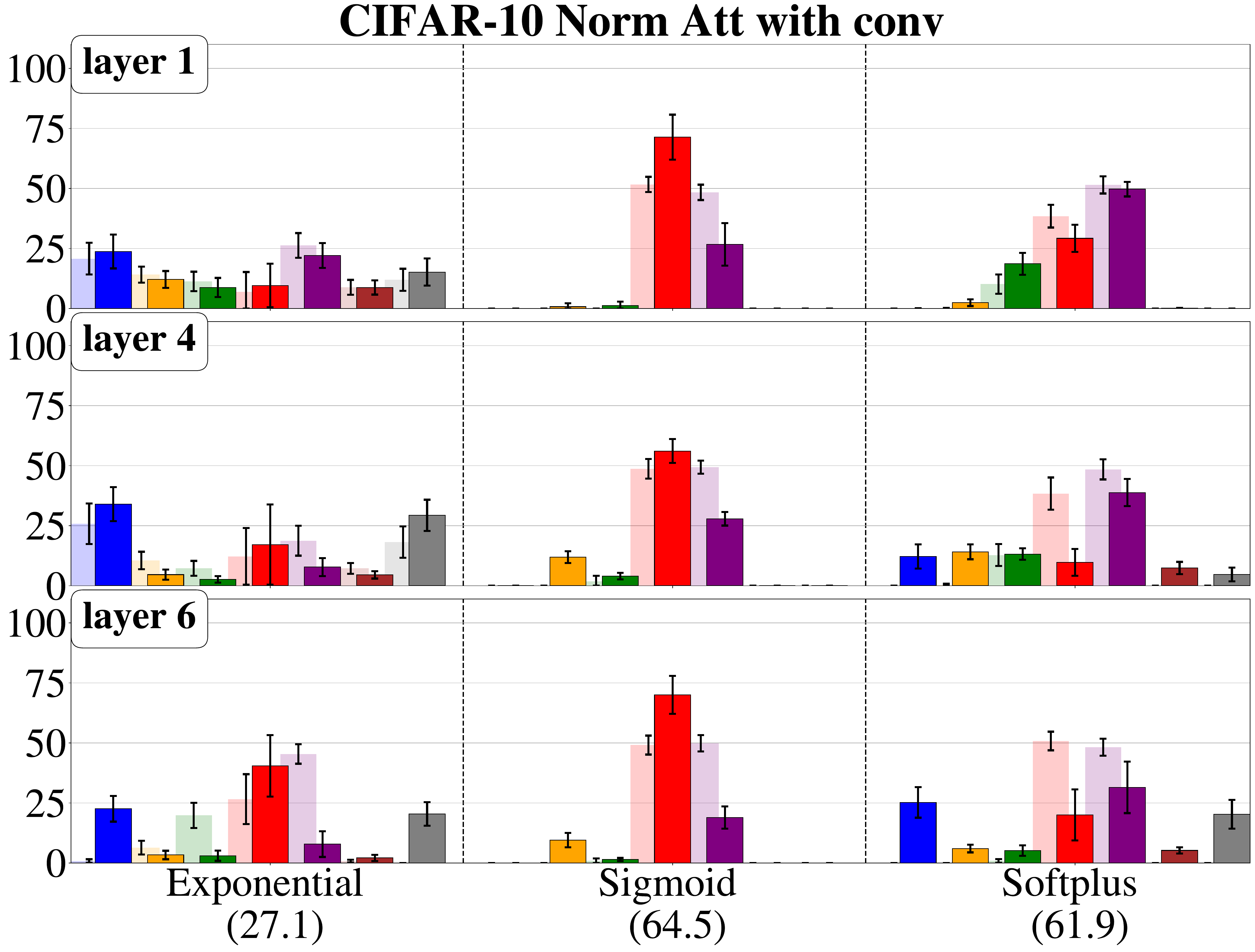}};

            \node (pic3) at (-0.3, 10.5)
                {\includegraphics[width=1.88cm]
                {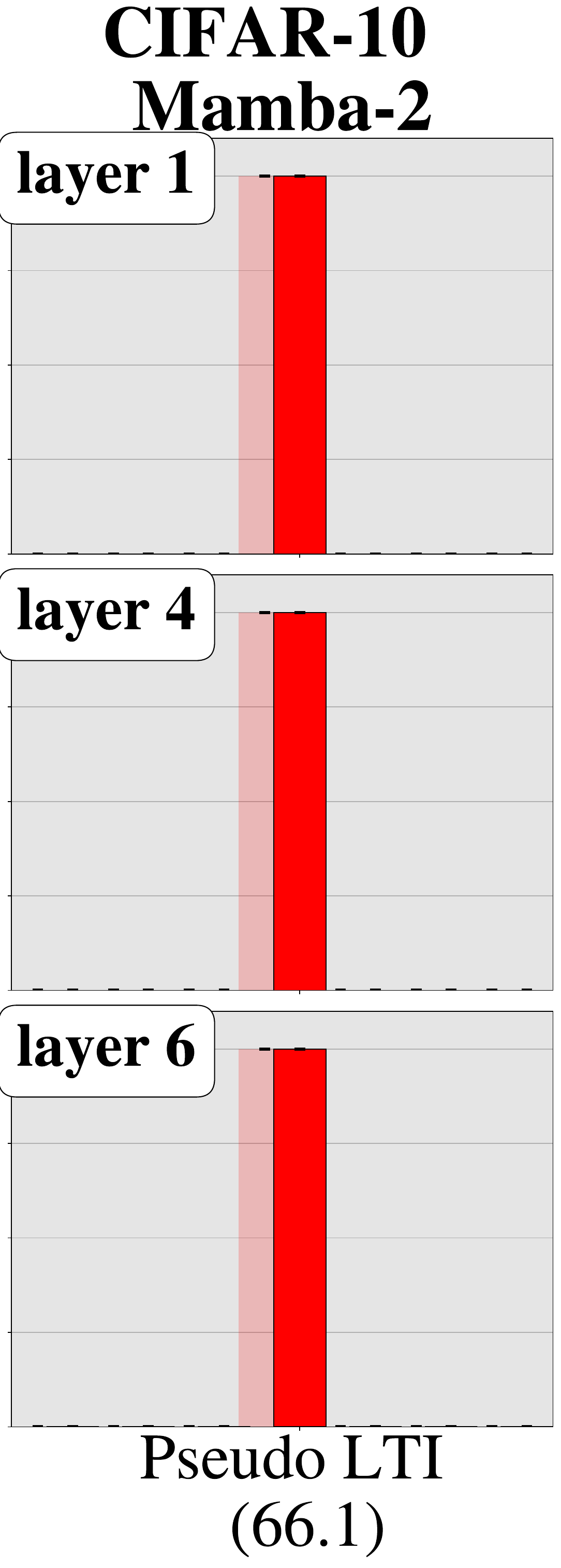}};

            \node (pic4) at (1.7, 10.5)
                {\includegraphics[width=1.88cm]
                {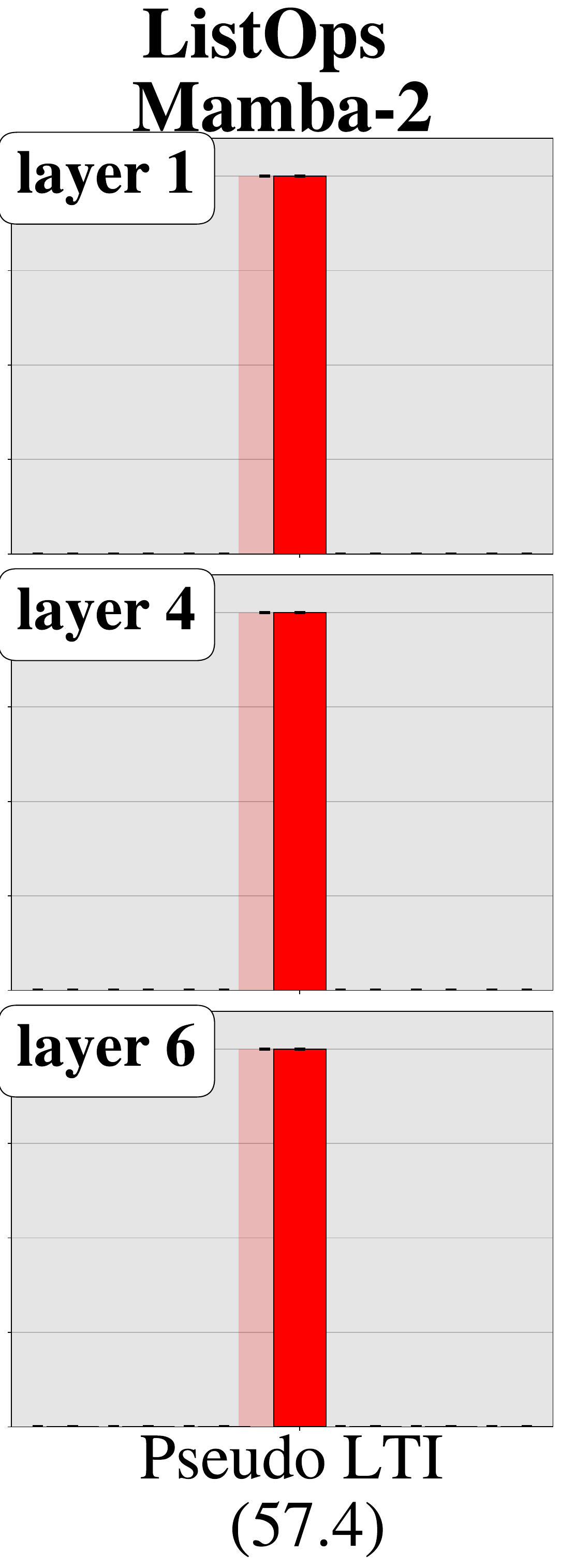}};

        \end{tikzpicture}
        \end{adjustbox}

        \captionof{figure}{(left) Eigenvalue distributions for one of the heads for  norm attention with  convolution and different normalization functions; (right) Mamba-2 as LTI. For complete plots of all layers and tasks, see Figures~\ref{fig:norm_init_appendix} and~\ref{fig:gating_conv_mamba2_appendix} in the Appendix.}
        \label{fig:normalization}
\end{figure}

\begin{figure}
    \centering
        \includegraphics[width=\linewidth]{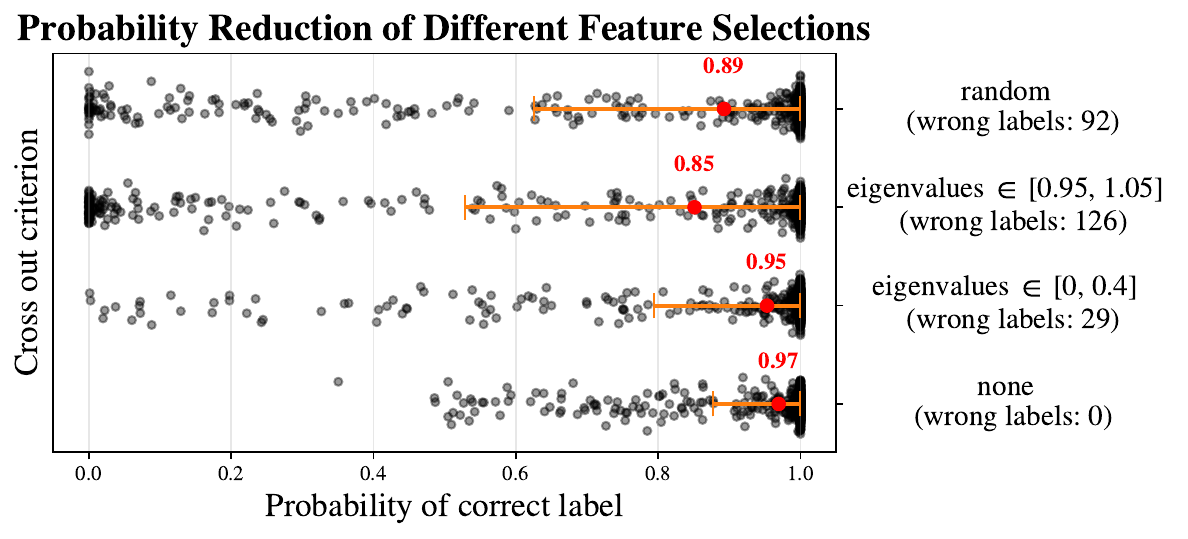}

        \captionof{figure}{Likelihood of correct classification after occluding an identical number of pixels for three pixel selection criteria based on their corresponding eigenvalues. Red numbers indicate mean values, red error bars one standard deviation and counts beneath each criterion the
resulting misclassifications.}
        \label{fig:feature_selection}
\end{figure}

\subsection{Guidance of Eigenvalue Spectra}
\label{subsec:guidance_of_spectra}

We next show that eigenvalue spectra can be steered through simple architectural choices, and doing so predictably affects memory dynamics and task performance.

{\color{black} \paragraph{Normalization of Norm Attention:} As shown in \cite{sieber2024understanding} and summarized in the Appendix, the transition dynamics of attention-based models
are governed entirely by the normalization term 
$\eta_i$ and the state transition matrix of the LPV
representation is $ \Lambda_{i} = \frac{\eta_{i-1}}{\eta_i}\, \mathbb{I}_{N} = \lambda_i\mathbb{I}_N$, so the eigenvalues are ratios of successive normalizer values. 
While for softmax
attention, $\eta$ is determined implicitly by the exponential sum over query and key
scores, for norm attention, $\eta$ is specified directly as a
nonlinear function of the projected input, for which our formulation
nominally adopts \texttt{softplus}. This explicit input dependence makes
norm attention particularly amenable to spectral intervention: replacing
the normalization function directly reshapes the attainable range of
$\lambda_i$, and thus the eigenvalue placement both
at initialization and during training. We therefore substitute the nominal
\texttt{softplus} in norm attention with explicit convolution by two
alternatives, \texttt{exponential} and \texttt{sigmoid}.} As illustrated by Figure \ref{fig:normalization}, different normalization functions lead to different eigenvalue distributions, reflecting trade-offs between memory and selectivity. Exponential normalization places eigenvalues that promote selectivity, sigmoid keeps them close to one, emphasizing memory retention, and softplus exhibits an intermediate pattern combining both regimes. In summary, the choice of normalization function matters and, when chosen in line with the task requirements, it leads to increased performance (e.g., IMDb and CIFAR-10 in combination with sigmoid or softplus). \textcolor{black}{Another way to start training from a more favourable spectrum configuration is pretraining, which is investigated in the Appendix.}

\paragraph{Mamba-2 as a Pseudo-LTI:} Motivated by the strong performance of SSMs with LTI structure on LRA tasks, and by the high variance observed in input-dependent eigenvalues (particularly pronounced in Mamba-2 on ListOps), we construct a pseudo-LTI variant of Mamba-2. Concretely, we modify the discretization step so that the state matrix $\Lambda$ is computed with respect to a fixed sampling interval, rather than an input-dependent one. This ensures that $\Lambda$ remains constant across inputs, while allowing input-dependencies to be absorbed into the input matrix $B$. According to Figure~\ref{fig:normalization}, this modification of Mamba-2 yields a behavior analogous to S4, both in performance on ListOps and eigenvalue distribution, with all eigenvalues remaining close to one, while the time-varying dynamics are captured through the input matrix $B$. More fine-grained plots are provided in Figure~\ref{fig:gating_conv_mamba2_appendix} of the Appendix, revealing that eigenvalues still shift from their initial distribution, though only within a narrow range.

\subsection{Feature Reduction}
\label{subsec:feature_reduction}

While analyzing eigenvalue spectra across layers and heads provides a general view of how information is processed for a given task, we observe that examining  eigenvalues corresponding to each token offers insight into input token importance. Since the influence of input tokens with low eigenvalues rapidly diminishes within the corresponding dynamical system representation of each layer, we expect that the removal of tokens with consistently low eigenvalues across all heads of the initial layer would have a less pronounced effect on model performance than the removal of tokens with high eigenvalues in the initial layer, motivating a principled criterion for feature reduction. To investigate our hypothesis, we consider a total of 890 images within the CIFAR-10 test set, correctly classified by a norm attention model. For each image, we record the likelihood change of the correct class label, when changing the same number of pixel values to an uninformative gray color (inspired by the occlusion sensitivity analysis proposed in \cite{zeiler2014visualizing}). The experiment is performed for pixels masked based on different criteria: a low assigned eigenvalue, an eigenvalue close to one (both accross all heads in the first layer), and random choice. The results are reported in Figure \ref{fig:feature_selection}, supporting our hypothesis that masking pixels with low eigenvalues does not significantly affect the prediction, while masking those with eigenvalues close to one leads to approximately 5 times more misclassifications. This experiment suggests that eigenvalues in attention models could be used as an indicator of the input token's importance. Note that the computation of this importance score does not require retraining to perform a sensitivity analysis, and is reproducible for different cross-out methods as shown in Figure \ref{fig:additional_features} in the Appendix. Both these aspects address some of the main challenges of explainability and interpretability metrics as proposed in \cite{zhang2025towards}, suggesting eigenvalues as a promising metric in this area. Exact experiment details are provided in the Appendix.

\section{Conclusion}

We leveraged a unified dynamical systems perspective on SSMs and attention to study eigenvalue analysis as a generalizable metric of memory dynamics for sequence models beyond SSMs. For this purpose, we conducted an extensive empirical analysis of eigenvalue distributions across various sequence models and tasks, revealing task-specific spectra and demonstrating the efficacy of eigenvalues in capturing essential aspects of selectivity and memory retention across all models, supporting our hypothesis. Building upon these findings, we then show how this novel metric can be employed to not only guide architectural choices, but also increase performance by steering the eigenvalue spectra favorably for the task at hand. On a token level, we finally demonstrate how eigenvalues can allow for feature importance selection and describe a potential feature selection criterion. However, this analysis is component-wise: it captures the
effects of several design decisions, but other components may also shape overall
performance and dynamics, and extending it to the full complexity of model
dynamics and further choices such as positional embeddings remains an important future
direction.

\bibliographystyle{plainnat}
\bibliography{bibliography}

\appendix
\onecolumn

\section{Additional Details on the DSF}
\label{app:dsf-details}
The LPV dynamics~\eqref{eqn:LTV} can also be written in a convolutional representation, i.e., $\mathbf{y} = \mathbf \Phi \mathbf{u}$, where the kernel $\mathbf \Phi$ is defined as
\begin{equation}\label{eq:conv-LTV}
    \mathbf \Phi = \begin{bmatrix}
        C_0B_0 + D_0 & & & \\ C_1\Lambda_1B_0 & C_1B_1 + D_1 &  &  \\ \vdots & \ddots & \ddots & \\ C_L\prod_{k=1}^{L} \Lambda_k B_0 & \dots & C_L\Lambda_{L}B_{L-1} & C_LB_L + D_L
    \end{bmatrix},
\end{equation}
and $\mathbf{u}$ and $\mathbf{y}$ stack the input and output tokens, respectively, over the full sequence length $i=0,\hdots, L$.
Note that the kernel $\mathbf \Phi$ has the same dimensions as the attention matrix, i.e., $\textrm{softmax}(\mathbf{q}\mathbf{k}^\top)$ for softmax attention, and that the two matrices are equivalent, up to a scaling factor $W_V$. Accordingly, using Lemma~1 in \citet{sieber2024understanding}, all considered masked attention can be written in the form given by~\eqref{eqn:LTV} as
\begin{subequations}\label{eqn:dsf_attention}
\begin{align}
    \Lambda_{i} &= \frac{ \eta(q_{i-1}, \mathbf{k}_{i-1})}{ \eta(q_i, \mathbf{k}_i)} \mathbb{I}_{N} \in \mathbb{R}^{N \times N}, \label{eqn:Lambda_app} \\
    B_i &= \left(\frac{1}{\eta(q_{i-1}, \mathbf{k}_{i-1})} \mathbb{I}_d \otimes \psi(k_i)\right) W_V \in \mathbb{R}^{N \times d}, \\
    C_i &= \mathbb{I}_d \otimes \phi(q_i)^\top \in \mathbb{R}^{d \times N},
\end{align}
\end{subequations}
retaining input-output equivalence. Thereby, $\otimes$ denotes the Kronecker product and the transition dynamics $\Lambda_i$ are solely governed by a scalar defined through normalization terms $\eta$ in \eqref{eqn:Lambda_app}. These normalization terms are given by $\eta(q_i, \mathbf{k}_i) = (\mathrm{elu}(q_i) + 1)^\top \sum_{j = 0}^{i}(\mathrm{elu}(k_j) + 1)$ for linear attention \cite{Katharopoulos2020} and as $\eta(q_i, \mathbf{k}_i) = \sum_{j=0}^i \exp(q_i^\top k_j)$ for softmax attention. In both cases, the normalization depends explicitly on the queries and keys. In contrast, norm attention resolves this dependence: its normalization can be defined directly as a general nonlinear function of $u$, for which we adopt the softplus function in our formulation. The detailed derivation of softmax attention as an LPV system is provided in Appendix~B of~\cite{sieber2024understanding}.

\section{Extended Related Work}
\label{app:related}

Modeling long-range dependencies is a central challenge across recurrent neural networks (RNNs), SSMs, and attention-based models. Due to their structural differences, separate methods have been developed for either class. 

\paragraph{RNNs and SSMs} 
\textit{Eigenvalue-based analyses:} Eigenvalue spectra provide a principled way to characterize stability and the relationship between remembering and forgetting, i.e., memory retention and information decay in dynamical systems. This perspective has a long history in RNNs, where eigenvalue normalization was introduced to mitigate vanishing gradients while preserving controlled memory decay~\cite{helfrich2019eigenvaluenormalizedrecurrentneural}. Later analyses investigated how different eigenvalue spectra encode solutions to temporal tasks, revealing that seemingly diverse eigenvalue distributions can correspond to functionally equivalent memory behaviors~\cite{Jarne_2022}. Additionally, \cite{naiman2023operator} shows that the eigenvalues of an approximate Koopman operator matrix of single-layer RNNs reflect task-specific structures. Early works on SSMs studied how careful initialization with eigenvalue placement can provide models with long-range dependency, capable of competing with RNNs and transformers on synthetic benchmarks~\cite{Gu2020,Fu2023}. Beyond initialization, subsequent works focused on structural re-parameterizations. In particular, the eigenvalues of the transition matrix have been shown to play a crucial role in governing the stability and memory length of SSMs~\cite{wang2024stablessmalleviatingcursememory,grazzi2025unlockingstatetrackinglinearrnns}. 

\textit{Memory function analyses:} A complementary line of work analyzes long-term dependency through the lens of memory functions~\cite{oppenheim1997signals}, which have a strong relation to the eigenvalues of the system. These functions quantify how much influence past inputs have on current outputs, typically formalized via norms of system response functions or spectral measures. For SSMs, memory functions provide a principled way to characterize expressivity and effective memory horizons~\cite{wang2023state}. For RNNs, memory functions are used to prevent rapid decay of state memory~\cite{Su_2019}.

\paragraph{Attention-based Models}
In attention-based architectures, analyses have largely focused on mechanisms for extending memory. Early work demonstrated that even simple feed-forward architectures equipped with attention can solve long-memory tasks, outperforming classical RNNs~\cite{raffel2016feedforwardnetworksattentionsolve}. Later, augmenting self-attention with persistent memory slots was proposed to extend the memory context~\cite{sukhbaatar2019augmentingselfattentionpersistentmemory}. Other works analyze the statistical properties of the attention score matrices themselves~\cite{bao2024self,bhojanapalli2021eigen}, revealing certain regularities in how attention distributes over inputs. \\

However, obtaining a unified metric remains challenging. While insightful, existing approaches for attention-based models remain tied to the score-matrix formulation and do not naturally translate into a framework that allows systematic comparison with SSMs. Furthermore, memory functions require architecture-specific design choices and do not generalize easily to linear parameter-varying (LPV) systems, such as attention-based models and Mamba~\cite{mamba2}. Using the duality between SSMs and attention-based models, established by the recently proposed DSF, eigenvalue spectra analysis presents a potential metric for systematic cross-architectural comparisons of memory behavior. We leverage this to provide the first comprehensive eigenvalue-based analysis of attention mechanisms, enabling a direct comparison with SSMs. 

\section{Proof of the Proposition 1}
\label{app:proof}

In the following, we provide proof of the Proposition~\ref{prop:gains}.

\begin{proof}
Throughout the proof, we write $\operatorname{diag}(\boldsymbol{\lambda}_i)$ for the diagonal matrix
whose diagonal is the eigenvalue vector $\boldsymbol{\lambda}_i$. By hypothesis, every representation
considered admits a single, time-invariant orthonormal eigenvector basis $V$ in which the
transition matrix is diagonal,
\[
  \Lambda_i \;=\; V^{-1}\operatorname{diag}(\boldsymbol{\lambda}_i)\,V,
  \qquad i = 0,1,\dots, \qquad V^{-1}=V^{\top}.
\]
In the LTI case, $\Lambda_i\equiv\Lambda$ and $\boldsymbol{\lambda}_i\equiv\boldsymbol{\lambda}$
are constant; in the diagonal LPV case $V=\mathbb{I}$ and
$\Lambda_i=\operatorname{diag}(\boldsymbol{\lambda}_i)$. In either case $V$ is independent of
$i$, which is the only structural fact used below.

Recall the state recurrence of~\eqref{eqn:LTV},
\[
  h_t = \Lambda_t h_{t-1} + B_t u_t,
  \qquad
  y_t = C_t h_t + D_t u_t,
  \qquad h_0 = B_0 u_0 .
\]
For $0\le j\le t$, we define the eigenvalue product
\[
  g_{t,j} \;:=\; \prod_{i=j+1}^{t}\operatorname{diag}(\boldsymbol{\lambda}_i)
           \;=\; \operatorname{diag}\!\Big(\textstyle\prod_{i=j+1}^{t}\boldsymbol{\lambda}_i\Big),
\]
with the empty-product convention $g_{t,t}=\mathbb{I}$; the second equality holds because
diagonal matrices commute (the product of the eigenvalue vectors is taken entry-wise).

\medskip
\noindent For every $t\ge 0$, we will show using mathematical induction, that
\begin{equation}
  h_t \;=\; \sum_{j=0}^{t} V^{-1} g_{t,j}\, V B_j u_j. \tag{$\star$}
\end{equation}

\noindent\emph{Base case ($t=0$):} The sum has the single term $j=0$ and $g_{0,0}=\mathbb{I}$,
so the right-hand side equals $V^{-1}\mathbb{I}\,V B_0 u_0 = B_0 u_0 = h_0$, hence the equality is satisfied.

\smallskip
\noindent\emph{Induction step.} Suppose $(\star)$ holds at $t-1$. Applying the recurrence and
$\Lambda_t = V^{-1}\operatorname{diag}(\boldsymbol{\lambda}_t)\,V$,
\[
  h_t
  = \Lambda_t h_{t-1} + B_t u_t
  = V^{-1}\operatorname{diag}(\boldsymbol{\lambda}_t)\,V
      \sum_{j=0}^{t-1} V^{-1} g_{t-1,j}\, V B_j u_j \;+\; B_t u_t .
\]
Cancelling $V V^{-1}=\mathbb{I}$ inside the sum and using
$\operatorname{diag}(\boldsymbol{\lambda}_t)\,g_{t-1,j}
   = \prod_{i=j+1}^{t}\operatorname{diag}(\boldsymbol{\lambda}_i) = g_{t,j}$, we have that 
\[
  h_t
  = \sum_{j=0}^{t-1} V^{-1} g_{t,j}\, V B_j u_j \;+\; B_t u_t .
\]
Since $g_{t,t}=\mathbb{I}$ gives $B_t u_t = V^{-1} g_{t,t}\, V B_t u_t$, the last term is exactly
the $j=t$ summand, so $h_t = \sum_{j=0}^{t} V^{-1} g_{t,j}\, V B_j u_j$, which is $(\star)$ at $t$. Therefore, $(\star)$ holds for every $t \geq 0$.

\medskip
Substituting $(\star)$ into $y_t = C_t h_t + D_t u_t$ yields
\[
  y_t = D_t u_t + \sum_{j=0}^{t} C_t V^{-1}
        \Big(\underbrace{\textstyle\prod_{i=j+1}^{t}\operatorname{diag}(\boldsymbol{\lambda}_i)}_{g_{t,j}}\Big)
        V B_j u_j ,
\]
which is the asserted identity. 
\end{proof}

\newpage
\section{Experimental Details}
\label{app:exp-details}

\textcolor{black}{This section contains details about experimental results provided in the main paper. The experiments were performed  on the long range arena~(LRA) benchmark~\cite{Tay2021}\footnote{\url{https://github.com/google-research/long-range-arena}}, the multi-query associative recall~(MQAR) benchmark~\cite{zoology2023}\footnote{\url{https://github.com/HazyResearch/zoology}} and the WikiText benchmark~\cite{merity2016pointer}. To obtain these results, we combined the  LRA and Zoology code bases. }

\paragraph{Model Details}
We evaluate the following three architecture classes on all tasks:
\begin{itemize}
    \item \textbf{Attention:} softmax attention~\cite{Transformer}, linear attention~\cite{Katharopoulos2020} and norm-attention~\cite{sieber2024understanding}. In all three settings, we employ a standard GPT-2–style multi-headed Transformer, where the attention block is replaced by the respective attention mechanism. Each attention block is followed by a task-dependent multi-layer perceptron. For fairness, all three variants share the same overall architecture and hyperparameters, differing only in the attention function, learning rate, dropout, and batch size. Furthermore, softmax and linear attention use learnable positional token embeddings, while norm attention does not.
    \item \textbf{Linear parameter-varying state space models:} Mamba-2~\cite{mamba2}. We use the same GPT-2 style multi-headed Transformer backbone with learnable token embeddings but no positional encoding, substituting the attention block with the SSD block, where we additionally apply a 1D short-convolution, with a filter of dimension 4 across sequence length to the input of each layer. The implementation follows the official codebase\footnote{\url{https://github.com/state-spaces/mamba}}.
    \item \textbf{Linear time-invariant state space and recurrent models:} S4~\cite{S4} and LRU~\cite{LRU}. We use one-hot encoding of the input, followed by a stacked encoder model architecture, which consists of a linear encoder followed by SSM blocks corresponding to the specific model. We adapt the S4\footnote{\url{https://github.com/state-spaces/s4}}
 and unofficial LRU\footnote{\url{https://github.com/NicolasZucchet/minimal-LRU}} code bases and integrate them into our framework. To ensure a fair comparison, we keep the architecture and all hyperparameters of both models constant, except for the learning rate, dropout and batch size.
\end{itemize}

All models are trained and evaluated over five different tasks (3 LRA tasks, MQAR and WikiText-103), while their architectures were fixed for each task (with the architecture details provided in Table~\ref{tab:architecture}), with the aim to keep the models comparable in size.  {The number of parameters per model is provided in Table \ref{tab:num_params}}. Furthermore, unless indicated otherwise, we fixed our seed to 1919 for all experiments. For the evaluation on additional seeds, we chose 1717, 2222, and 2929. We trained and evaluated all our models on a single NVIDIA A100 GPU (80GB). In the remainder of the section, we provide task-wise details of our experiments.

\subsection{LRA experiments}
\paragraph{Training details}

As outlined in the main paper, we restrict our experiments to a subset of LRA tasks: CIFAR-10, IMDb, and ListOps. For all LRA experiments, we follow the training protocol below:
\begin{itemize}
    \item \textbf{Optimizer and schedule:} Linear warmup with cosine annealing and AdamW optimizer~\cite{adamw}, trained for a constant number of epochs per task, as indicated in Table \ref{tab:architecture}.
    \item \textbf{Position information:} Positional embeddings~\cite{Brown2020} are used only for linear and softmax attention.
    \item \textbf{Data:} Each model is trained on the subset of standard datasets from the LRA benchmark, summarized below; full details can be found in~\cite{Tay2021}.
    \begin{itemize}
    \item Long List Operations (\texttt{ListOps}): This benchmark tests a model’s ability to handle hierarchical dependencies across long input sequences. The task involves predicting the outcome of mathematical expressions built from nested \emph{mean}, \emph{median}, \emph{max}, and \emph{min} operations.\footnote{For instance, $\texttt{input: max(4, 3, min(2, 3), 1, 0, median(1, 5, 8, 9, 2)},$ $\texttt{output:  5}.$} It is framed as a ten-class classification problem, with input lengths of up to 2k tokens. 
    \item Byte-Level Text Classification  (\texttt{IMDb}): This task measures a model’s ability to identify sentiment in long tokenized texts. The dataset is composed of IMDb movie reviews, each labeled as either positive or negative. It is formulated as a binary classification problem with input sequences of up to 4k tokens.
    \item Image Classification from Pixel Sequences (\texttt{CIFAR-10}): This benchmark assesses a model’s capacity to infer 2D spatial structure from linearized pixel sequences. The dataset contains vectorized images belonging to one of ten categories (e.g., horse, car). The task is framed as a ten-class classification problem with input sequences of up to 1k tokens.
\end{itemize}
\end{itemize}

\paragraph{Performed Experiments} We run all six models with task-wise parameters given in Table~\ref{tab:architecture} on the described subset of LRA tasks for four different random seeds specified above. For each of the runs, we  provide the eigenvalue distributions in the corresponding figures of the main text and the Appendix. Note that we do not optimize the hyperparameters for each model task combination, but as mentioned in the main text, we aim to obtain a middle ground between competitive performance relative to the best known
baselines and the comparable model sizes to make comparison sensible, and therefore the reported accuracies might be lower than in the original LRA paper~\cite{Tay2021}.

\subsection{MQAR experiments}
\paragraph{Training Details}
For all MQAR runs, we use the following training protocol:
\begin{itemize}
    \item \textbf{Optimizer and schedule:} Linear warm-up with duration of 10\% and cosine annealing, with AdamW optimizer~\cite{adamw}. For each run, we sweep the learning rates in $\texttt{logspace}(-4, -2, 4)$ and train for 40 000 steps with a global batch size of $64$. This is the same setup as in~\cite{zoology2023}.
    \item \textbf{Position information:} Positional embeddings~\cite{Brown2020} are used for linear and softmax attention models only, and not for the SSM architecture classes. This is the same setup as in~\cite{zoology2023}.
    \item \textbf{Data:} Each model is trained on an MQAR dataset with 100,000 datapoints and evaluated on 3,000 datapoints. This is the same setup as in~\cite{zoology2023}. The data and its order are constant for all runs, with the number of key-value pairs equal to 64 and an input sequence length of 512, corresponding to the most challenging task from~\cite{zoology2023}. 
\end{itemize}

\paragraph{Performed Experiments} We run all six models on the above-specified MQAR task, for 4 different seeds. For each model, we apply the architecture settings in Table~\ref{tab:architecture} and sweep over the learning rates. We only report the results of the best-performing learning rate in the corresponding figures of the main text and the Appendix.

\subsection{WikiText experiments}
\paragraph{Training Details}
For all WikiText experiments, we use the following training protocol:
\begin{itemize}
    \item \textbf{Optimizer and schedule:} Linear warm-up for $3000$ steps and cosine annealing, with AdamW optimizer~\cite{adamw}. We train for 130 000 steps with a global batch size of $8$. 
    \item \textbf{Position information:} Positional embeddings~\cite{Brown2020} are used for softmax and linear attention models only.
    \item \textbf{Data:} We use Wikitext-103, which contains 103 million tokens extracted from the set of verified articles on Wikipedia. Aside from featuring a large vocabulary, it also retains the original case, punctuation and numbers. Given the fact that it is composed of full articles, this dataset is well-suited for evaluating a model's ability to take advantage of long term dependencies. 
\end{itemize}

\paragraph{Performed Experiments} 
\begin{itemize}
    \item We run all models on the above-specified WikiText task for four different seeds. For each model, we apply the architecture settings in Table~\ref{tab:architecture} and report the results in the corresponding figures of the main text and the Appendix.
    \item Additionally, we evaluate the pretrained Olmo-3-7B-Think model on the WikiText task and provide the resulting eigenvalue spectrum in Figure~\ref{fig:olmo3} of the Appendix.
\end{itemize}

\begin{table}[h]
 \caption{Common architectures for different tasks. Hidden state dimension in SSMs refers to $N$, while in attention models it refers to $m$, i.e., the dimension of the queries and keys. Eval BSZ corresponds to the size of the batch used for the computation of eigenvalues of attention models and Mamba-2. 
    }
\centering
\begin{adjustbox}{max height=\textheight, max width=0.98\textwidth}
\begin{tabular}{lccccccccccc}
\toprule
 & Depth & Model dim $(d)$ & Hidden state dim $(N/m)$ & Heads & Mixer dim & Iterations & Eval BSZ \\
\midrule
CIFAR-10 & 6 & 512 & 64 &  4 & 128 & 50 epochs &  64\\
IMDb & 4 & 128 & 64 & 4 & 512 & 30 epochs & 32 \\
ListOps & 6 & 128 & 64 & 4 & 256 & 50 epochs &  32 \\
\midrule
MQAR & 2 & 128 & 128 & 1 & - & 40k steps & 64\\
\midrule
WikiText & 6 & 512 & 512 & 8 & 512 & 130k steps & 8\\
\bottomrule
\end{tabular}
\end{adjustbox}
    \label{tab:architecture}
\end{table}

 {
\begin{table}[H]
 \caption{ {Total number of parameters (in millions) excluding the encoder, across different tasks.}}
\centering
\begin{tabular}{lccccccccccc}
\toprule
 & S4 & LRU & Mamba-2 & Lin Att & Sm Att & Norm Att &\\
\midrule
CIFAR-10 & 4.74 & 3.95 & 6.73 &  4.41 & 4.41 & 4.42 \\
IMDb & 0.40 & 0.27 & 0.34 & 0.79 & 0.79 & 0.79 \\
ListOps & 0.60 & 0.40 & 0.51 & 0.73 & 0.73 & 0.73 \\
\midrule
MQAR & 1.39 & 1.26 & 1.26 & 1.18 & 1.18 & 1.18 \\
\midrule
WikiText & 41.53 & 35.24 & 35.30 & 35.19 &  35.19 & 35.22\\
\bottomrule
\end{tabular}
    \label{tab:num_params}
\end{table}
}

\clearpage
\newpage

\section{Additional Results on Spectral Pattern Analysis Across Models and Tasks}\label{app:additional_results}

\begin {figure}[H]
\centering
\begin{adjustbox}{max height=\textheight, max width=0.75\textwidth}
\begin{tikzpicture}[scale=1.0]
    
    \node (pic1) at (-10.0, 10) {\includegraphics[width=22cm]
    {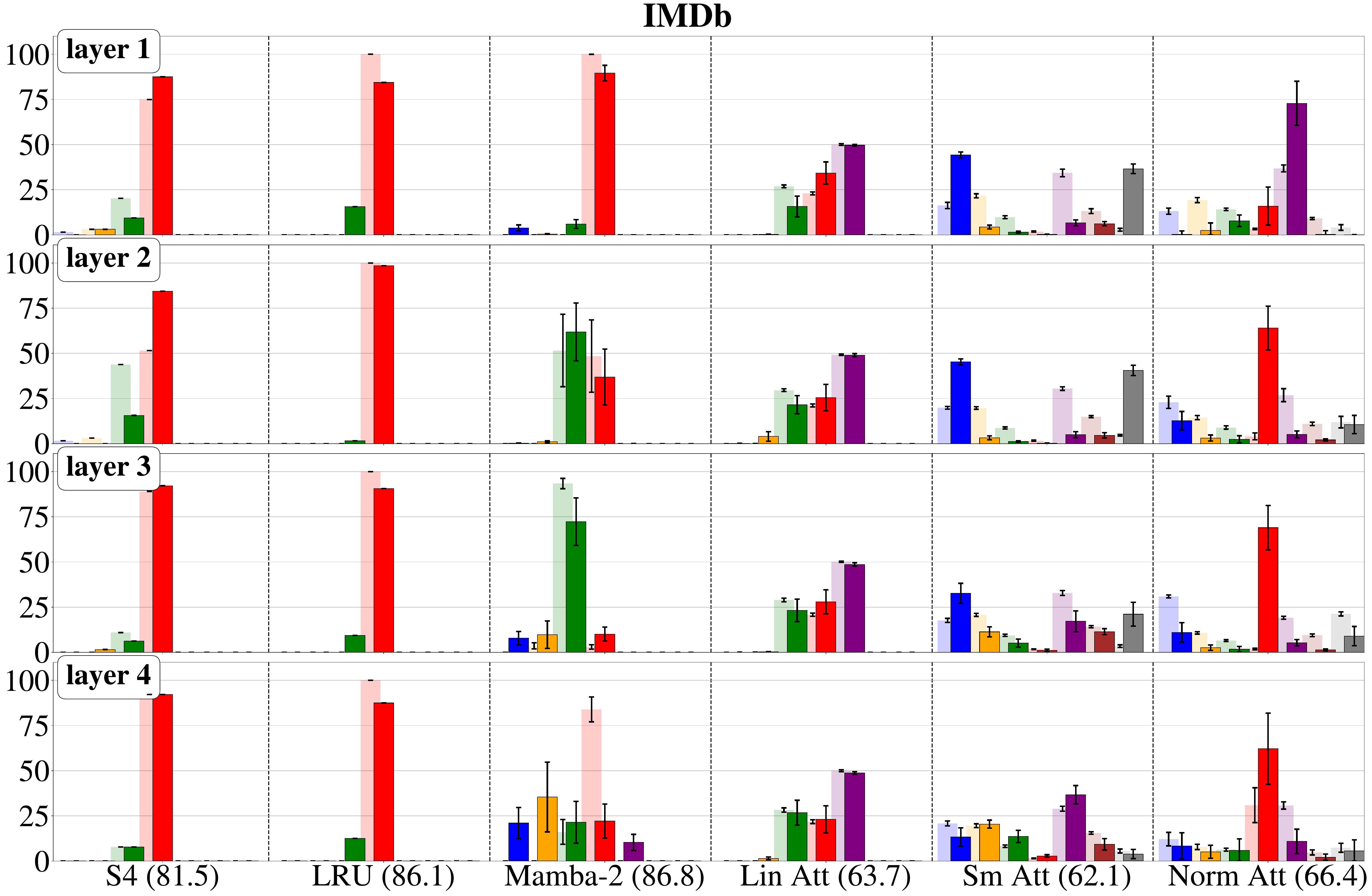}};
    
    \node (pic3) at (-10.0, -7) {\includegraphics[width=22cm]
    {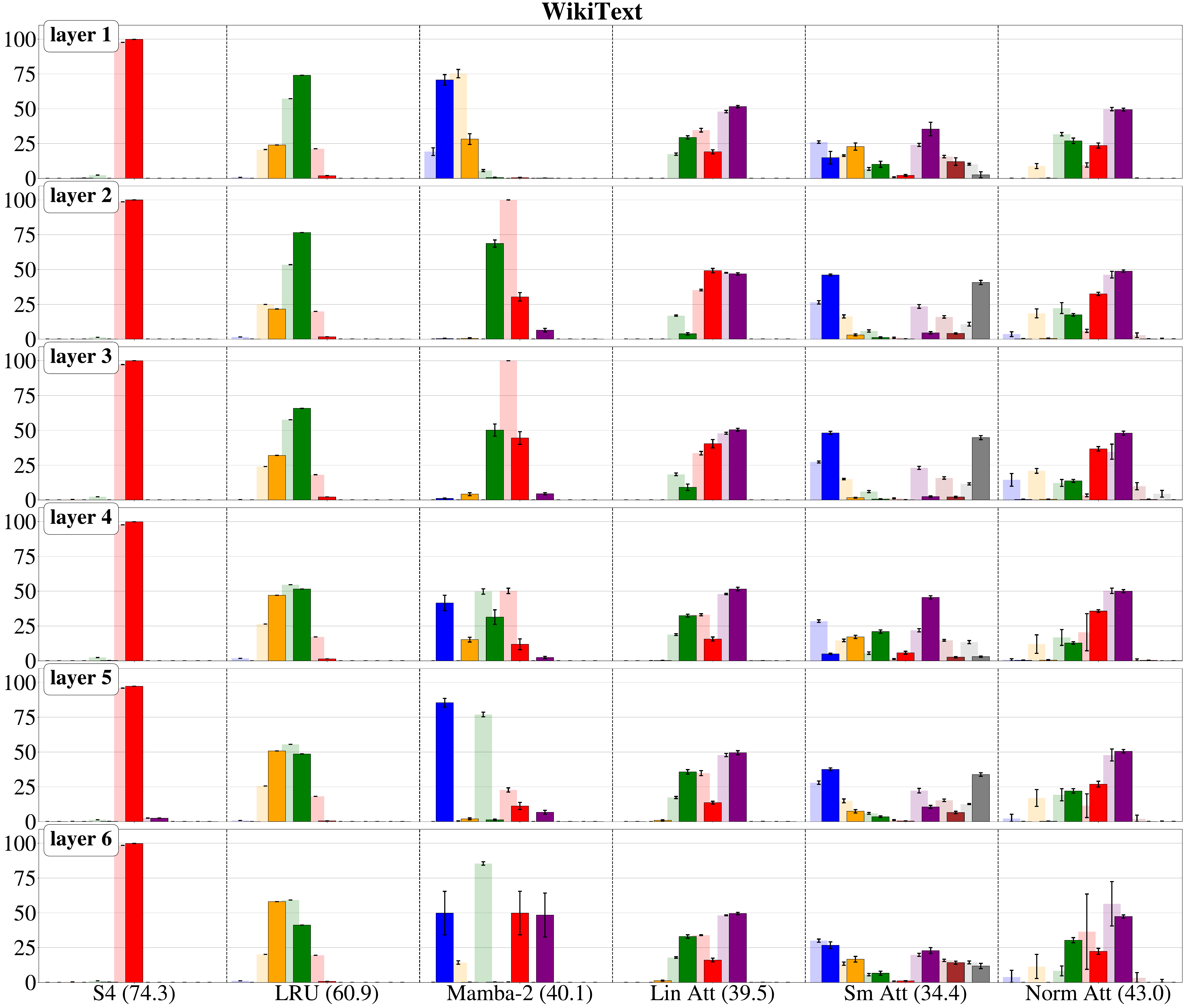}};
    
\end{tikzpicture}
\end{adjustbox}
\caption{Eigenvalue distributions for one head, across all models, layers, and tasks (excluding MQAR).
}
  \label{fig:main_plot_appendix}
\end{figure}

\begin {figure}[H]
    \centering
    \begin{adjustbox}{max height=0.95\textheight, max width=0.90\textwidth}
        \begin{tikzpicture}[scale=1.0]
            
            \node (pic1) at (-10.0, 10) {\includegraphics[width=16cm]
            {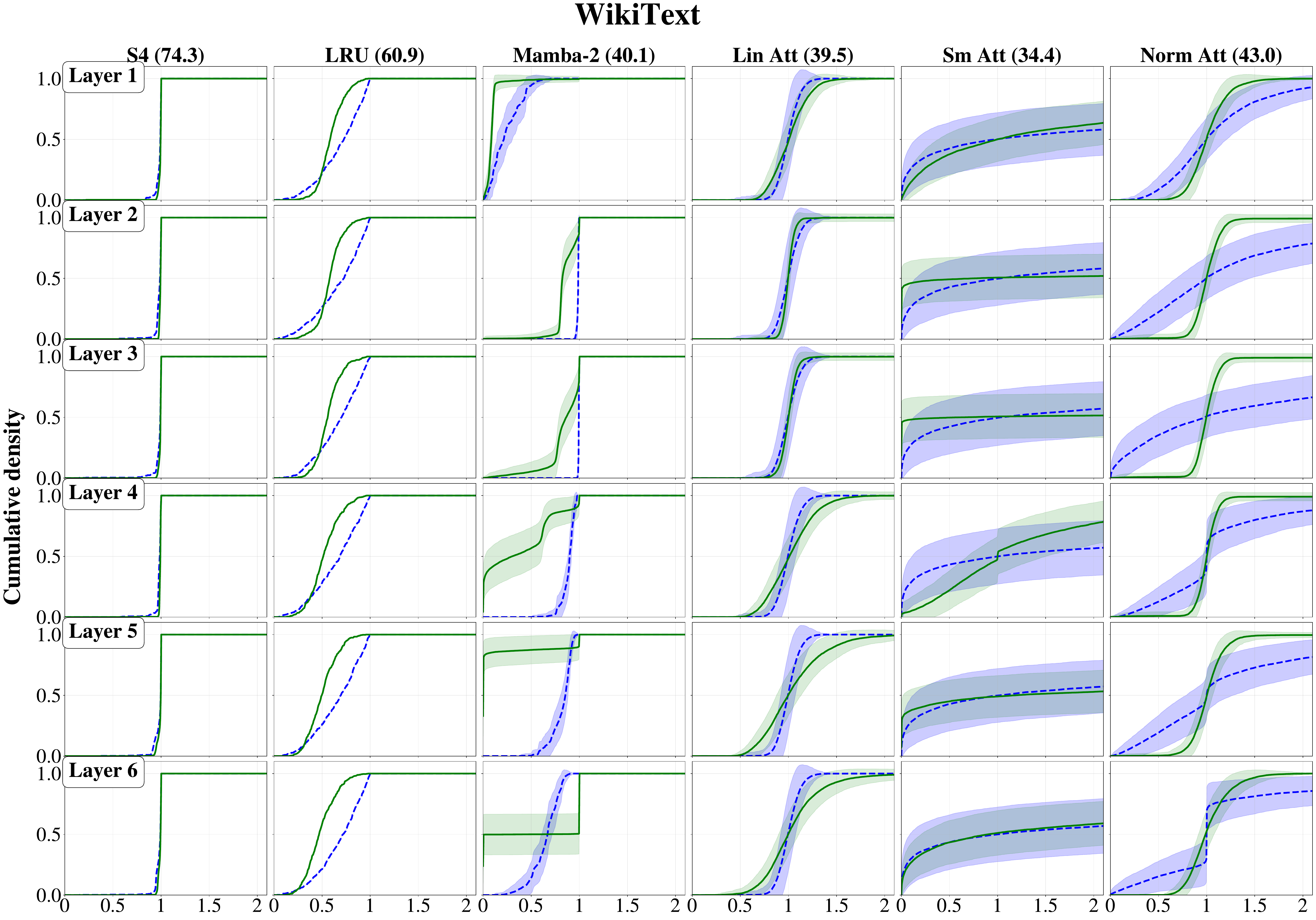}};
            
            \node (pic3) at (-10.0, 0.6) {\includegraphics[width=16cm]
            {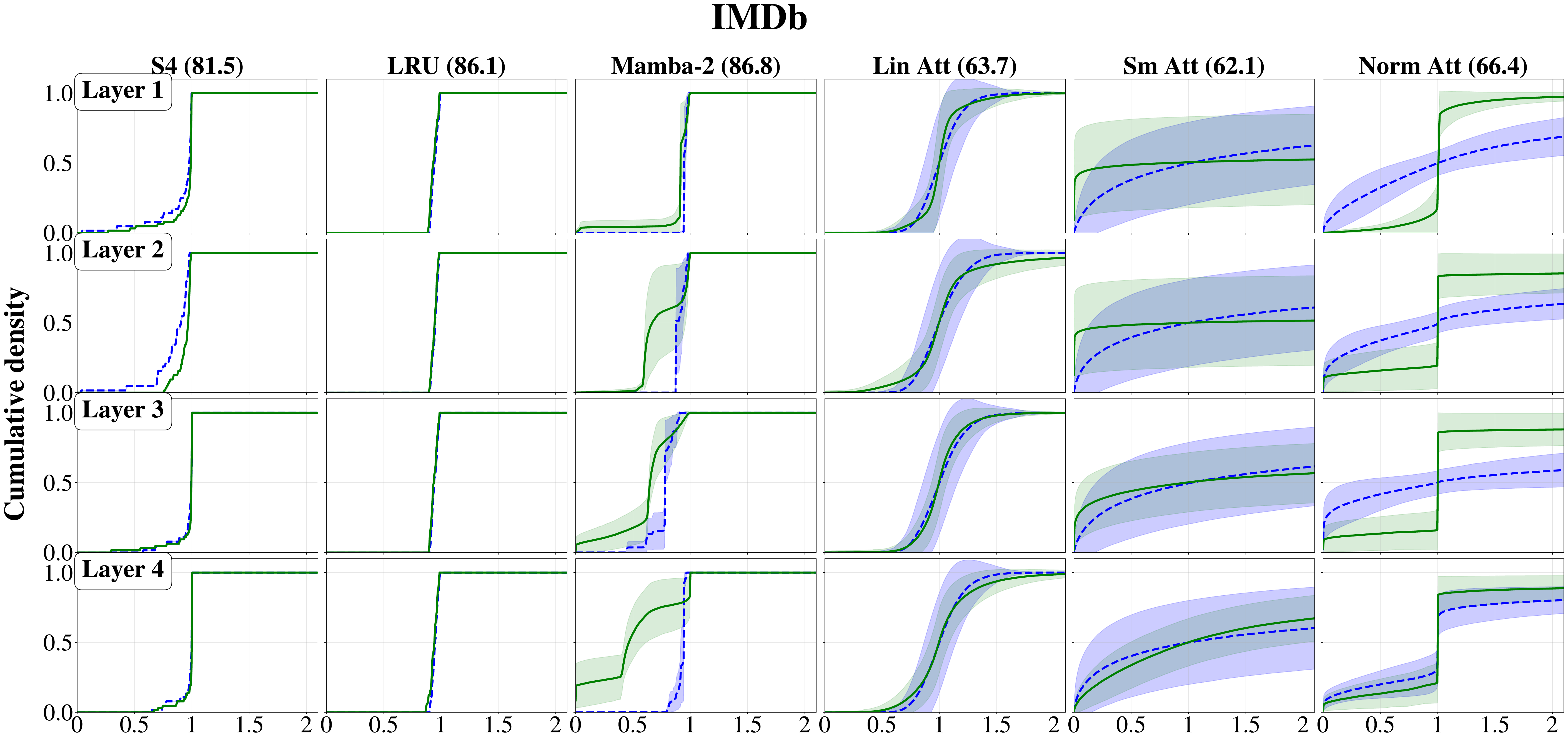}};

            \node (pic3) at (-10.0, -5.5) {\includegraphics[width=16cm]
            {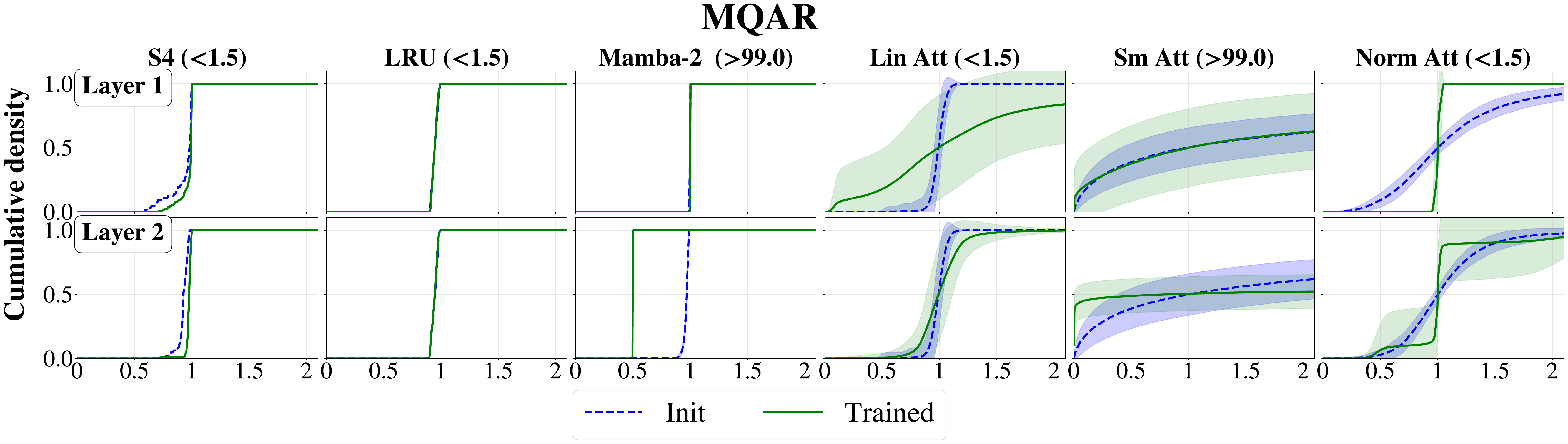}};
            
        \end{tikzpicture}
    \end{adjustbox}
    \caption{Eigenvalue distributions as a cumulative density plot for one head, across all models, layers, and tasks.The distribution at initialization is shown as a dashed blue line, while the distribution after training is shown as a solid green line. Shaded regions of the corresponding colors indicate one standard deviation. 
    }
  \label{fig:cdf_appendix}
\end{figure}

\begin {figure}[H]
\centering
\begin{adjustbox}{max height=0.9\textheight, max width=0.95\textwidth}
\begin{tikzpicture}[scale=1.0]

    \node (pic1) at (-10.0, 8.5) {\includegraphics[width=11.85cm]
    {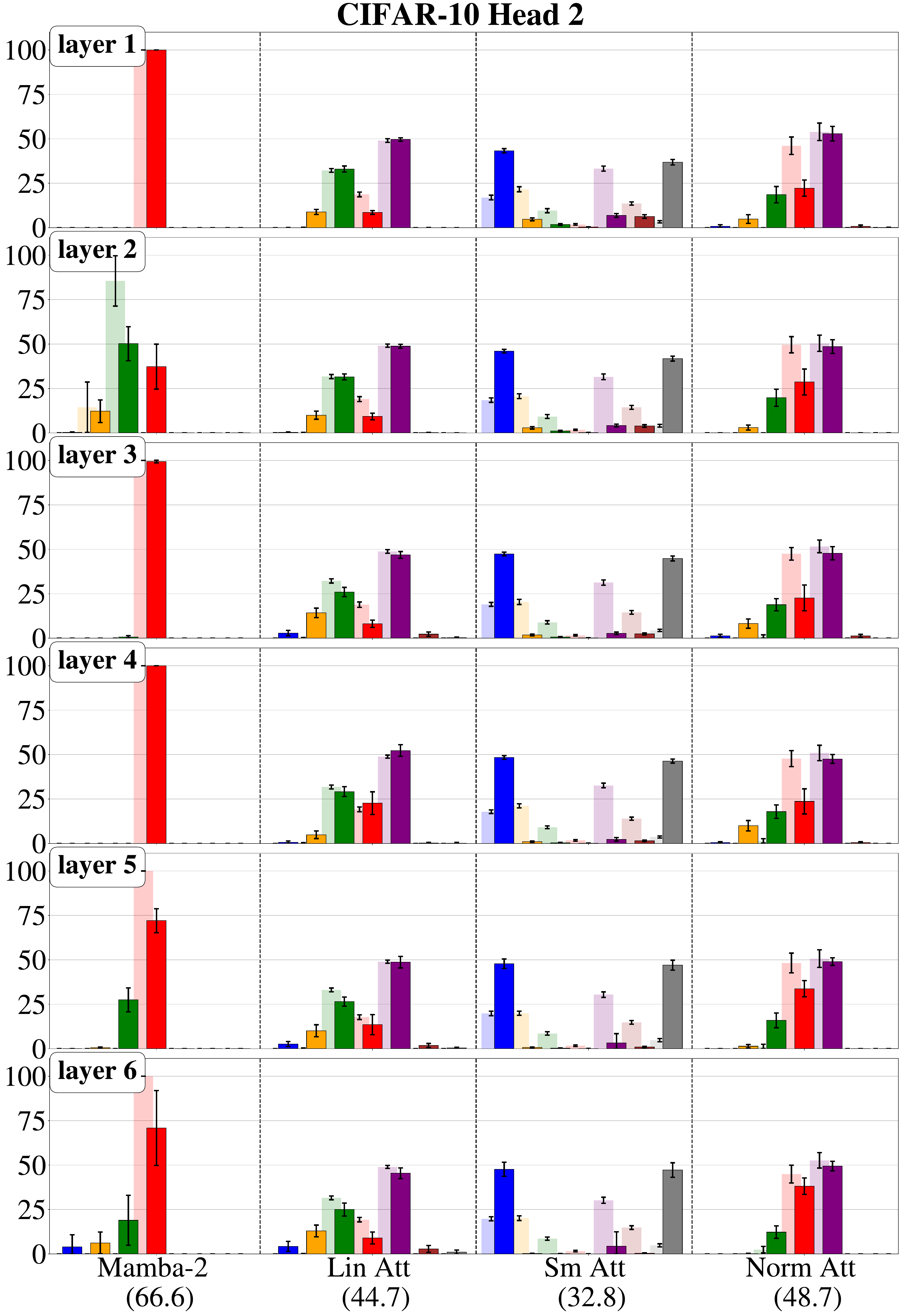}};

    \node (pic2) at (1.65, 8.5) {\includegraphics[width=11.3cm]
    {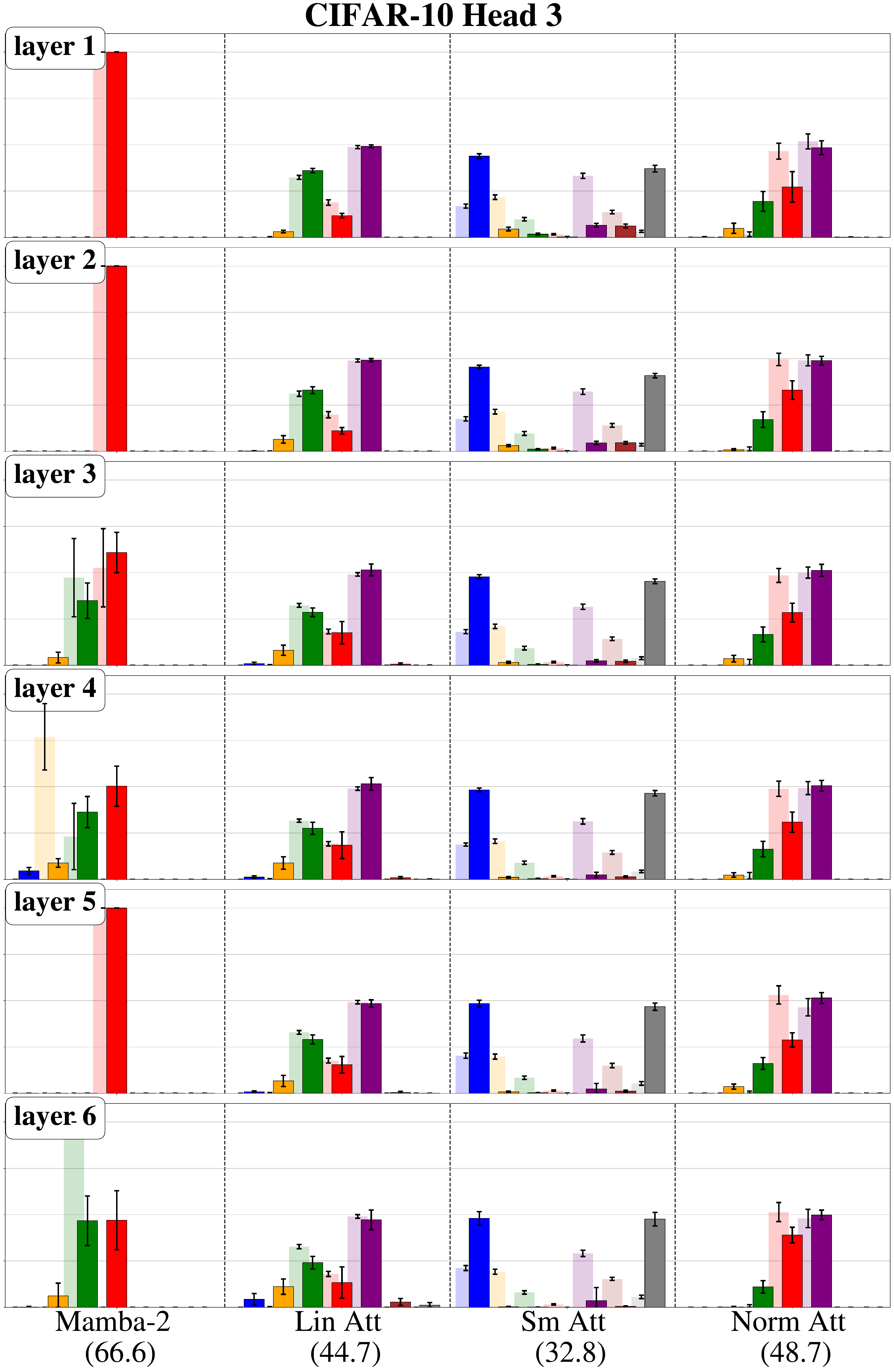}};

    \node (pic3) at (-4.5, -8.9) {\includegraphics[width=11.85cm]
    {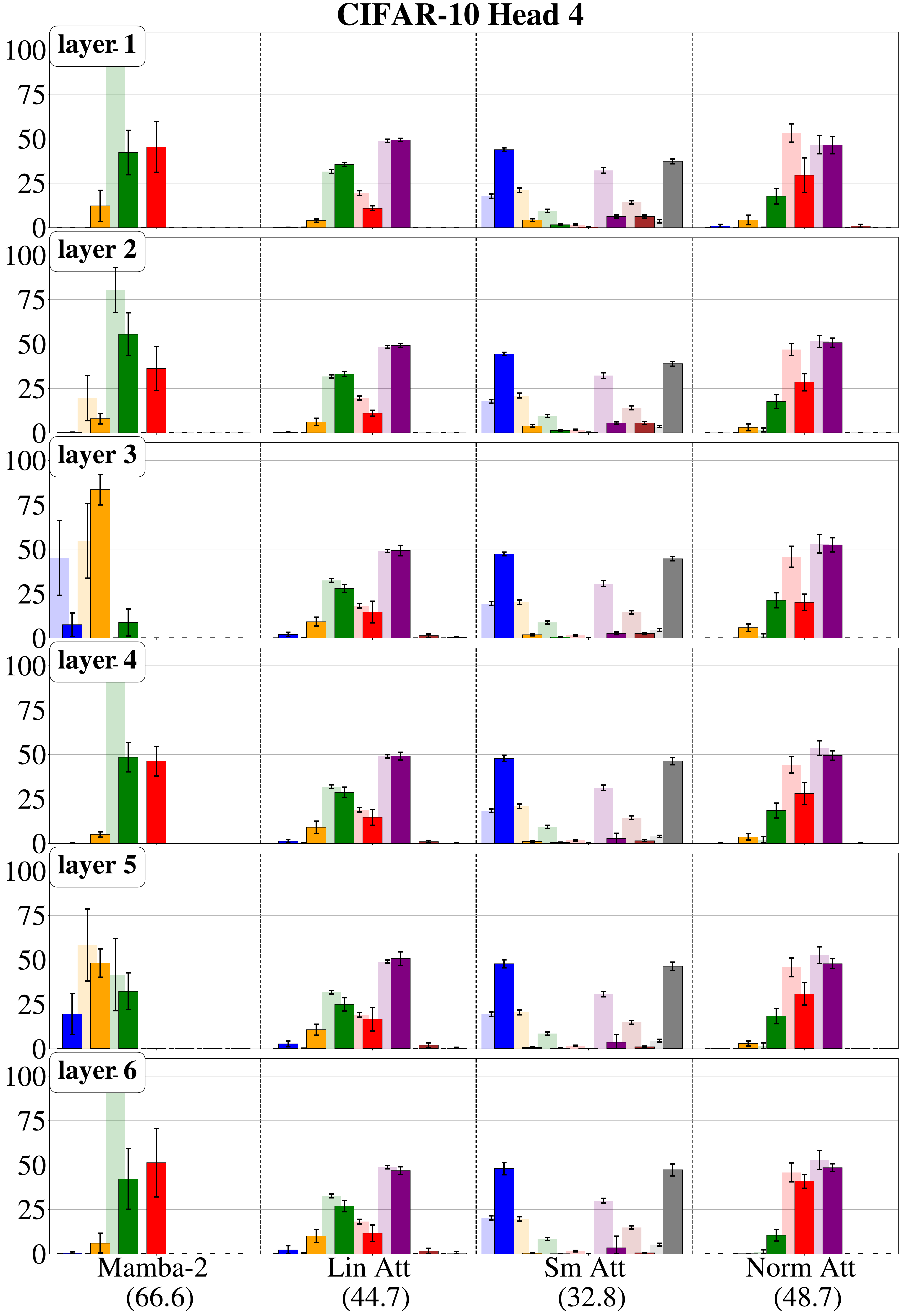}};
    
\end{tikzpicture}
\end{adjustbox}
\caption{Eigenvalue distributions across remaining heads and all layers for CIFAR-10.}
 \label{fig:heads_cifar_plot}
\end{figure}

\begin {figure}[H]
\centering
\begin{adjustbox}{max height=0.95\textheight, max width=0.95\textwidth}
\begin{tikzpicture}[scale=1.0]

    \node (pic1) at (-10.0, 8.5) {\includegraphics[width=11.85cm]
    {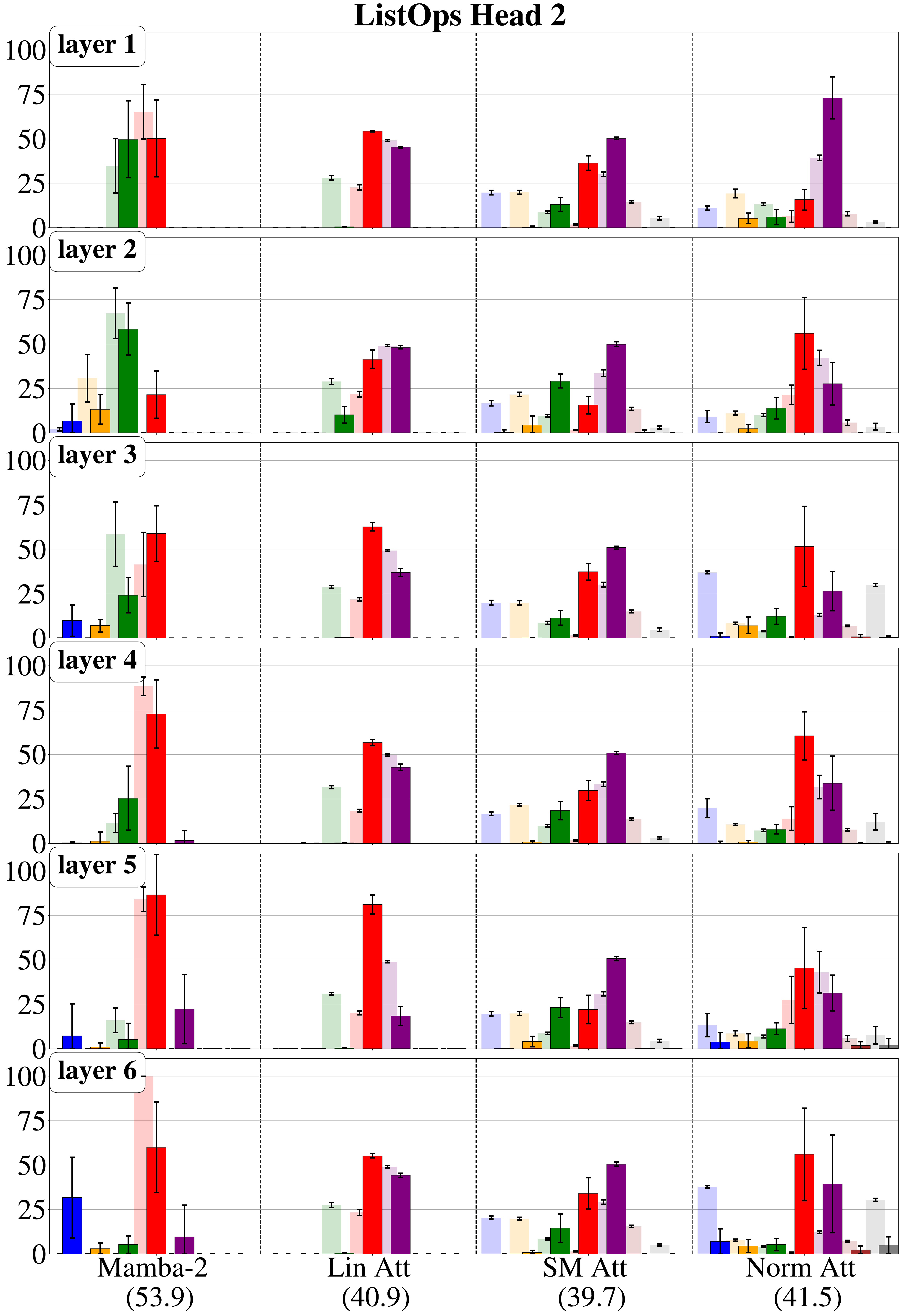}};

    \node (pic2) at (1.65, 8.5) {\includegraphics[width=11.3cm]
    {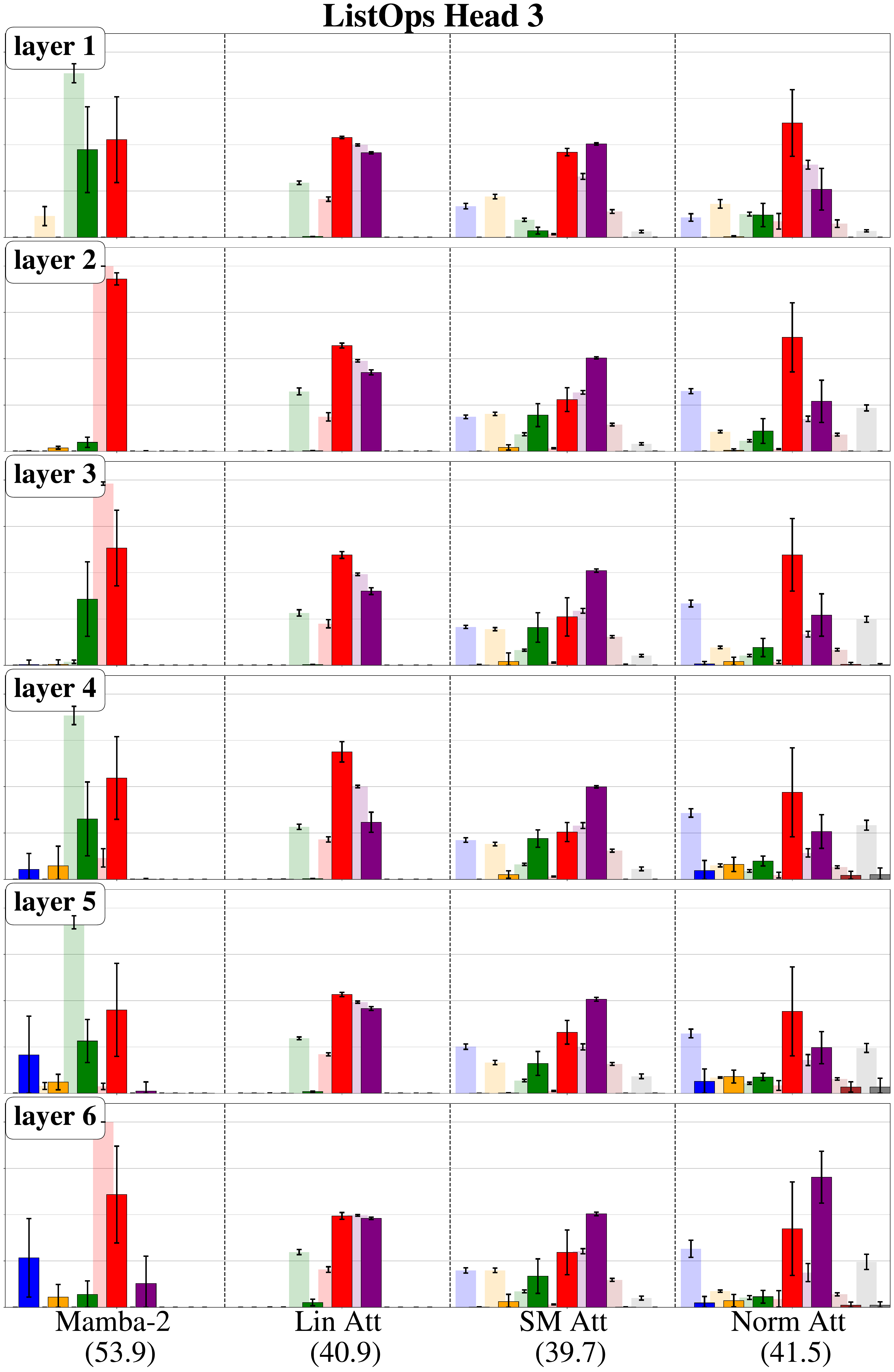}};

    \node (pic3) at (-4.5, -8.9) {\includegraphics[width=11.85cm]
    {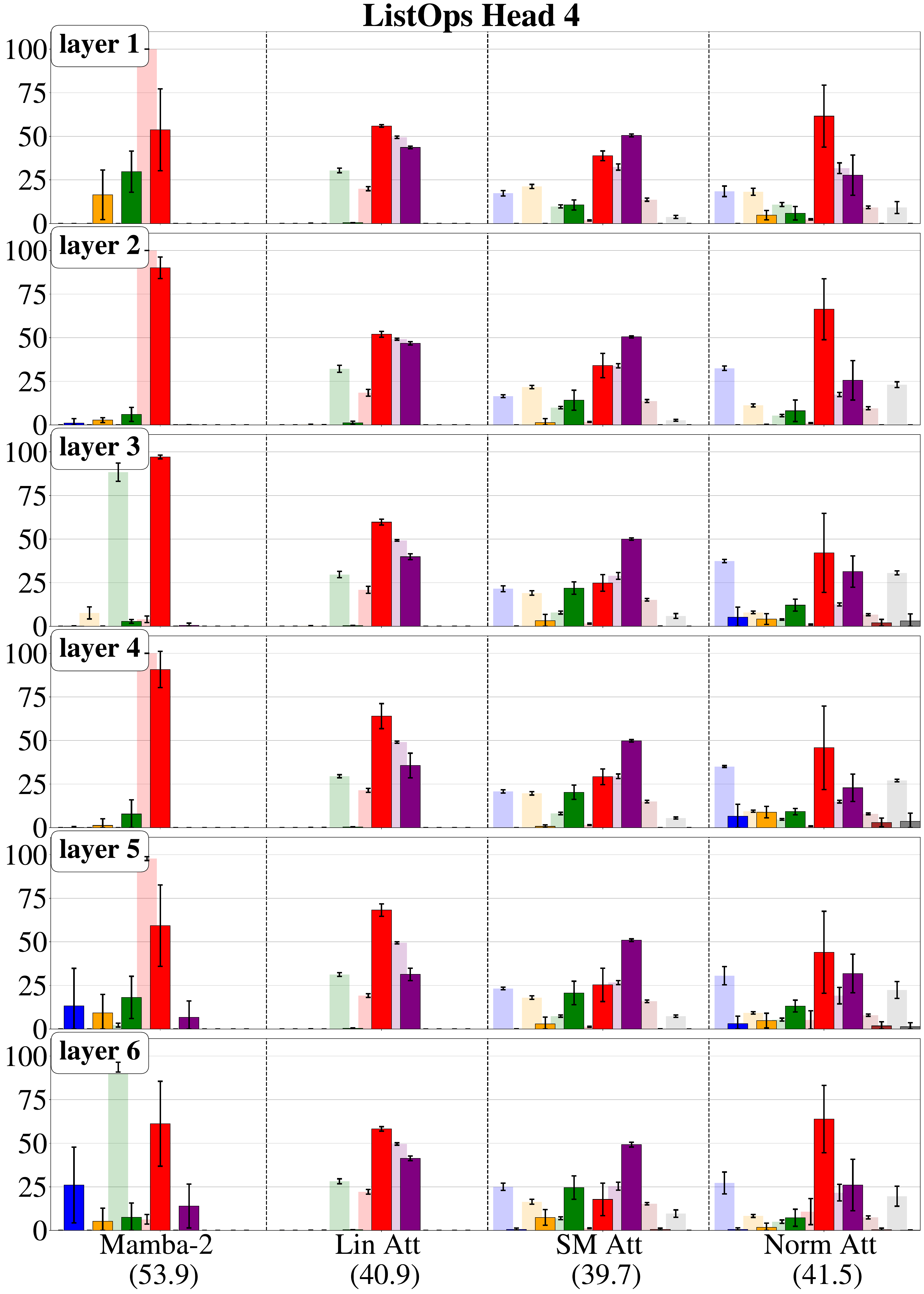}};
    
\end{tikzpicture}
\end{adjustbox}
\caption{Eigenvalue distributions across remaining heads and all layers for ListOps.}
 \label{fig:heads_listops_plot}
\end{figure}

\begin {figure}[H]
\centering
\begin{adjustbox}{max height=0.95\textheight, max width=0.95\textwidth}
\begin{tikzpicture}[scale=1.0]

    \node (pic1) at (-10.0, 8.5) {\includegraphics[width=11.85cm]
    {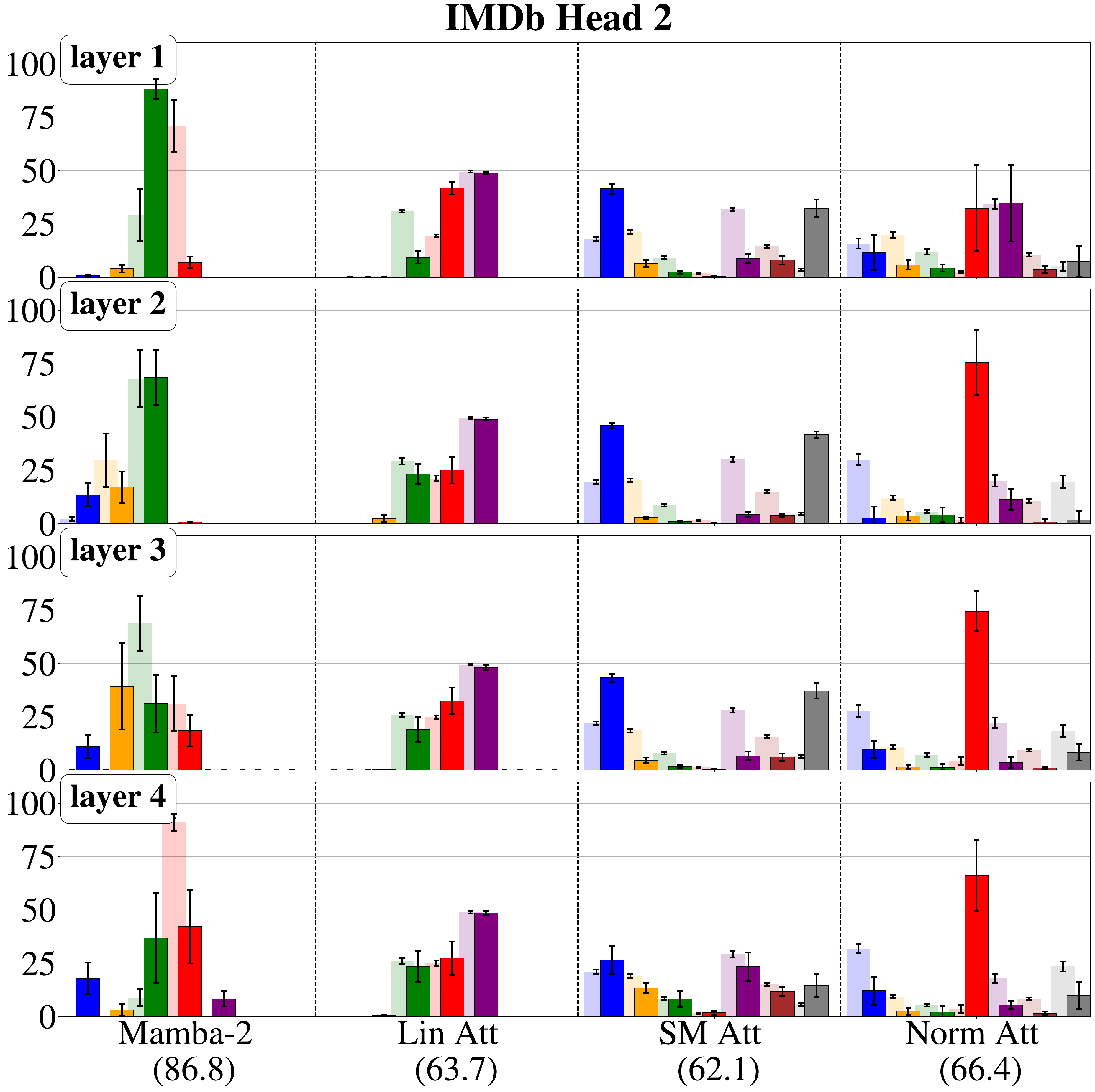}};

    \node (pic2) at (1.65, 8.5) {\includegraphics[width=11.3cm]
    {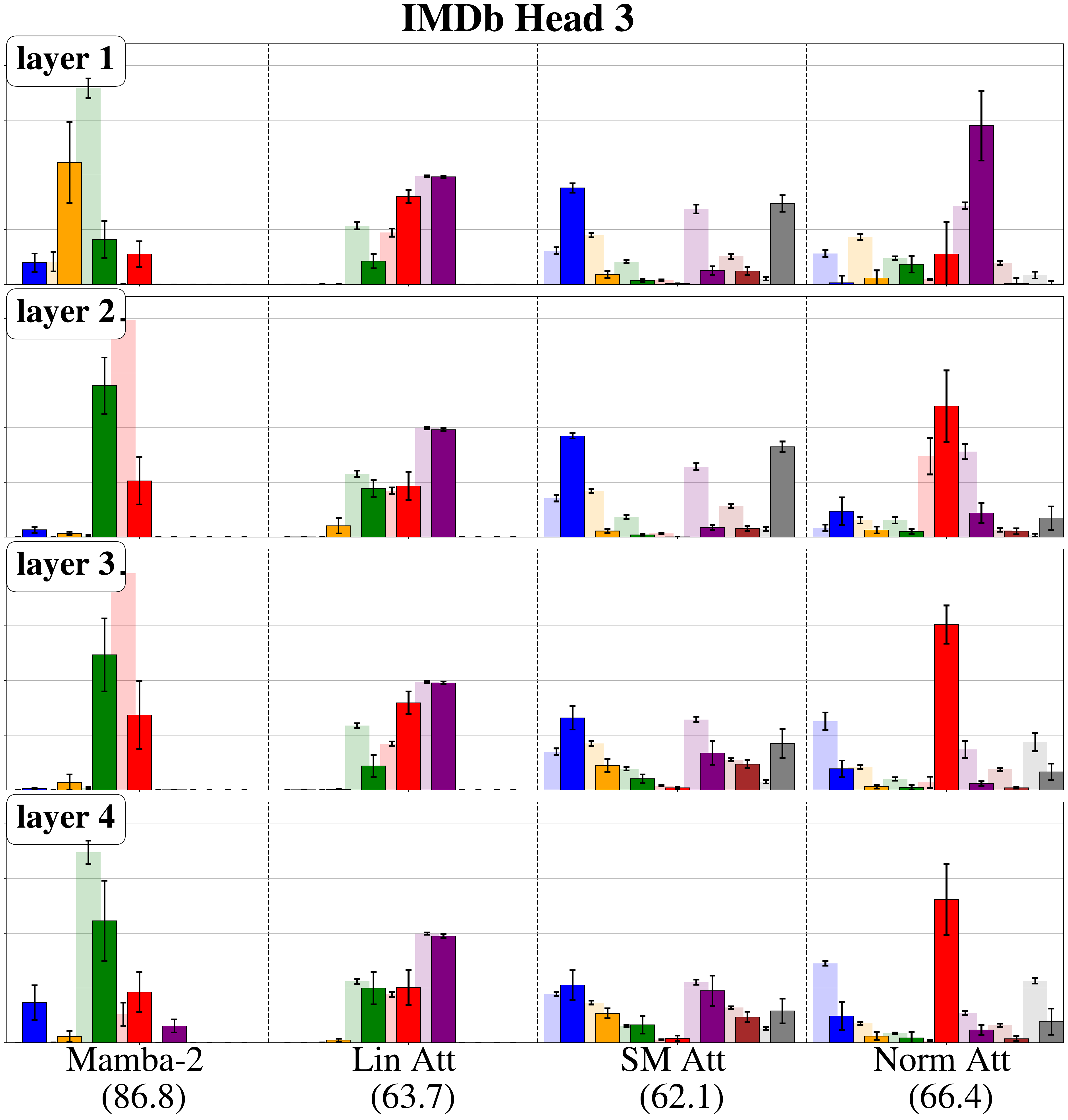}};

    \node (pic3) at (-4.5, -3.9) {\includegraphics[width=11.85cm]
    {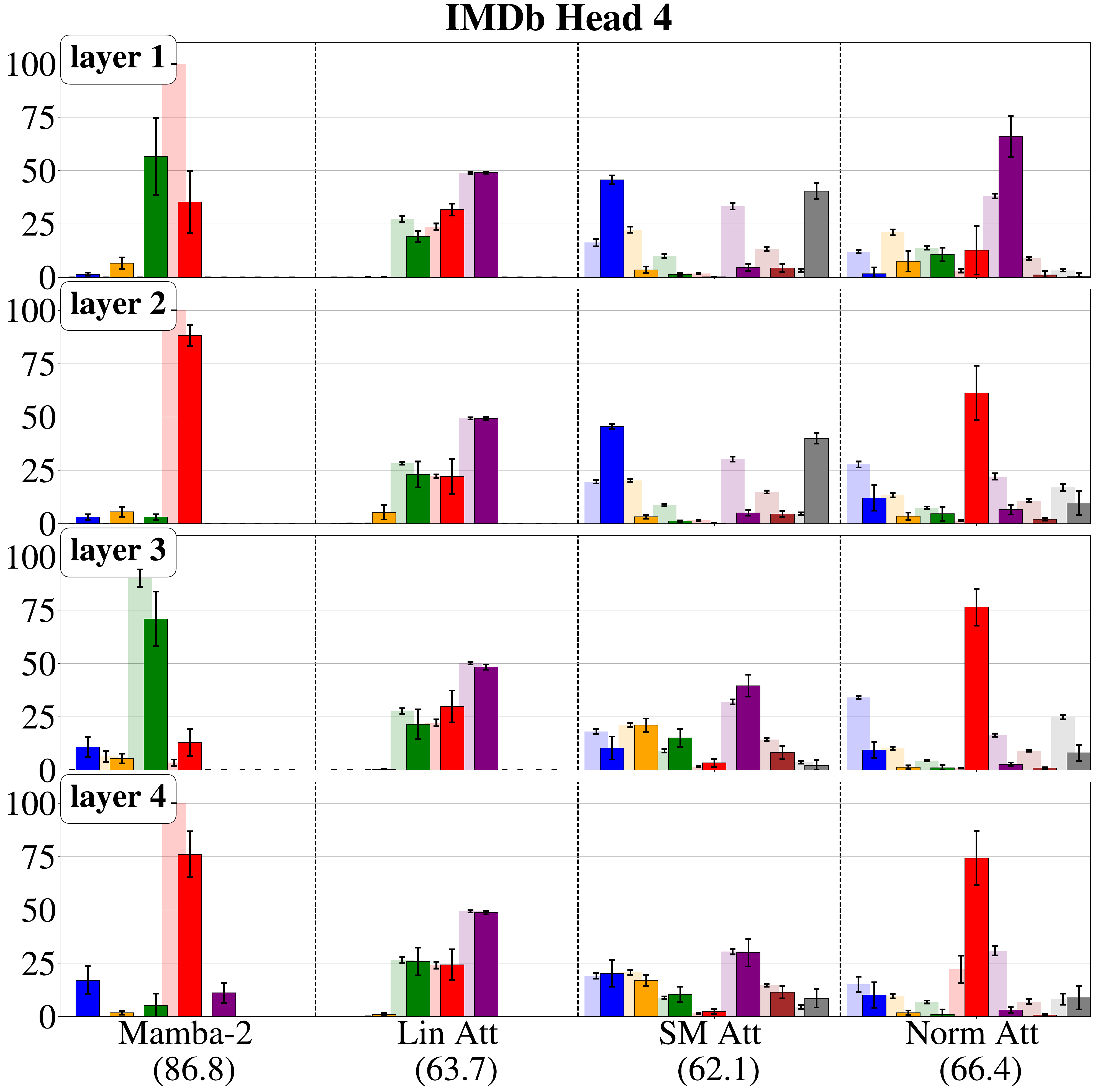}};
    
\end{tikzpicture}
\end{adjustbox}
\caption{Eigenvalue distributions across remaining heads and all layers for IMDb.}
 \label{fig:heads_imdb_plot}
\end{figure}

\begin {figure}[H]
\centering
\begin{adjustbox}{max height=0.95\textheight, max width=0.95\textwidth}
\begin{tikzpicture}[scale=1.0]

    \node (pic1) at (-10.0, 8.5) {\includegraphics[width=11.85cm]
    {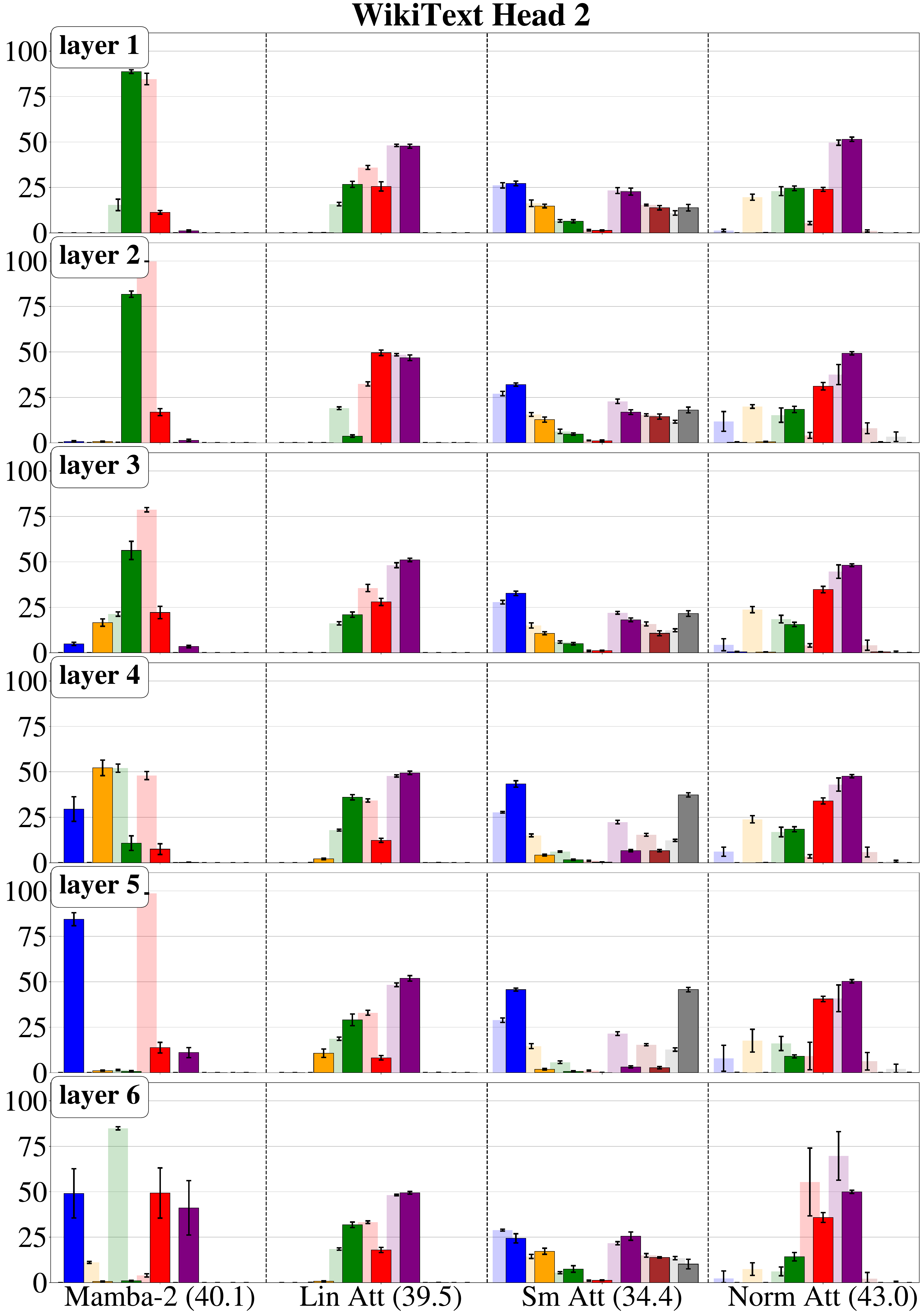}};

    \node (pic2) at (1.65, 8.5) {\includegraphics[width=11.3cm]
    {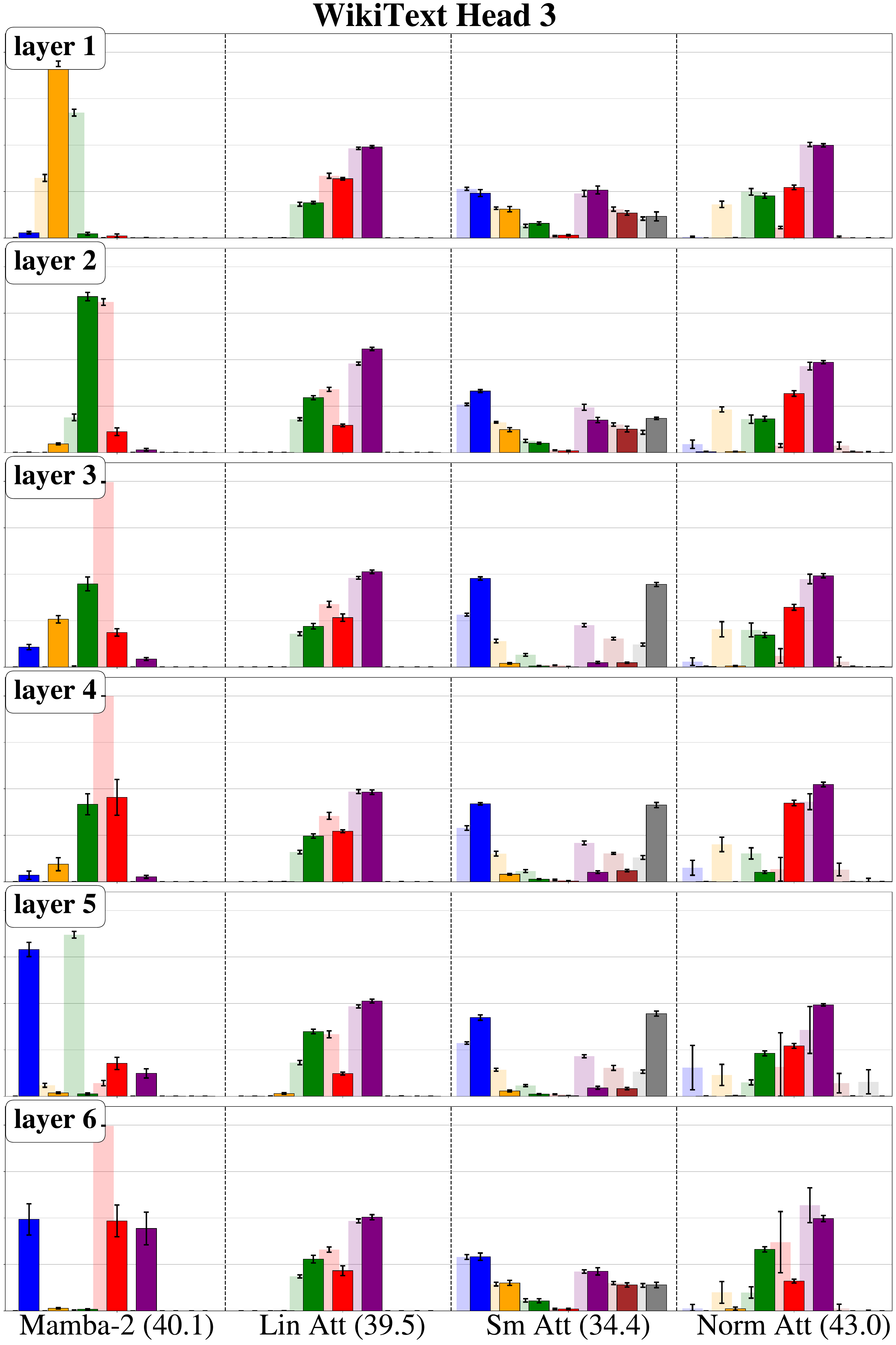}};

    \node (pic3) at (-10.0, -8.9) {\includegraphics[width=11.85cm]
    {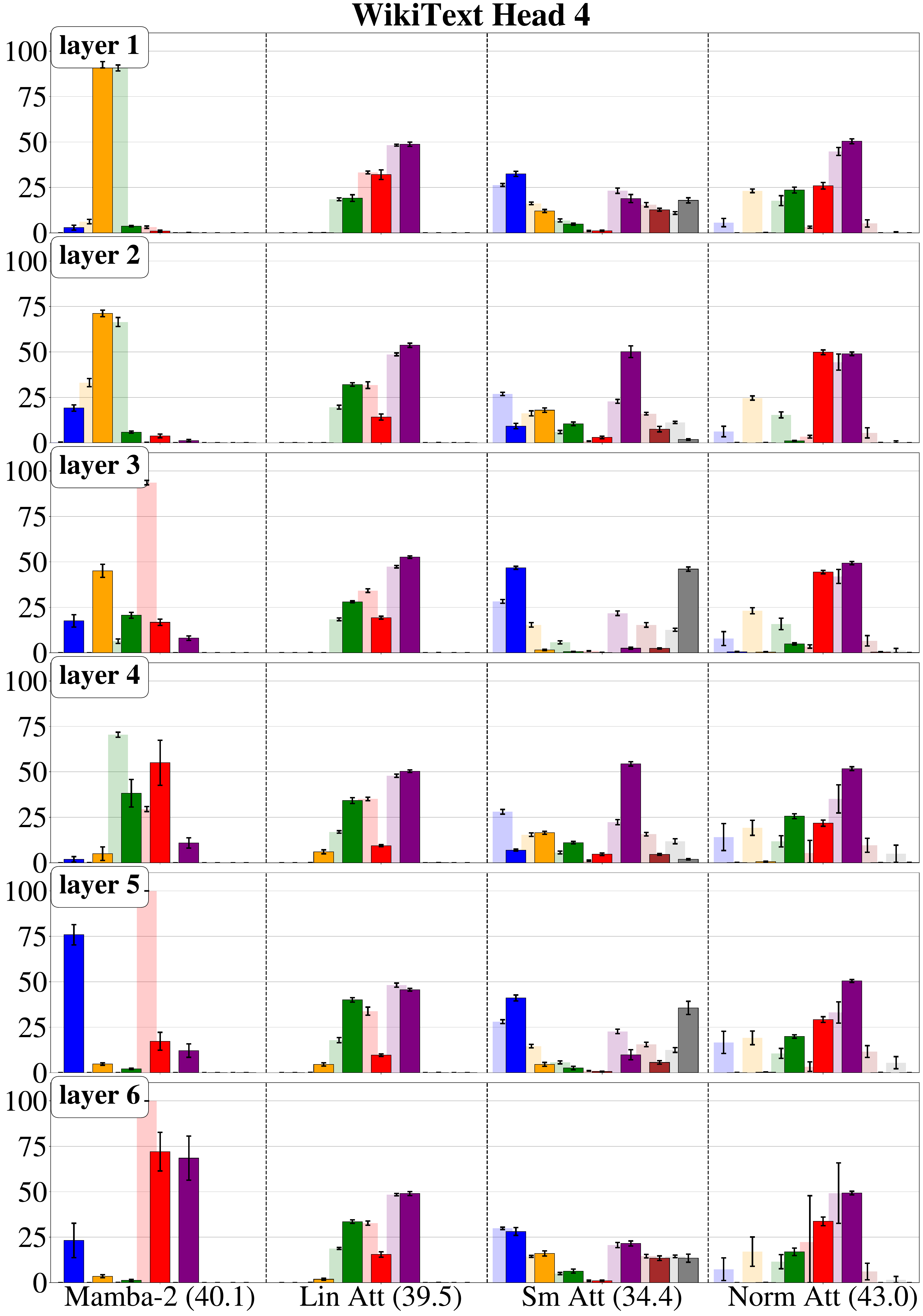}};

    \node (pic3) at (1.65, -8.9) {\includegraphics[width=11.3cm]
    {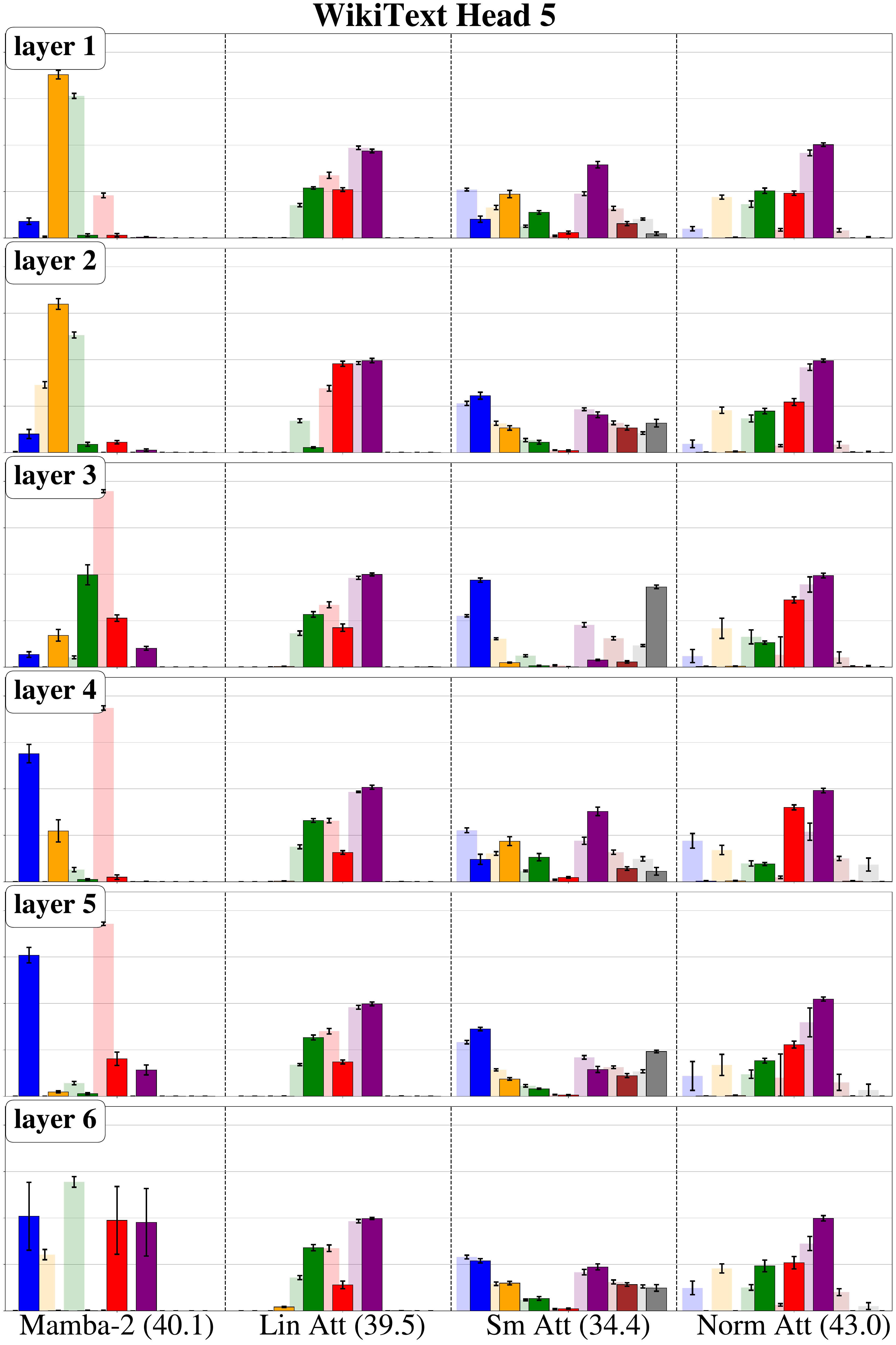}};
    
\end{tikzpicture}
\end{adjustbox}
\caption{Eigenvalue distributions across 4 additional heads and all layers for WikiText.}
 \label{fig:heads_wikitext_plot}
\end{figure}

\begin {figure}[H]
\centering
\begin{adjustbox}{max height=0.95\textheight, max width=0.95\textwidth}
\begin{tikzpicture}[scale=1.0]

    \node (pic1) at (-10.0, 8.5) {\includegraphics[width=11.85cm]
    {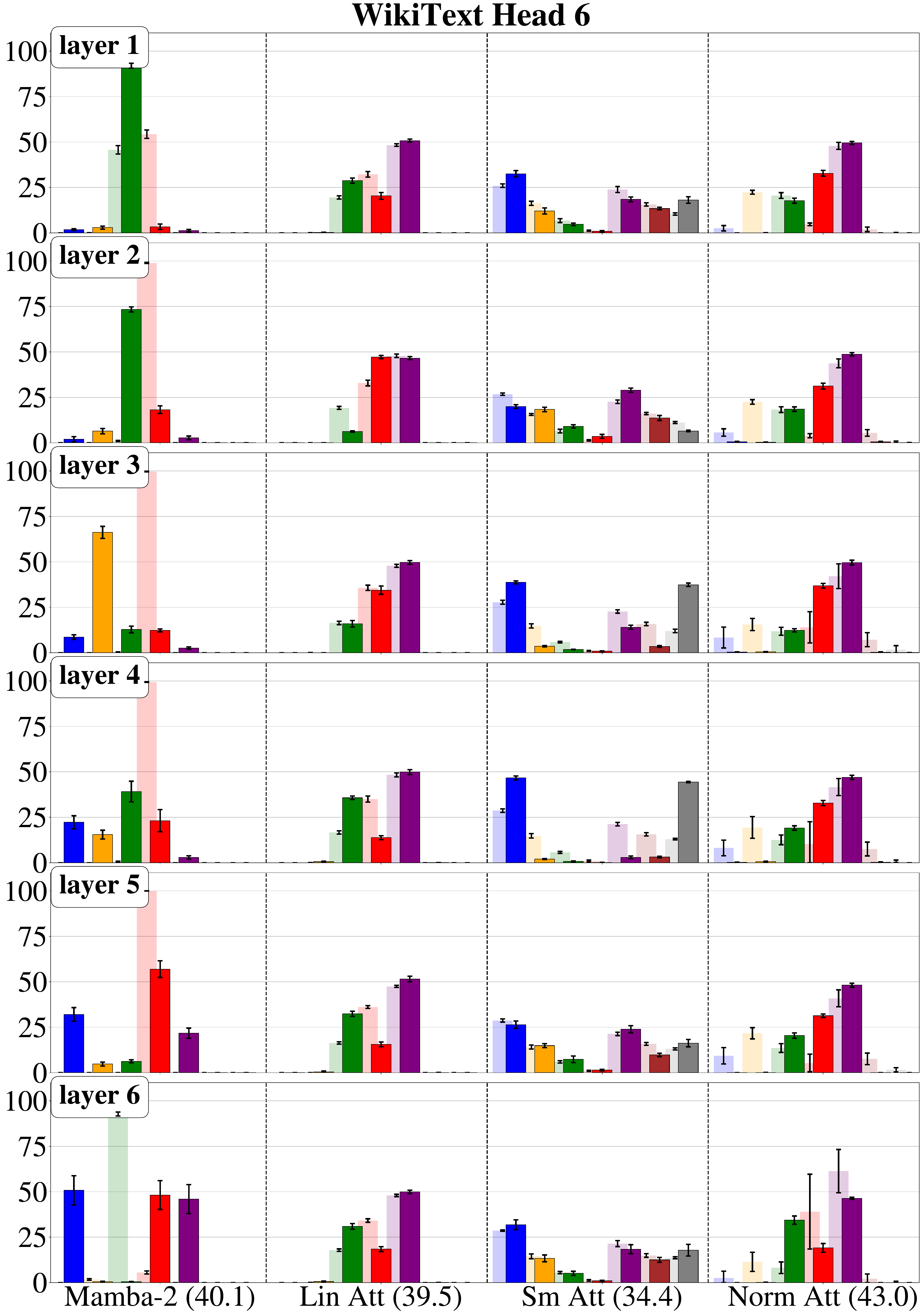}};

    \node (pic2) at (1.65, 8.5) {\includegraphics[width=11.3cm]
    {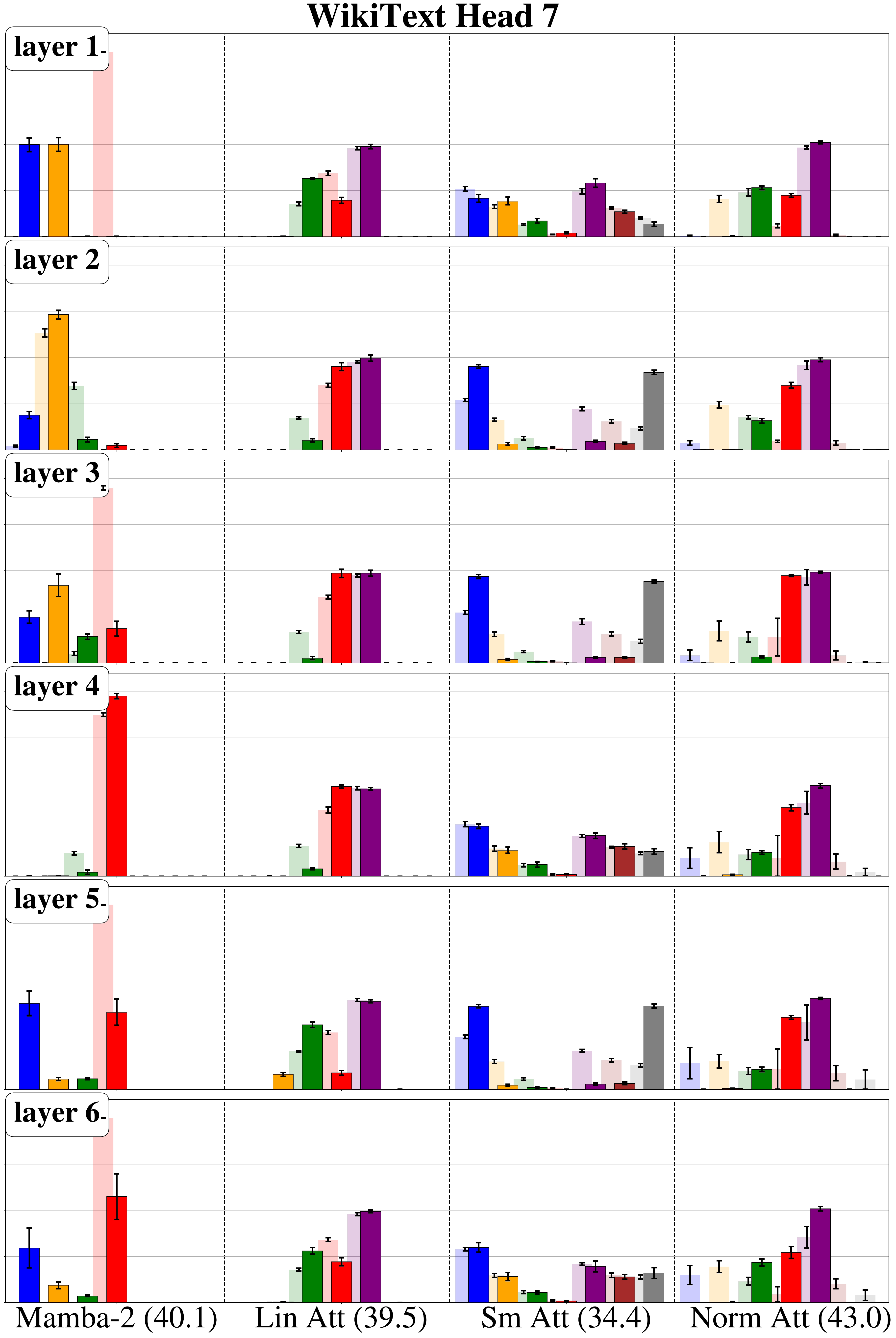}};

    \node (pic3) at (-4.5, -8.9) {\includegraphics[width=11.85cm]
    {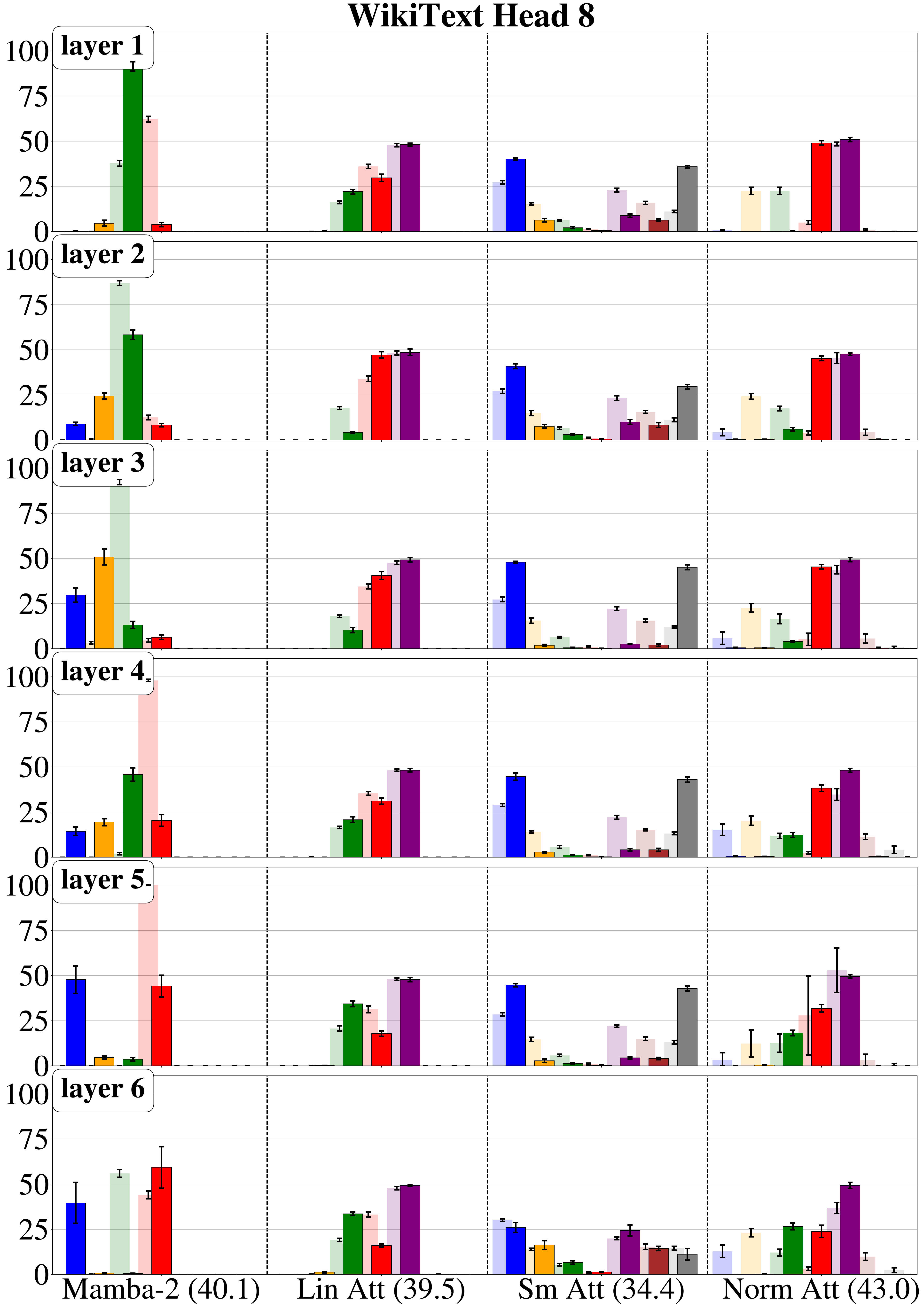}};
    
\end{tikzpicture}
\end{adjustbox}
\caption{Eigenvalue distributions across remaining heads and all layers for WikiText.}
 \label{fig:heads_wikitext_plot2}
\end{figure}

\begin {figure}[H]
\centering
\begin{adjustbox}{max height=0.9\textheight, max width=0.98\textwidth}
\begin{tikzpicture}[scale=1.0]

    \node (pic1) at (-10.0, 8.5) {\includegraphics[width=11.76cm]
    {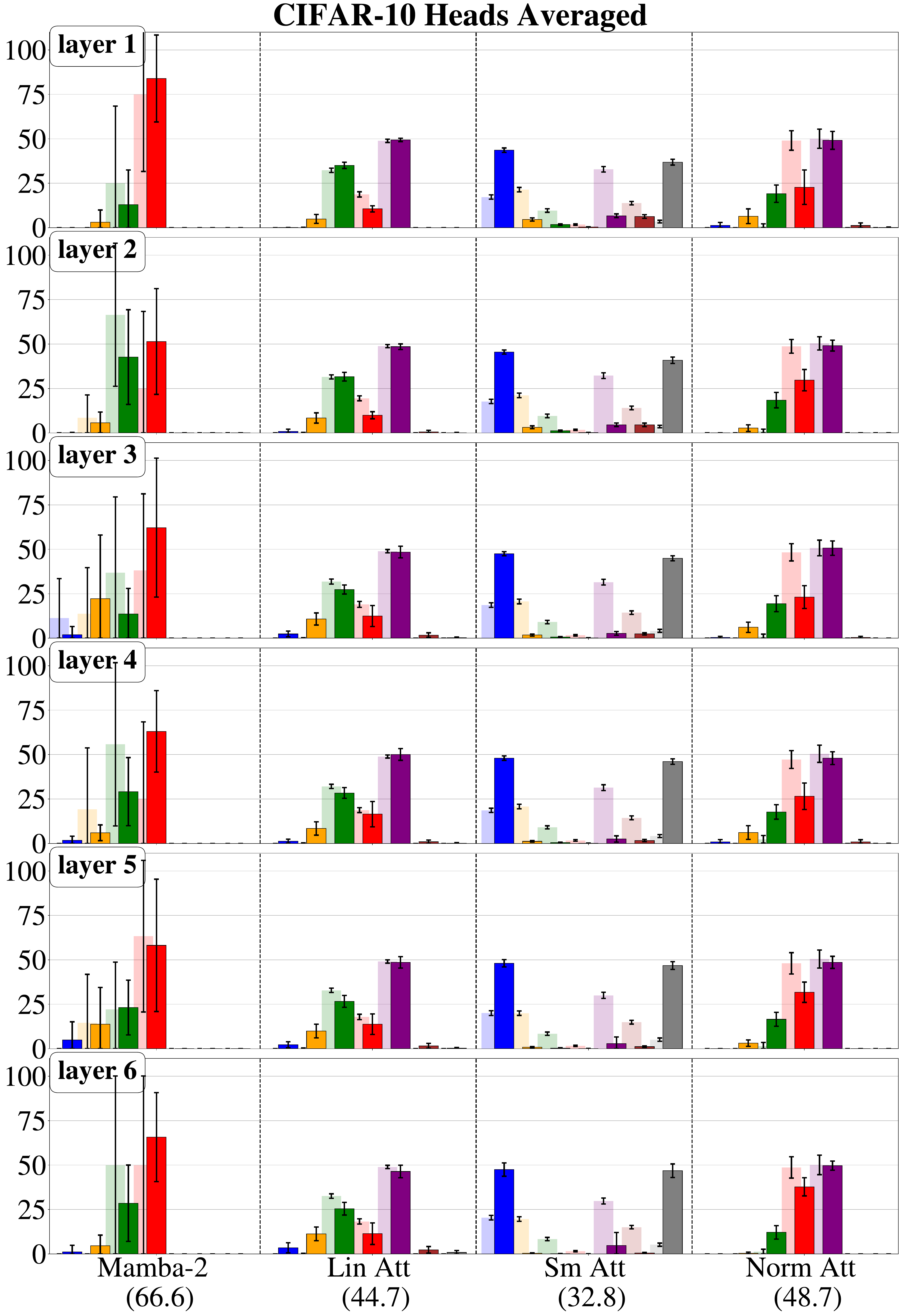}};

    \node (pic2) at (1.65, 8.5) {\includegraphics[width=11.2cm]
    {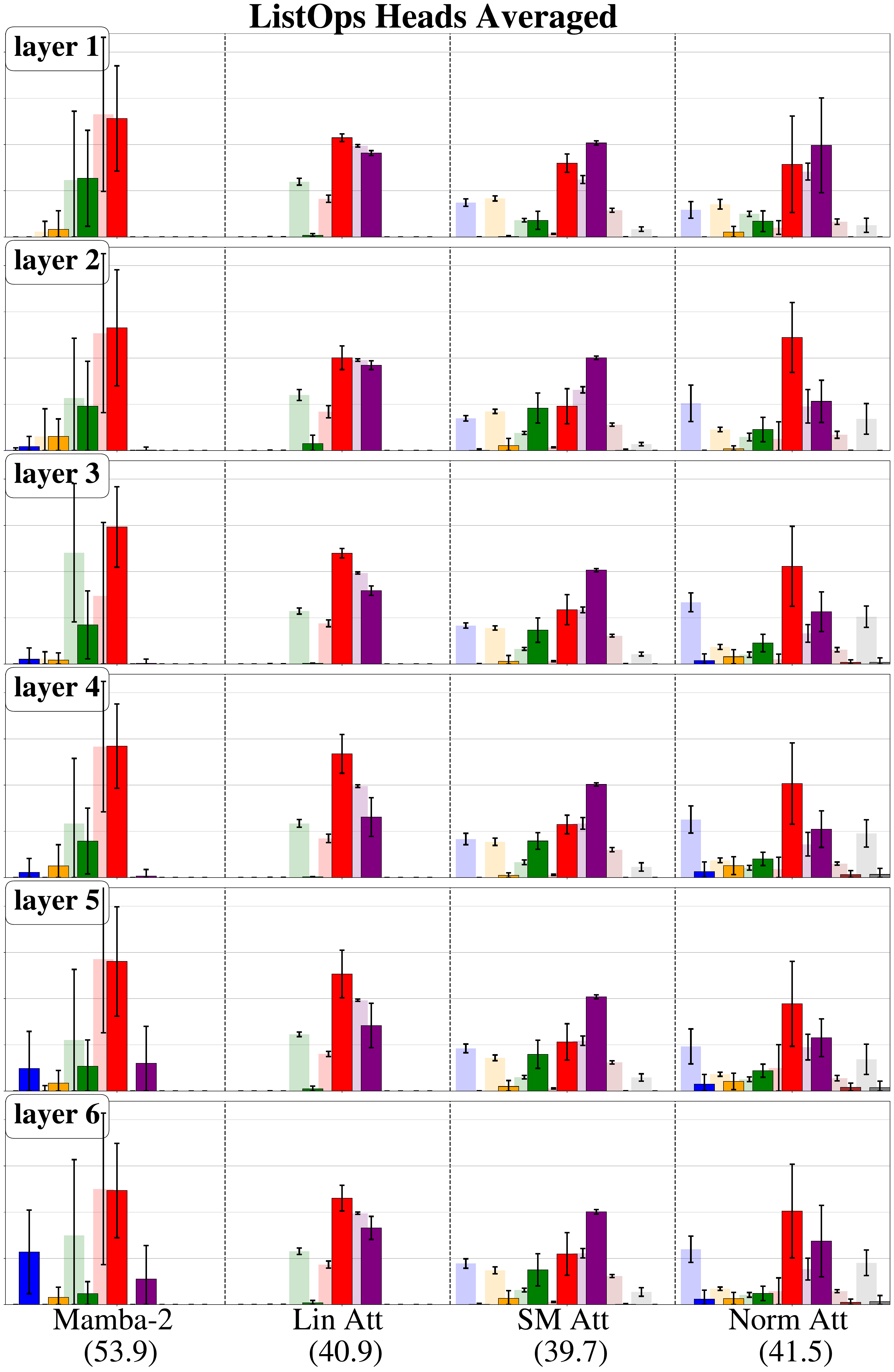}};

    \node (pic3) at (-10.0, -8.9) {\includegraphics[width=11.76cm]
    {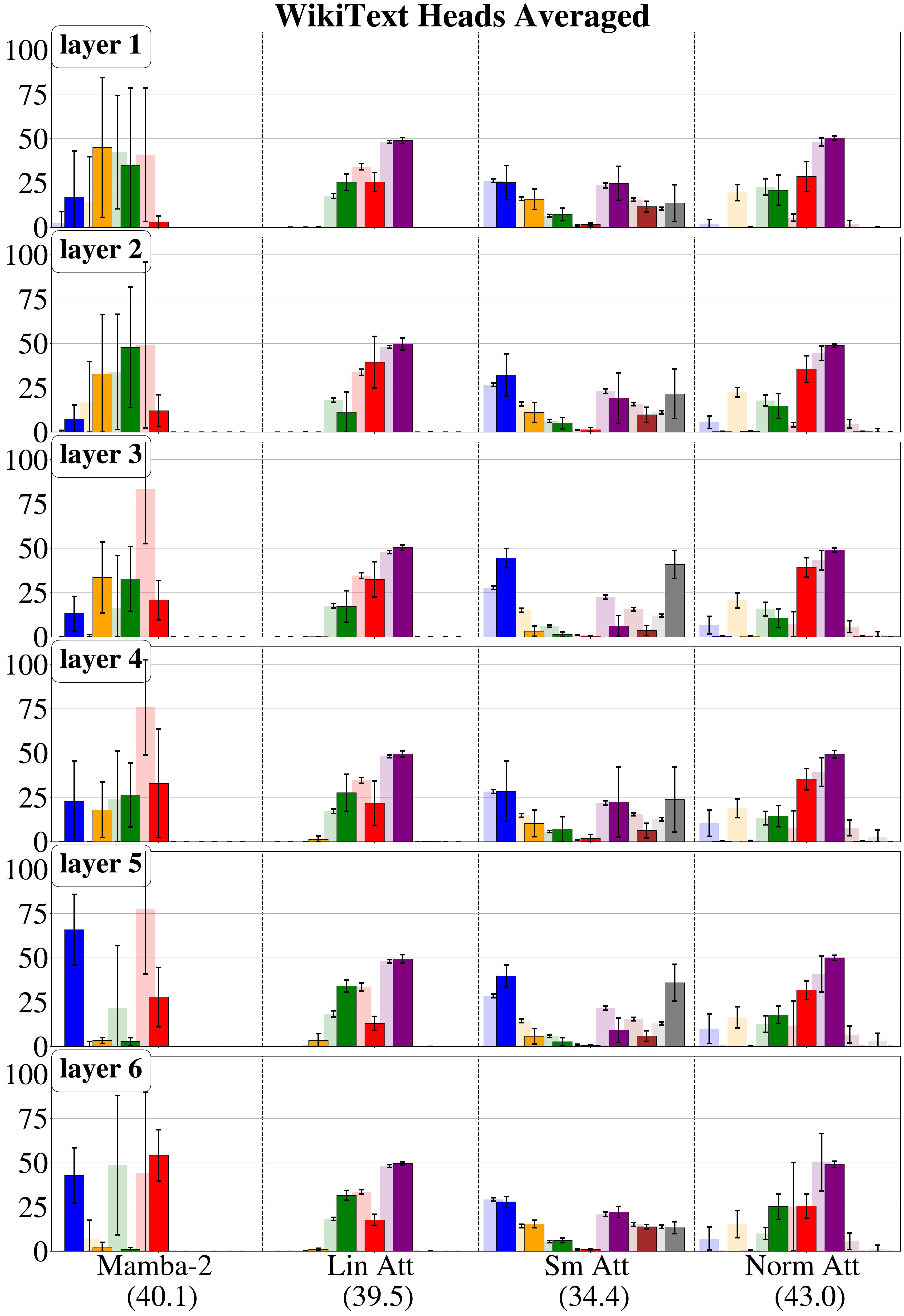}};

    \node (pic3) at (1.65, -6.18) {\includegraphics[width=11.2cm]
    {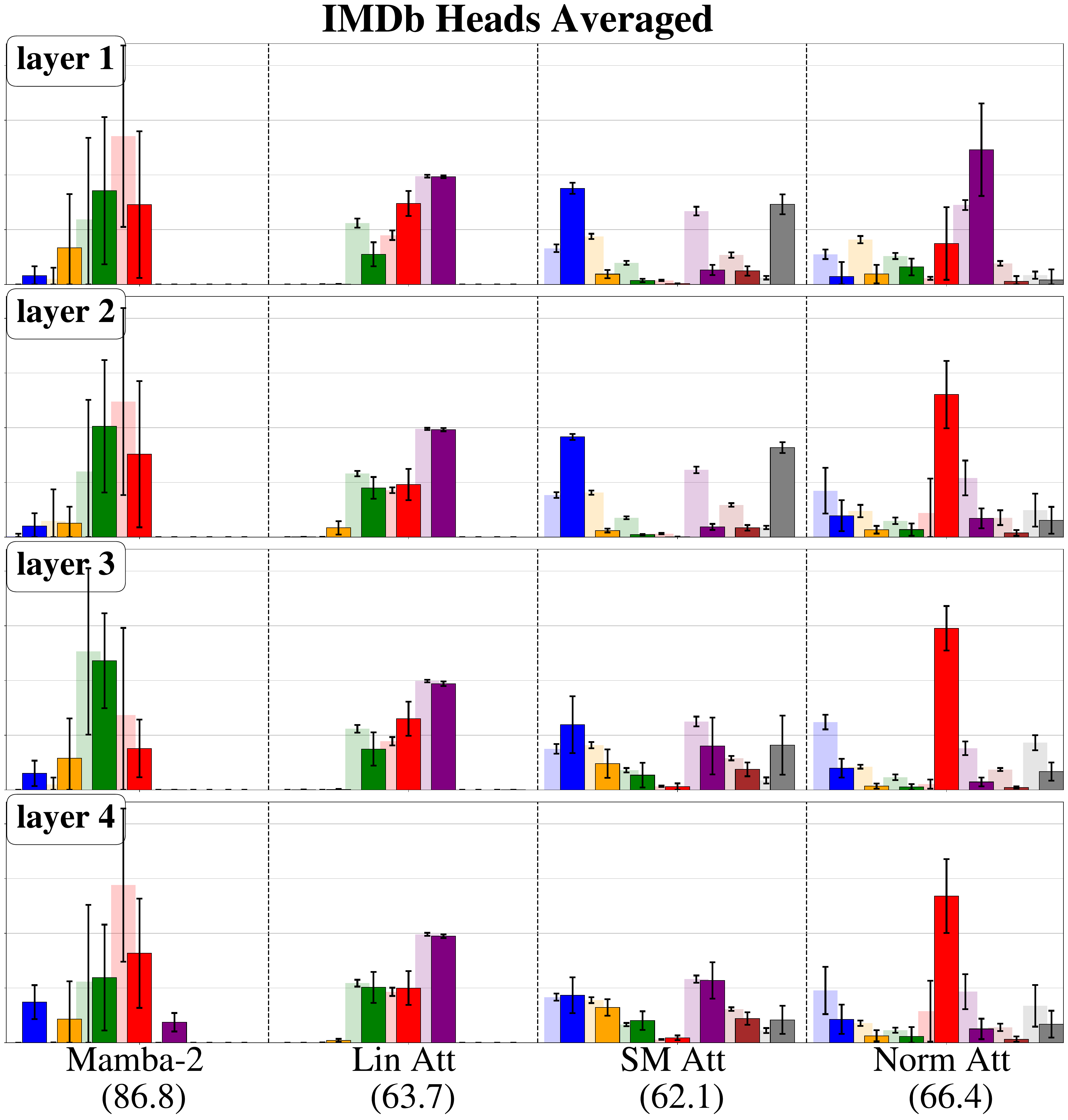}};
    
\end{tikzpicture}
\end{adjustbox}
\caption{ { Distributions of eigenvalues averaged across heads for all layers and tasks.}}
 \label{fig:avg_heads}
\end{figure}

\begin {figure}[H]
\centering
\begin{adjustbox}{max height=0.9\textheight, max width=0.82\textwidth}
\begin{tikzpicture}[scale=1.0]

    \node (pic1) at (-10.0, 10) {\includegraphics[width=16cm]
    {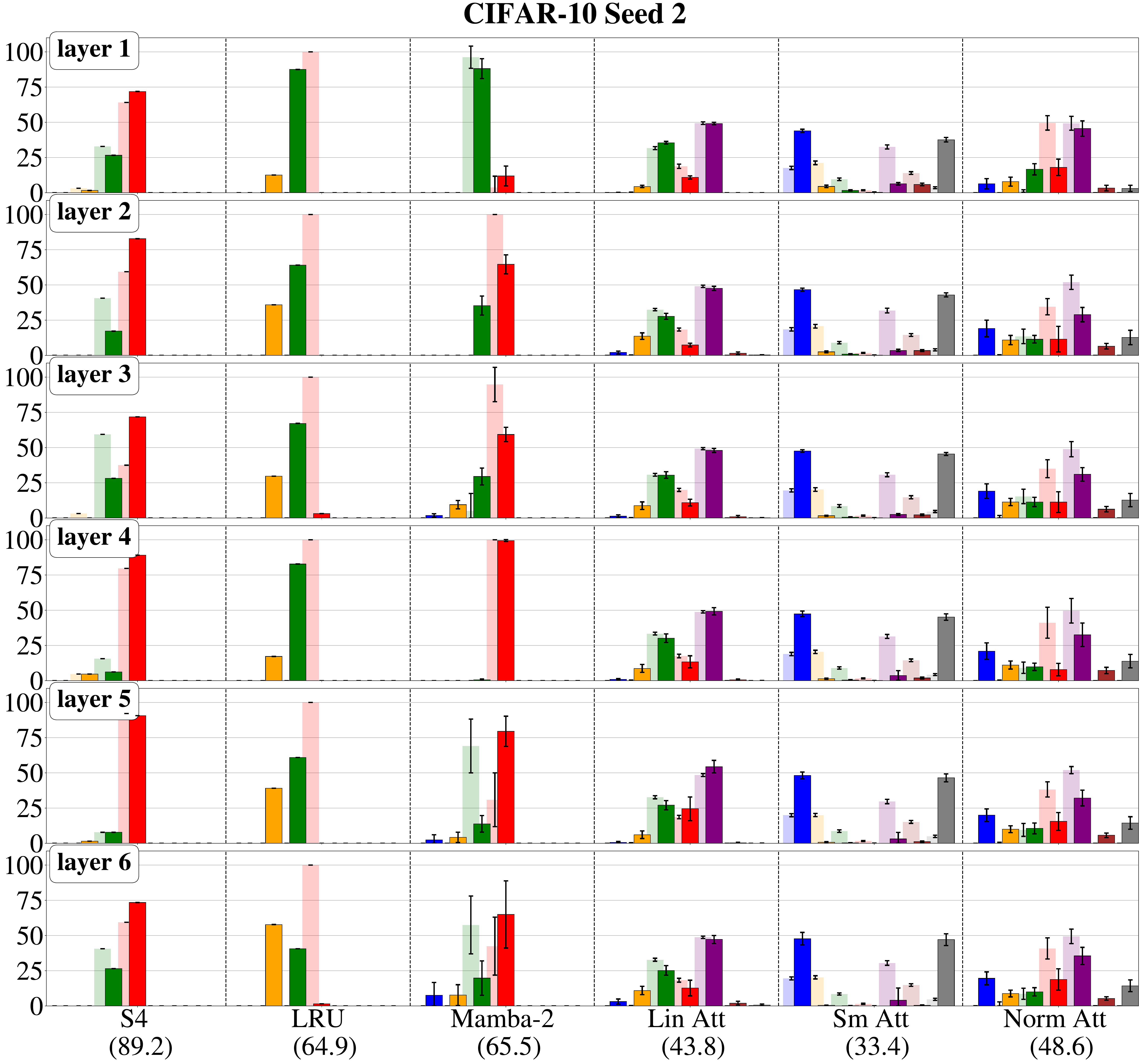}};
    
    \node (pic3) at (-10.0, -5) {\includegraphics[width=16cm]
    {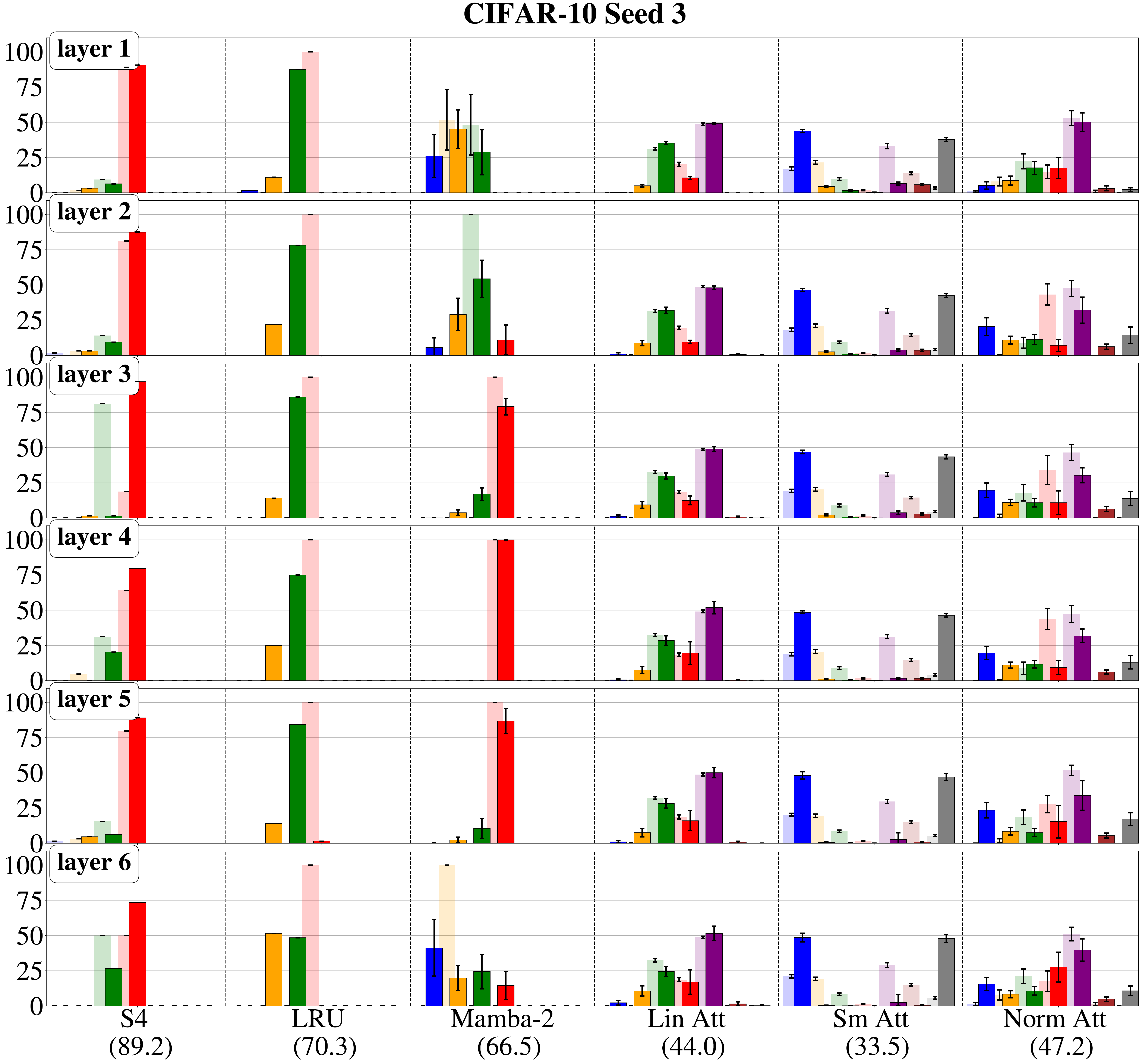}};
    
\end{tikzpicture}
\end{adjustbox}
\caption{Eigenvalue distributions across models, layers, and two out of three additional random seeds for \mbox{CIFAR-10}.}
\label{fig:cifar_seeds1}
\end{figure}

\begin {figure}[h]
\centering
\begin{adjustbox}{max height=0.95\textheight, max width=0.70\textwidth}
\begin{tikzpicture}[scale=1.0]

    \node (pic1) at (-10.0, 10) {\includegraphics[width=16cm]
    {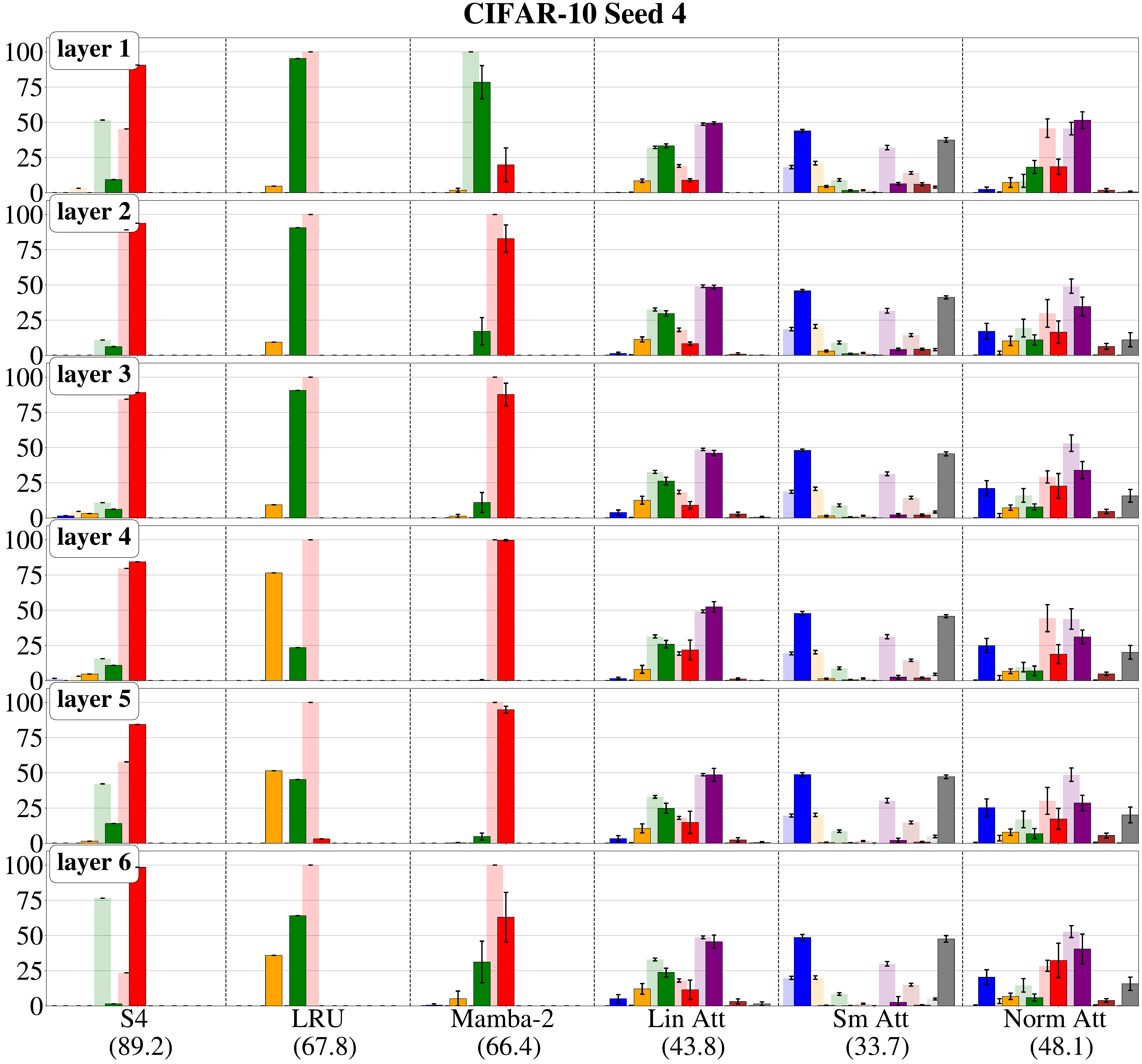}};
\end{tikzpicture}
\end{adjustbox}
\caption{Eigenvalue distributions across models, layers, and one remaining additional random seed for CIFAR-10.}
\label{fig:cifar_seeds2}
\end{figure}

\begin {figure}[H]
\centering
\begin{adjustbox}{max height=0.93\textheight, max width=0.82\textwidth}
\begin{tikzpicture}[scale=1.0]

    \node (pic1) at (-10.0, 10) {\includegraphics[width=16cm]
    {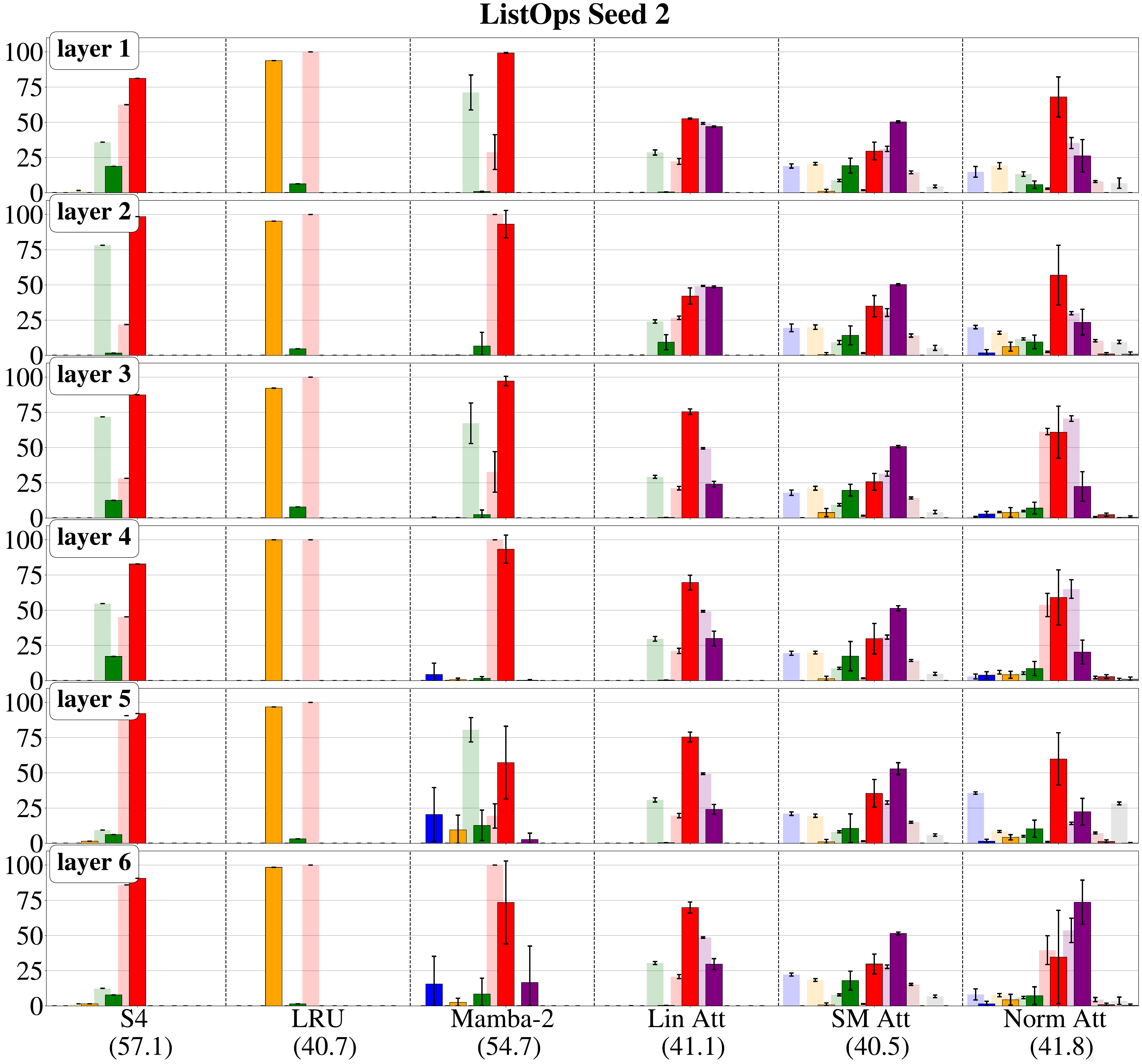}};
    
    \node (pic3) at (-10.0, -5) {\includegraphics[width=16cm]
    {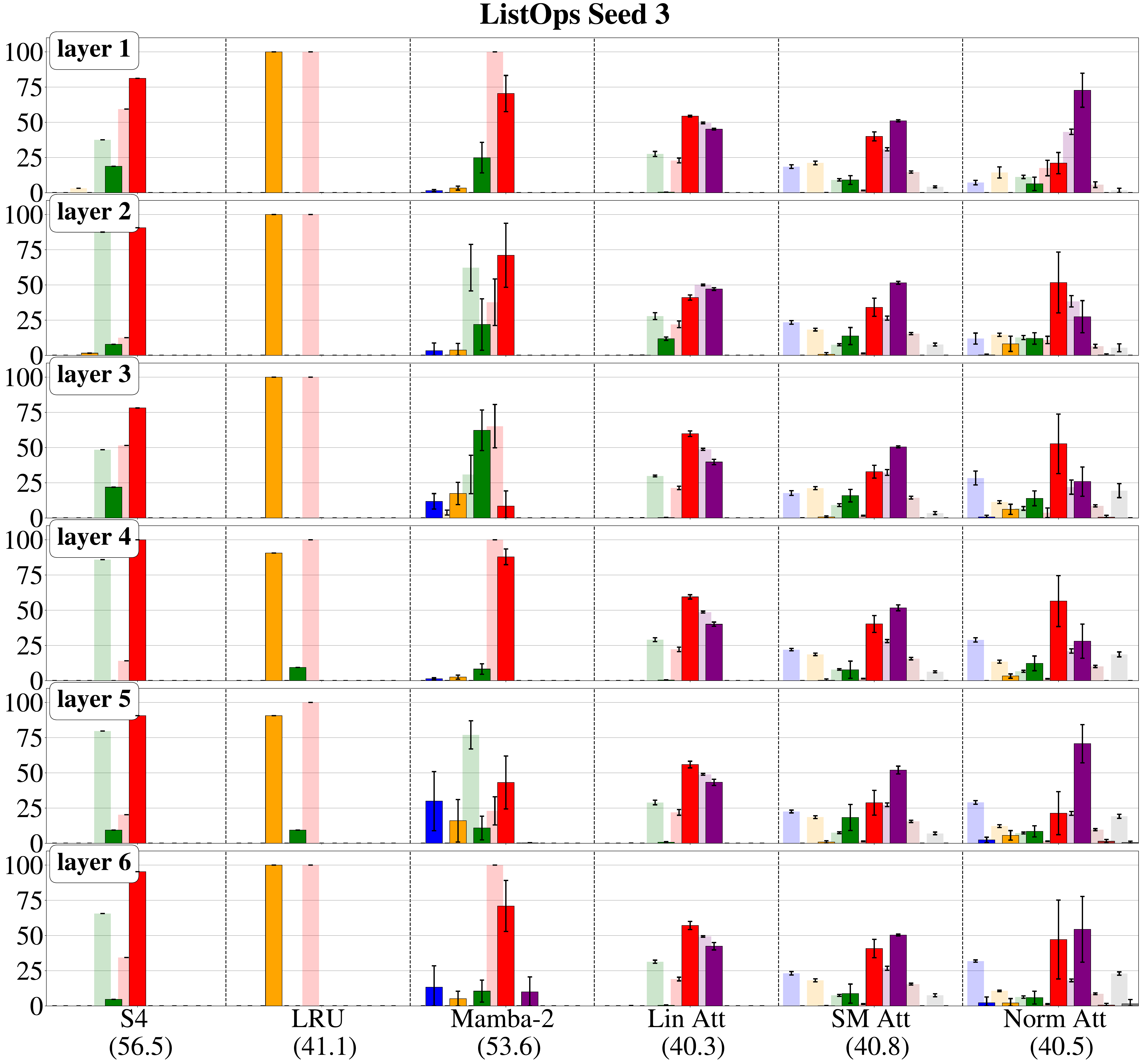}};
\end{tikzpicture}
\end{adjustbox}
\caption{Eigenvalue distributions across models, layers, and two out of three additional random seeds for ListOps.}
\label{fig:listops_seeds1}
\end{figure}

\begin {figure}[H]
\centering
\begin{adjustbox}{max height=0.95\textheight, max width=0.70\textwidth}
\begin{tikzpicture}[scale=1.0]

    \node (pic1) at (-10.0, 10) {\includegraphics[width=16cm]
    {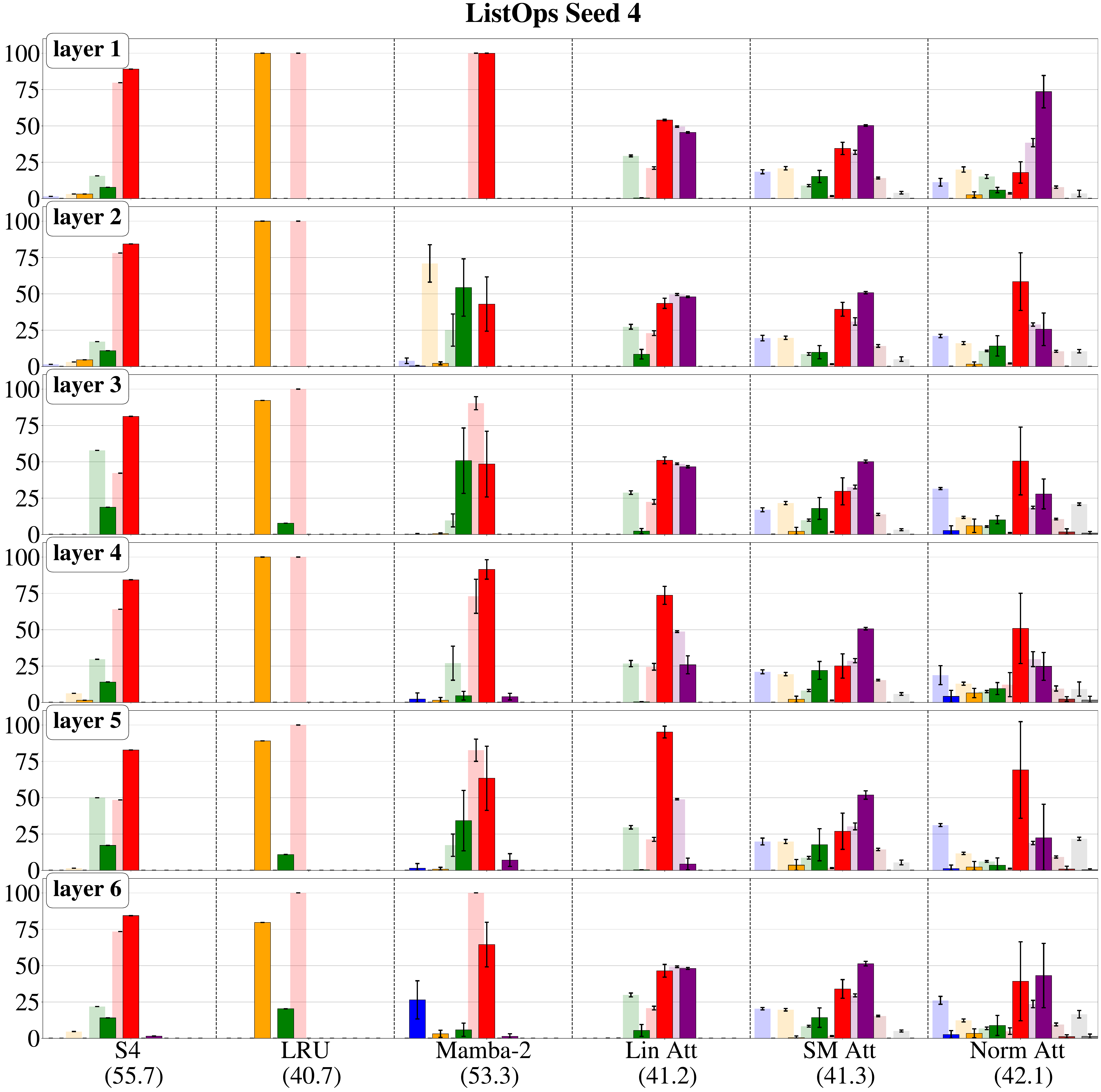}};
\end{tikzpicture}
\end{adjustbox}
\caption{Eigenvalue distributions across models, layers, and one remaining additional random seed for ListOps.}
\label{fig:listops_seeds2}
\end{figure}

\begin {figure}[H]
\centering
\begin{adjustbox}{max height=0.97\textheight, max width=0.82\textwidth}
\begin{tikzpicture}[scale=1.0]

    \node (pic1) at (-10.0, 10.5) {\includegraphics[width=14.2cm]
    {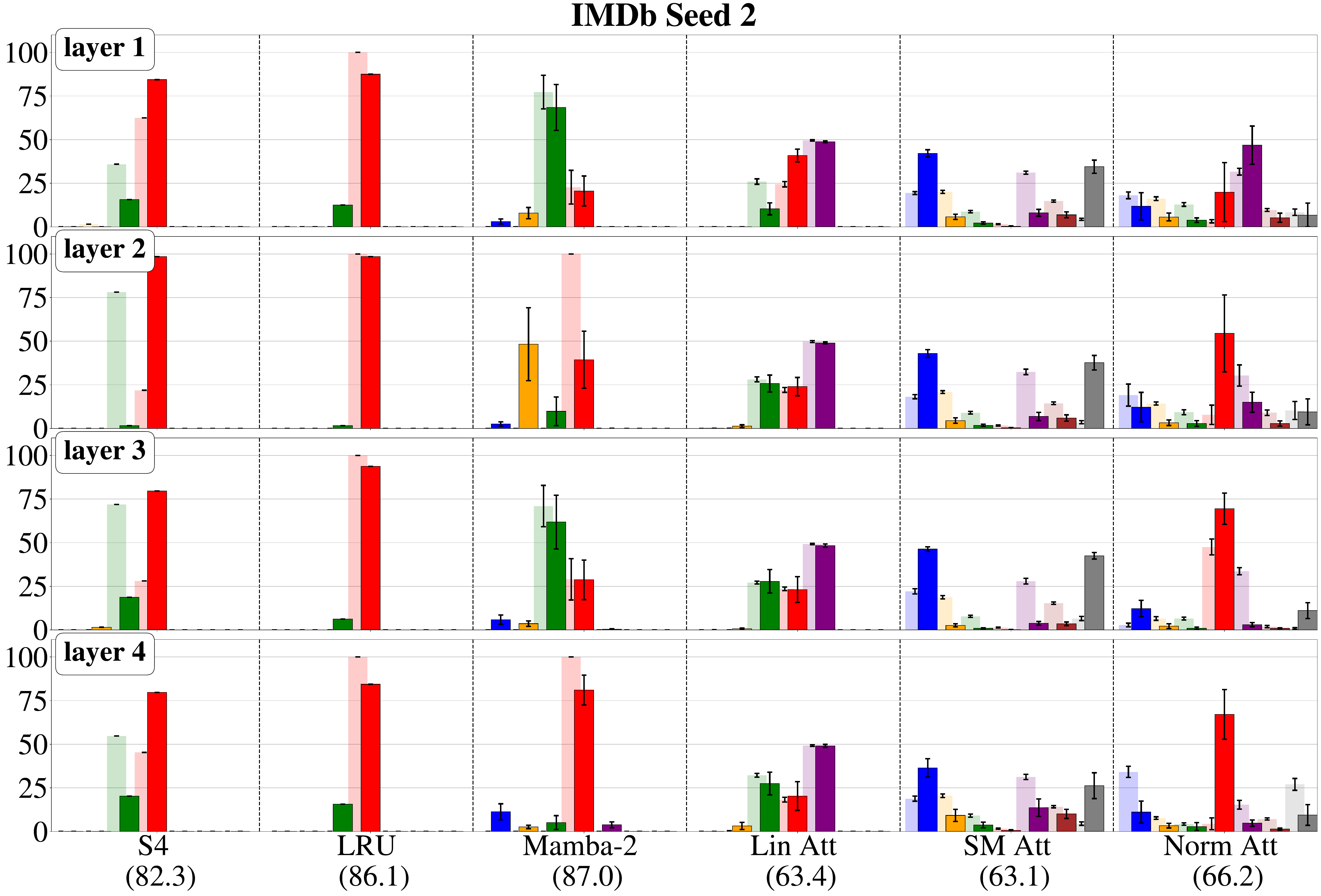}};
    
    \node (pic3) at (-10.0, 0.8) {\includegraphics[width=14.2cm]
    {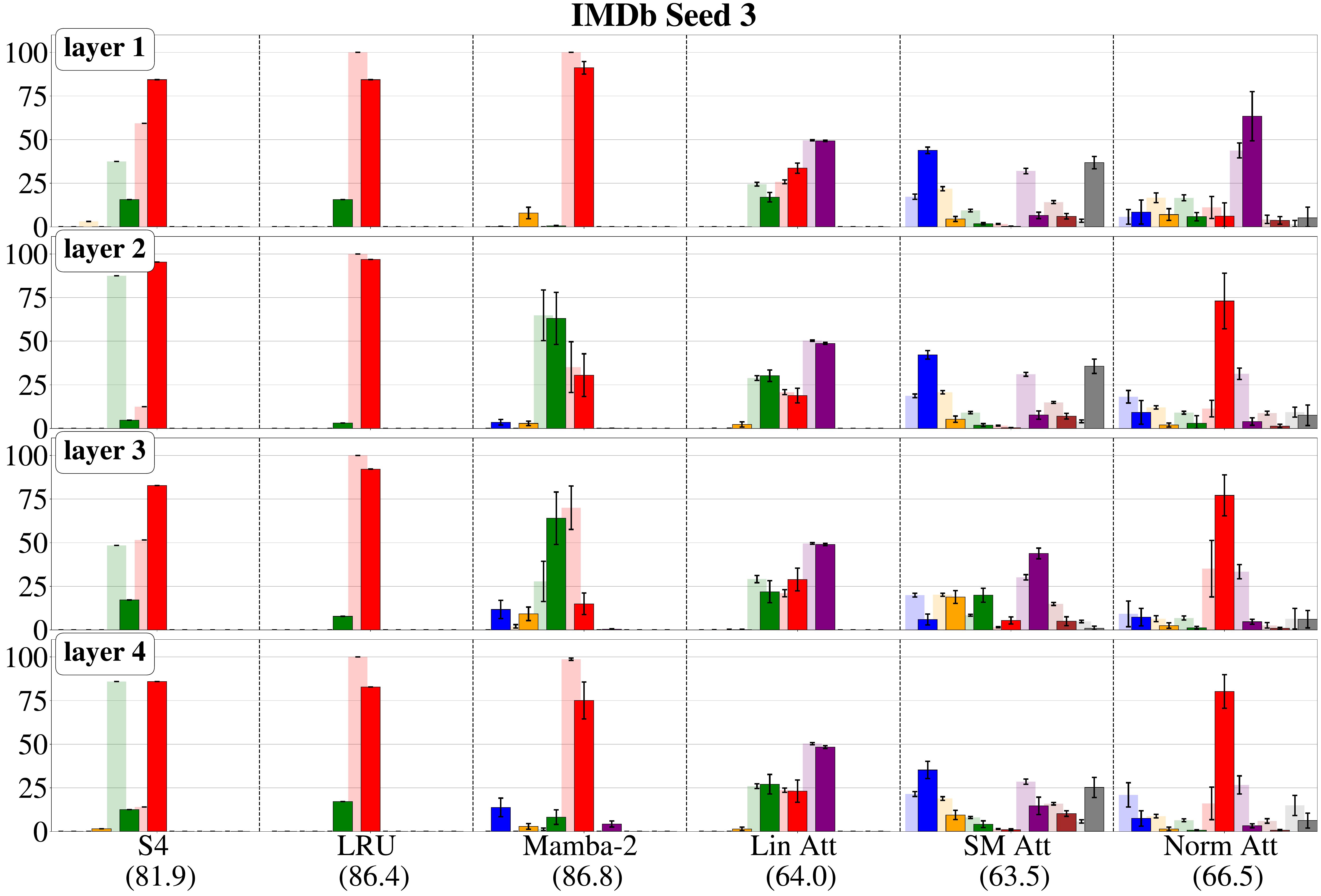}};

    \node (pic4) at (-10.0, -8.9) {\includegraphics[width=14.2cm]
    {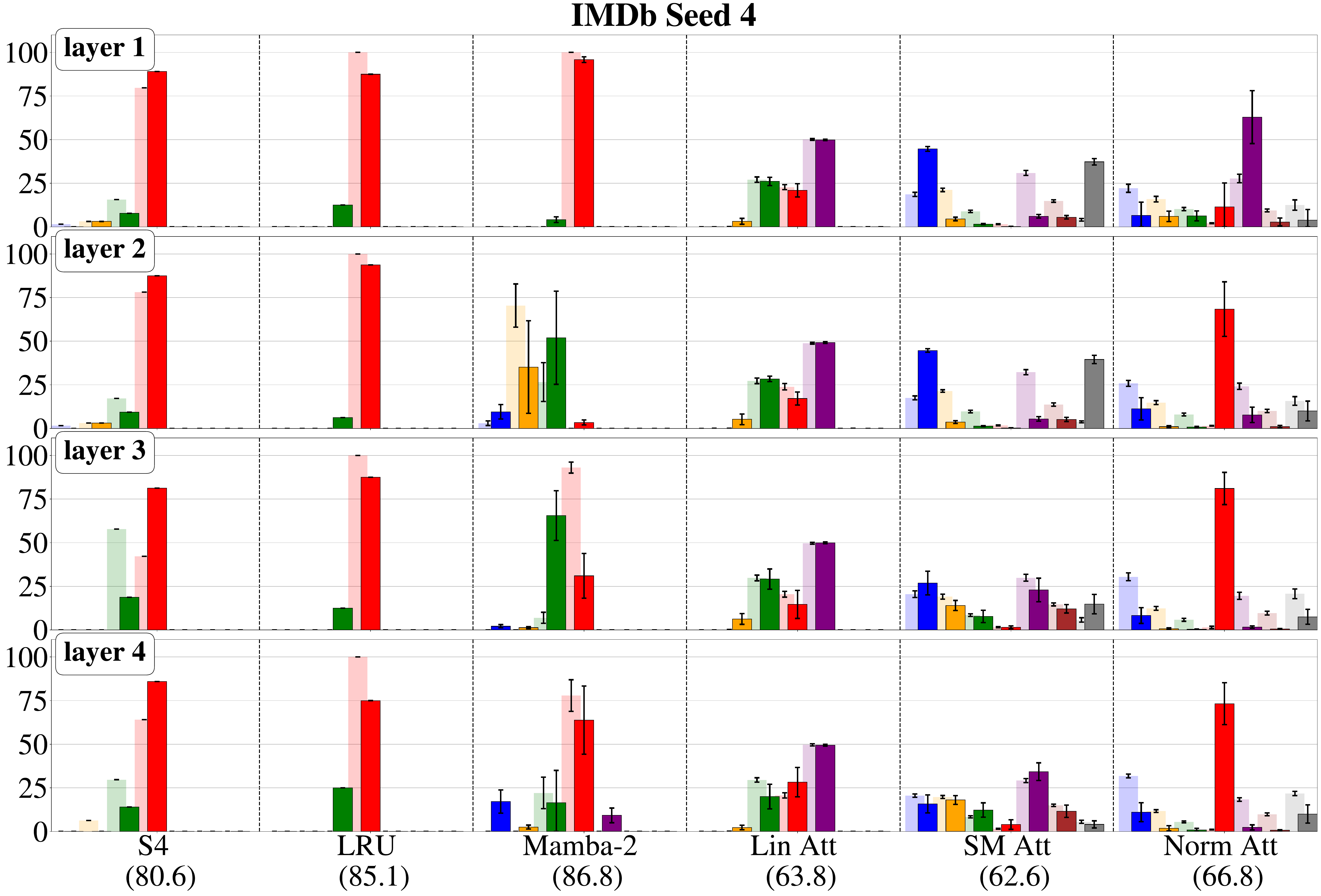}};
    
\end{tikzpicture}
\end{adjustbox}
\caption{Eigenvalue distributions across models, layers, and three additional random seeds for IMDb.}
\label{fig:imdb_seeds}
\end{figure}

\begin {figure}[H]
\centering
\begin{adjustbox}{max height=\textheight, max width=0.72\textwidth}
\begin{tikzpicture}[scale=1.0]

    \node (pic1) at (-10.0, 6.5) {\includegraphics[width=16cm]
    {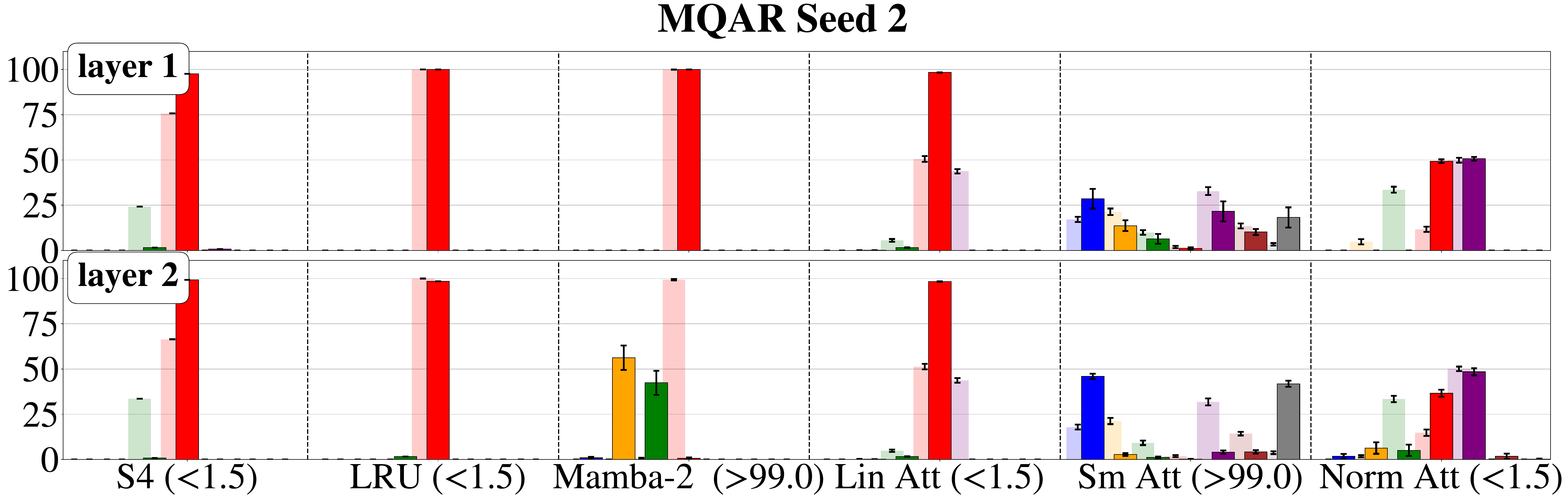}};
    
    \node (pic3) at (-10.0, 0) {\includegraphics[width=16cm]
    {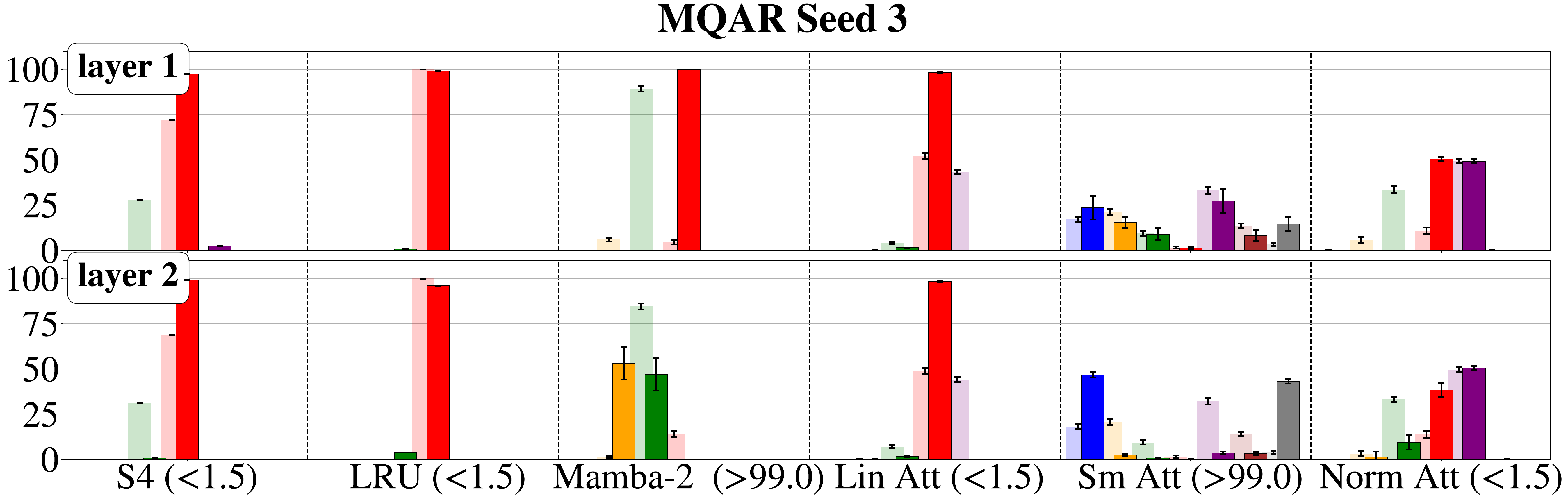}};

    \node (pic4) at (-10.0, -6.5) {\includegraphics[width=16cm]
    {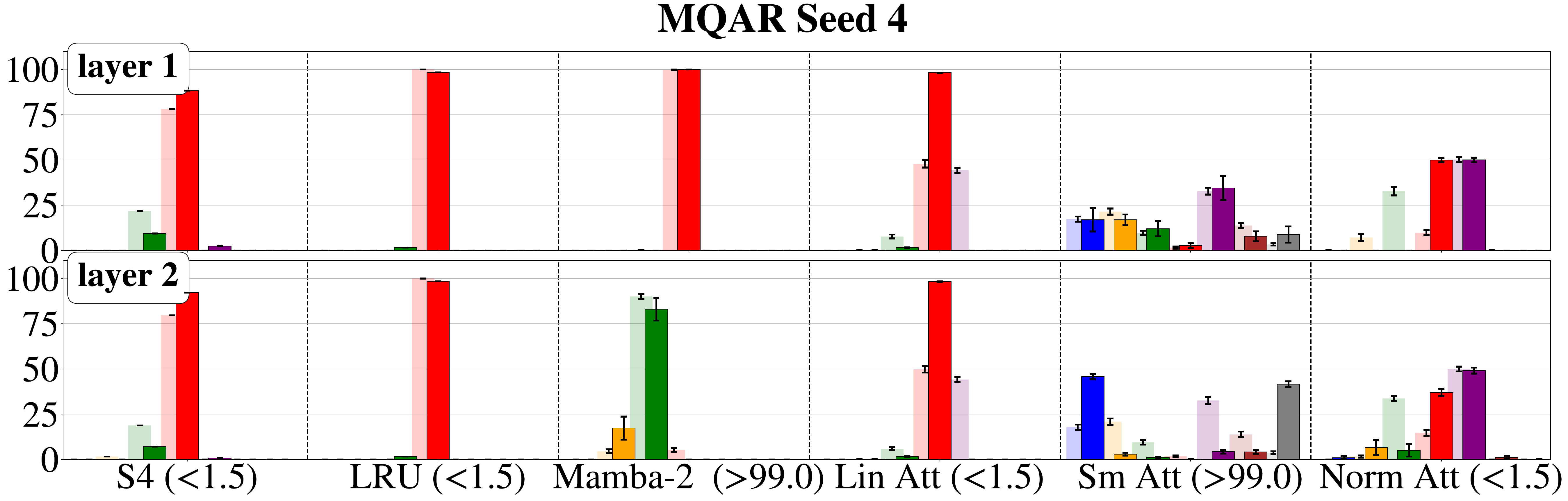}};
    
\end{tikzpicture}
\end{adjustbox}
\caption{Eigenvalue distributions across models, layers, and three additional random seeds for MQAR.}
\label{fig:mqar_seeds}
\end{figure}

\begin {figure}[H]
\centering
\begin{adjustbox}{max height=0.97\textheight, max width=0.82\textwidth}
\begin{tikzpicture}[scale=1.0]

    \node (pic1) at (-10.0, 10) {\includegraphics[width=16cm]
    {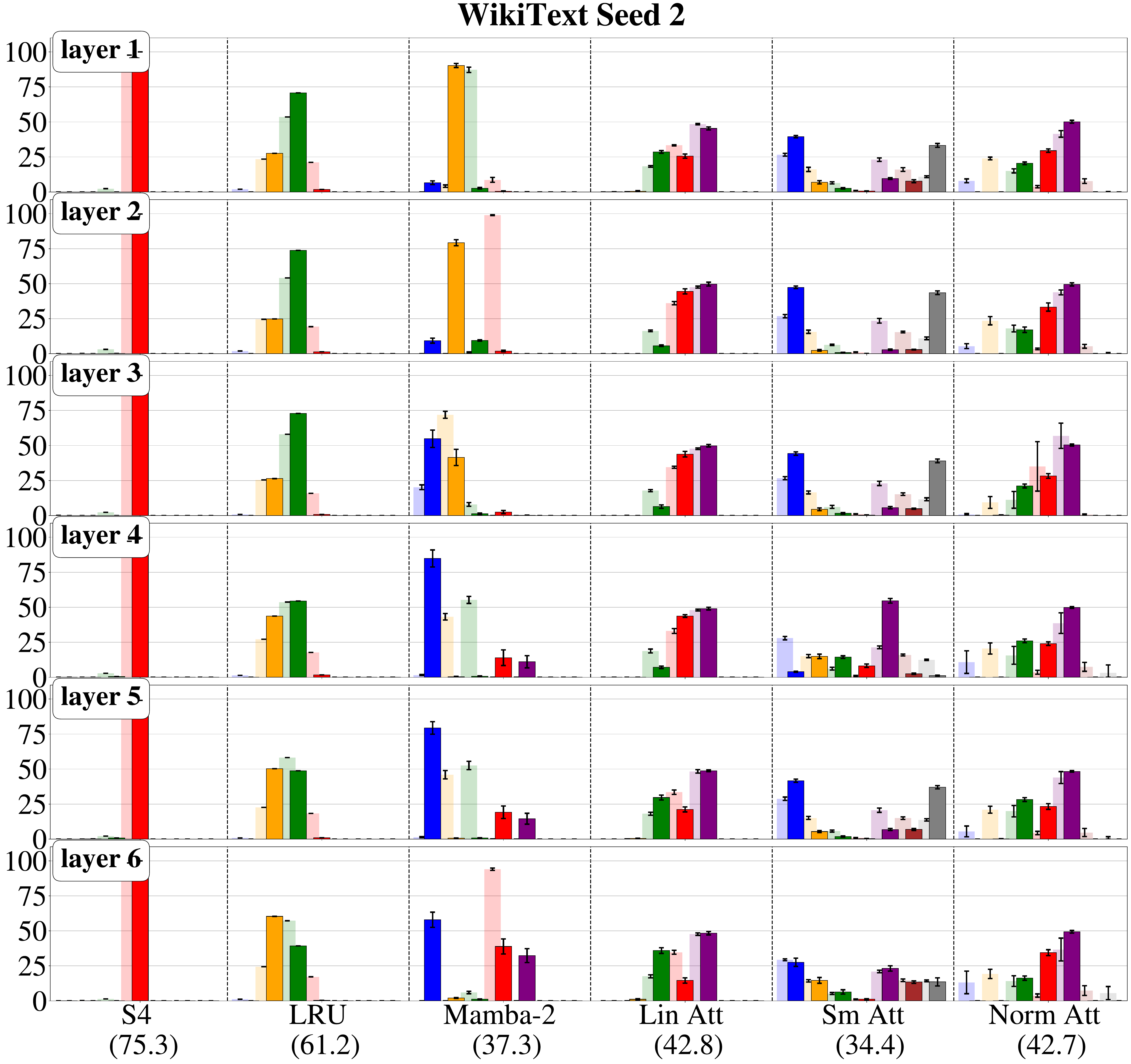}};
    
    \node (pic3) at (-10.0, -5) {\includegraphics[width=16cm]
    {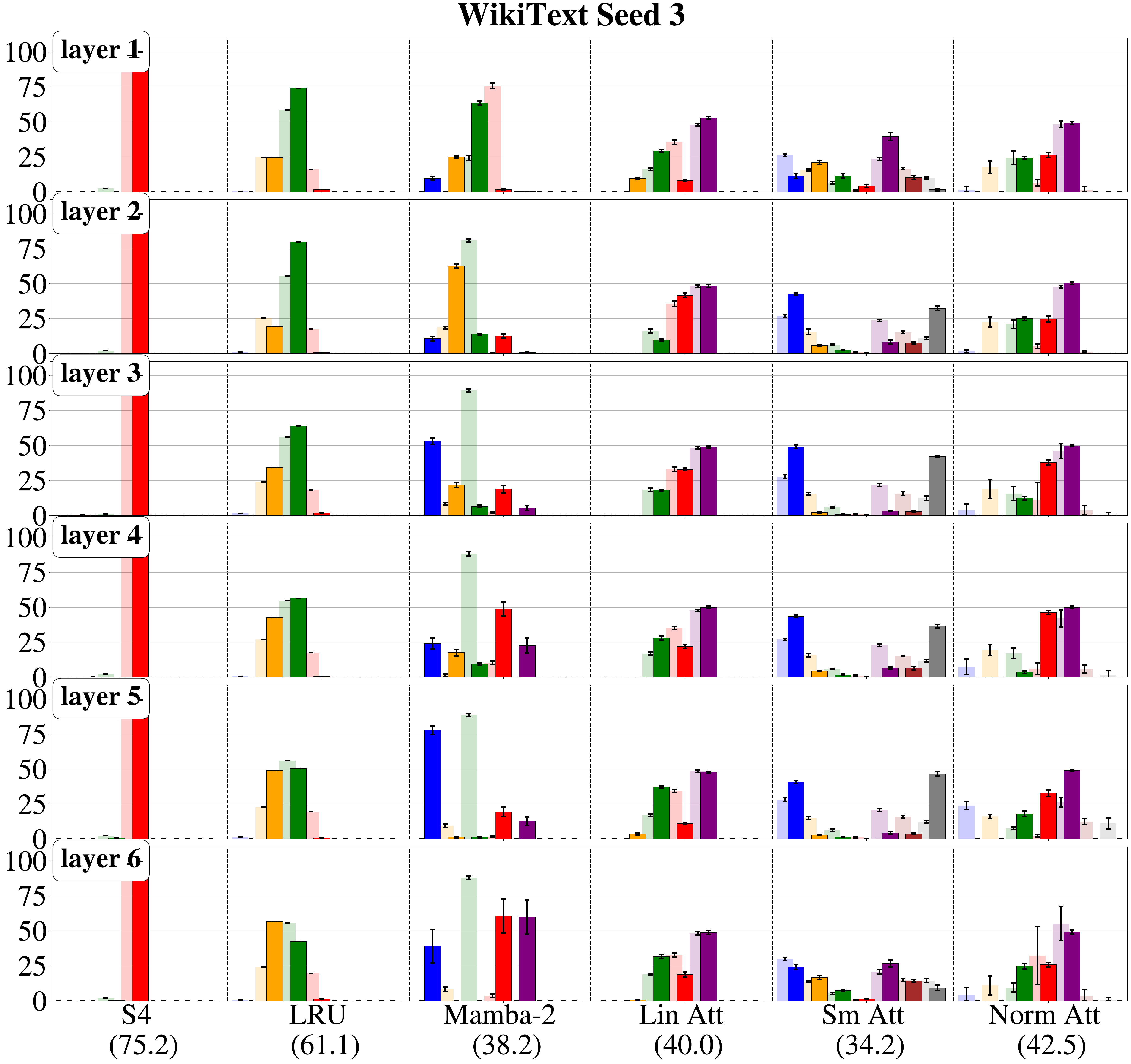}};
    
\end{tikzpicture}
\end{adjustbox}
\caption{Eigenvalue distributions across models, layers, and two out of three additional random seeds for \mbox{WikiText}.}
\label{fig:wiki_seeds1}
\end{figure}

\begin {figure}[H]
\centering
\begin{adjustbox}{max height=0.95\textheight, max width=0.70\textwidth}
\begin{tikzpicture}[scale=1.0]
    
    \node (pic1) at (-10.0, 10) {\includegraphics[width=16cm]
    {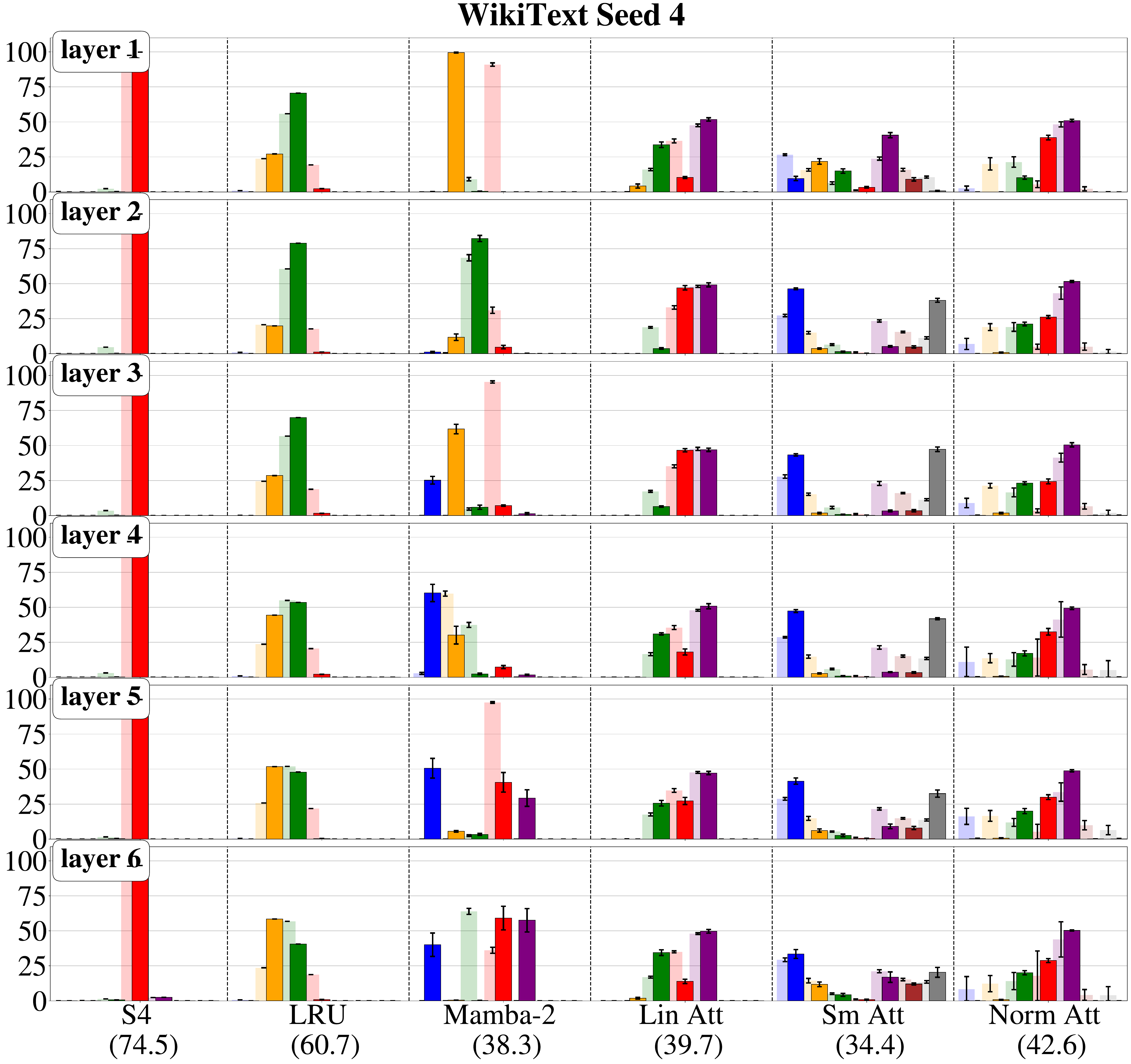}};
\end{tikzpicture}
\end{adjustbox}
\caption{Eigenvalue distributions across models, layers, and one remaining additional random seed for WikiText.}
\label{fig:wiki_seeds2}
\end{figure}

\begin {figure}[H]
\centering
\begin{adjustbox}{max height=0.9\textheight, max width=0.98\textwidth}
\begin{tikzpicture}[scale=1.0]

    \node (pic1) at (-10.0, 8.5) {\includegraphics[width=10cm]
    {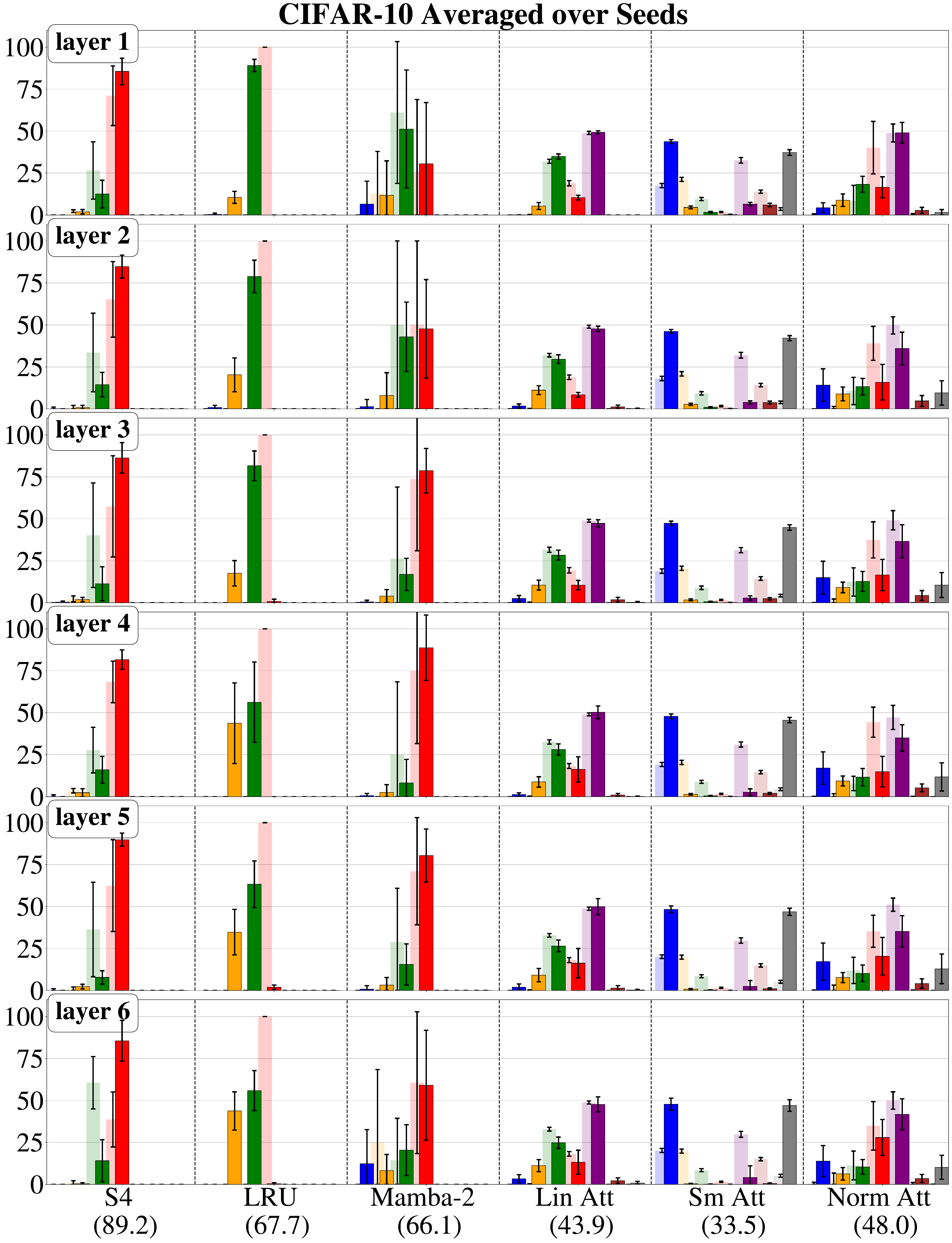}};

    \node (pic3) at (-10.0, -2.5) {\includegraphics[width=10cm]
    {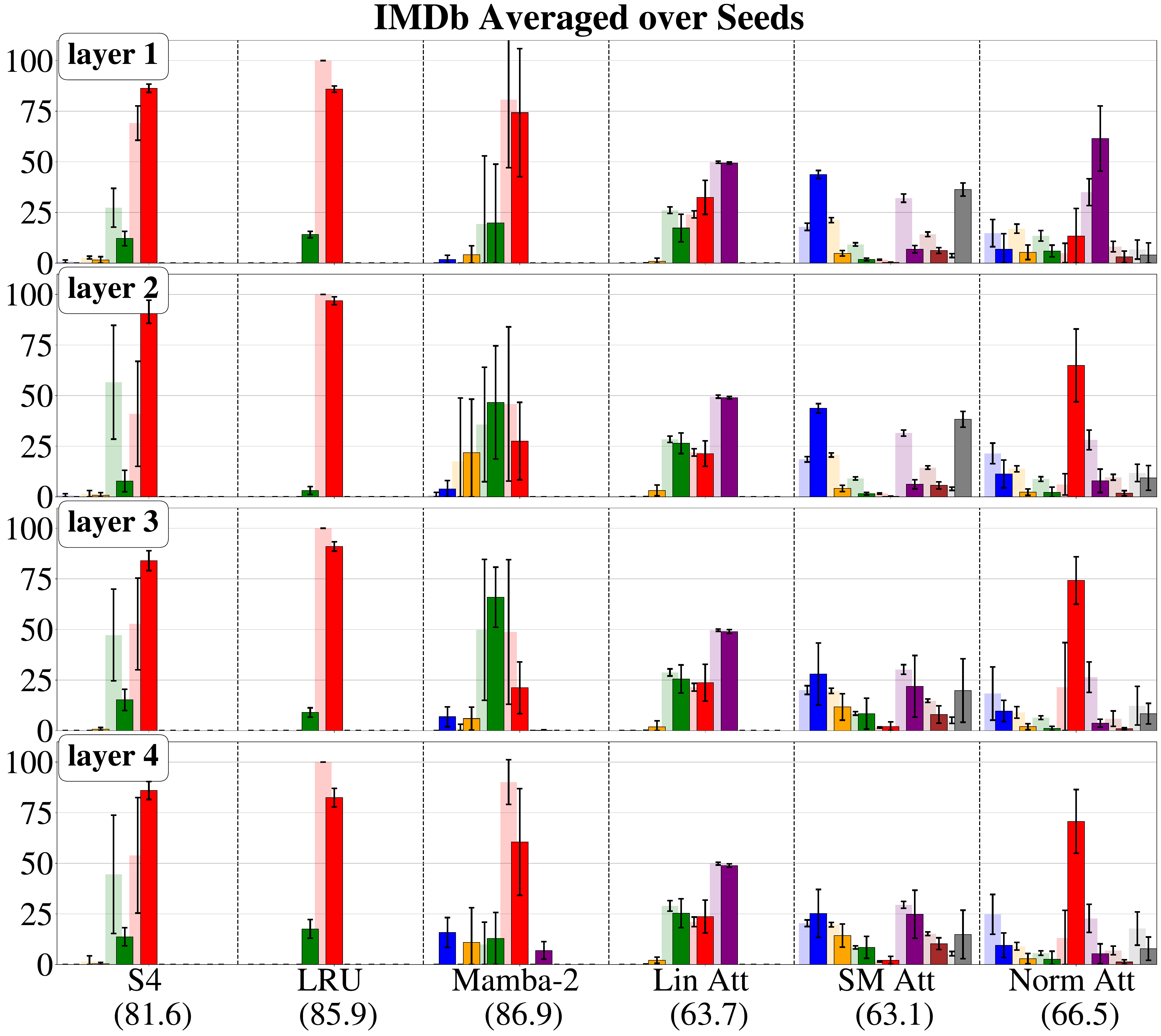}};

\end{tikzpicture}
\end{adjustbox}
\caption{ {Distributions of eigenvalues averaged across seeds for one of the heads on CIFAR and IMDb tasks.}}
 \label{fig:avg_seeds1}
\end{figure}

\begin {figure}[H]
\centering
\begin{adjustbox}{max height=0.95\textheight, max width=0.98\textwidth}
\begin{tikzpicture}[scale=1.0]

    \node (pic1) at (-10.0, 8.5) {\includegraphics[width=10cm]
    {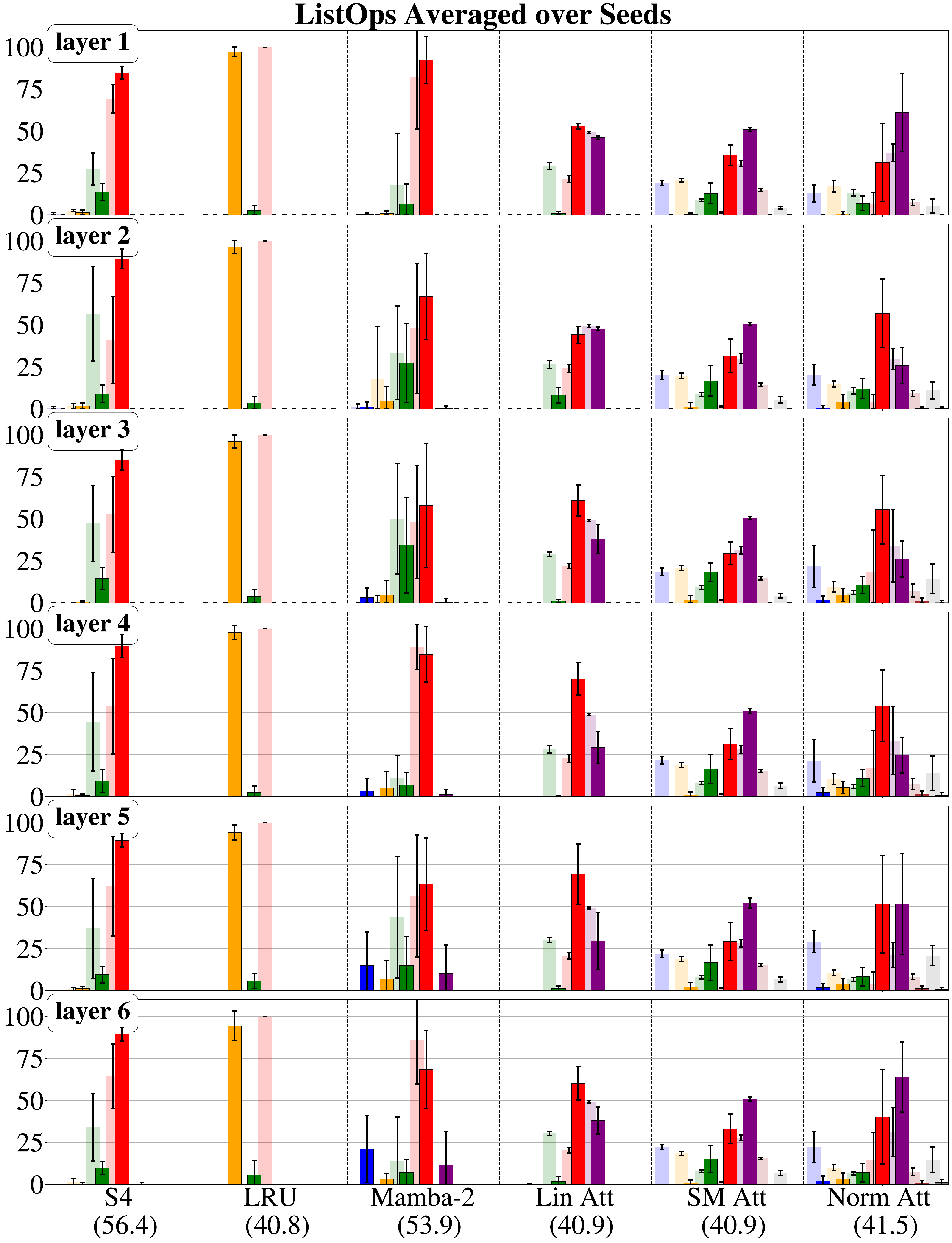}};

    \node (pic3) at (-10.0, -0.5) {\includegraphics[width=10cm]
    {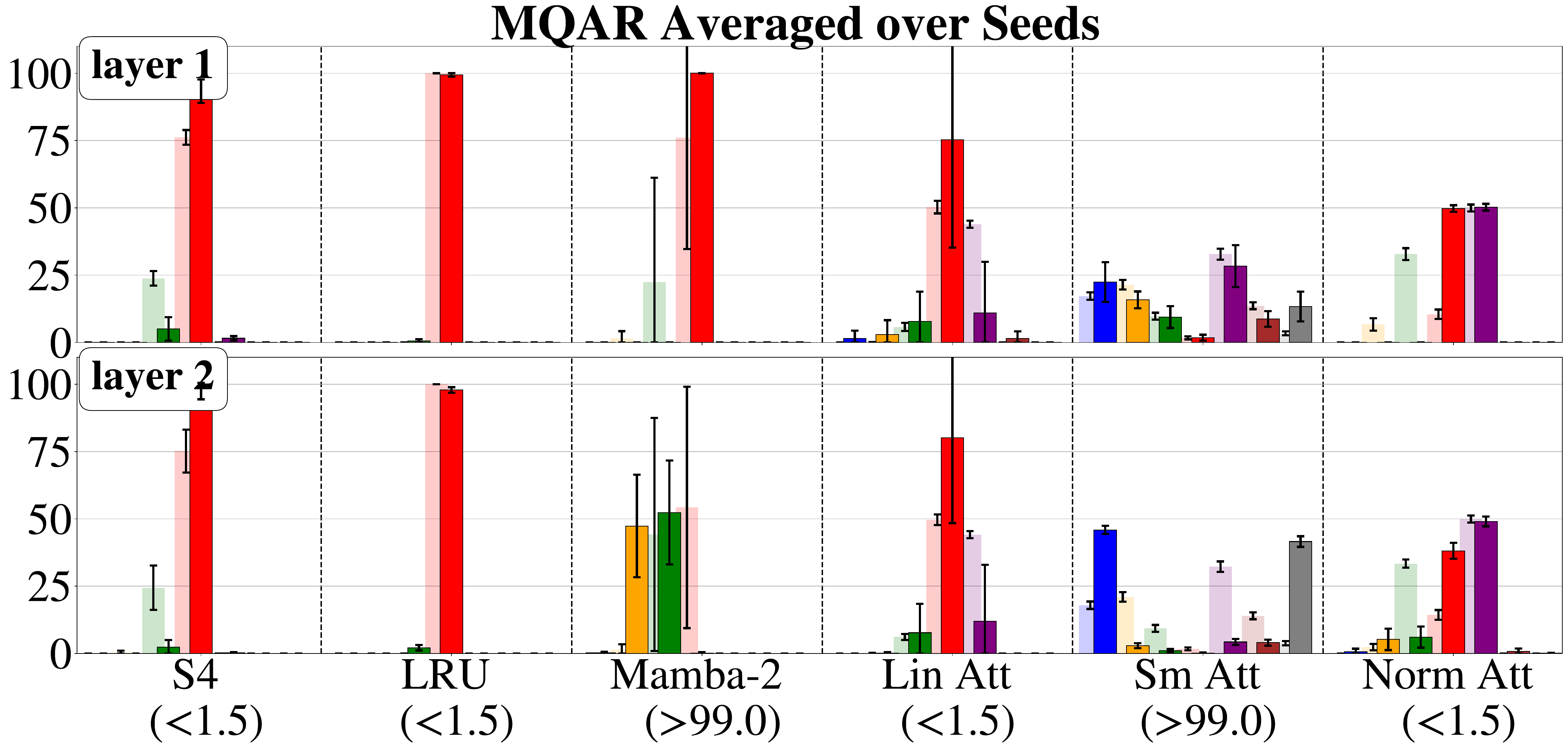}};

\end{tikzpicture}
\end{adjustbox}
\caption{ {Distributions of eigenvalues averaged across seeds for one of the heads on ListOps and MQAR tasks.}}
 \label{fig:avg_seeds2}
\end{figure}

\begin {figure}[H]
\centering
\begin{adjustbox}{max height=0.95\textheight, max width=0.98\textwidth}
\begin{tikzpicture}[scale=1.0];

    \node (pic1) at (-10.0, 8.5) {\includegraphics[width=10cm]
    {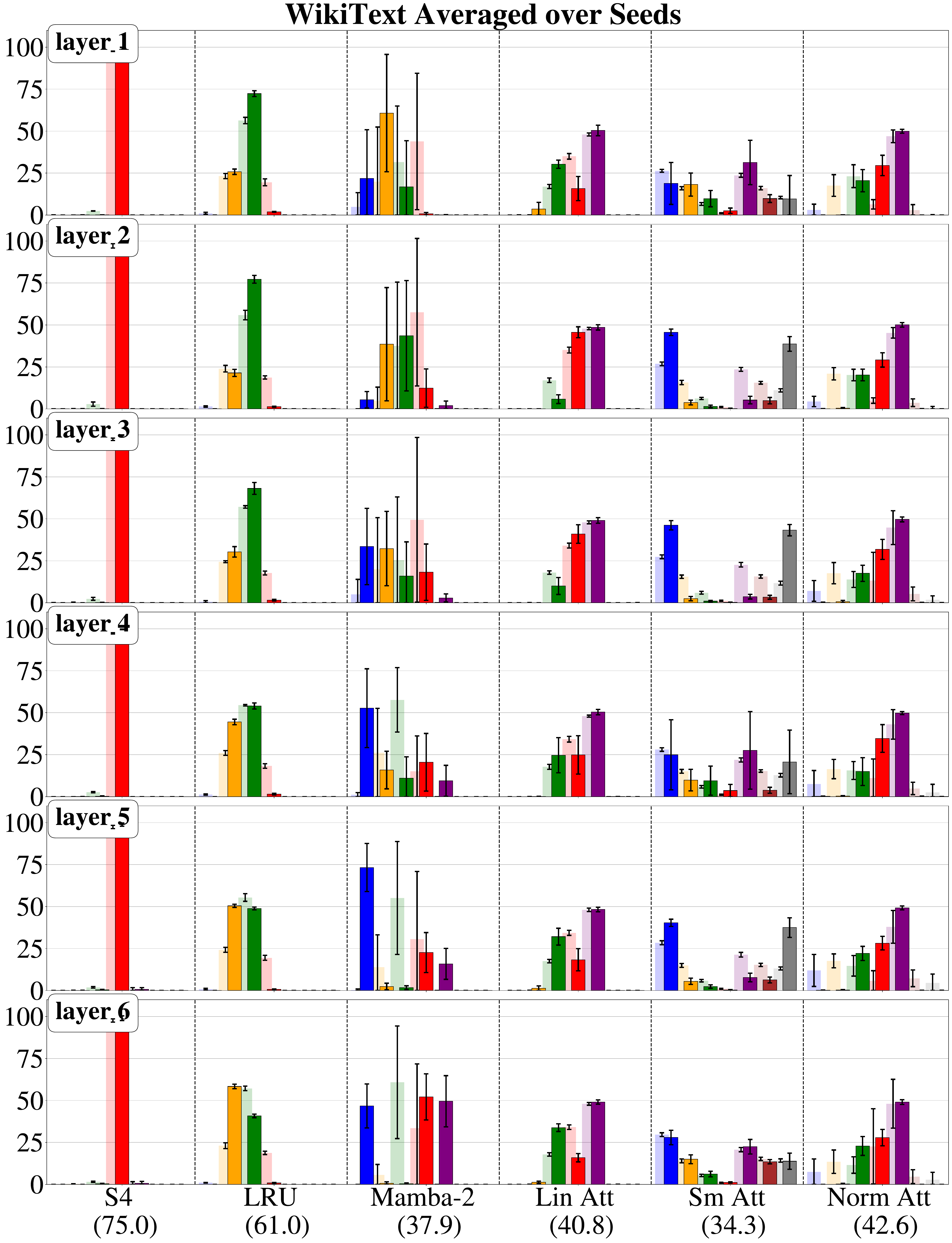}};
    
\end{tikzpicture}
\end{adjustbox}
\caption{ {Distributions of eigenvalues averaged across seeds for one of the heads on WikiText task.}}
 \label{fig:avg_seeds3}
\end{figure}

\begin {figure}[H]
\centering
\begin{adjustbox}{max height=0.9\textheight, max width=0.90\textwidth}
\begin{tikzpicture}[scale=1.0]

    \node (pic1) at (-10.0, 8.5) {\includegraphics[width=11.85cm]
    {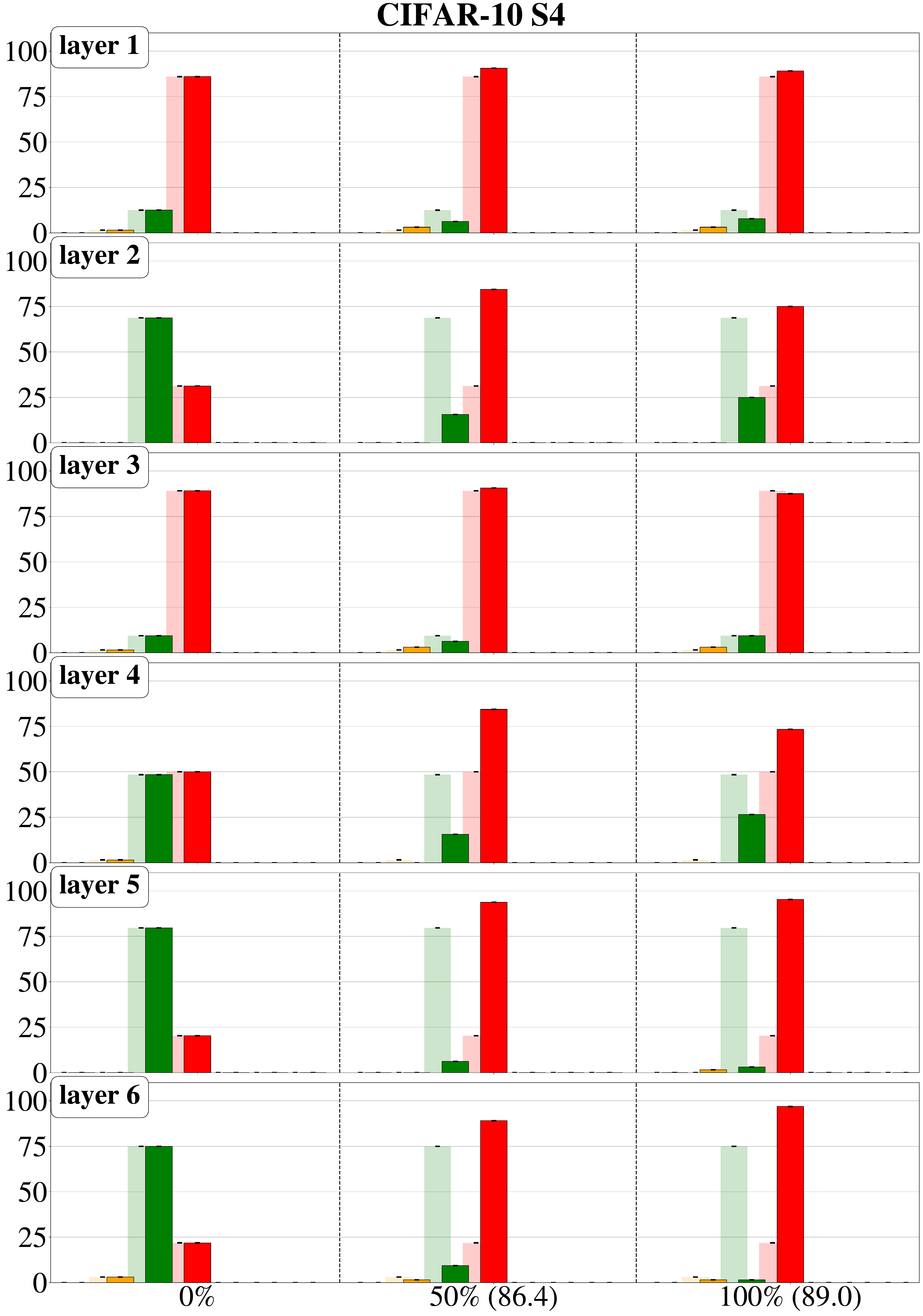}};

    \node (pic2) at (1.65, 8.5) {\includegraphics[width=11.27cm]
    {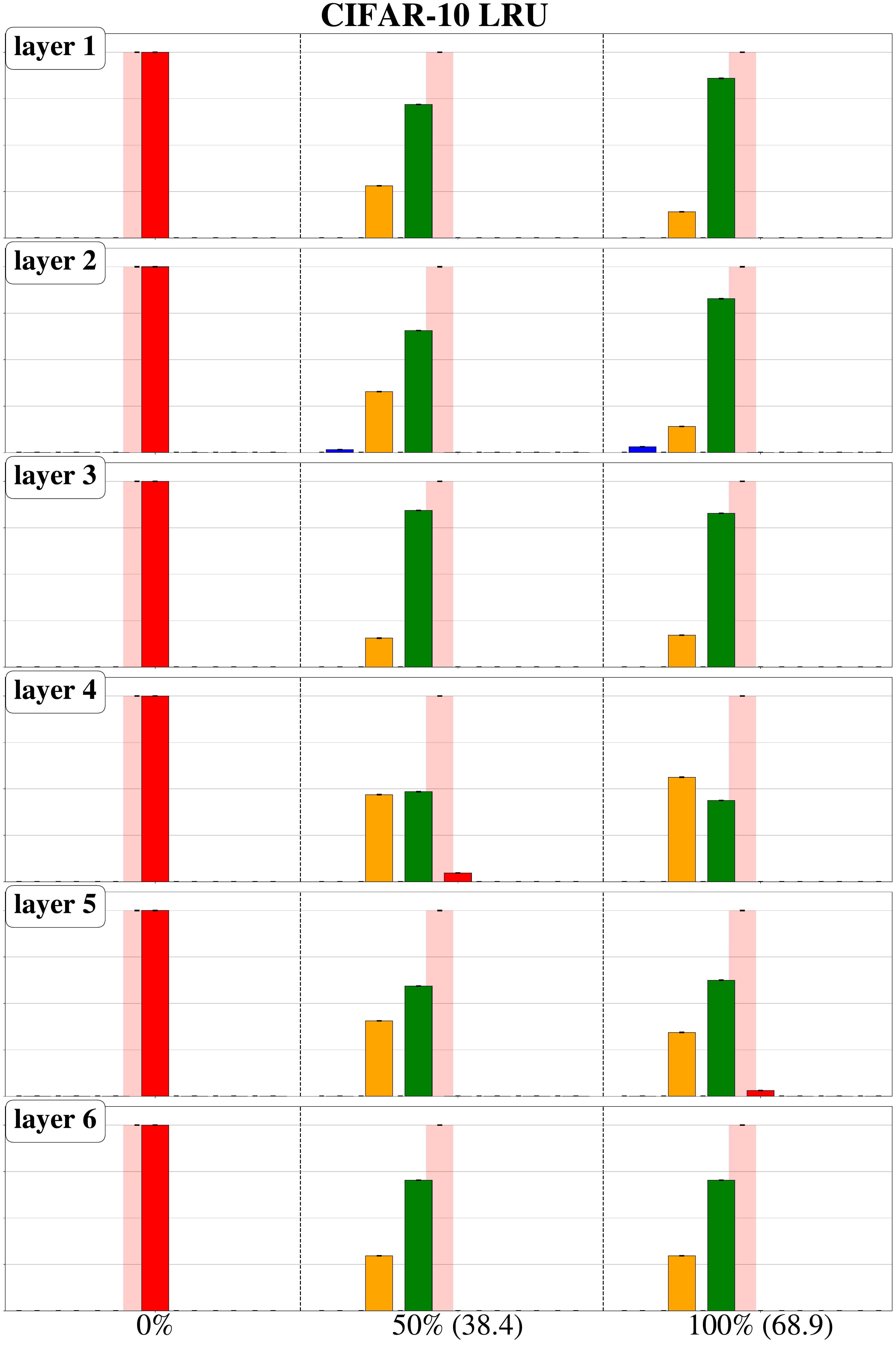}};

    \node (pic3) at (-4.5, -8.9) {\includegraphics[width=11.85cm]
    {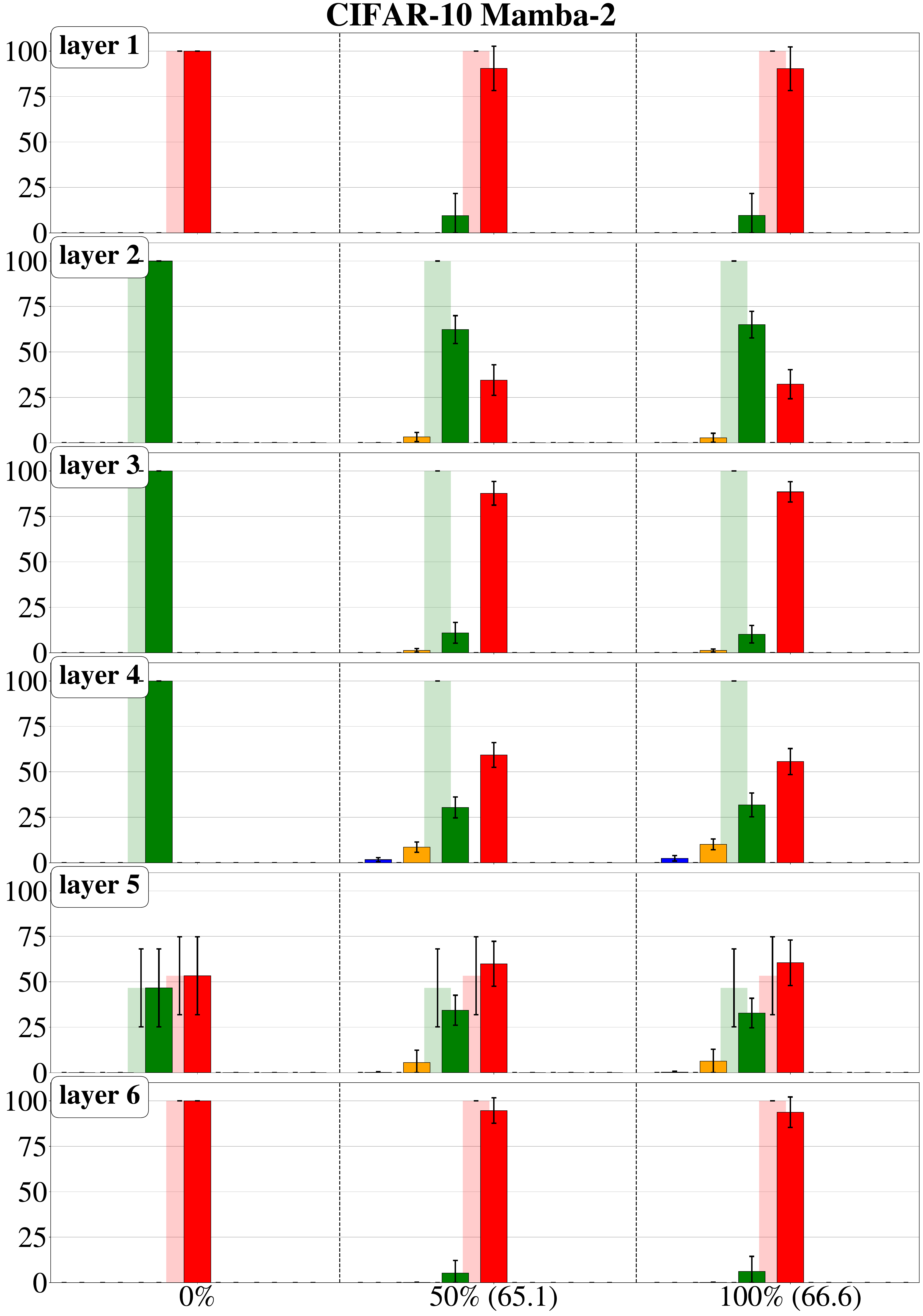}};
    
\end{tikzpicture}
\end{adjustbox}
\caption{ {Evolution of eigenvalues from initialization, over training half-time, to the final distribution for CIFAR-10.}}
 \label{fig:cifar_evolution}
\end{figure}

\begin {figure}[H]
\centering
\begin{adjustbox}{max height=\textheight, max width=0.98\textwidth}
\begin{tikzpicture}[scale=1.0]

    \node (pic1) at (-10.0, 8.5) {\includegraphics[width=11.85cm]
    {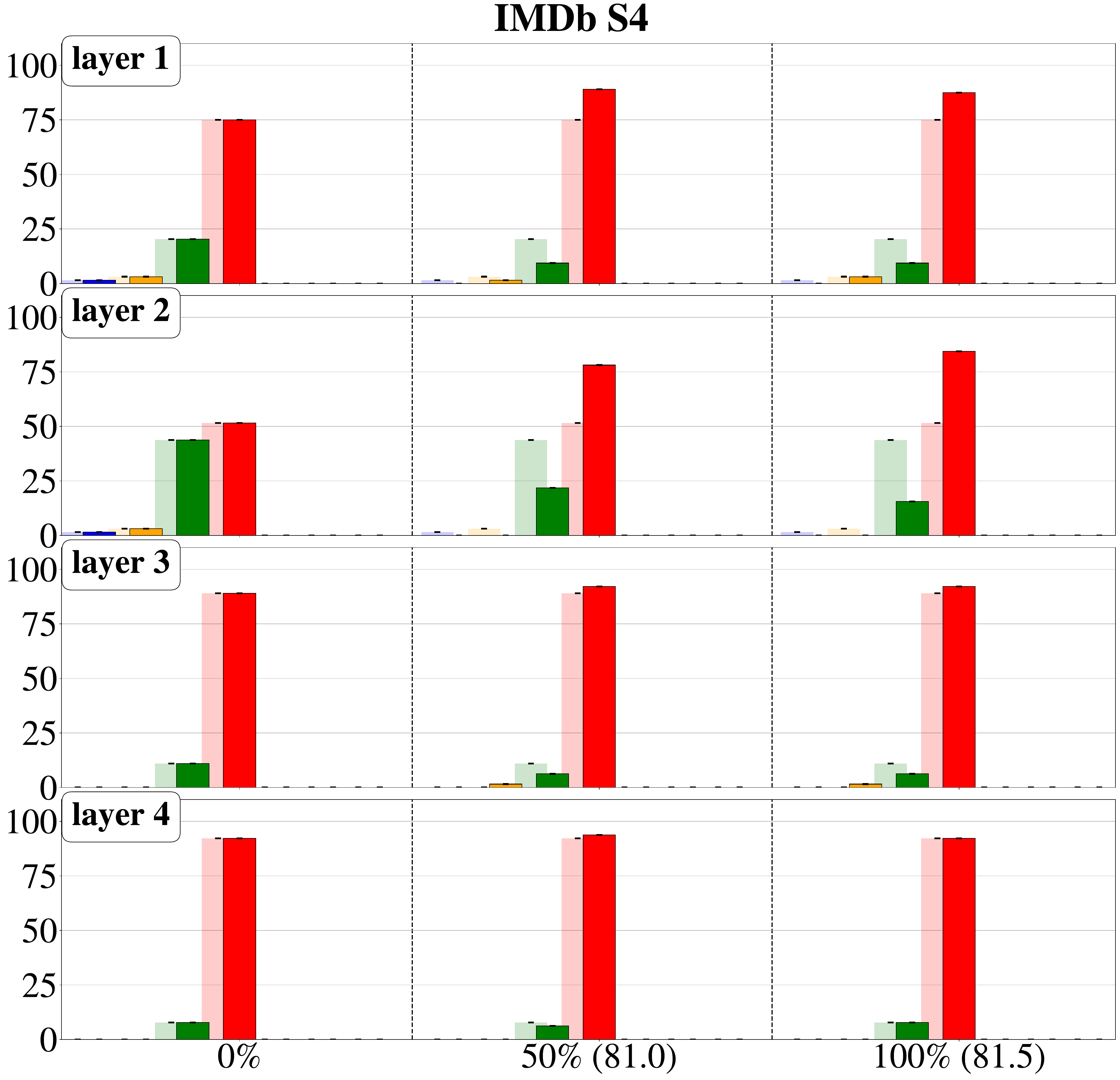}};

    \node (pic2) at (1.65, 8.5) {\includegraphics[width=11.30cm]
    {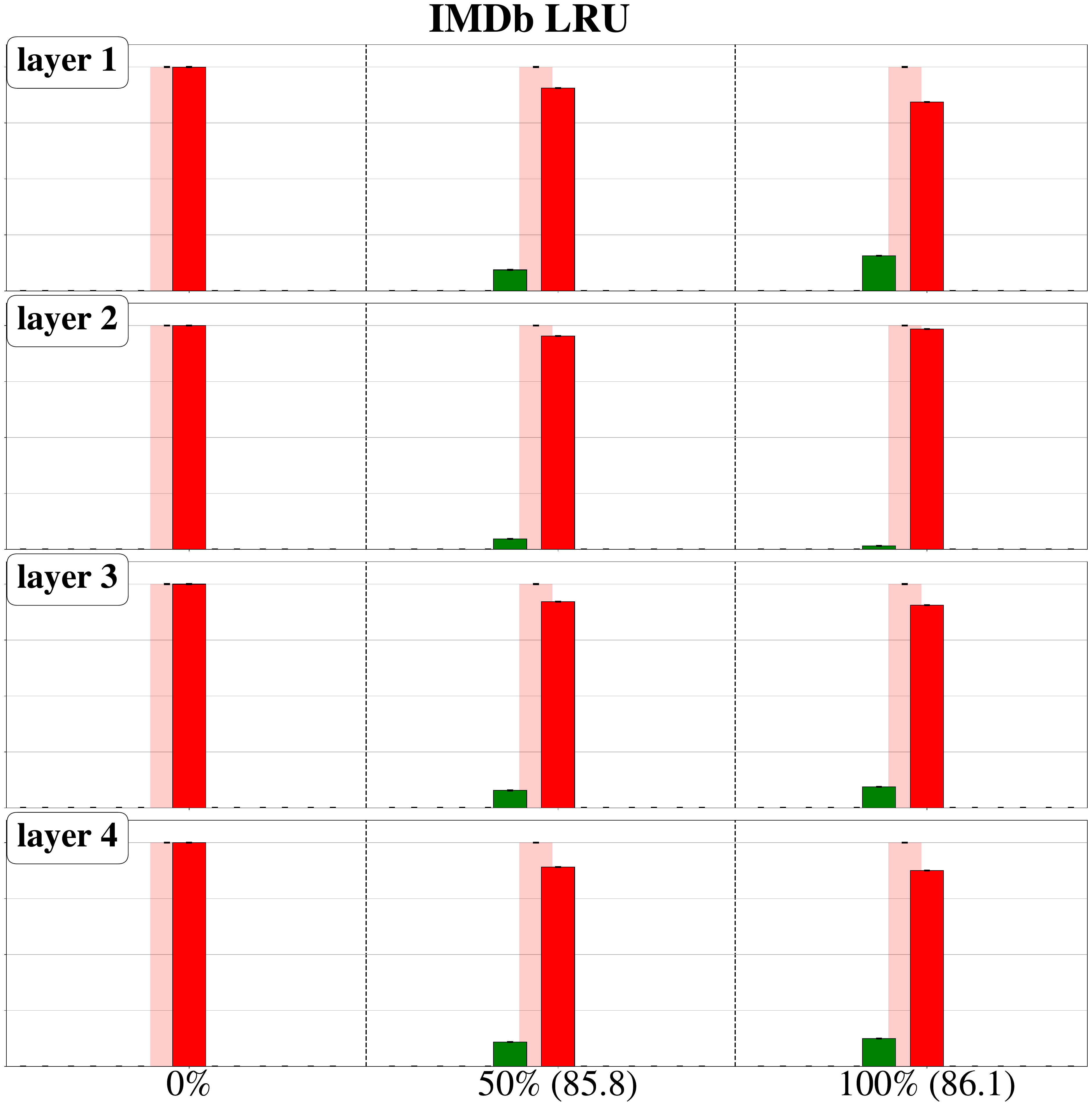}};

    \node (pic3) at (-4.5, -4.9) {\includegraphics[width=11.85cm]
    {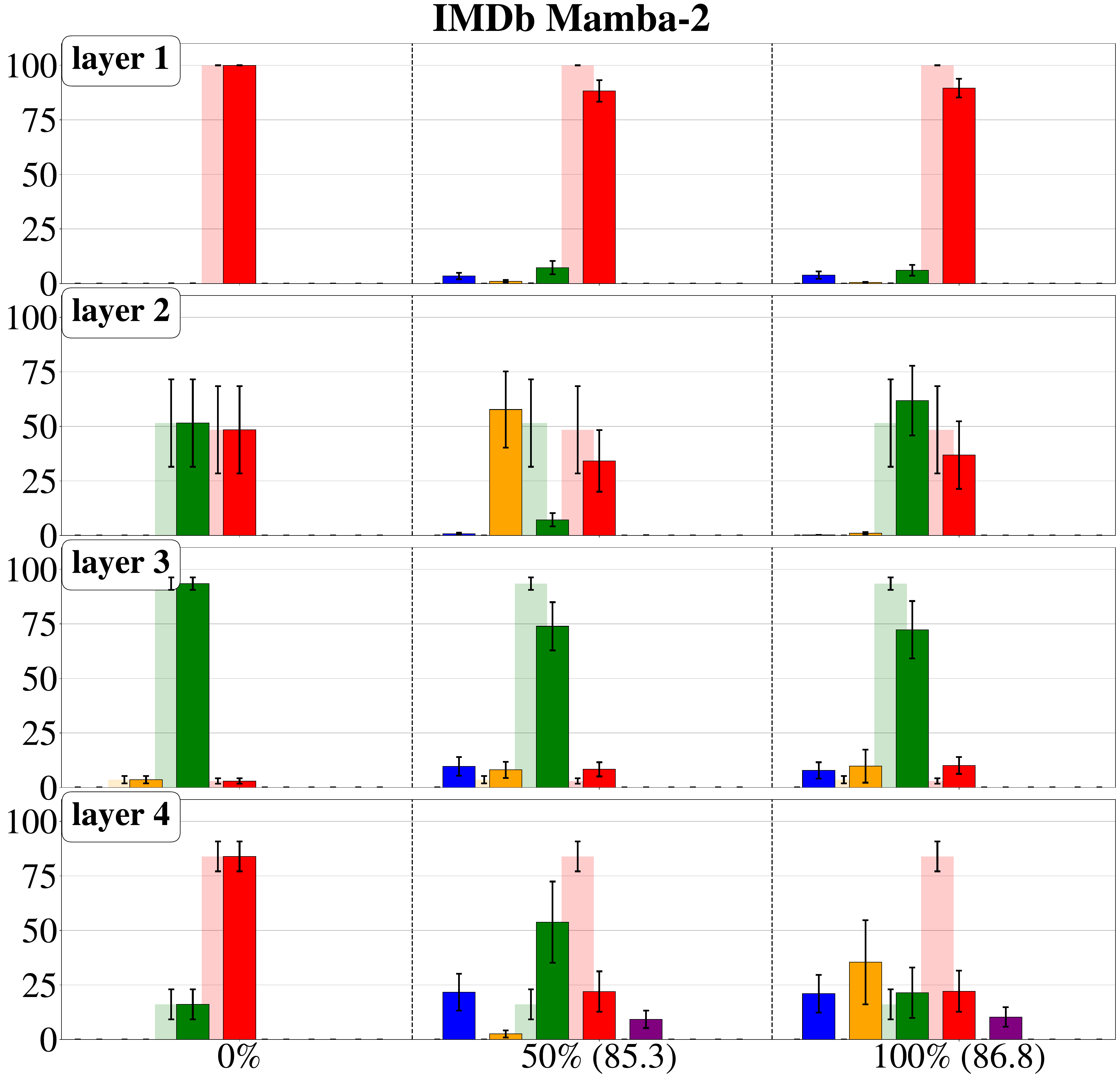}};
    
\end{tikzpicture}
\end{adjustbox}
\caption{ {Evolution of eigenvalues from initialization, over training half-time, to the final distribution for IMDb.}}
 \label{fig:imdb_evolution}
\end{figure}

\begin {figure}[h] 
\centering
\begin{adjustbox}{max height=0.88\textheight, max width=0.90\textwidth}
\begin{tikzpicture}[scale=1.0]

    \node (pic1) at (-13.0, 14.5) {\includegraphics[width=7.90cm]{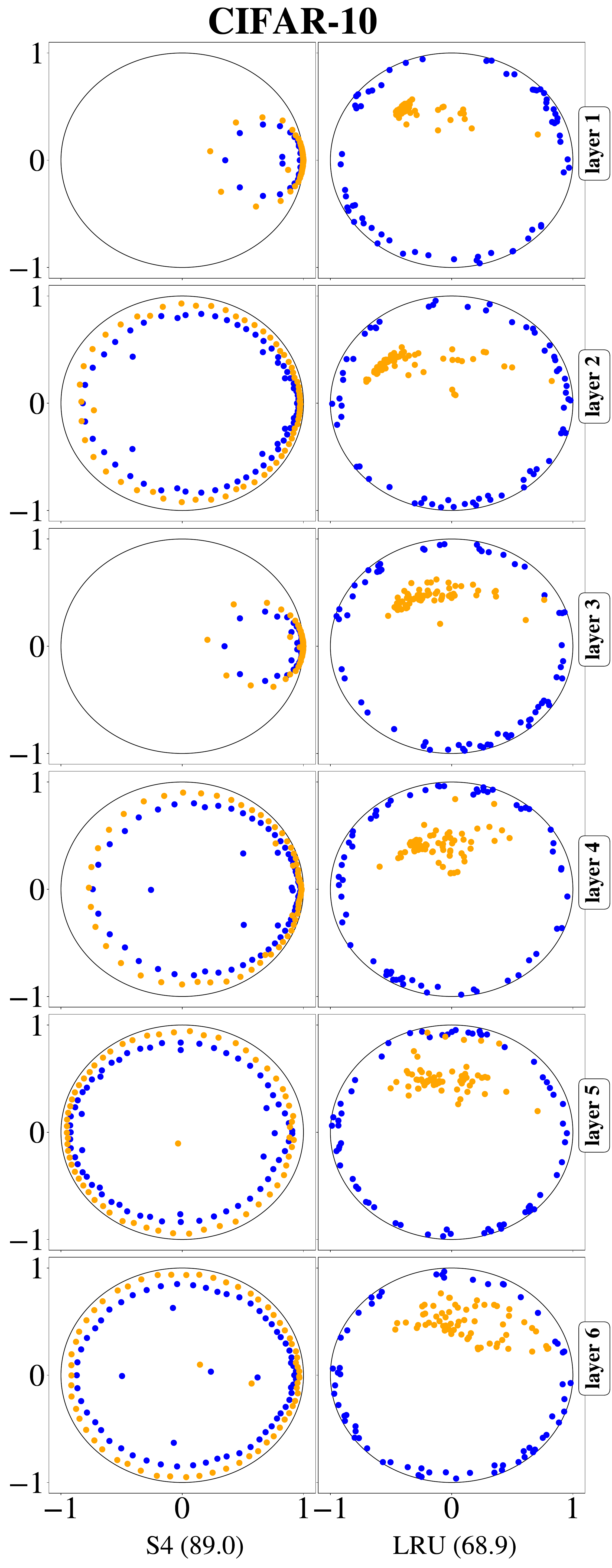}};

    \node (pic2) at (-5.1, 14.5) {\includegraphics[width=7.90cm]    {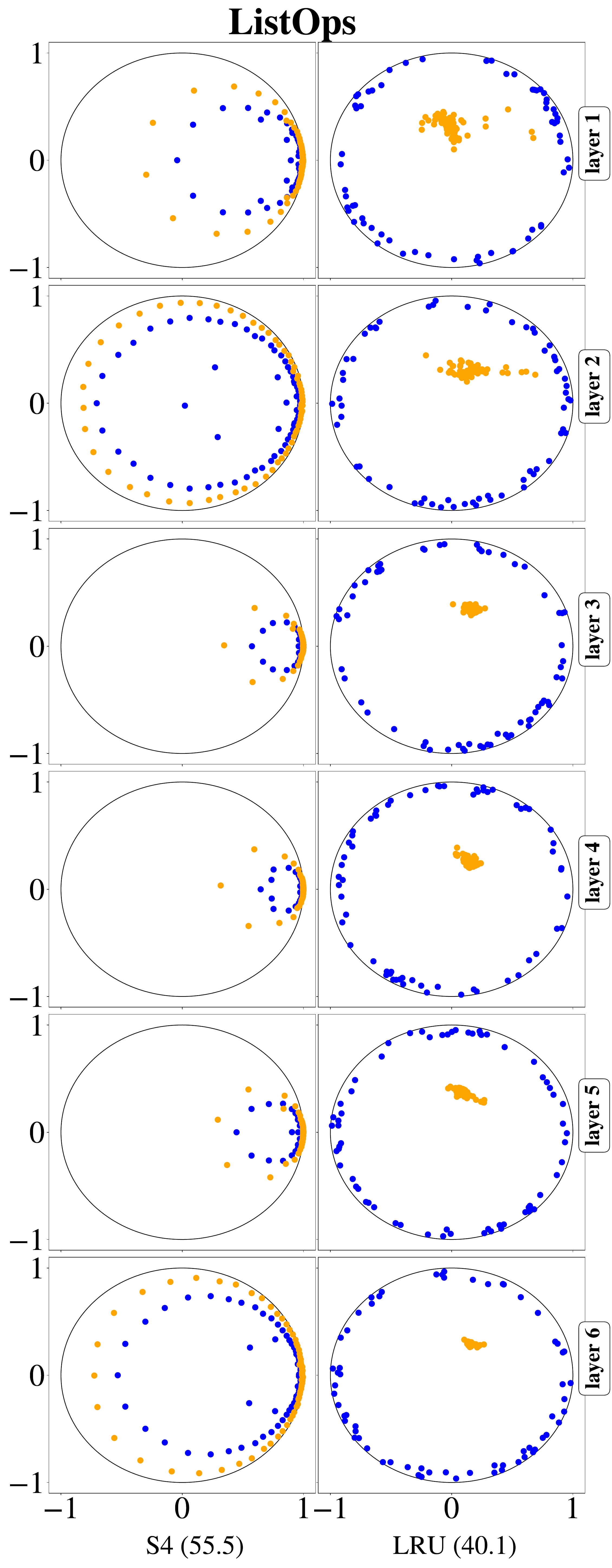}};

    \node (pic3) at (2.8, 17.7) {\includegraphics[width=7.90cm]
    {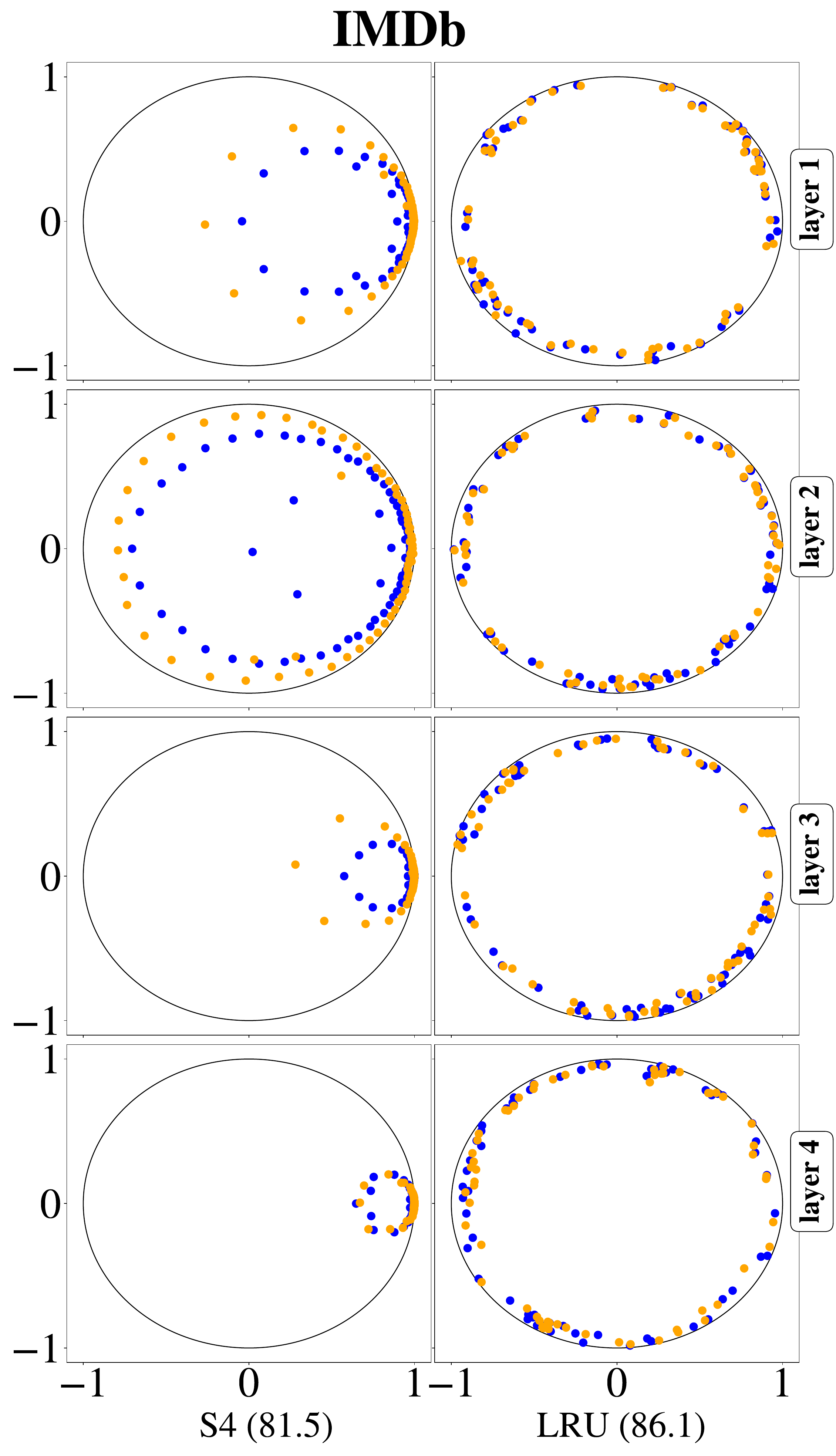}};

\end{tikzpicture}
\end{adjustbox}
\caption{ {Visualization of complex eigenvalues across LRA task, SSMs, and all layers. The trained values are indicated in orange and the initialization in blue. While for LRU there appears to be a shift from initialization to the trained pattern, which is consistent across layers, this behaviour is not as pronounced in S4. There, conversely, eigenvalues remain spread around a circle with a specific radius. While the results are presented for completeness, they would require further investigation for more concrete statements.} 
}
  \label{fig:complex_eigenvalues_appendix}
\end{figure}

\begin{figure}[H]
    \centering
    \includegraphics[width=0.7
\linewidth]{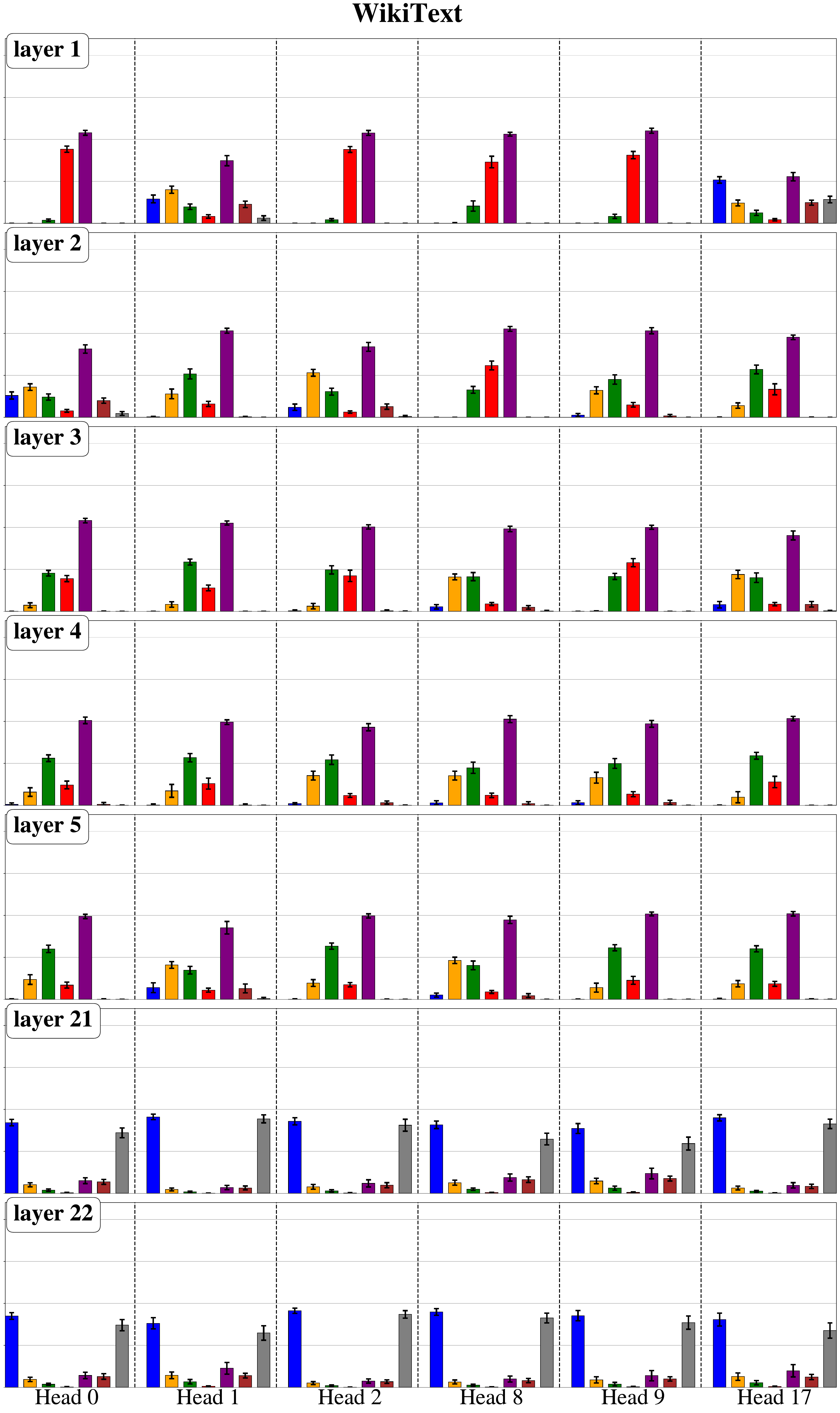}
    \caption{The eigenvalues of the Olmo-3 model for selected 6 out of 32 heads and 7 out of 32 layers. The layers were selected such that they reflect the two prevailing modes: memory retention (early layers) and selective forgetting (later layers).}
    \label{fig:olmo3}
\end{figure}

In this section, we present additional results and visualizations that complement and extend the empirical analysis in the main paper. To enable a simpler, however, less detailed, assessment, we present correlation plots between model performance and the proportions of high and low eigenvalues. Furthermore, we provide cumulative density plots, which offer a more fine-grained representation that is independent of the choice of histogram binning.  
With results in the main text being limited to a single head, we also extend our empirical analysis to further head dimensions of all investigated multi-head models (Mamba-2 and attention models) and benchmark tasks. Note that since MQAR has been trained with a single head, no further results are displayed in the following. To further strengthen our empirical investigation, we subsequently display results for three additional random seeds, all models, and all benchmark tasks. Additionally, we provide evolution plots, displaying the movement of the eigenvalue spectra during training, as well as a visualization of complex eigenvalues. Finally, we provide a visualization of the eigenvalue spectra of Olmo-3-7B-Think. The overview of all presented results is given below:

\begin{itemize}
    \item Figure~\ref{fig:correlation_plot_appendix}: contains all simplified (i.e., head and layer averaged) correlation plots corresponding to Figure~\ref{fig:main_plot};
    \item Figure~\ref{fig:main_plot_appendix}: contains all missing layers from Figure~\ref{fig:main_plot}, and the additional results on the WikiText benchmark;
    \item Figure~\ref{fig:cdf_appendix}: contains cumulative density plots corresponding to the histograms in Figures~\ref{fig:main_plot} and \ref{fig:main_plot_appendix};
    \item Figures~\ref{fig:heads_cifar_plot}-\ref{fig:heads_wikitext_plot2}: show remaining heads for the same models from Figure~\ref{fig:main_plot} on LRA tasks and on the WikiText benchmark;
    \item  {Figure~\ref{fig:avg_heads}: shows eigenvalue distributions averaged over all heads for all multi-headed models and all tasks, with the corresponding standard deviations;}
    \item Figures~\ref{fig:cifar_seeds1}-\ref{fig:wiki_seeds2}: illustrate eigenvalue spectra for one of the heads over three additional seeds for all models and tasks;
    \item  {Figures~\ref{fig:avg_seeds1}-\ref{fig:avg_seeds3}: show eigenvalue distributions averaged over all seeds for one of the heads, all models and all tasks, with the corresponding standard deviations;}
    \item  {Figures~\ref{fig:cifar_evolution}-\ref{fig:imdb_evolution}: show evolution of eigenvalues from initialization, over training half-time, to the final distribution for CIFAR-10 and IMDb tasks.}
    \item  {Figure~\ref{fig:complex_eigenvalues_appendix}: visualizes the complex eigenvalues across LRA task, SSMs, and all layers. The trained values are indicated in blue and the initialization in orange.}
    \item{Figure~\ref{fig:olmo3}: shows eigenvalues for selected, representative layers and heads of the Olmo-3-7B-Think \cite{olmo2025olmo3} model evaluated on the WikiText benchmark.}
\end{itemize}

The provided figures showcase the robustness of our conclusions and their extension to different seeds, model heads, and tasks not directly provided in the main paper.

\begin{figure}[htb]
    \centering
    \includegraphics[width=0.95\linewidth]{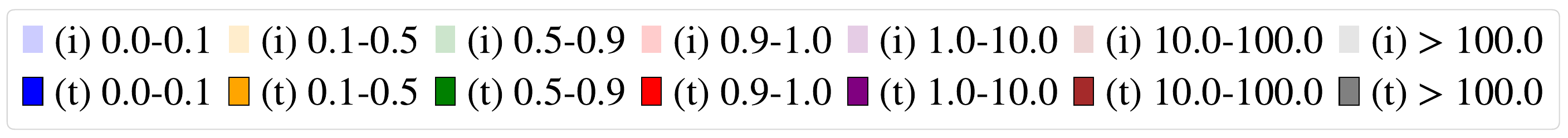}
    \caption{Plot legend of all subsequent histogram figures.  }
    \label{fig:legend_1}
\end{figure}

All figures with histograms make use of the same legend as Figure \ref{fig:main_plot}, once again displayed in Figure \ref{fig:legend_1}, below. Thereby, bars show the proportion of eigenvalues falling within discretized ranges. Lightly shaded and darker bars indicate the distribution at initialization and after training, respectively. Error bars denote standard deviation across input sequences. Model performance, measured as perplexity for WikiText
(lower is better) and percentage of correct output sequences for the remaining tasks (higher is better), is indicated in the parentheses. 
\clearpage
\newpage
\subsection{Performance Correlation Plots}
\label{subapx:performance_correlation_plots}

\begin{figure}[H]
    \centering
    \begin{adjustbox}{max height=0.9\textheight, max width=0.98\textwidth}
        \begin{tikzpicture}[scale=1.0]

            \node (pic1) at (-10.0, 10) {\includegraphics[width=20cm]
            {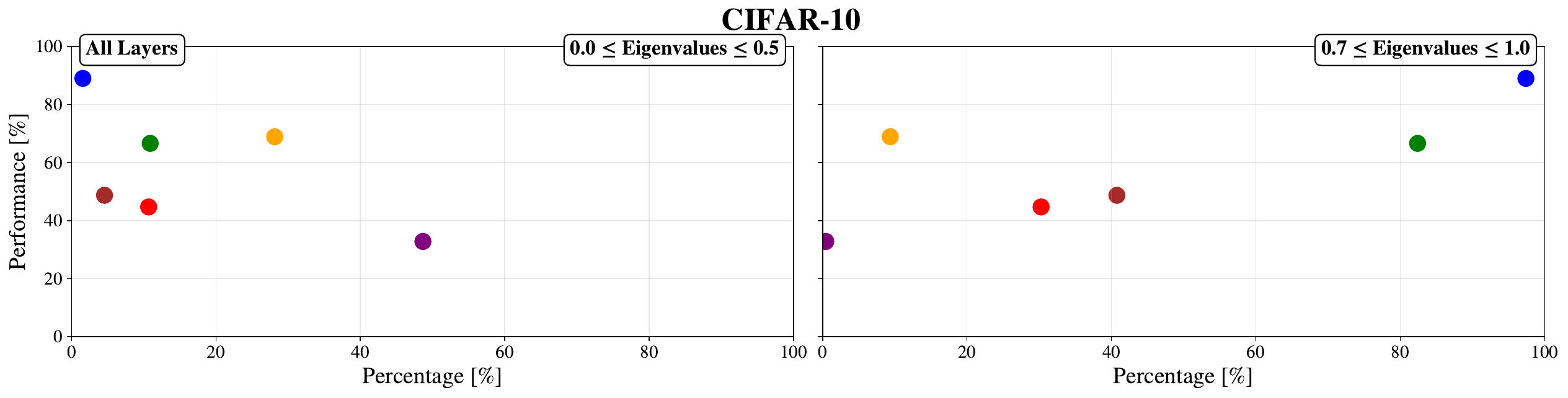}};
            
            \node (pic2) at (-10.0, 5) {\includegraphics[width=20cm]
            {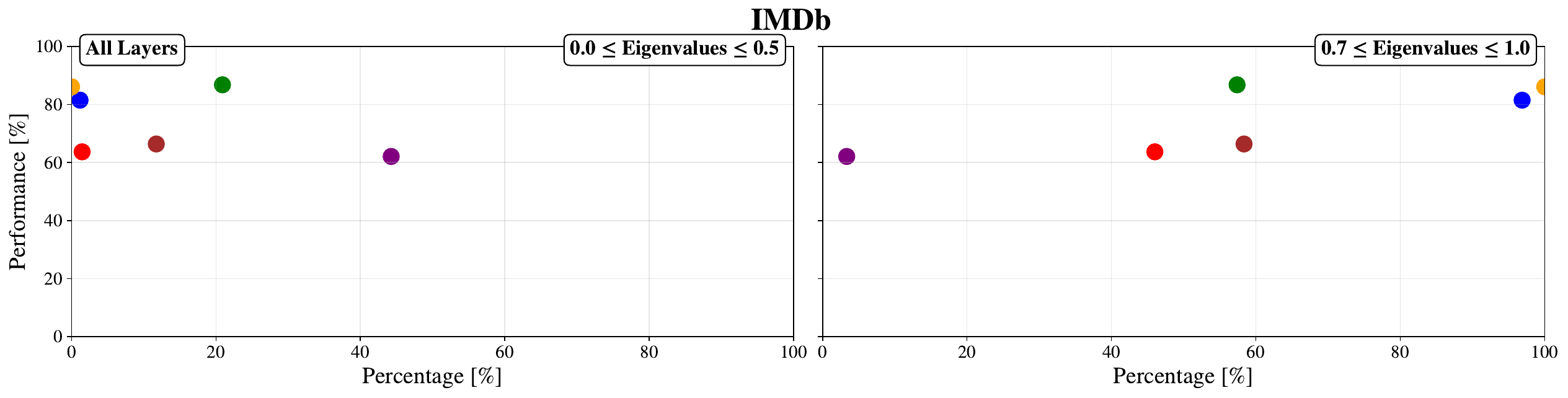}};

            \node (pic3) at (-10.0, 0) {\includegraphics[width=20cm]
            {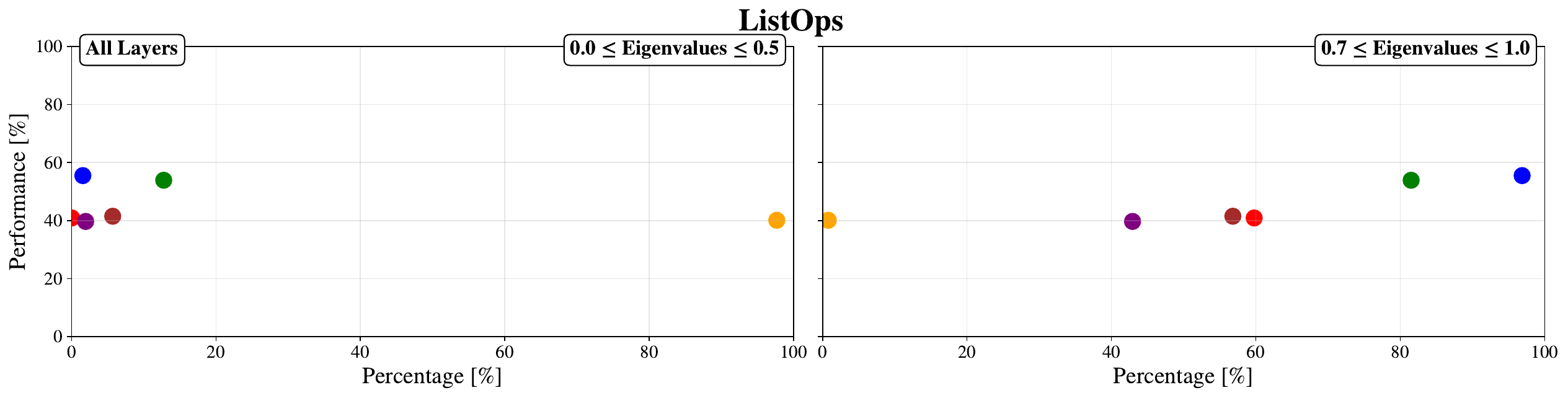}};

            \node (pic4) at (-10.0, -5) {\includegraphics[width=20cm]
            {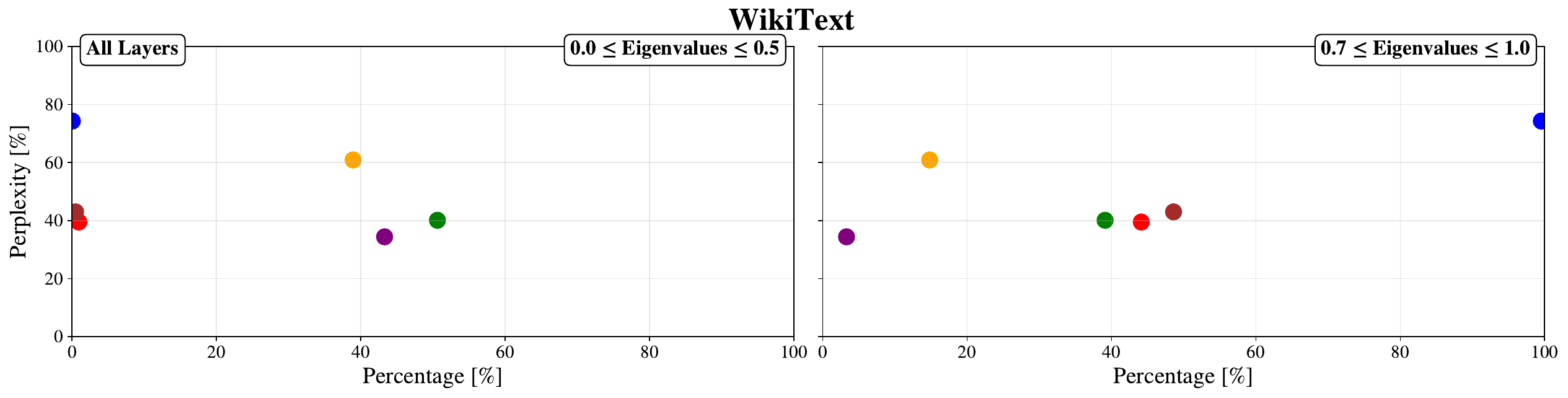}};

            \node (pic5) at (-10.0, -10.5) {\includegraphics[width=20cm]
            {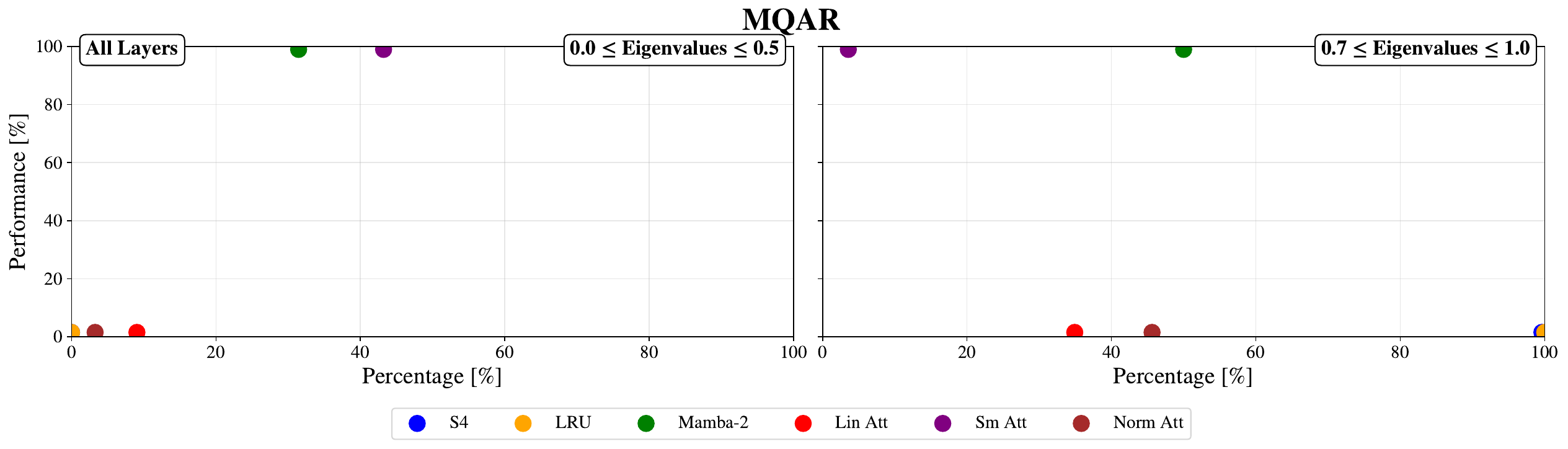}};

        \end{tikzpicture}
    \end{adjustbox}
    \caption{Correlation plots of performance with respect to percentages of small eigenvalues (forgetting, left plots) and eigenvalues close to 1 (rentention, right plots), averaged over all layers and heads.
}
  \label{fig:correlation_plot_appendix}
\end{figure}

While Figure~\ref{fig:main_plot} and its complete version in Figure~\ref{fig:main_plot_appendix} show the per-bin eigenvalue distributions, Figure~\ref{fig:correlation_plot_appendix} aggregates the per-bin correlation plots into two coarse ranges: the proportion of small
eigenvalues ($0.0$--$0.5$, associated with selective forgetting) and the
proportion of eigenvalues close to one ($0.7$--$1.0$, associated with
retention), averaged over all layers and heads. On the memory-intensive LRA
tasks (CIFAR-10, IMDb, ListOps), higher performance shows a consistent
association with a larger retention fraction and a smaller forgetting fraction,
in line with our hypothesis that eigenvalues near one support long-term memory.
This trend is most visible on CIFAR-10, where the weakest models are precisely
those placing the largest share of eigenvalues in the forgetting range.
Attention-based models sit apart from this pattern: consistent with their gating
behavior, they retain a non-negligible fraction of small eigenvalues even on
these tasks, and therefore deviate from the association followed by the SSMs. On
MQAR and WikiText, where only specific information must be preserved, the picture
reverses, and the successful models are those that tolerate or exploit a larger
forgetting fraction. Taken together, the plots indicate that performance tracks
the alignment between a model's spectrum and the memory demand of the task,
rather than any single eigenvalue range in isolation, which is consistent with
Proposition~\ref{prop:gains}, where the input-output gain is governed by the
product $g_{t,j}$ of eigenvalues across the sequence.

\FloatBarrier
\subsection{Eigenvalue Spectra for Additional Layers}
\label{subapx:additional_layers}

\begin{figure}[H]
    \centering
    \begin{adjustbox}{max height=0.95\textheight, max width=0.98\textwidth}
        \begin{tikzpicture}[scale=1.0]

            \node (pic1) at (-10.0, 10) {\includegraphics[width=20cm]
            {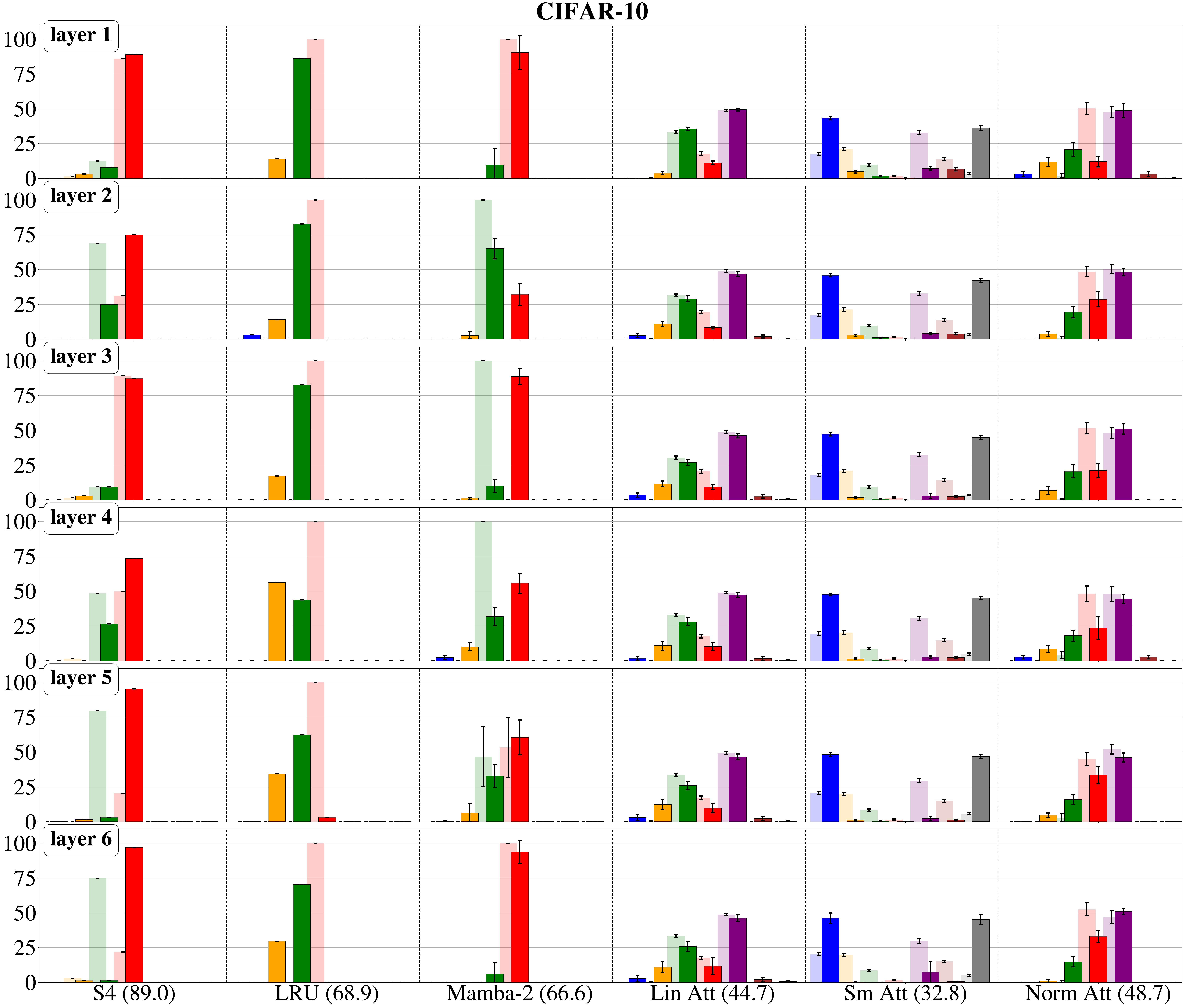}};
            
            \node (pic3) at (-10.0, -8) {\includegraphics[width=20cm]
            {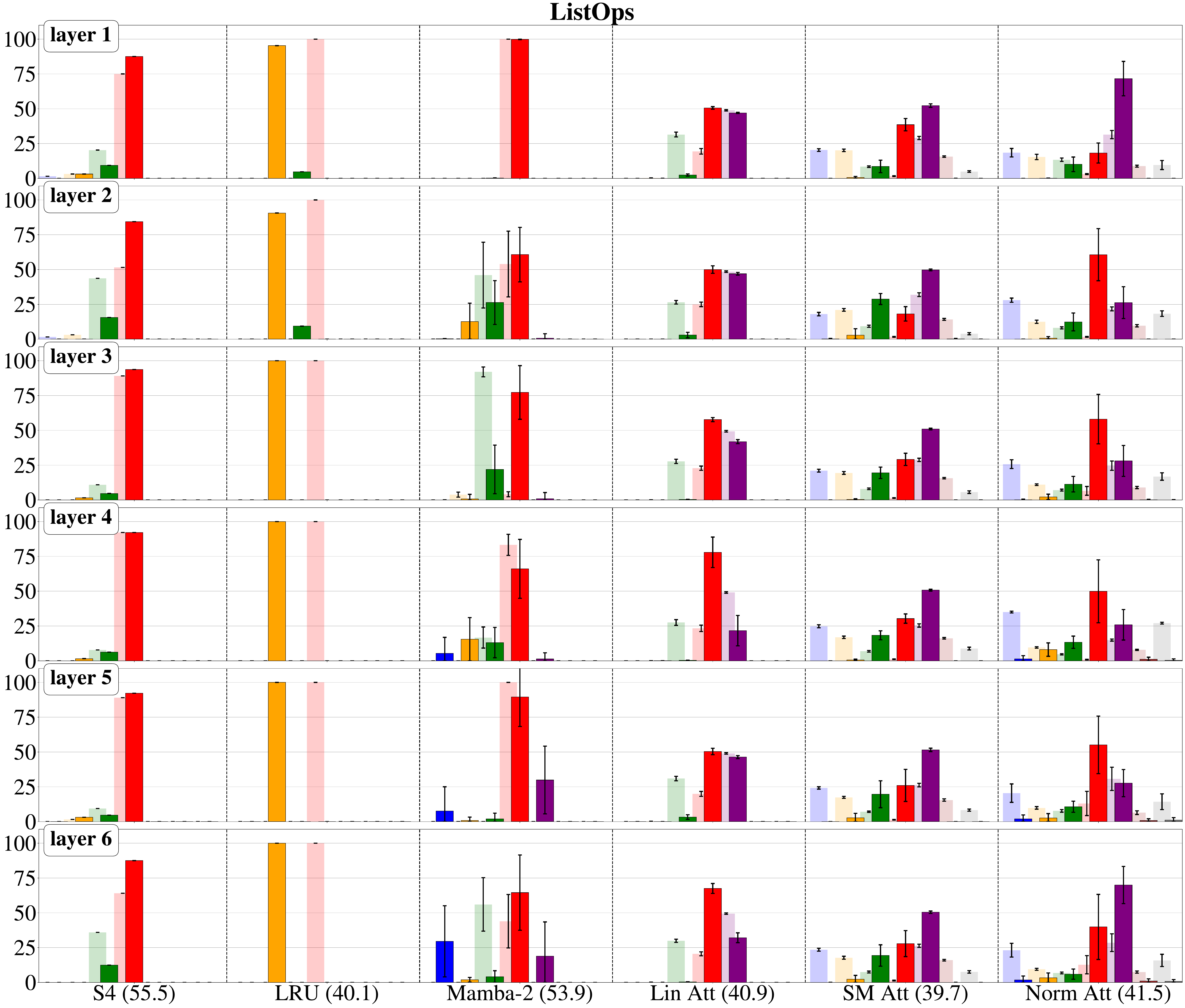}};

        \end{tikzpicture}
    \end{adjustbox}
\end{figure}

\FloatBarrier
\subsection{Eigenvalue Spectra Cumulative Density Representation}
\label{subapx:cdf}

\begin{figure}[H]
    \centering
    \begin{adjustbox}{max height=0.95\textheight, max width=0.95\textwidth}
        \begin{tikzpicture}[scale=1]

            \node (pic1) at (-10.0, 10) {\includegraphics[width=20cm]
            {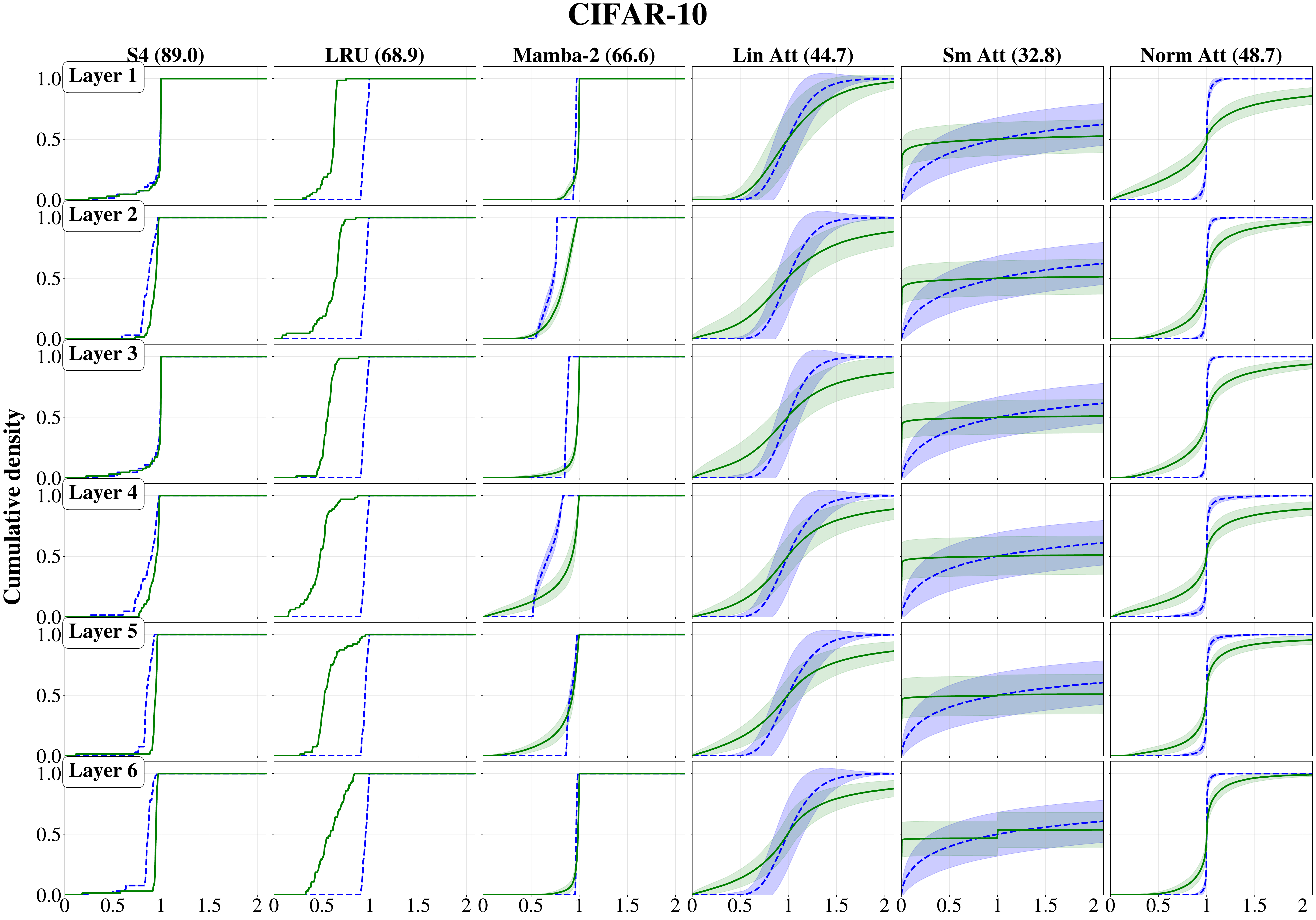}};
            
            \node (pic3) at (-10.0, -4.5) {\includegraphics[width=20cm]
            {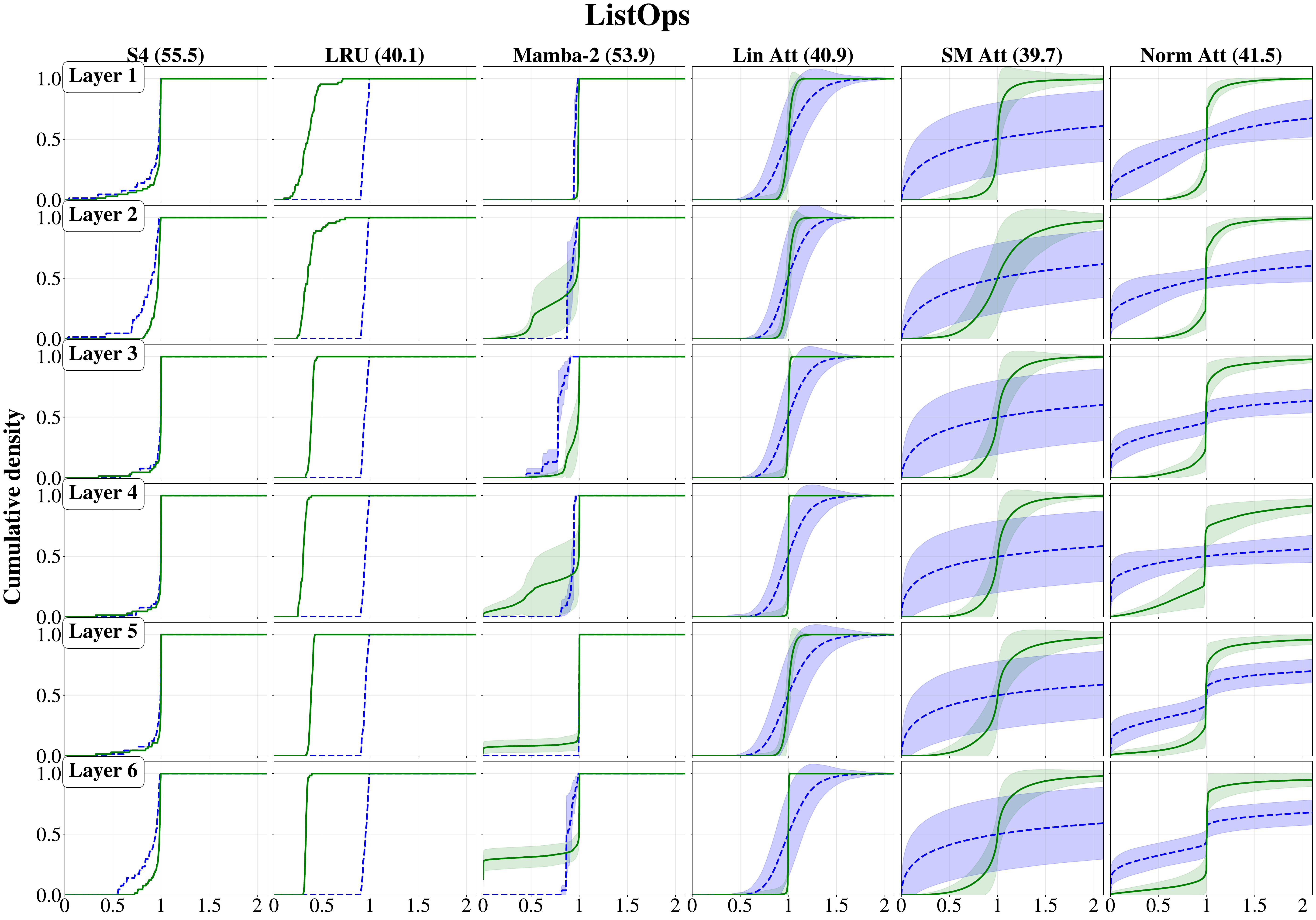}};

        \end{tikzpicture}
    \end{adjustbox}
\end{figure}

\FloatBarrier
\subsection{Eigenvalue Spectra for Additional Heads}
\label{subapx:additional_heads}

\FloatBarrier
\subsection{Eigenvalue Spectra Averaged over Heads}
\label{subapx:averaged_over_heads}

\FloatBarrier
\subsection{Eigenvalue Spectra for Additional Seeds}
\label{subapx:additional_seeds}

\clearpage
\newpage

\FloatBarrier
\subsection{Eigenvalue Spectra Averaged over Seeds}
\label{subapx:averaged_over_seeds}

\FloatBarrier

\subsection{Evolution of Eigenvalues During Training}
\label{subapx:evolution_of_eigenvalues}

\FloatBarrier
\clearpage
\subsection{Phase and Radius Plots for Complex Eigenvalues}
\label{subapx:complex_eigenvalues}

\FloatBarrier
\subsection{Eigenvalues for Olmo-3-7B-Think}
\label{subapx:eigenvalues_olmo_3}

\section{Additional Results on Model Analysis and Design}\label{app:mad_additional_results}

\begin {figure}[H] 
\centering
\begin{adjustbox}{max height=0.93\textheight, max width=0.80\textwidth}
\begin{tikzpicture}[scale=1.0]
    
    \node (pic1) at (-13.0, 6.5) {\includegraphics[width=7.95cm]{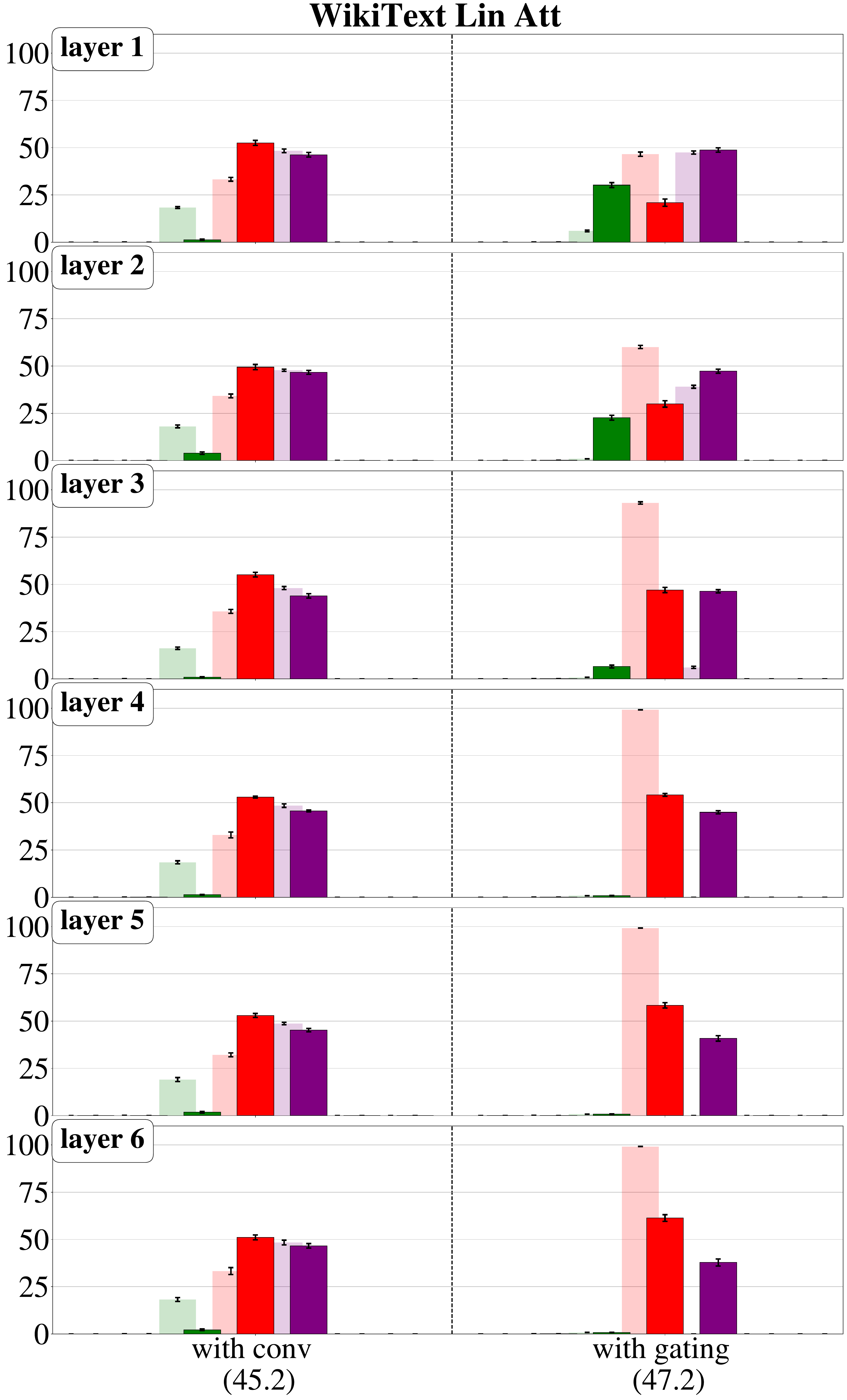}};

    \node (pic2) at (-4.9, 6.5) {\includegraphics[width=7.5cm]    {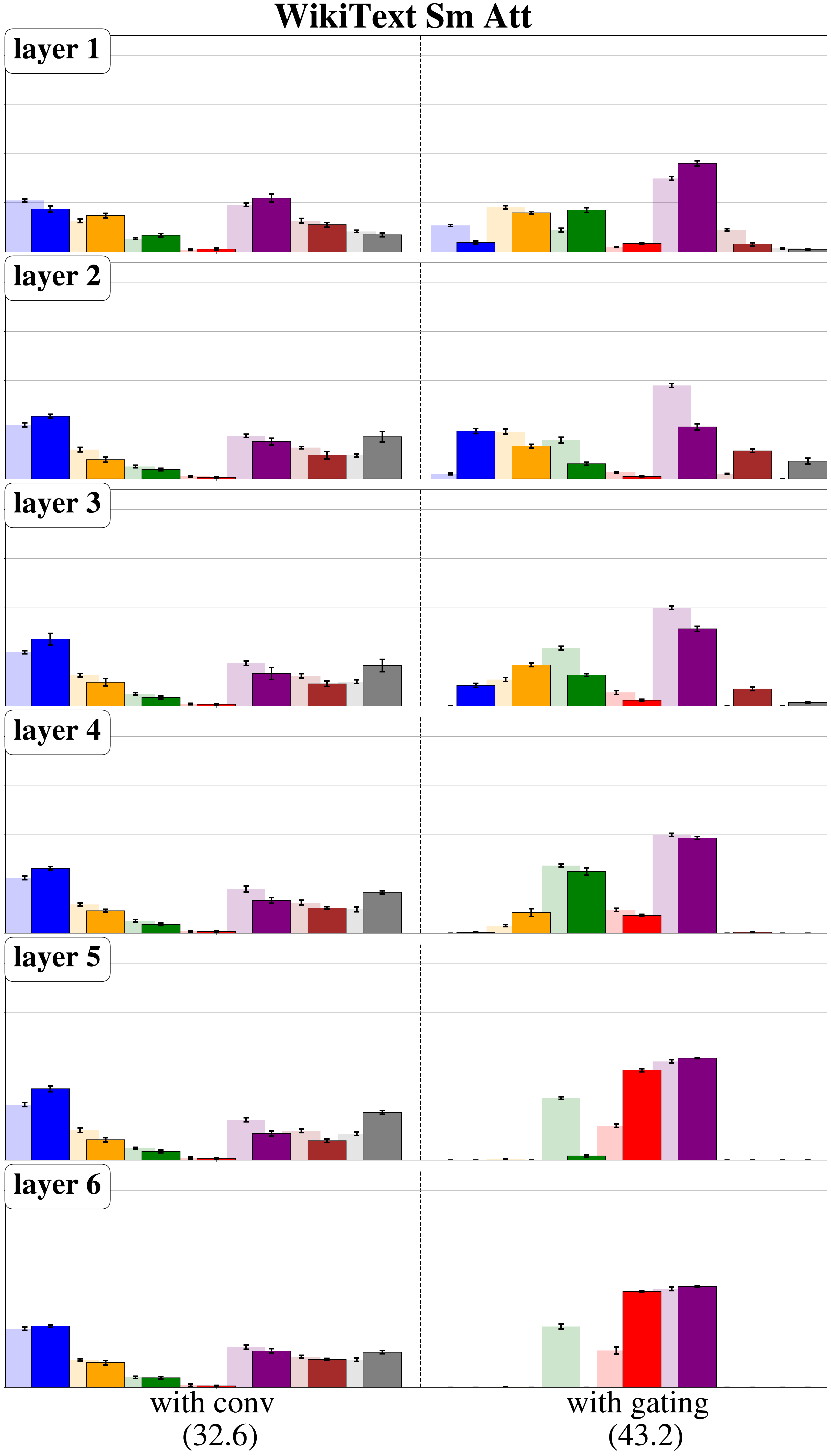}};

    \node (pic3) at (2.8, 6.5) {\includegraphics[width=7.5cm]
    {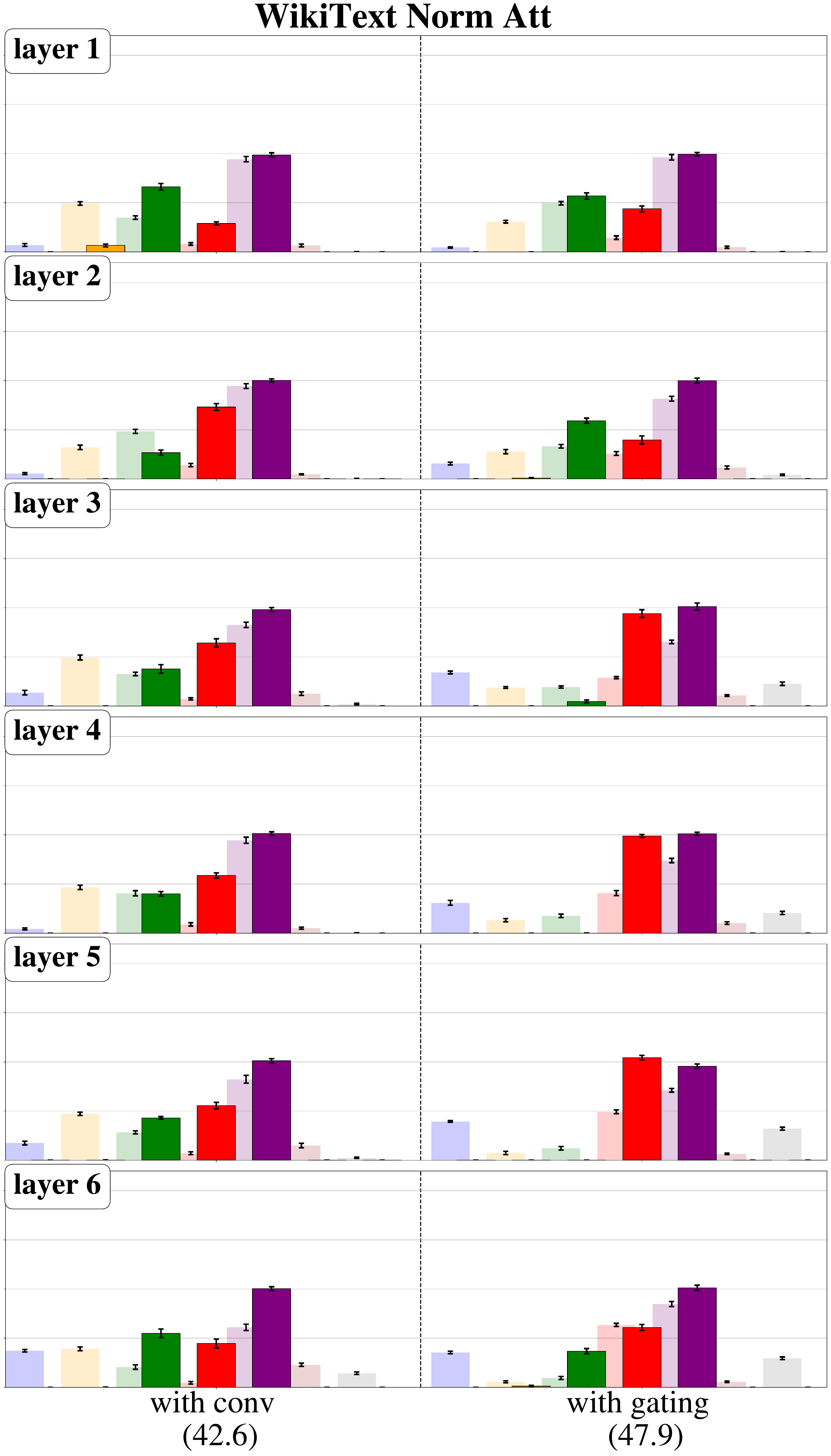}};

\end{tikzpicture}
\end{adjustbox}
\caption{Comparison of the effects of gating and convolution on the eigenvalue spectra for one head, WikiText, the three investigated attention models, and all layers.
}
  \label{fig:gating_conv_wikitext_appendix}
\end{figure}

\begin {figure}[H]
\centering
\begin{adjustbox}{max height=0.91\textheight, max width=0.90\textwidth}
\begin{tikzpicture}[scale=1.0]

    \node (pic1) at (-10.0, 8.5) {\includegraphics[width=11.85cm]
    {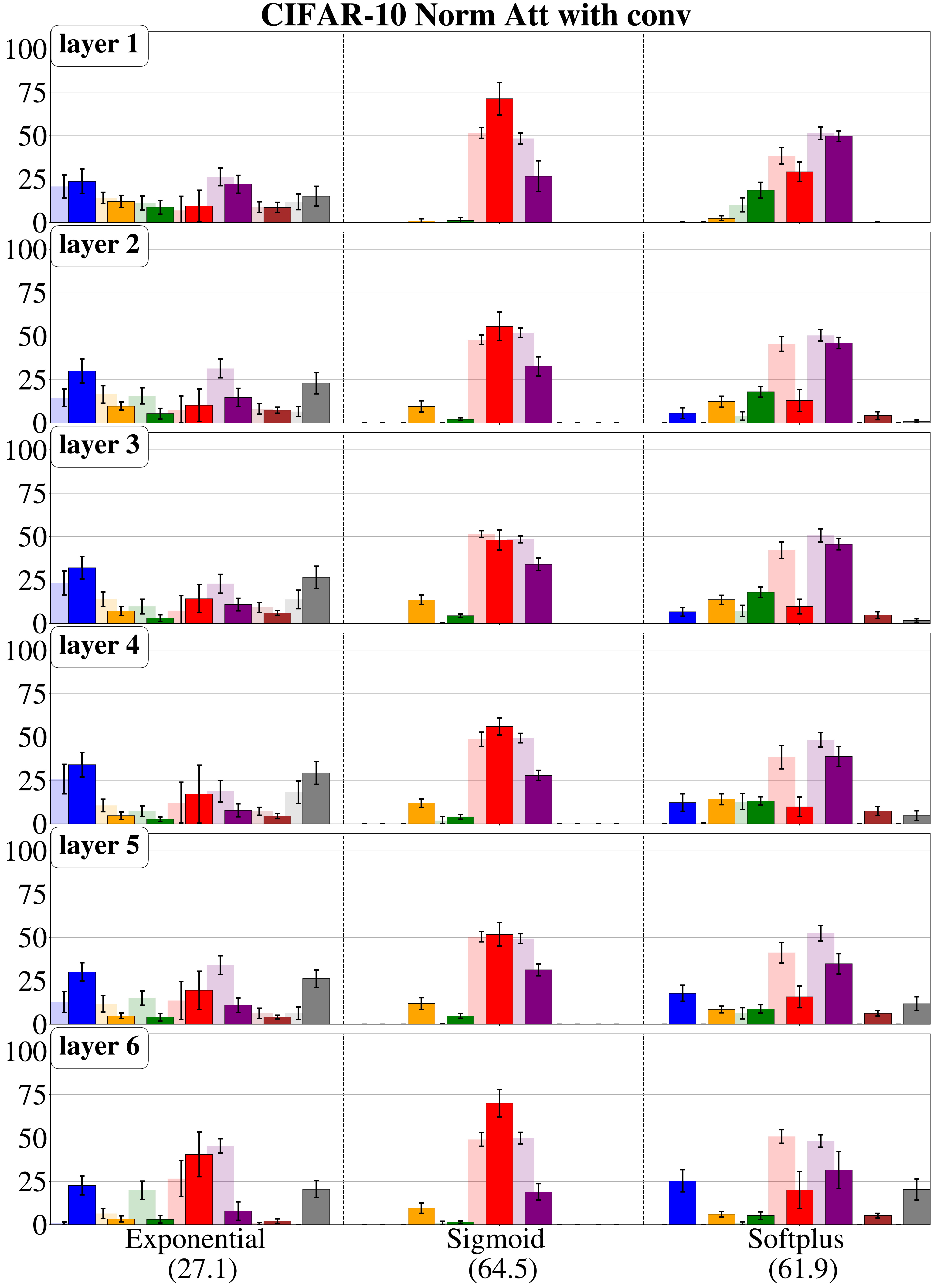}};

    \node (pic2) at (1.65, 8.5) {\includegraphics[width=11.3cm]
    {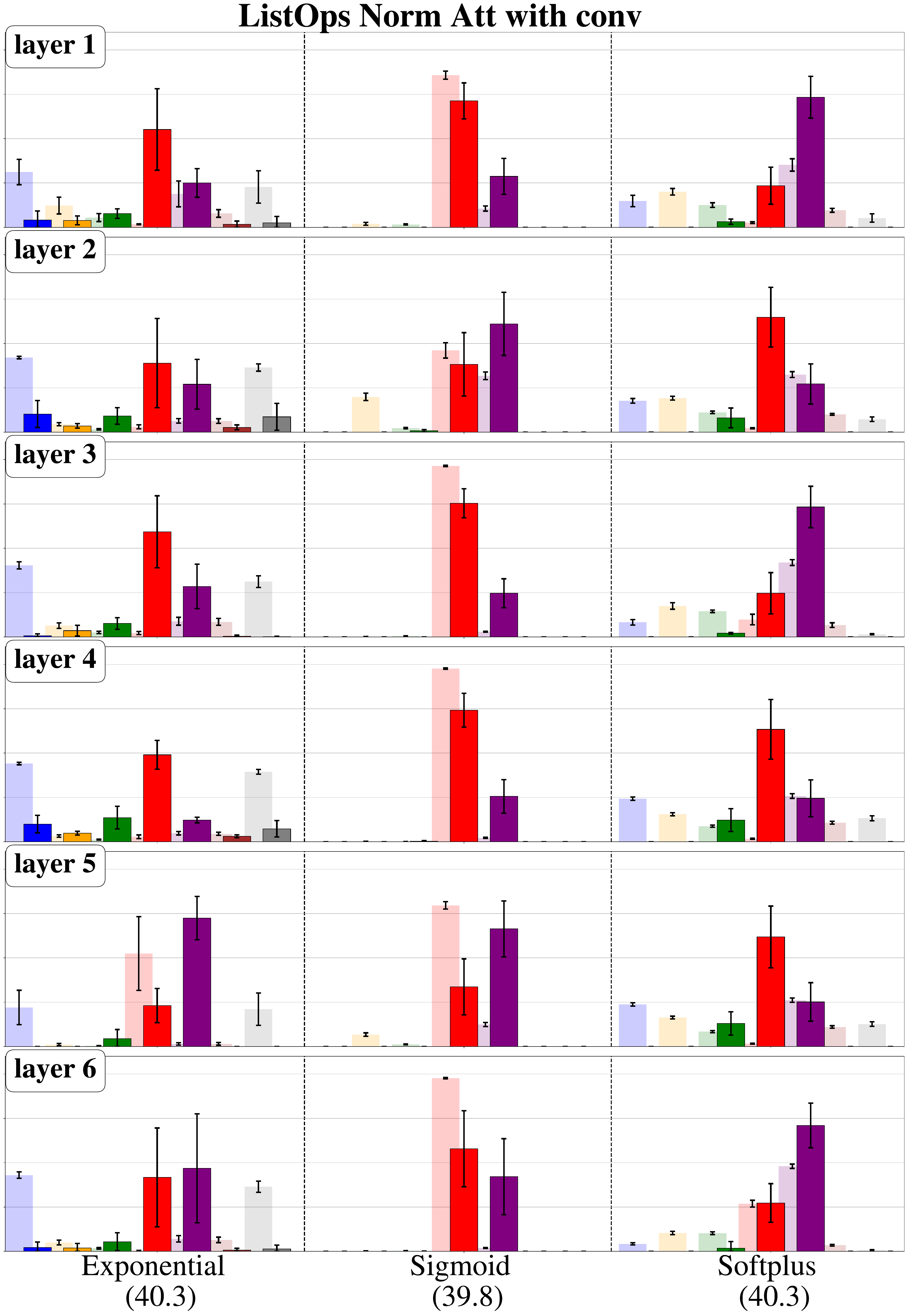}};

    \node (pic3) at (-10.0, -8.2) {\includegraphics[width=11.85cm]
    {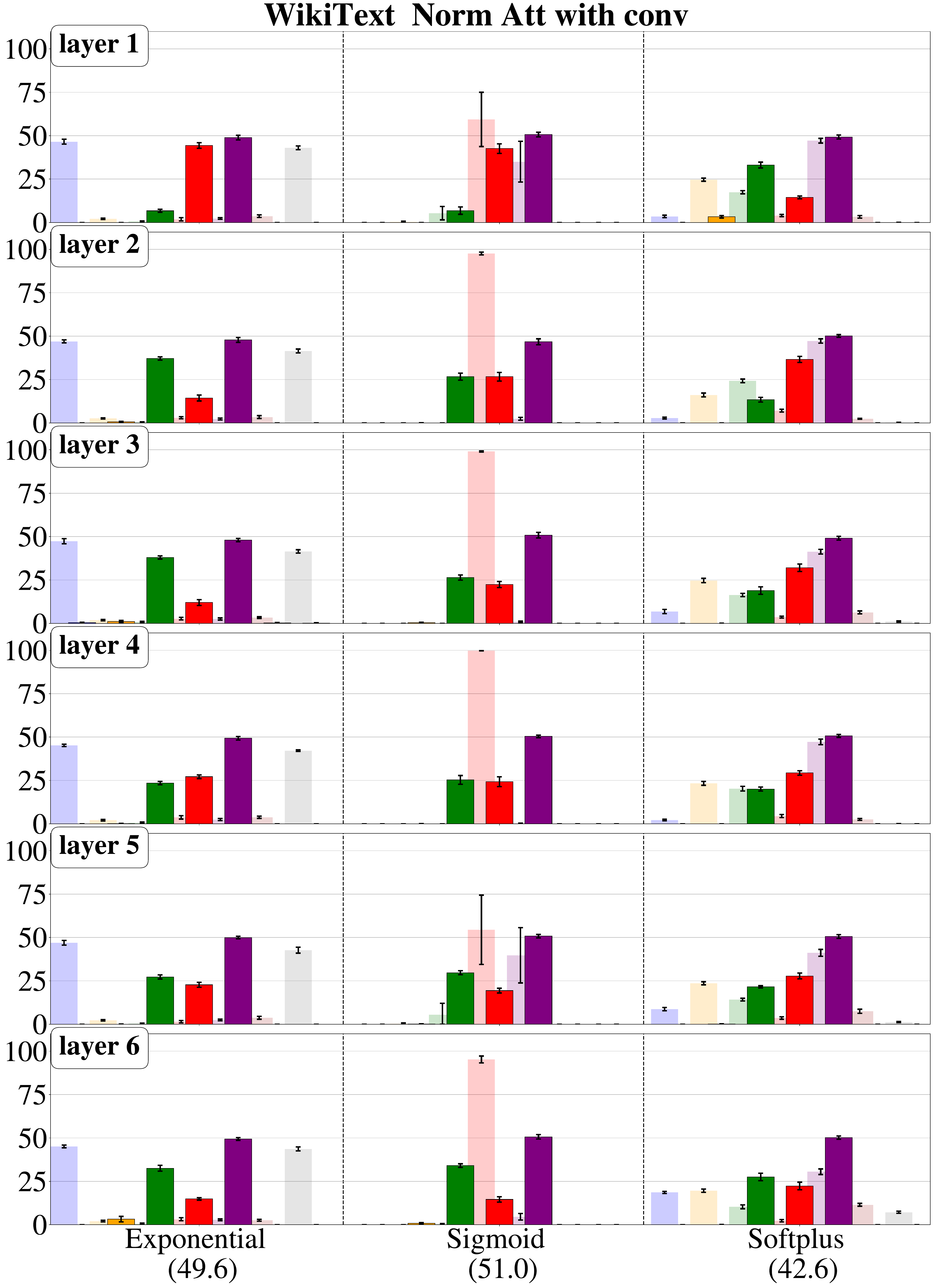}};

    \node (pic6) at (1.65, -5.7) {\includegraphics[width=11.5cm]
    {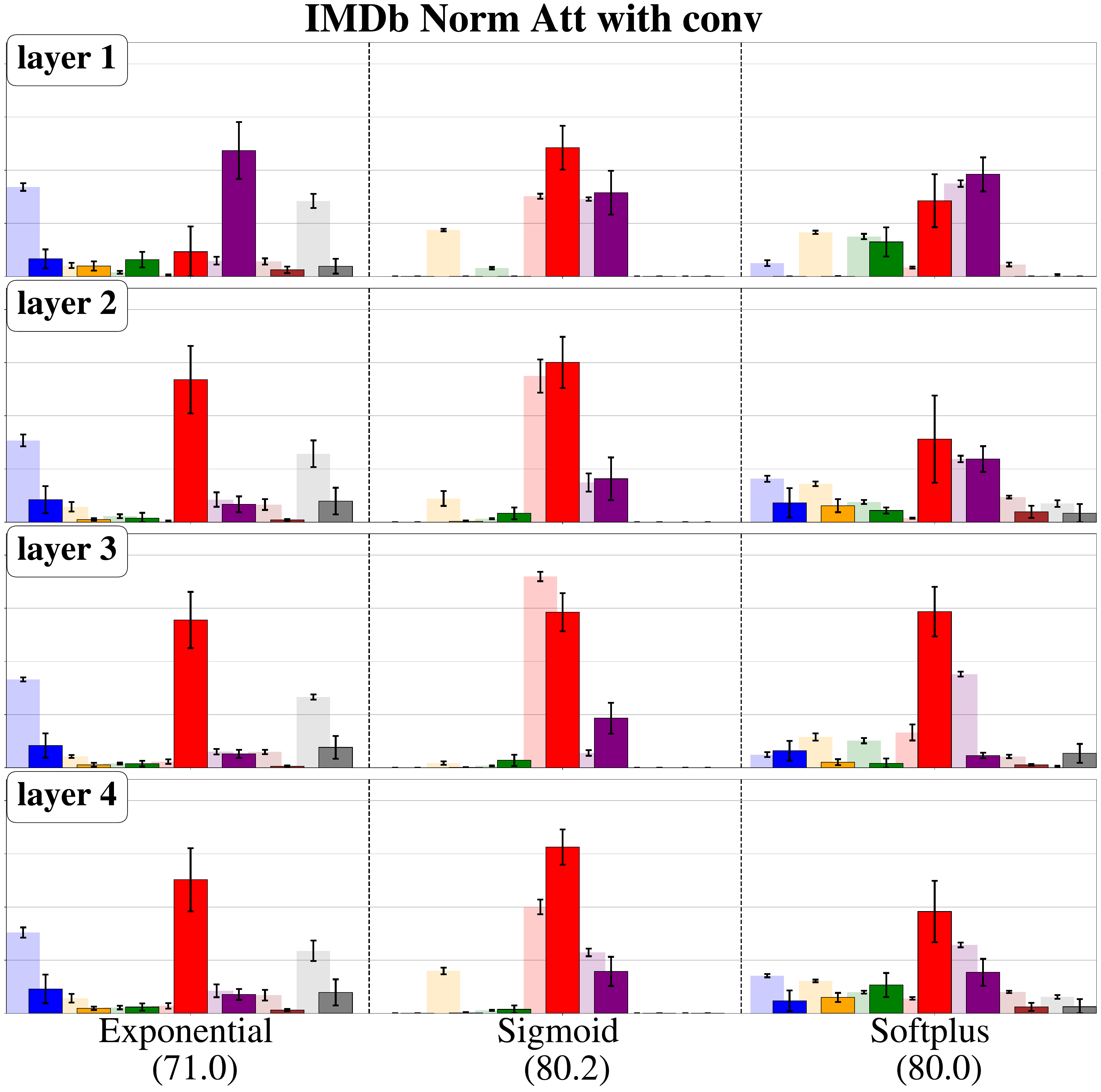}};

    \node (pic4) at (1.65, -13.95) {\includegraphics[width=11.5cm]
    {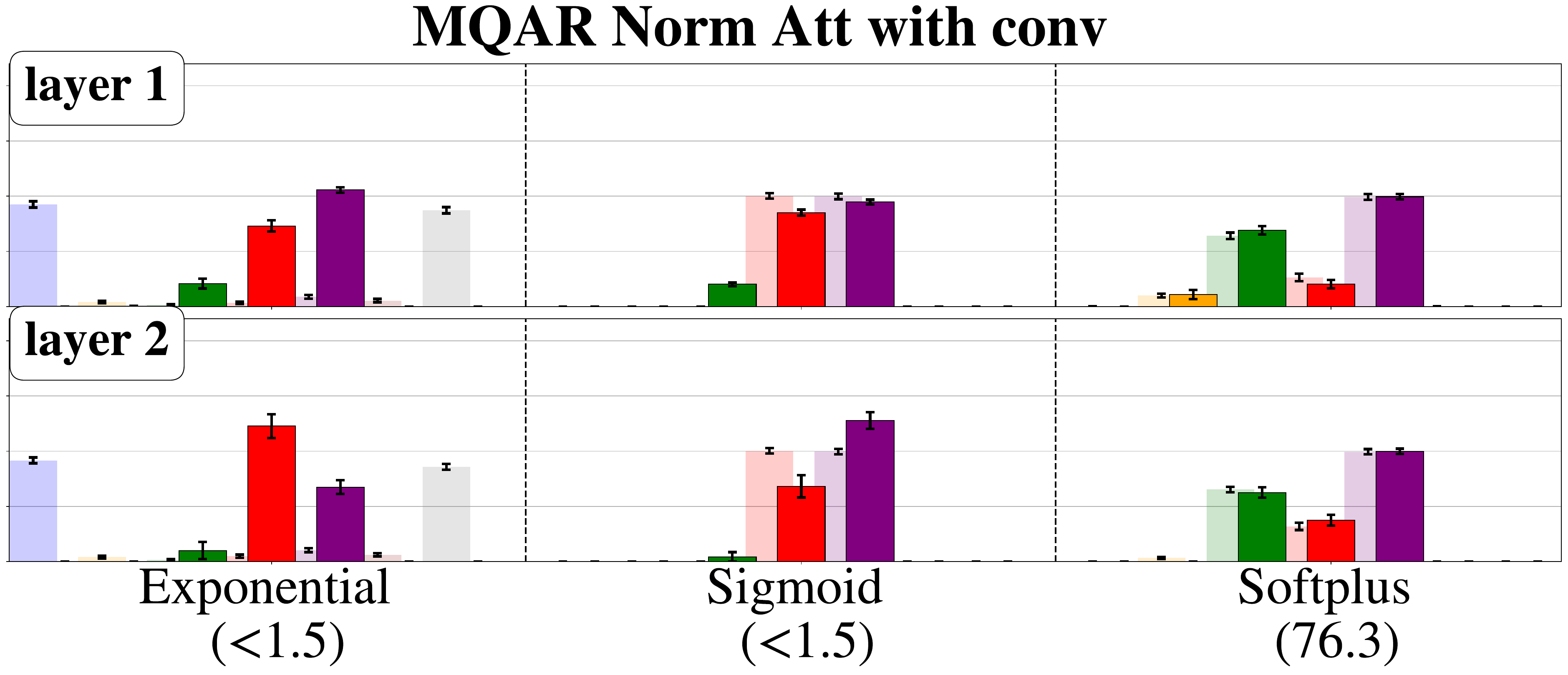}};
    
\end{tikzpicture}
\end{adjustbox}
\caption{Eigenvalue distributions for all tasks and one of the heads for norm attention with convolution, using different normalization functions.}
 \label{fig:norm_init_appendix}
\end{figure}

\begin {figure}[H] 
\centering
\begin{adjustbox}{max height=0.90\textheight, max width=0.90\textwidth}
\begin{tikzpicture}[scale=1.0]

    \node (pic1) at (-13.0, 9.5) {\includegraphics[width=7.95cm]{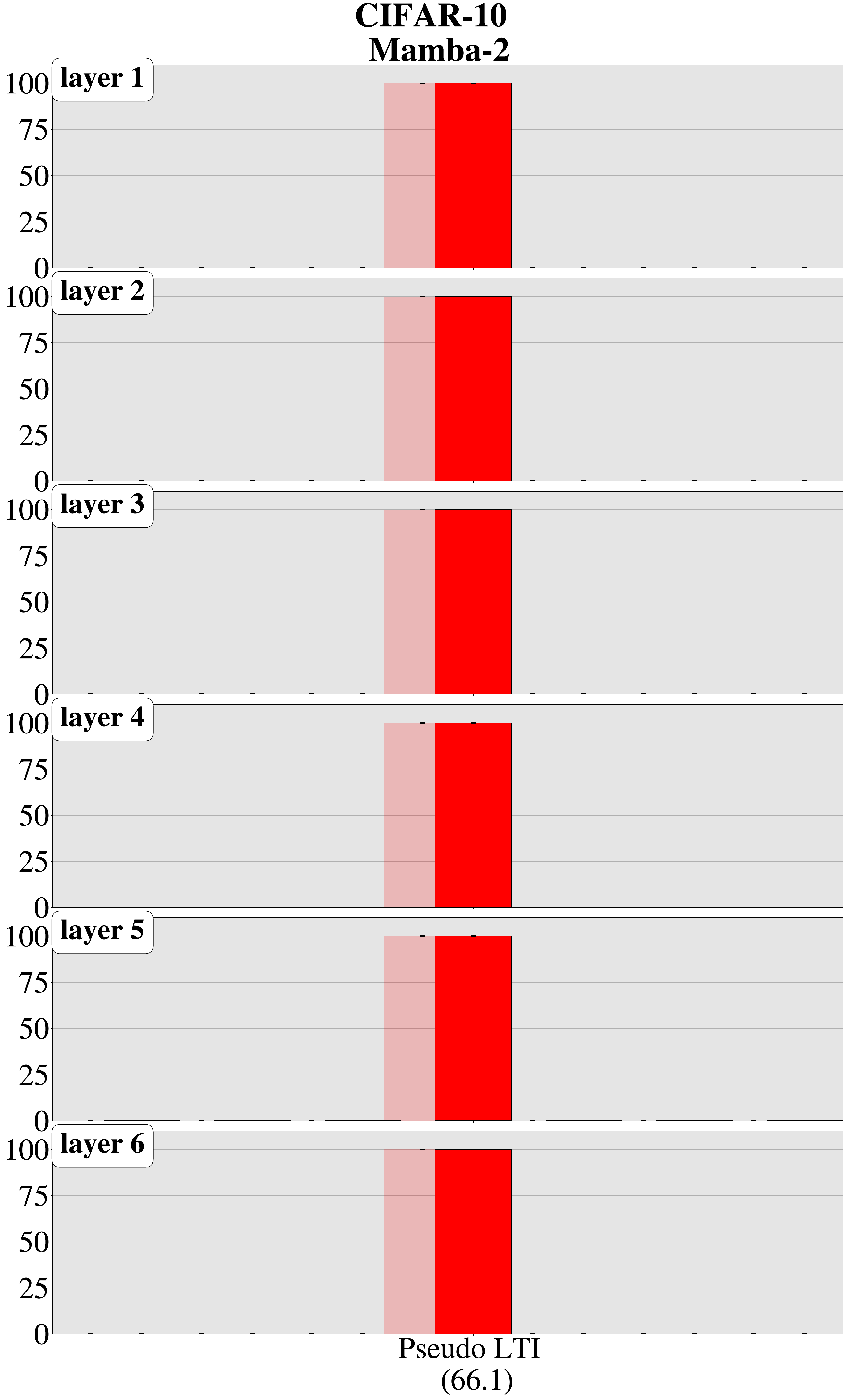}};

    \node (pic2) at (-4.9, 9.5) {\includegraphics[width=7.5cm]    {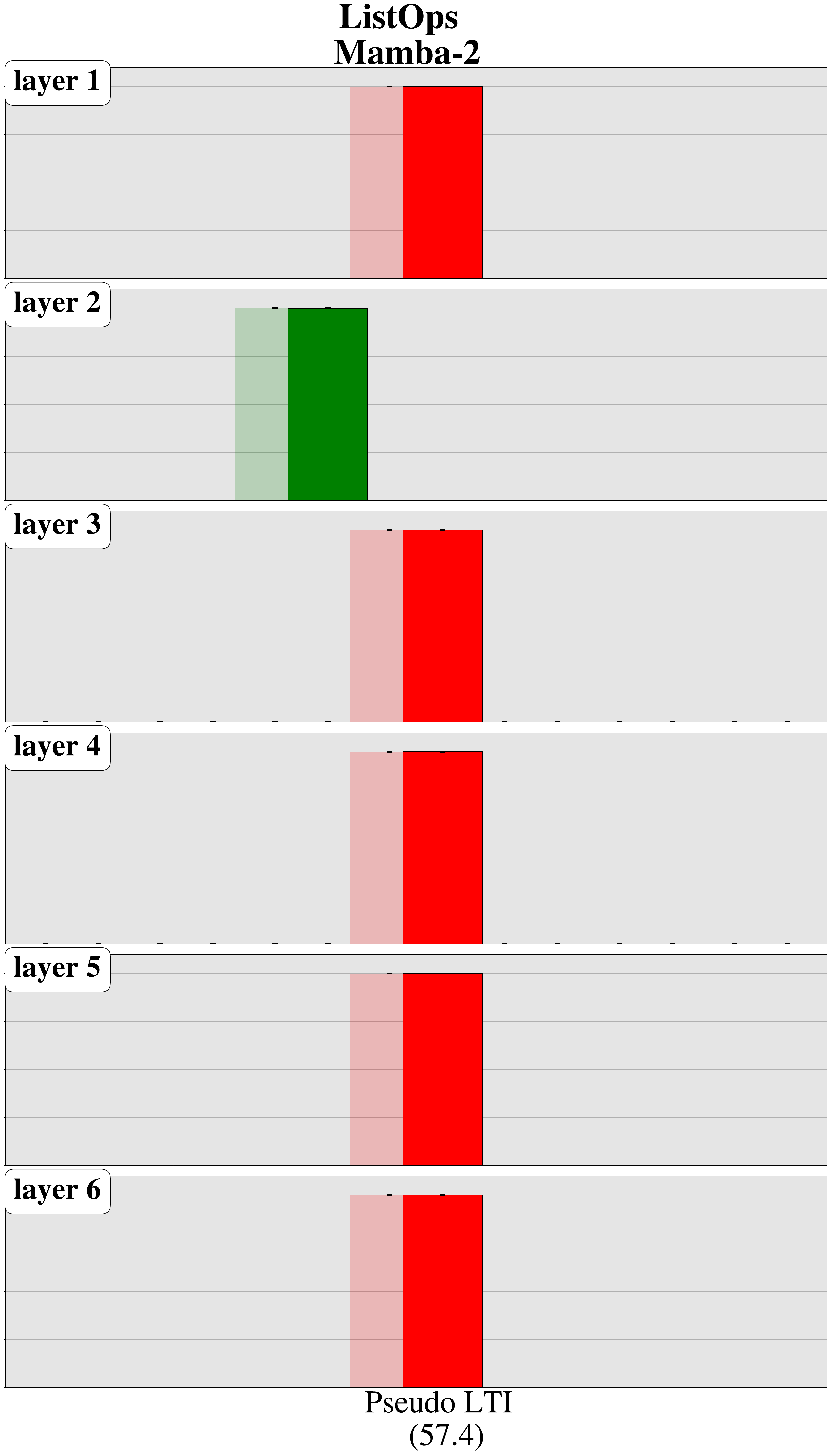}};

    \node (pic3) at (2.9, 11.5) {\includegraphics[width=7.6cm]
    {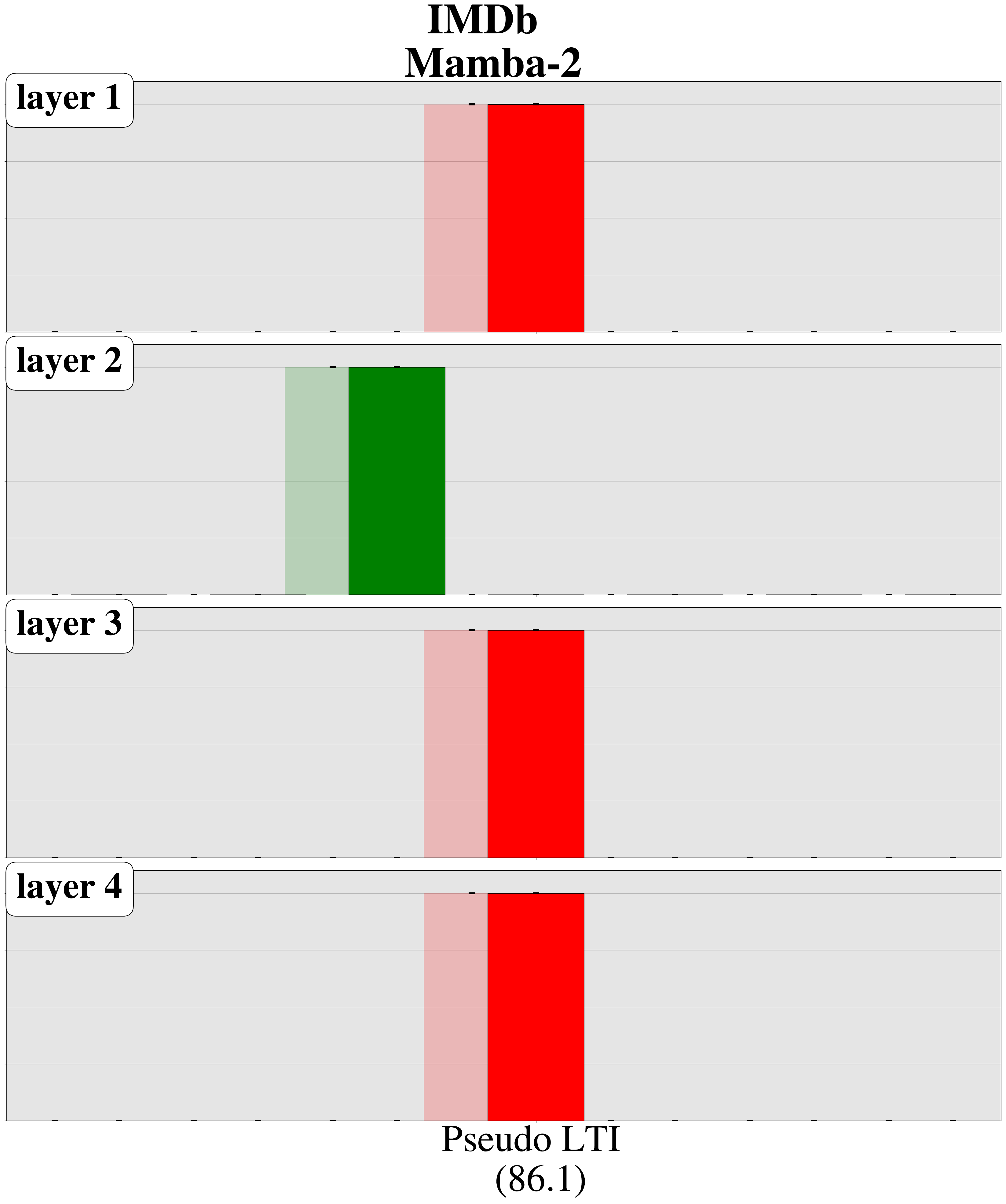}};

    \node (pic0) at (-4.9, 1.0) {\includegraphics[width=22.5cm]{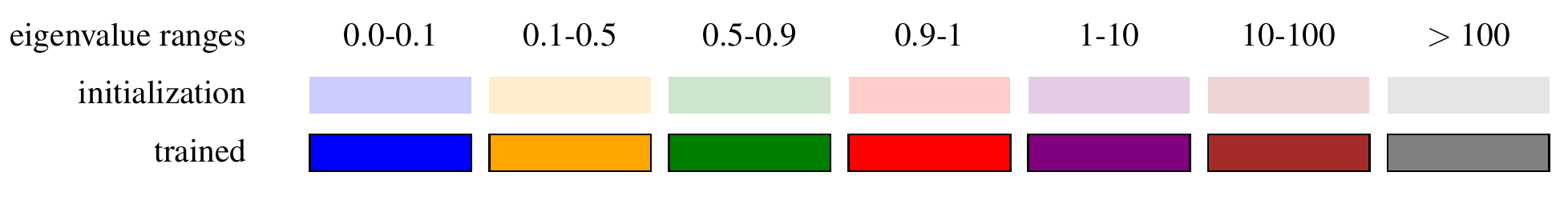}};

    \node (pic4) at (-13.0, -7.5) {\includegraphics[width=7.95cm]{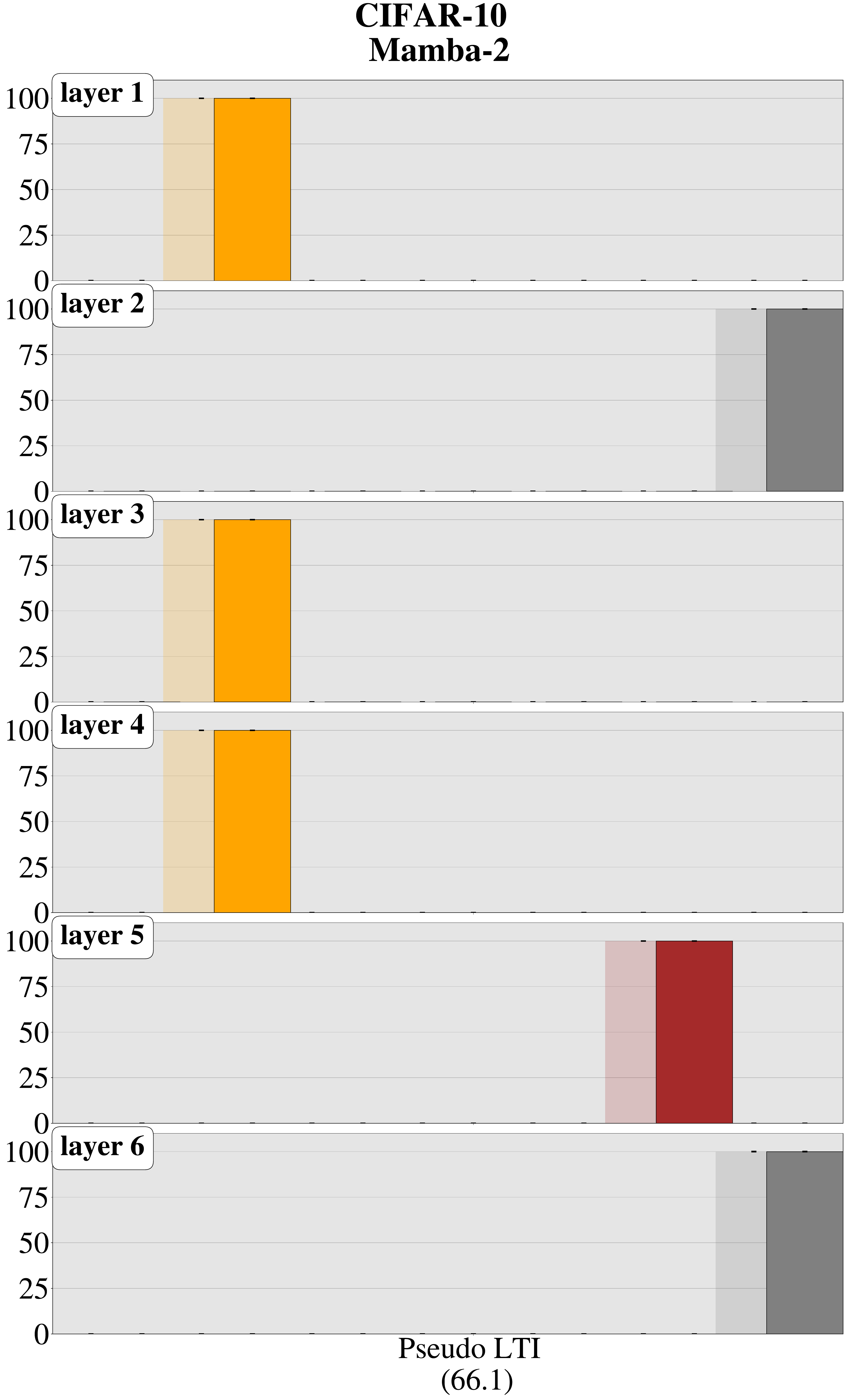}};

    \node (pic5) at (-4.9, -7.5) {\includegraphics[width=7.5cm]    {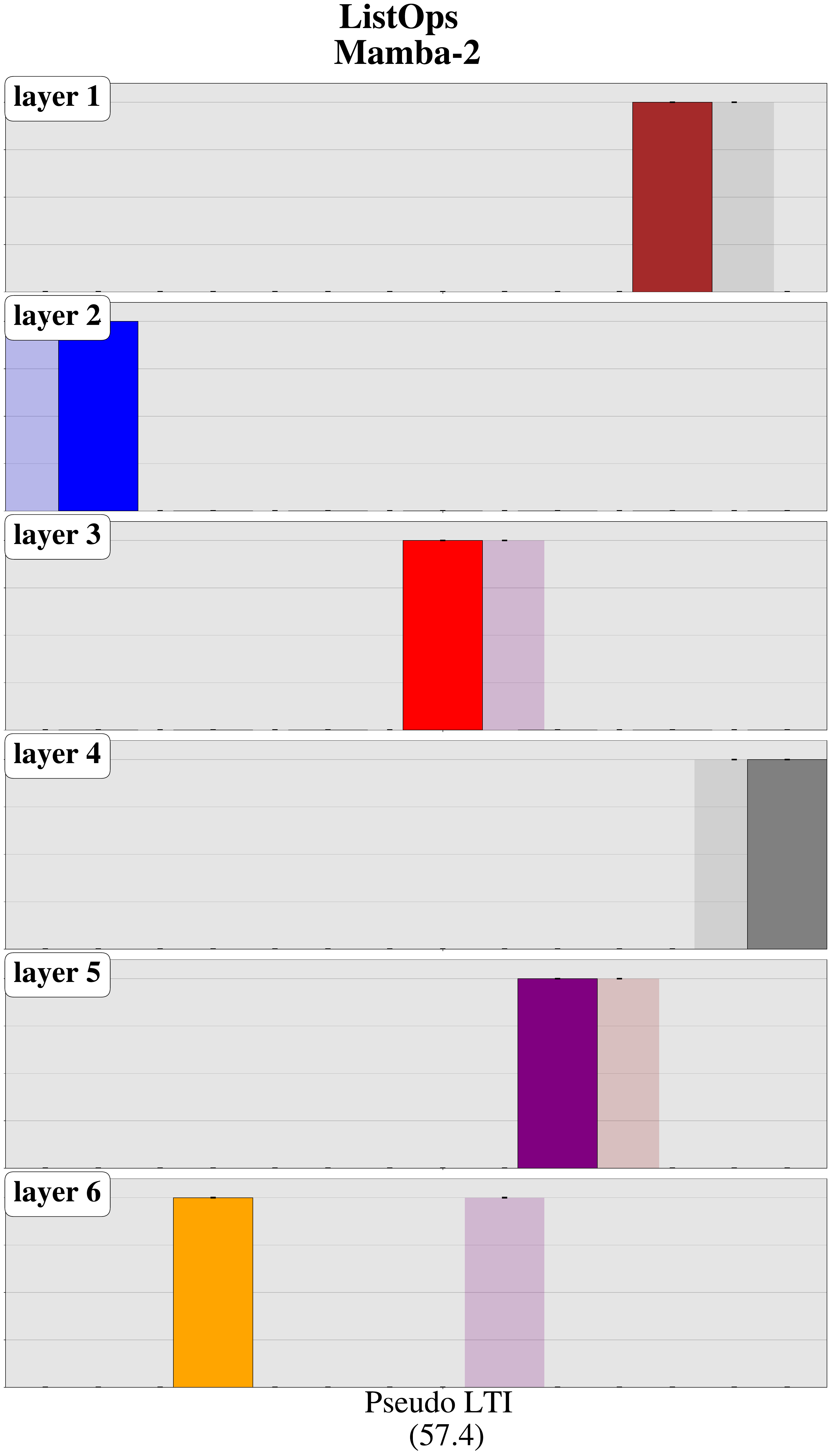}};

    \node (pic6) at (2.9, -5.5) {\includegraphics[width=7.6cm]
    {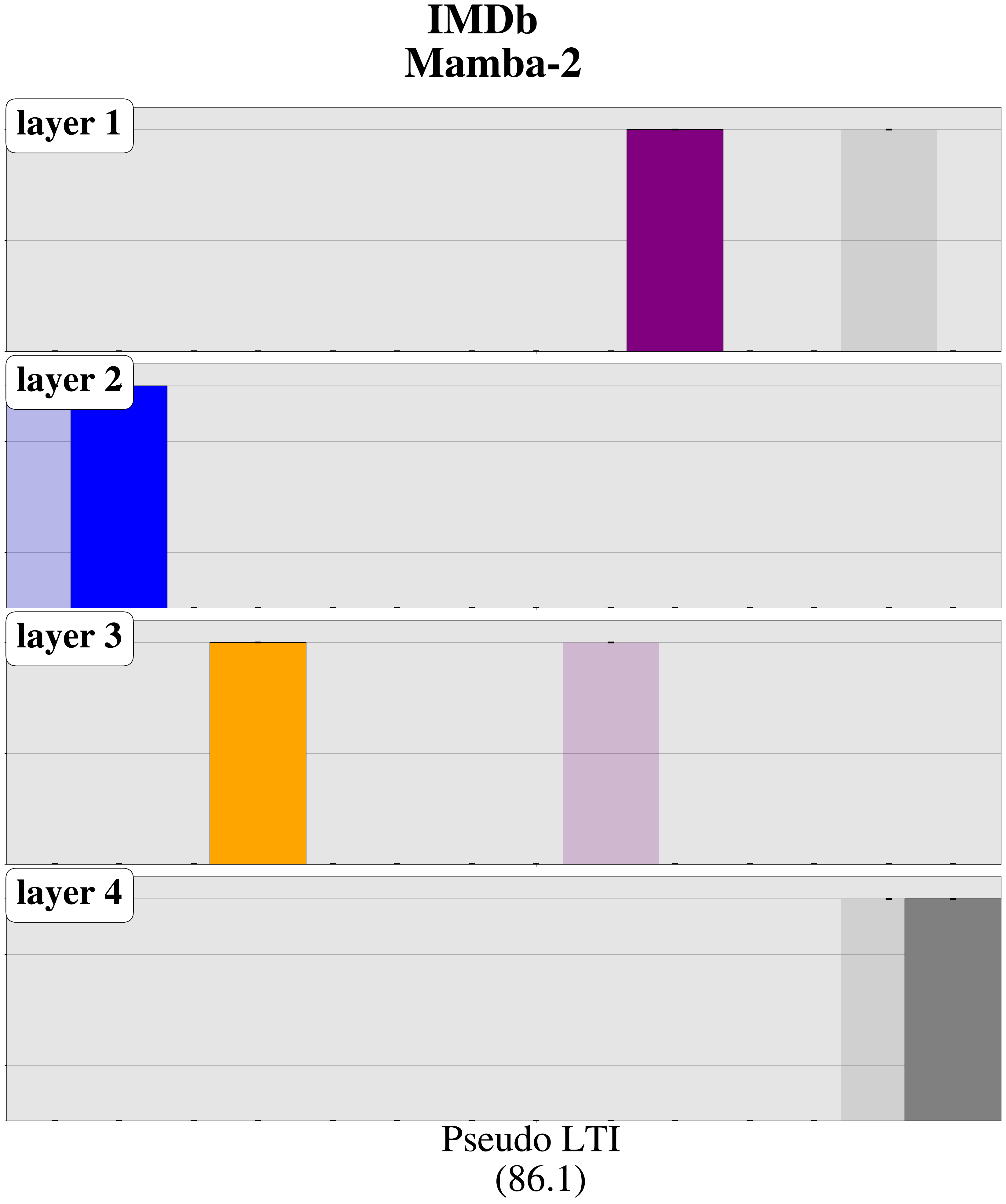}};

    \node (pic0) at (-4.9, -16.0) {\includegraphics[width=22.5cm]{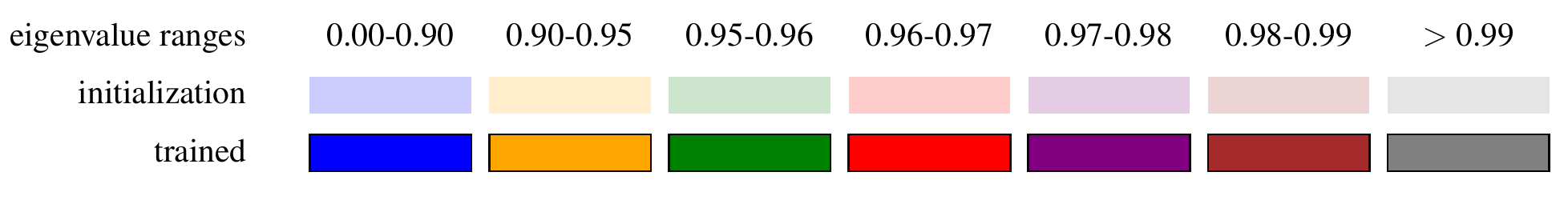}};
    
\end{tikzpicture}
\end{adjustbox}
\caption{Eigenvalue spectra of Mamba-2 Pseudo LTI, one head, across all layers and LRA models. Standard binning is presented in the top row of plots, followed by a more fine-grained visualization around one. The corresponding legends are provided below the respective plots. 
}
  \label{fig:gating_conv_mamba2_appendix}
\end{figure}

\begin {figure*}[h]
\centering
\begin{adjustbox}{max height=0.7\textheight, max width=0.96\textwidth}
\begin{tikzpicture}[scale=1.0]
    
        \node (pic4) at (-3.8, 56.5) {\includegraphics[width=50cm]
    {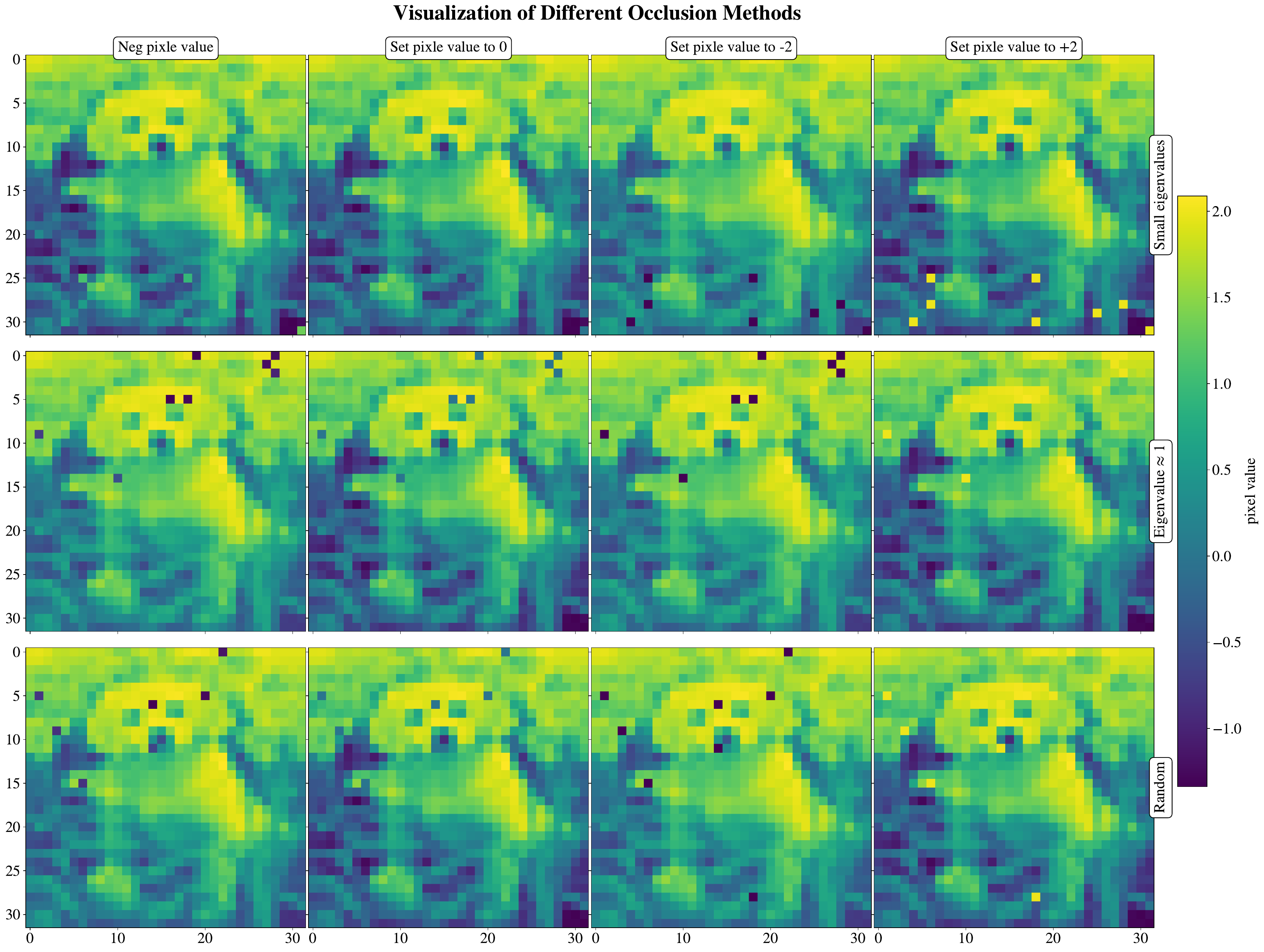}};

\end{tikzpicture}

\end{adjustbox}
\caption{Visualization of the feature importance experiment for different masking criteria (horizontally) and masking operation (vertically) on an example image. 
}
    \label{fig:occlusion_visualization}
\end{figure*}

As the main text reports results for only a subset of models, tasks, and layers, we use this section to present additional findings on the effects of architectural changes, spectral guidance, and feature reduction. These results demonstrate that the conclusions drawn in the main paper also hold for the models, tasks, and layers not covered in the main text. The overview of all presented results is given below:

\begin{itemize}
    \item Figure~\ref{fig:gating_conv_lra_appendix} and Figure~\ref{fig:gating_conv_wikitext_appendix}: comparison of attention models with explicit convolution and gating for one of the heads on LRA tasks and on WikiText benchmark (extension of Figure~\ref{fig:gating_conv});
    \item Figure~\ref{fig:norm_init_appendix}: comparison of different normalization functions for norm attention models on all tasks (extension of Figure~\ref{fig:normalization});
    \item Table~\ref{tab:performance} and Figure~\ref{fig:heatmaps}: summarized performance of nominal models from Figure~\ref{fig:main_plot} and the absolute change in performance after introducing the architectural changes (introduction of explicit gating and convolution, as well as the change of normalization function for norm attention);
    \item Figure~\ref{fig:gating_conv_mamba2_appendix}: contains eigenvalue distributions for all layers of one of the heads for Mamba-2 pseudo-LTI variation on all LRA tasks (extension of Figure~\ref{fig:normalization}), as well as fine-grained eigenvalue distribution plots showing that the eigenvalues shift after training, though almost negligibly.
    \item Figure~\ref{fig:gating_conv_mamba2_appendix} provides results on the influence of pretraining on the eigenvalue spectra of the pre- and post-trained model. 
    \item Figure~\ref{fig:occlusion_visualization} and Figure~\ref{fig:additional_features}: visualization of different occlusion methods and their effect on performance, respectively.
\end{itemize}

\begin{figure}[h]
    \centering
    \includegraphics[width=0.95\linewidth]{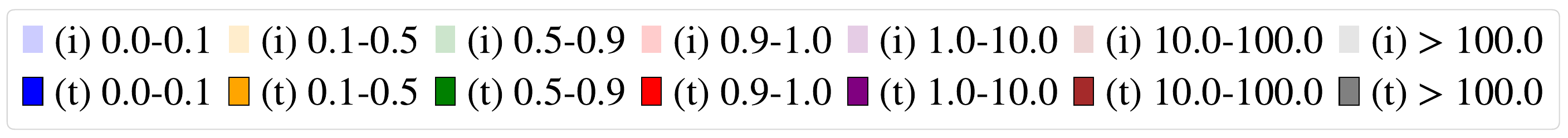}
    \caption{Plot legend of all subsequent histogram figures.  }
    \label{fig:legend}
\end{figure}

All figures make use of the same legend as Figure \ref{fig:main_plot}, once again displayed in Figure \ref{fig:legend}, below. Thereby, bars show the proportion of eigenvalues falling within discretized ranges. Lightly shaded and darker bars indicate the distribution at initialization and after training, respectively. Error bars denote standard deviation across input sequences. Model performance, measured as perplexity for WikiText
(lower is better) and percentages of correct output sequences for the remaining tasks (higher is better), is indicated in the parentheses. 

\subsection{Attention-Based Models with Explicit Convolution or Gating}
\label{subapx:convolution_and_gating}

  \begin {figure}[H] 
\centering
\begin{adjustbox}{max height=0.91\textheight, max width=0.90\textwidth}
\begin{tikzpicture}[scale=1.0]

    \node (pic1) at (-13.0, 6.5) {\includegraphics[width=7.95cm]{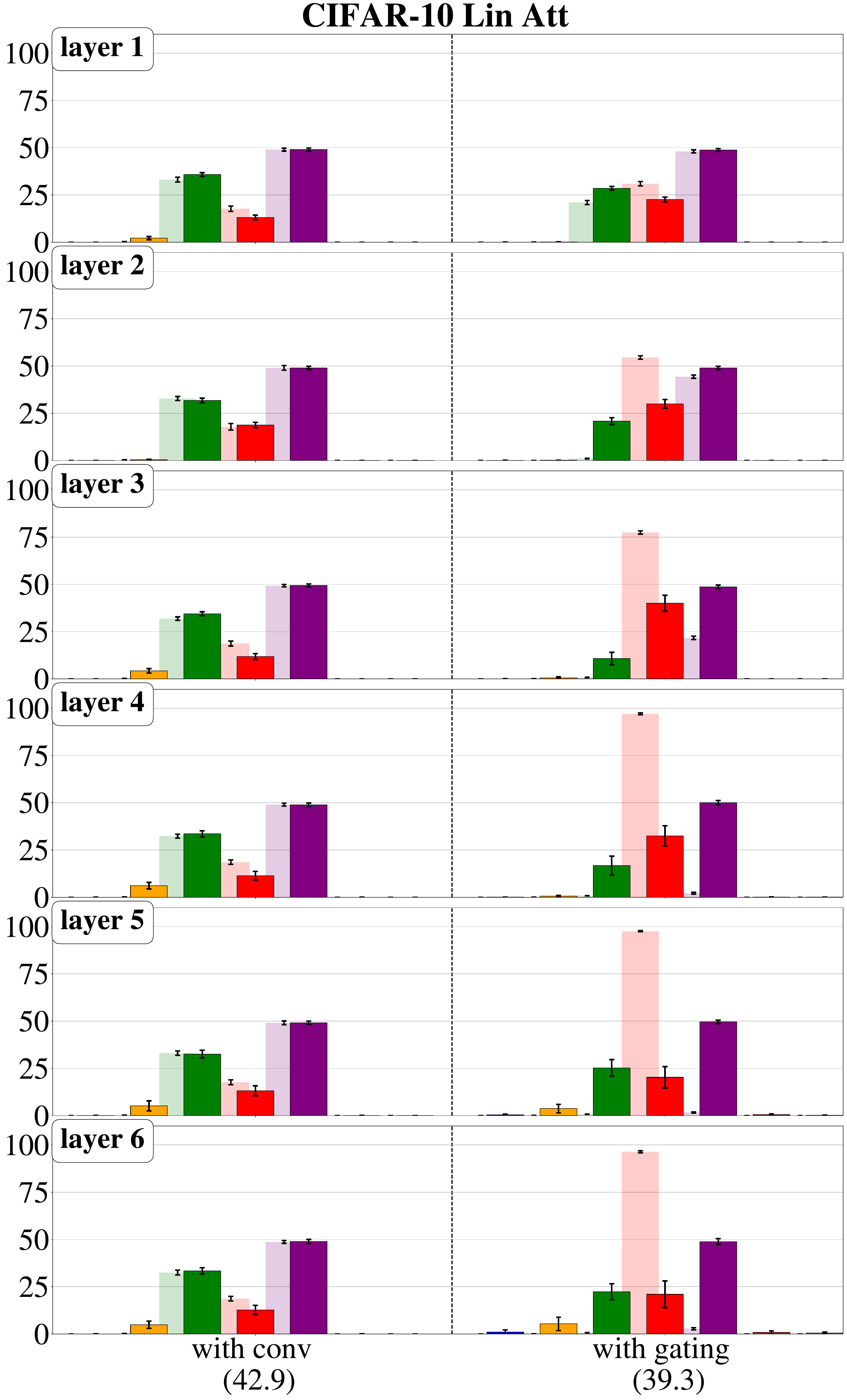}};

    \node (pic2) at (-4.9, 6.5) {\includegraphics[width=7.5cm]    {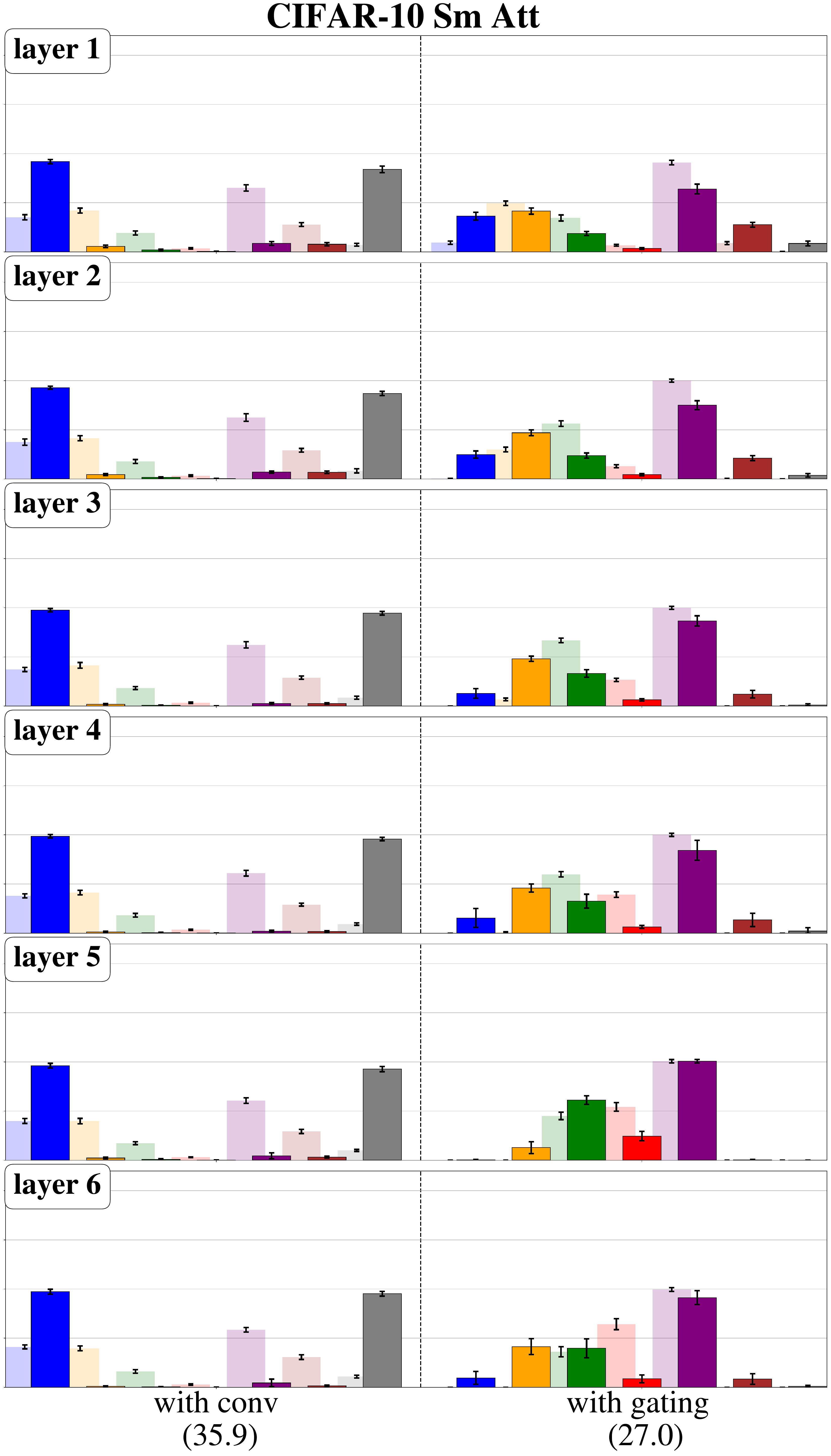}};

    \node (pic3) at (2.8, 6.5) {\includegraphics[width=7.5cm]
    {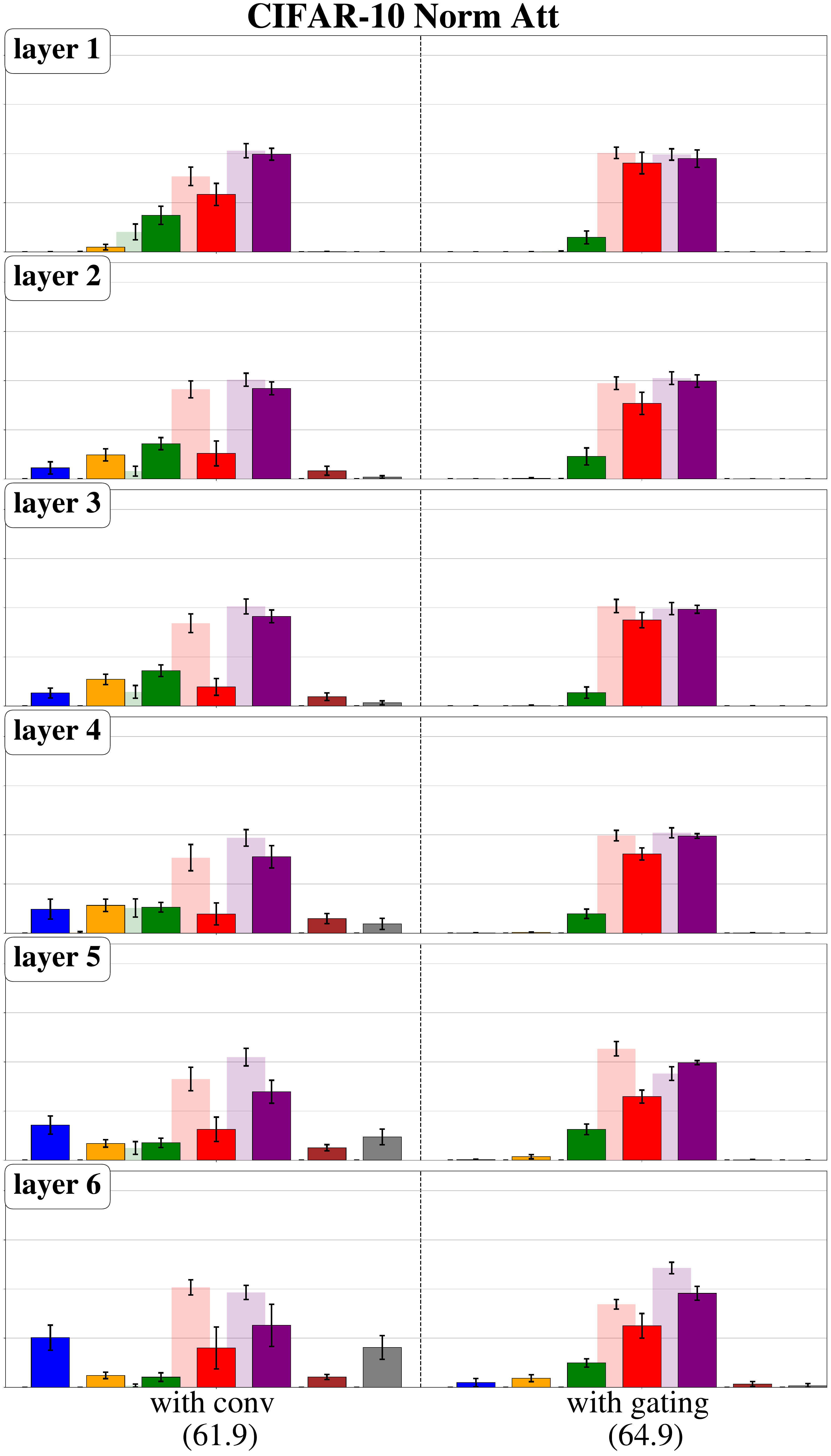}};

    \node (pic5) at (-13.0, -4.5) {\includegraphics[width=7.95cm]
    {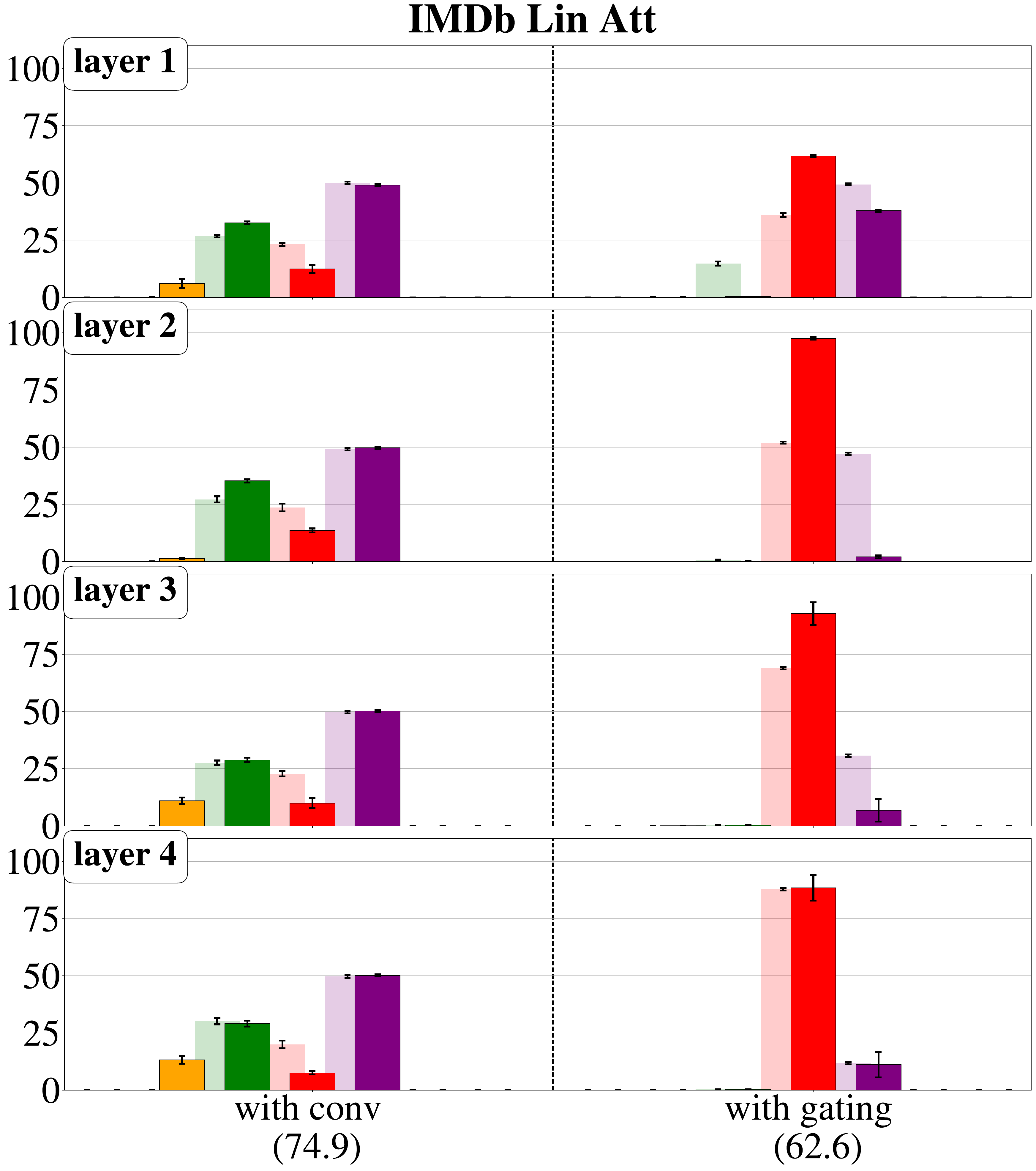}};

    \node (pic6) at (-4.9, -4.5) {\includegraphics[width=7.5cm]
    {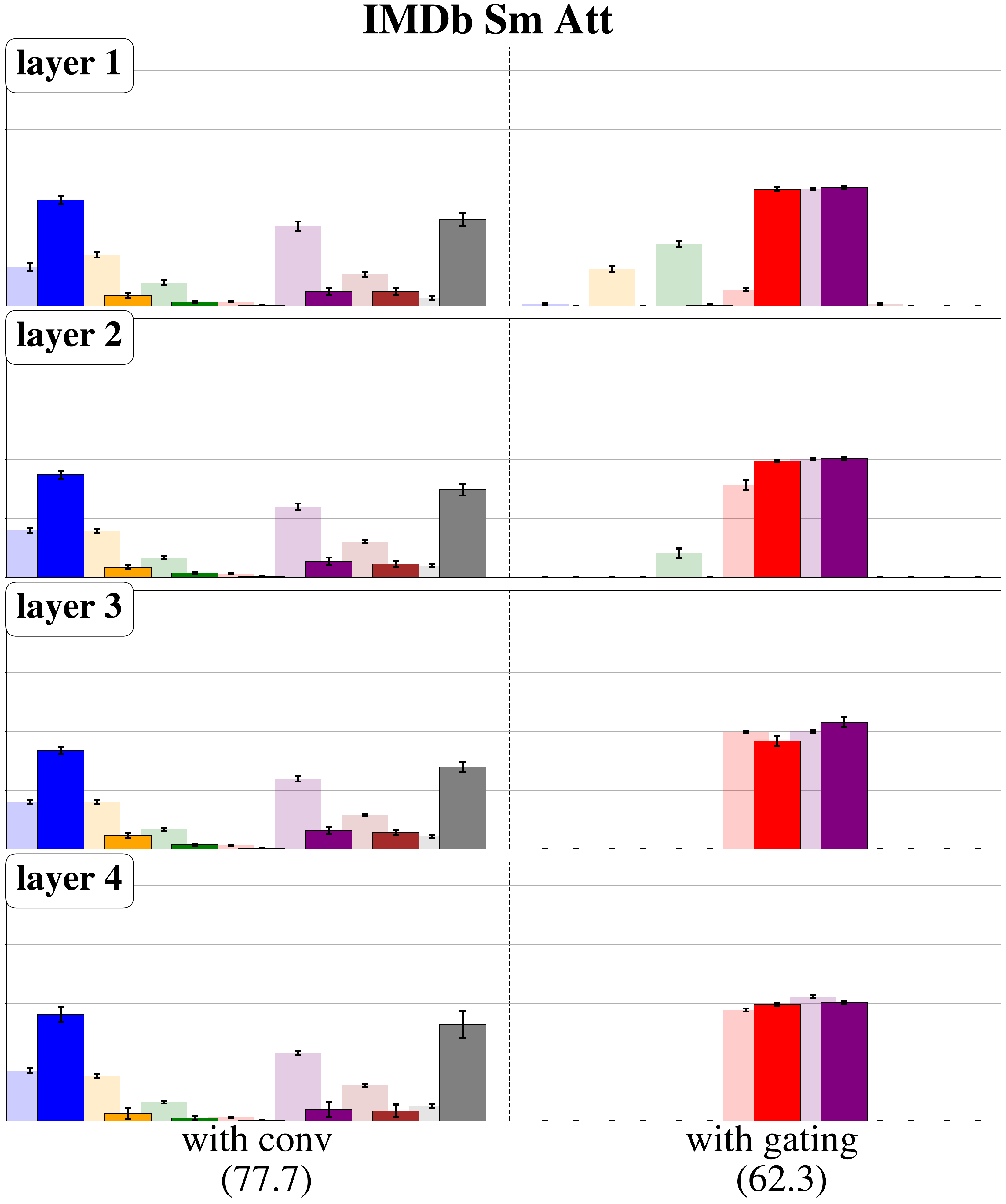}};

    \node (pic7) at (2.8, -4.5) {\includegraphics[width=7.5cm]
    {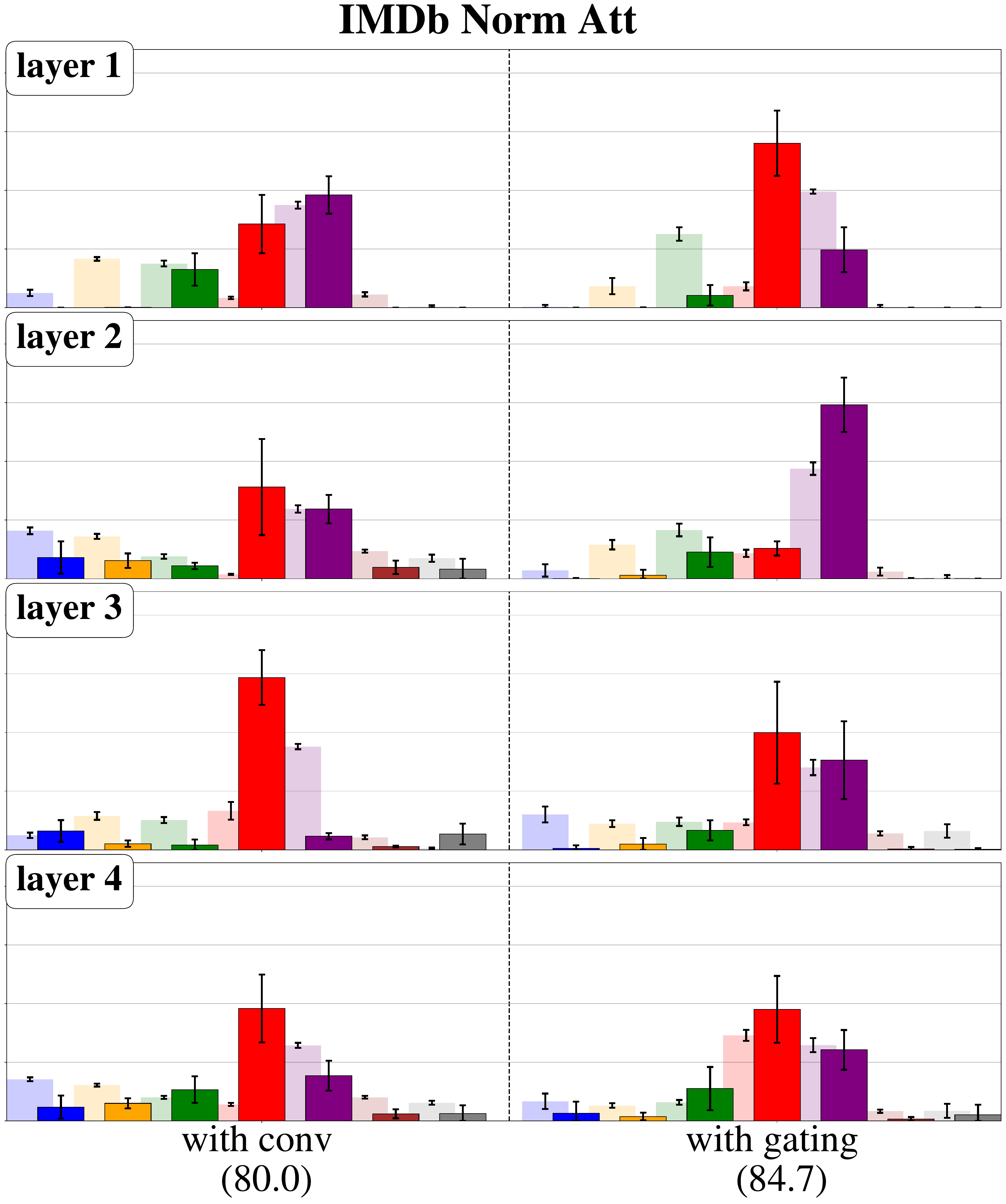}};

    \node (pic8) at (-13.0, -15.5) {\includegraphics[width=7.95cm]
    {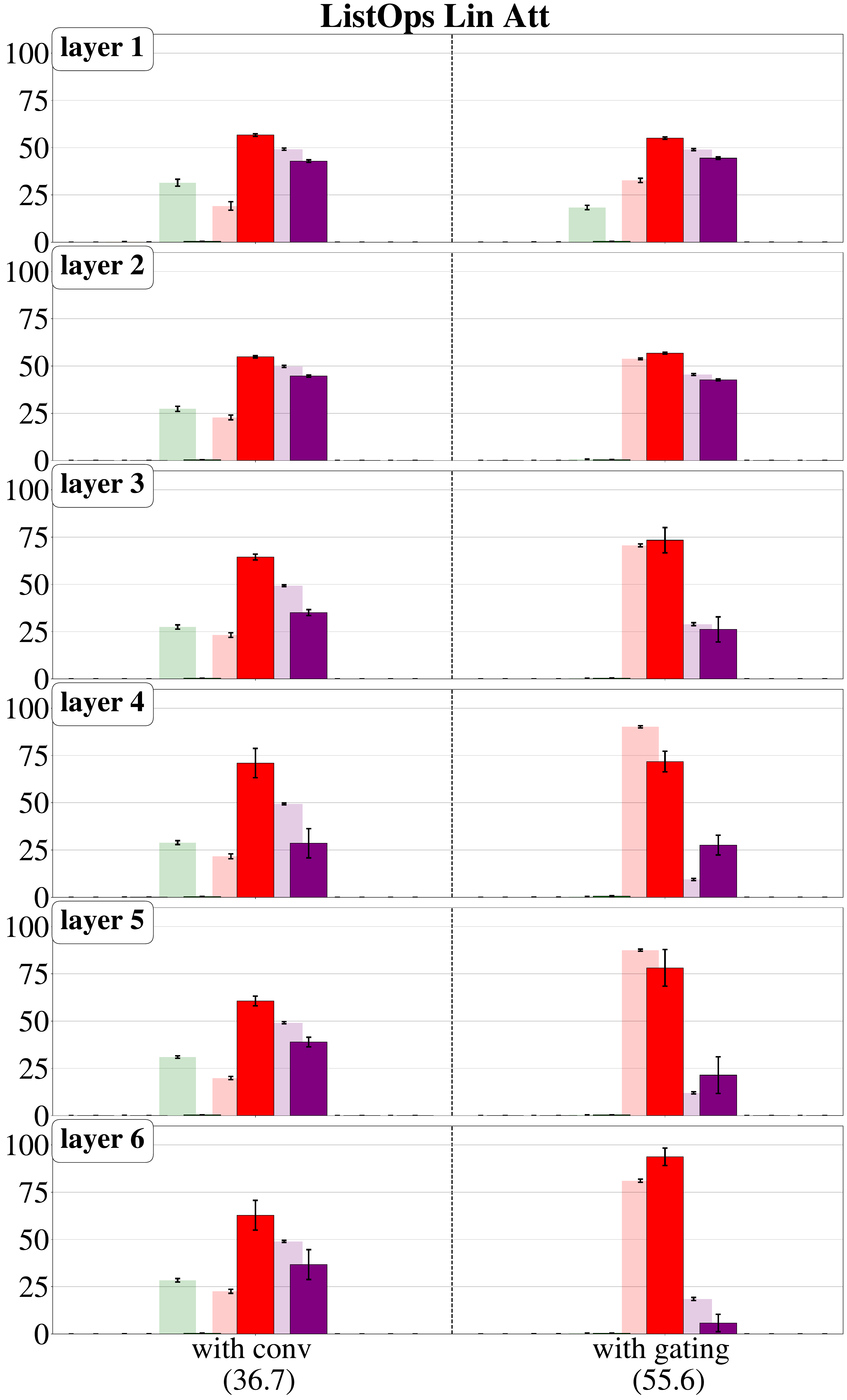}};

    \node (pic9) at (-4.9, -15.5) {\includegraphics[width=7.5cm]
    {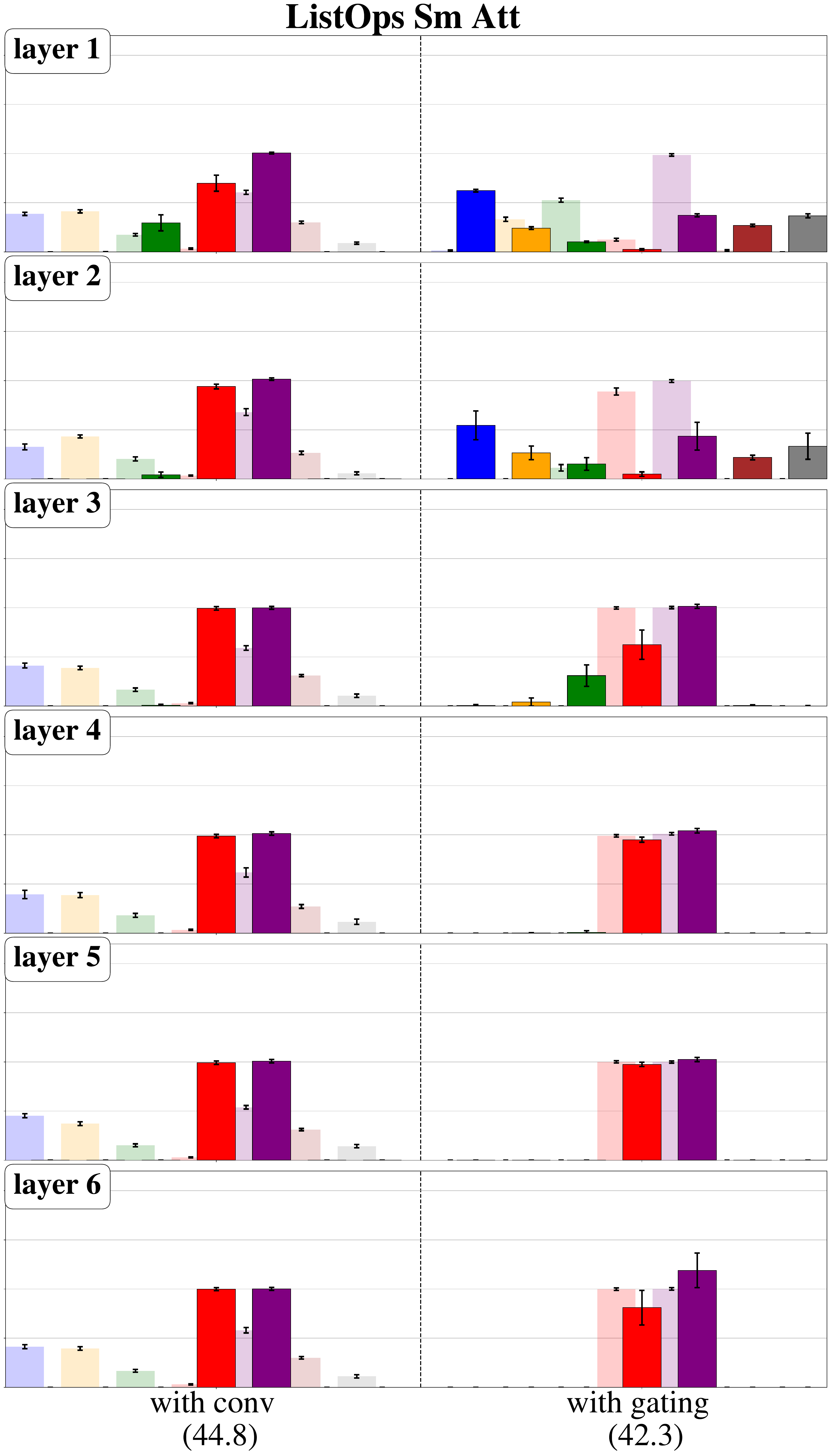}};

    \node (pic10) at (2.8, -15.5) {\includegraphics[width=7.5cm]
    {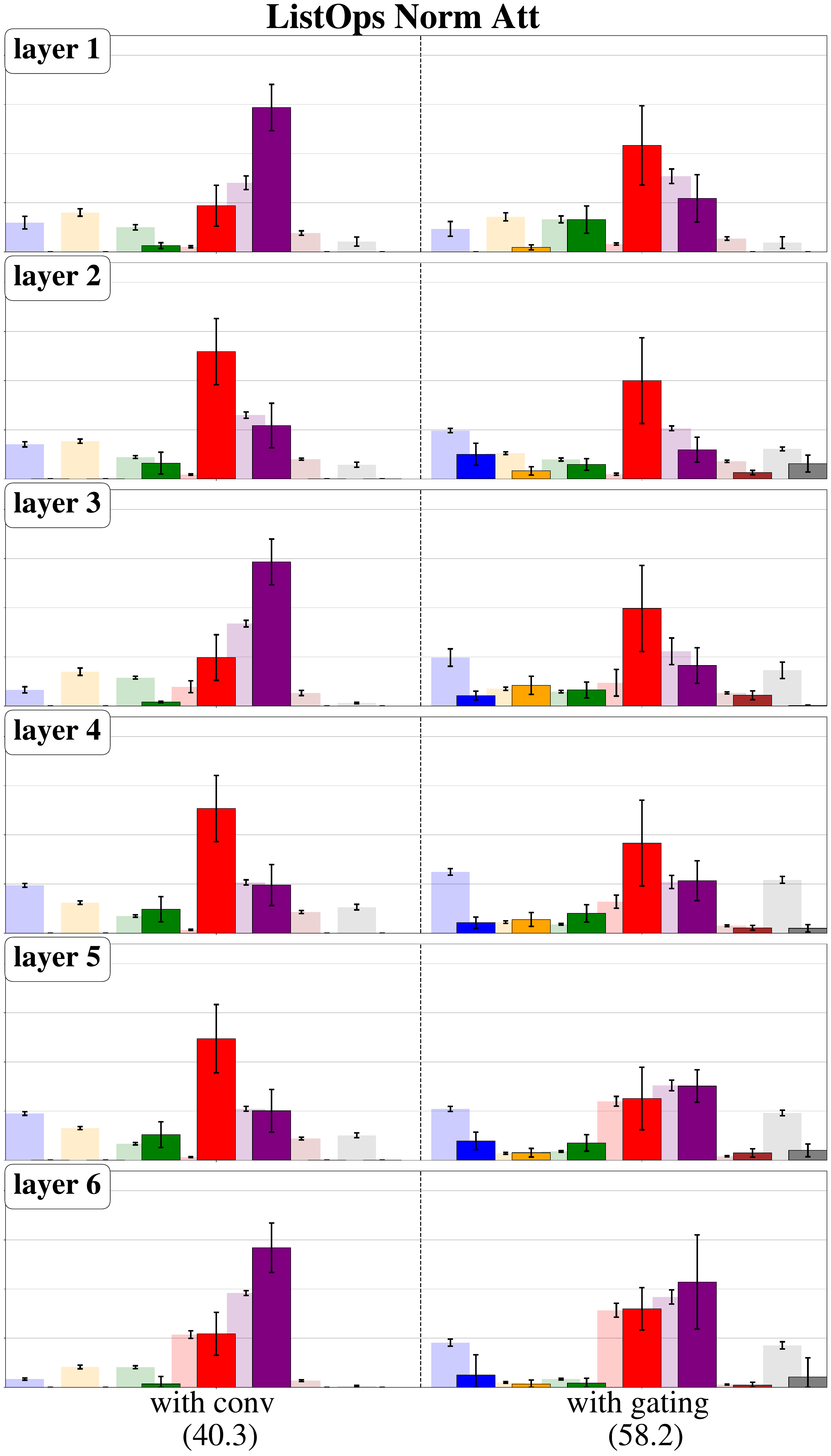}};

\end{tikzpicture}
\end{adjustbox}
\caption{Comparison of the effects of gating and convolution on the eigenvalue spectra for one head, across CIFAR-10, IMDb, and ListOps, the three investigated attention models, and all layers. 
}
  \label{fig:gating_conv_lra_appendix}
\end{figure}

\FloatBarrier
\subsection{Guidance of Eigenvalue Spectra Through Normalization Function}
\label{subapx:normalization_function}

\FloatBarrier
\subsection{Resulting Performance Changes}
\label{subapx:resulting_performance_changes}

\begin{figure}[H]
    \centering
    \includegraphics[width=0.99\linewidth]{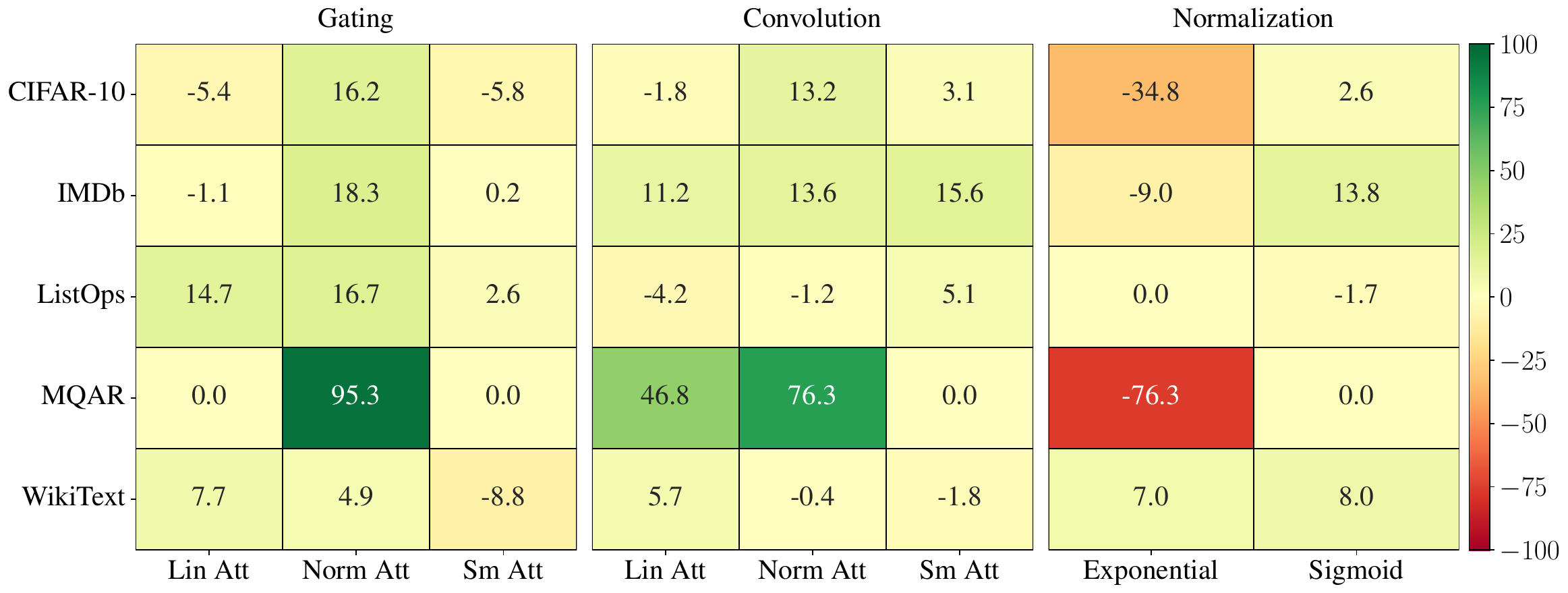}
    \caption{The absolute change in performance after each architectural change across models and tasks. For gating and convolution, the difference is computed in comparison to the nominal models shown in Figure~\ref{fig:main_plot}, whose performance is summarized in Table \ref{tab:performance}. For normalization, the difference is computed with respect to the performance of norm attention with convolution and softplus as normalization function, shown in Figure~\ref{fig:norm_init_appendix}. For Wikitext, where we indicate perplexity, the sign of the perplexity change is flipped to align with the color scheme, i.e., a positive value indicates decreasing perplexity.}
    \label{fig:heatmaps}
\end{figure}

In this subsection, we examine how model performance across all tasks is affected by the introduction of convolution and gating, as well as by modifying the normalization function in norm attention. Table~\ref{tab:performance} reports model performance on the tasks from the initial study described in main paper. Figure~\ref{fig:heatmaps} then illustrates how these performances change as a result of these modifications. We observe numerous improvements, particularly across all tasks for norm attention, and across all models on ListOps when gating is introduced. In addition, adding convolution to linear and norm attention yields significant improvements on MQAR, while all considered models achieve improved performance on IMDb.

\begin{table}[H]
 \caption{Performance of models across different tasks for the nominal seed shown in Figure \ref{fig:main_plot}.
 \label{tab:performance}
    }
\centering
\begin{tabular}{lccccccccccc}
\toprule
 & S4 & LRU & Mamba-2 & Lin Att & Sm Att & Norm Att &\\
\midrule
CIFAR-10 & $\mathbf{89.0}$ & 68.9 & 66.6 &  44.7 & 32.8 & 48.7 \\
IMDb & 81.5 & 86.1 & $\mathbf{86.8}$ & 63.7 & 62.1 & 66.4 \\
ListOps & $\mathbf{55.5}$ & 40.1 & 53.9 & 40.9 & 39.7 &  41.5\\
\midrule
MQAR & $<1.5$ & $<1.5$ & $\mathbf{>99.0}$ & $<1.5$ &  $\mathbf{>99.0}$ & $<1.5$ \\
\midrule
WikiText & $74.3$ & $60.9$ & $40.1$ & $39.5$ &  $\mathbf{34.4}$ & $43.0$\\
\bottomrule
\end{tabular}

\end{table}

\FloatBarrier
\subsection{Mamba-2 as an LTI}
\label{subapx:mamba2_as_lti}

\subsection{Pretraining} It was shown in \cite{amos2023never} that attention-based models achieve significantly better performance on the LRA benchmark and can even outperform SSMs when they are first pretrained on standard denoising tasks, and subsequently fine-tuned on the same task data. We analyze the eigenvalue spectra of a transformer  with softmax attention from \cite{amos2023never} trained on CIFAR-10 and IMDb (shown in Figure~\ref{fig:normalization}), leading to two main observations. First, the eigenvalue distributions of the pretrained and fine-tuned models are nearly identical. Second, despite their markedly higher performance compared to the models from our initial study in the main paper, their resulting spectra closely resemble each other. This suggests that the eigenvalue distribution reflects an inner mechanism of softmax attention, and that pretraining ensures this mechanism is tuned properly, i.e., that the forgetting (reflected in setting eigenvalues to zero) is done when necessary and placed correctly.

  \begin {figure*}[h]
\centering
\begin{adjustbox}{max height=0.7\textheight, max width=0.94\textwidth}
\begin{tikzpicture}[scale=1.0]

    \node (pic2) at (-2.6, 10.5) {\includegraphics[width=3.5cm]
    {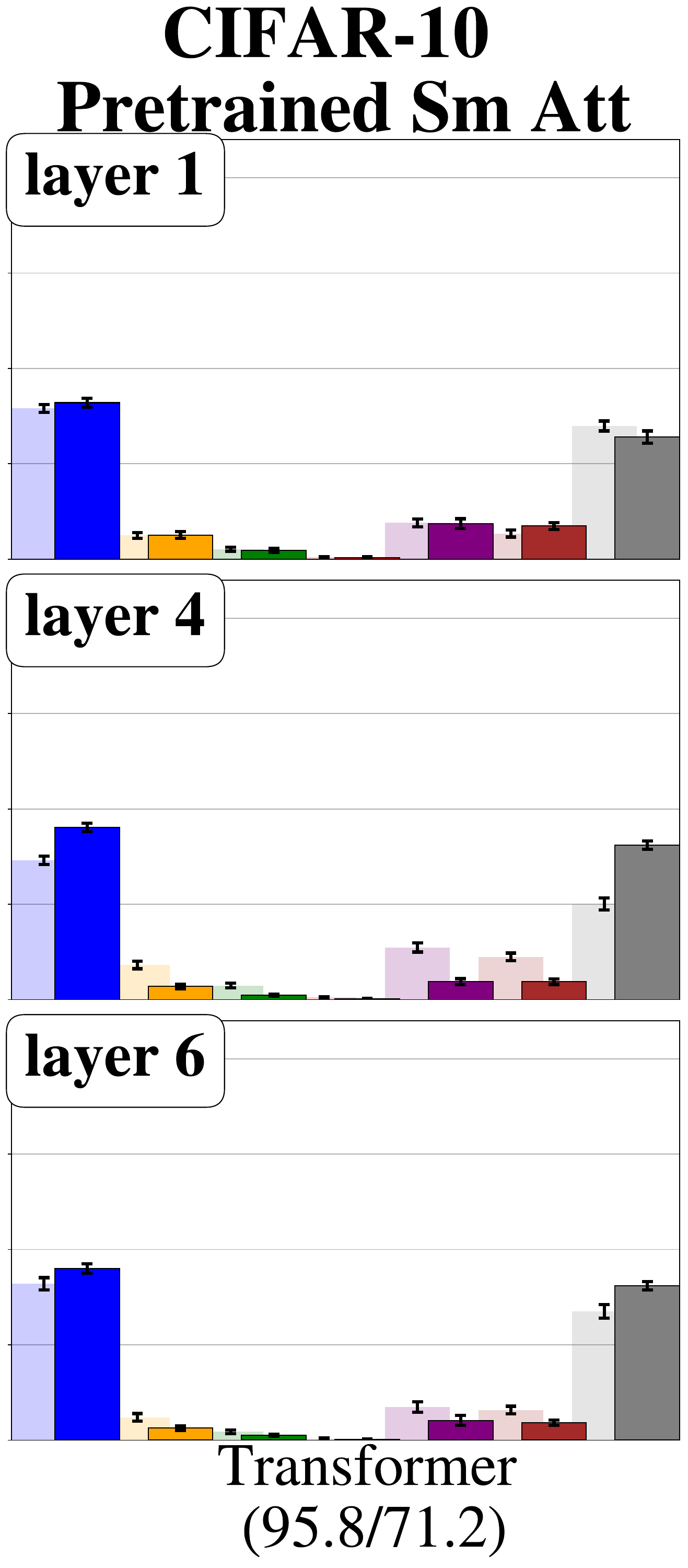}};

    \node (pic21) at (2.6, 10.5) {\includegraphics[width=3.5cm]
    {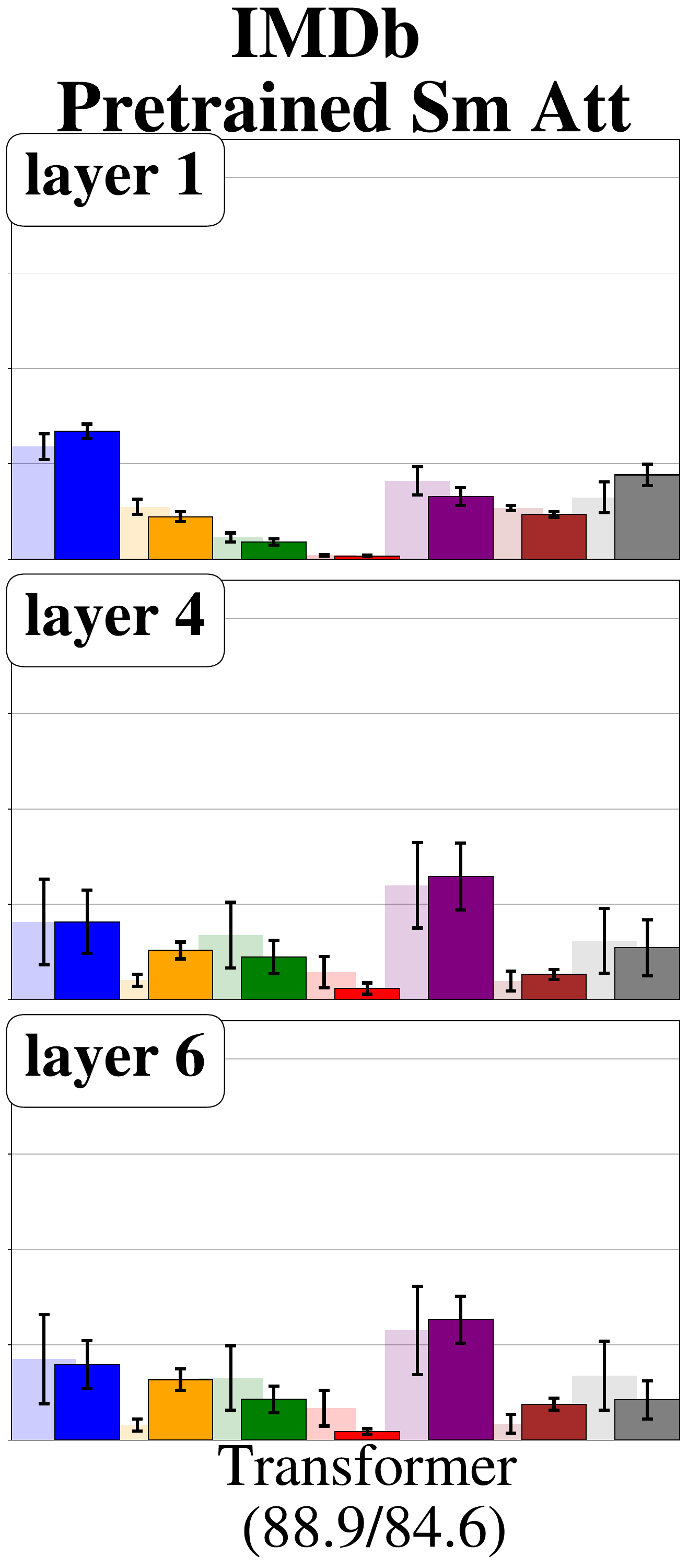}};

\end{tikzpicture}

\end{adjustbox}
\caption{Eigenvalue distributions for softmax attention for pretrained and finetuned models (represented as initial/ final distribution and performance)
}
    \label{fig:pretraining}
\end{figure*}

\subsection{Further Details on Feature Reduction Experiments}
\label{subapx:feature_importance}

\begin {figure*}[h]
\centering
\begin{adjustbox}{max height=0.7\textheight, max width=0.96\textwidth}
\begin{tikzpicture}[scale=1.0]

        \node (pic4) at (-3.8, 56.5) {\includegraphics[width=50cm]
    {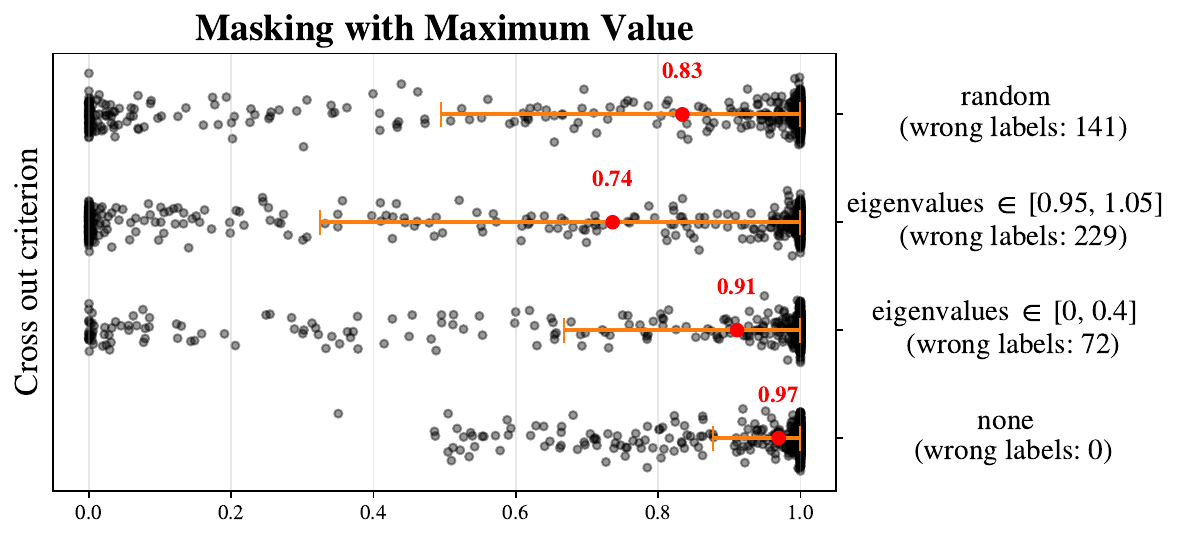}};

    \node (pic3) at (-3.8, 33.5) {\includegraphics[width=50cm]
    {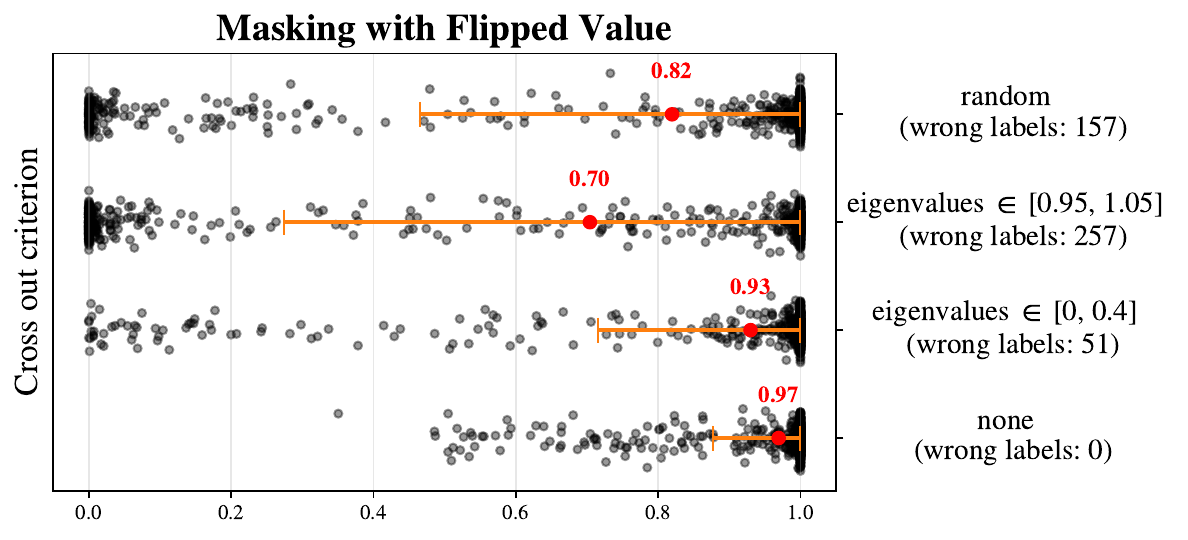}};

    \node (pic21) at (-3.8, 10.5) {\includegraphics[width=50cm]
    {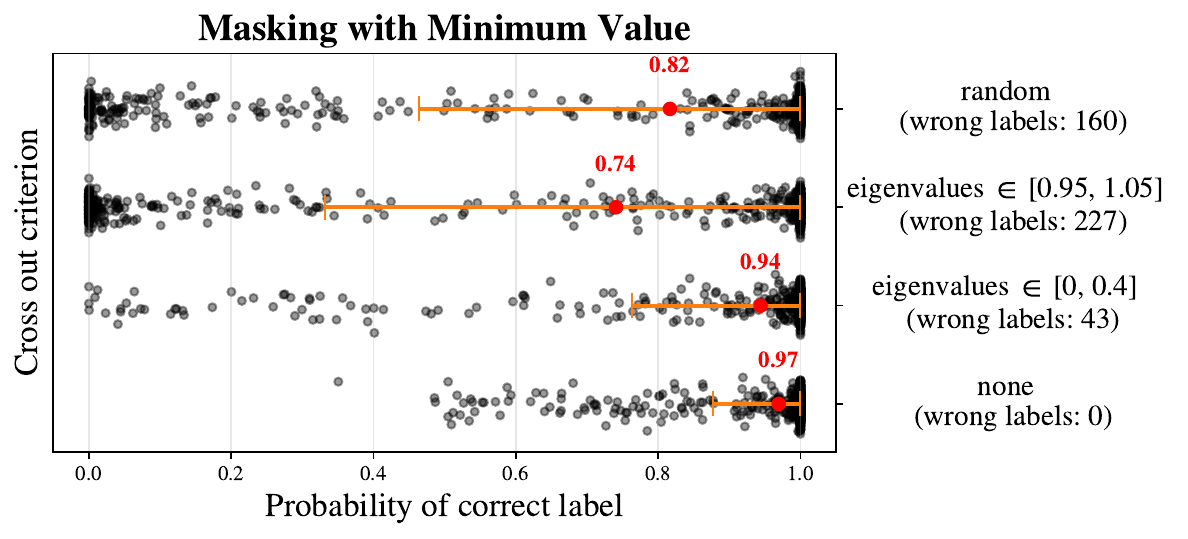}};

\end{tikzpicture}

\end{adjustbox}
\caption{Results of the feature importance experiment for different masking criteria.  Mean values are indicated and reported in red, error bars depict one standard deviation and the number of images for which pixel occlusion resulted in wrongful classification is reported beneath each selection criterion.
}
    \label{fig:additional_features}
\end{figure*}

We select norm-attention model with convolution and elu normalization (achieving 61.8\% accuracy) as one of the best-performing attention models on CIFAR-10 in our study. From the CIFAR-10 test set, we collect all images that this model classifies correctly and then sample a class-balanced subset of 790 images for analysis. \\
We evaluate several pixel-masking criteria and masking strategies. The number of occluded pixels is thereby chosen for each image by the smallest number of pixels that fall into each of the investigated masking criteria. The masking criteria specify how pixels are selected: (i) uniformly at random (as a baseline), or (ii) based on whether their eigenvalues fall within particular ranges over all heads. We consider two such ranges: low eigenvalues (when averaged across all heads of the first layer) and eigenvalues close to 1 (across all heads of the first layer), corresponding to pixel groups we associate with forgetting and remembering, respectively. \\
To ensure that our results are not sensitive to the specific masking operation, we employ four masking strategies commonly used in the literature (see \cite{sokol2024interpretable} for an overview): replacing pixels with a gray value \cite{zeiler2014visualizing}, white, black, or the flipped pixel value \cite{bach2015pixel}. \\
For each masked image, we recompute the model’s prediction and record (i) the change in the predicted probability of the correct class and (ii) whether the prediction becomes incorrect. The results, shown in Figures~\ref{fig:feature_selection} and~\ref{fig:additional_features}, demonstrate that masking pixels with low eigenvalues produces few misclassifications, whereas masking pixels with eigenvalues close to 1 causes a substantial increase in errors, regardless of the masking strategy.

\end{document}